%% file: main.tex
\documentclass{article}

\PassOptionsToPackage{numbers, compress}{natbib}
\usepackage[preprint]{neurips_2022}

\usepackage[utf8]{inputenc} % allow utf-8 input
\usepackage[T1]{fontenc}    % use 8-bit T1 fonts
\usepackage[colorlinks=true]{hyperref}       % hyperlinks
\usepackage{url}            % simple URL typesetting
\usepackage{booktabs}       % professional-quality tables
\usepackage{amsfonts}       % blackboard math symbols
\usepackage{nicefrac}       % compact symbols for 1/2, etc.
\usepackage{microtype}      % microtypography
\usepackage{xcolor}         % colors

\usepackage[accsupp]{axessibility}  % Improves PDF readability for those with disabilities.

\usepackage{multirow}
\usepackage{booktabs}
\usepackage[abs]{overpic}
\usepackage{makecell}
\usepackage{parskip}
\usepackage{xspace}

\usepackage{adjustbox}
\usepackage{algorithm}
\usepackage{algpseudocode}
\usepackage{subcaption}
\usepackage{mdframed}
\usepackage{xcolor}
\usepackage{wrapfig}

\usepackage{tikz}
\usetikzlibrary{positioning}

\usepackage[bottom]{footmisc}
\usepackage{placeins}

\newcommand{\mask}[0]{\texttt{[MASK]}\xspace}

\newcommand{\set}[4]{\ensuremath{{\{#1_{#2}\}}_{#2=#3}^{#4}}\xspace}

\newcommand{\bed}[0]{\textsc{Bedroom}\xspace}
\newcommand{\lib}[0]{\textsc{Library}\xspace}
\newcommand{\liv}[0]{\textsc{Living}\xspace}
\newcommand{\din}[0]{\textsc{Dining}\xspace}

\usepackage{pifont}% http://ctan.org/pkg/pifont
\newcommand{\cmark}{\ding{51}}%
\newcommand{\xmark}{\ding{55}}%
\newcommand{\para}[1]{\noindent\textbf{#1}:}

\definecolor{pastel_pink}{HTML}{F535AA}

\include{symbols}

\title{COFS: Controllable Furniture layout Synthesis}

\author{%
    \textbf{Wamiq Reyaz Para}$^1$ \quad   \textbf{Paul Guerrero}$^2$  \quad \textbf{Niloy Mitra}$^{2,4}$  \quad  \textbf{Peter Wonka}$^1$ \\ \\
    $^1$ KAUST \quad $^2$ Adobe Research \quad  $^3$ University College London 
}

\begin{document}

\maketitle

\begin{abstract}
Scalable generation of furniture layouts is essential for many applications in virtual reality, augmented reality, game development and synthetic data generation. Many existing methods tackle this problem as a sequence generation problem which imposes a specific ordering on the elements of the layout making such methods impractical for interactive editing or scene completion. Additionally, most methods focus on generating layouts unconditionally and offer minimal control over the generated layouts. We propose COFS, an architecture based on standard transformer architecture blocks from language modeling. The proposed model is invariant to object order by design, removing the unnatural requirement of specifying an object generation order. Furthermore, the model allows for user interaction at multiple levels enabling fine grained control over the generation process. Our model consistently outperforms other methods which we verify by performing quantitative evaluations. Our method is also faster to train and sample from, compared to existing methods.

% \keywords{Generative Modelling, Transformers, Masked Language Models, Layout Synthesis, Indoor Scene Synthesis}
\end{abstract}

\input{sections/introduction}
\input{sections/related_work}
\input{sections/method}

\input{sections/results}

\input{sections/conclusion}

\section{Acknowledgements}
We would like to thank Visual Computing Center (VCC), KAUST for support, gifts from Adobe Research and the UCL AI Centre.

{
\small
\bibliographystyle{plain}
\bibliography{main}
}
%%%%%%%%%%%%%%%%%%%%%%%%%%%%%%%%%%%%%%%%%%%%%%%%%%%%%%%%%%%%

%%%%%%%%%%%%%%%%%%%%%%%%%%%%%%%%%%%%%%%%%%%%%%%%%%%%%%%%%%%%
\appendix

\clearpage

\standalonetitle{COFS: Controllable Furniture layout Synthesis \\ {Supplementary Material}}

\begin{abstract}
    In this supplementary document accompanying our main submission, we describe our system in more detail. In particular, we detail each of the components of our architecture, including the training protocol. We discuss our method in comparison to ATISS. We describe the metrics and how they were evaluated and compared against the baselines. We provide details on the sampling strategy that we employ. We perform an ablation study justifying our design choices. We conclude with additional quantitative and qualitative results. We also provide a table of key notation used in the main paper. Additional details can be found on the \href{http://cofs-neurips.co}{project page}.
\end{abstract}

\section{Detailed Architecture and Training Setup}
We base our architecture on ATISS~\cite{Paschalidou2021NEURIPS} in order to ensure a fair comparison to our closest competitor, using the same underlying library~\cite{katharopoulos_et_al_2020}. Consequently, most of the building blocks are shared. However, we would like to point out major differences in this section.

\paragraph{Layout sequence:} During training, we construct the sequence $S$ corresponding to the layout by arranging the object bounding boxes in a random order with a permutation $\pi$ and concatenating their bounding box attributes as individual tokens.
\begin{equation}
\begin{aligned}
     s_2 &= \tau_{\pi_1}, \; s_3 = (\tau_{\pi_1})_x, \; s_4 = (\tau_{\pi_1})_y , \; s_5 = (\tau_{\pi_1})_z ,  \; s_6 = (e_{\pi_1})_x ,\; \cdots \; s_9 = \tau_{\pi_2}\; \cdots \; \\
     s_1 &= \texttt{SOS}, \; s_k = \texttt{EOS}
\end{aligned}
\end{equation}
where $\tau_{\pi_1}$ and $(t_{\pi_1})_x$ represent the class and $x-$translation of the first object after permutation, $\tau_{\pi_2}$ represents the class of the second object after permutation and so on.

Object attributes are always flattened the same way in our implementation, although in principle the attribute order can itself be permuted. We use the same attribute order for ease of implementation.
% To construct the condition $C$, we drop the \texttt{SOS} and \texttt{EOS} tokens and randomly mask out a random ratio of the tokens in the remaining sequence. We prepend the boundary representation $\cI$ from the Boundary Encoder to the other embeddings.

\paragraph{Embeddings:} We described how we generate embeddings for the tokens in $C$ and $S$. We use a learnable matrix $E_{class}$ of dimension $n_{\tau} \times 256$ to encode the type $\tau_i$, with each row corresponding to one class. We use an additional \mask class. For the other attributes of translation ($t_i$), size ($e_i$) and rotation ($r_i$), we use sinuosoidal positional encodings~\cite{vaswani, Paschalidou2021NEURIPS}  with 128 levels ($L = 128$). We call these embeddings $\gamma$:
\begin{equation}
    \gamma(b) =
    \begin{cases}
                (\sin(2^0\pi b), \cos(2^0\pi b), \dots, \sin(2^{L-1}\pi b), \cos(2^{L-1}\pi b)) & \qquad \text{if \quad} b \in \{t, e, r\} \\
                E_{class}[\tau, :] & \qquad \text{if \quad} b \in \{\tau\}
    \end{cases}
\end{equation}
For the encoder, the embeddings of $\mathcal{R}^i$ are a learned matrix $E_r$ of dimension $8 \times 256$. Each row corresponds to a different type of attribute - one for type, 3 each for translation and size, and one for the rotation. The embedding of $\mathcal{O}^i$ are again a learned matrix $E_o$ of dimension $k \times 256$, where $k$ is the maximum number of objects. For the decoder, the embeddings of $\mathcal{P}^i$ are also a learned matrix $E_p$ of size $n \times 256$.
The final embeddings are the \textit{sum} of the corresponding embeddings:
\begin{equation}
\begin{aligned}
    \gamma_e(b_i) &=  \gamma(b_i)  + E_r[\mathcal{R}^i, :] + E_o[\mathcal{O}^i, :] \\
    \gamma_d(b_i) & = \gamma(b_i)  + E_p[\mathcal{P}^i, :] 
\end{aligned}
\end{equation}
where $\gamma_e$ and $\gamma_d$ are the encoder and decoder embeddings respectively.

\paragraph{Optimizer:} We use the PyTorch~\cite{pytorch} implementation of the AdamW~\cite{loshchilov2018decoupled_adamw} optimizer with the default parameters for our model with a constant learning rate of $10^{-4}$ and weight decay set to $10^{-3}$. We linearly warmup the learning rate for 2000 steps. In addition, we found gradient clipping\footnote{ We use \texttt{torch.nn.utils.clip\_grad\_norm\_
}} to be necessary to ensure convergence. We set the maximum gradient norm to be $30$. Empirically, we found that setting the gradient norm to be low led to slower convergence. %We show the effect of training with and without gradient clipping in Fig. TODO.

We train with a batch size of 128, and train for 1000 epochs. We perform validation every 5 epochs. We save the model with the best performance on the validation set. We use random rotation augmentation by randomly rotating each scene between 0 and 360 degrees. 

We wish to clarify that while we used the AdamW optimizer for our model, we used the vanilla Adam optimizer for ATISS, as described in \cite{Paschalidou2021NEURIPS}.

\paragraph{Parameter Probability Distributions:} We need to predict object attributes from the final transformer decoder outputs. To this end, we use use MLPs to go from the embedding dimension to the parameters of the distribution describing the attributes. For the class $\tau$, we use a linear layer from the embedding dimension to the number of classes. For the other attributes, we use  MLPs with one-input layer ($256, 512$), one hidden-layer ($512, 256$), and one output-layer ($256, 30$) and ReLU activations. The output size reflects that we use a mixture distribution with 10 components, and each component-distribution is parameterized by 3 values.

% \paragraph{Condition Encoder ($g_\phi$):} The condition encoder is a bidirectional Transformer Encoder. We use 4 encoder layers

\paragraph{Transfer Learning:} The datasets \liv, \din, \lib are much smaller compared to the \bed dataset. Thus, we use a transfer learning approach, where we first train on the \bed dataset, and use those weights as an initialization, when training on the smaller datasets. This reduces the training time significantly, as well as combats overfitting on the smaller datasets. %We compare the effect of this strategy in Fig. TODO.

We note that the datasets have a slightly different number of classes, thus any weights associated with the number of classes are not transferred, but instead sampled from a Normal Distribution, with mean $0$ and standard deviation $0.01$.

\subsection{Metrics}
The evaluation protocol follows ATISS closely, but we describe it here for the sake of completeness.

To compute the FID, we render both the ground-truth and generated layouts from a top-down view into a $256 \times 256$ images with an orthographic camera using Blender v3.1.0~\cite{blender}.  Following ATISS, the FID is computed using the code from Parmar et al.~\cite{parmar2021cleanfid}~\footnote{ \href{https://github.com/GaParmar/clean-fid}{https://github.com/GaParmar/clean-fids}, commit \texttt{fca6718}}  We will release the \texttt{.blend}-file used for rendering upon acceptance.
To compute the Classifier Accuracy Score (CAS), we use an AlexNet~\cite{alexnet}~\footnote{ \texttt{torchvision.models.alexnet} \href{https://download.pytorch.org/models/alexnet-owt-7be5be79.pth}{Weights}} model pretrained on ImageNet\cite{deng2009imagenet} to classify the orthographic renderings as real or fake.
The Synthesis Time numbers for the competing models are lifted from ATISS~\cite{Paschalidou2021NEURIPS}. We train our models on an nVIDIA A100 GPU. To ensure fair comparison to the numbers in ATISS, we ran inference on a GTX 1080 GPU which is the same GPU used in \cite{Paschalidou2021NEURIPS}. To compute the KL-divergence, we simply create a histogram of object categories in the generated layouts $g_i$ and the ground-truth $gt_i$, where $1 \leq i \leq n_{class}$ and use the the formula for the categorical KL-Divergence:
\begin{equation}
    KL(gt_i \| g_i) = \sum_i gt_i \log\left( \frac{gt_i + \epsilon}{g_i + \epsilon}\right)     
\end{equation}
where $\epsilon=10^{-6}$ is a small constant for numerical stability.

We use the same \textit{train-val-test} splits as ATISS. To compute the aforementioned metrics, we generate layouts for all floorplans in the \textit{test}-set and compare each with the corresponding ground-truth layout.

\section{3D-Front Dataset}
\begin{figure}[t!]
    \centering
    
    \begin{tikzpicture}
        \node (x)%
            {
            \includegraphics[width=0.45\linewidth, trim=50 200 20 300, clip]{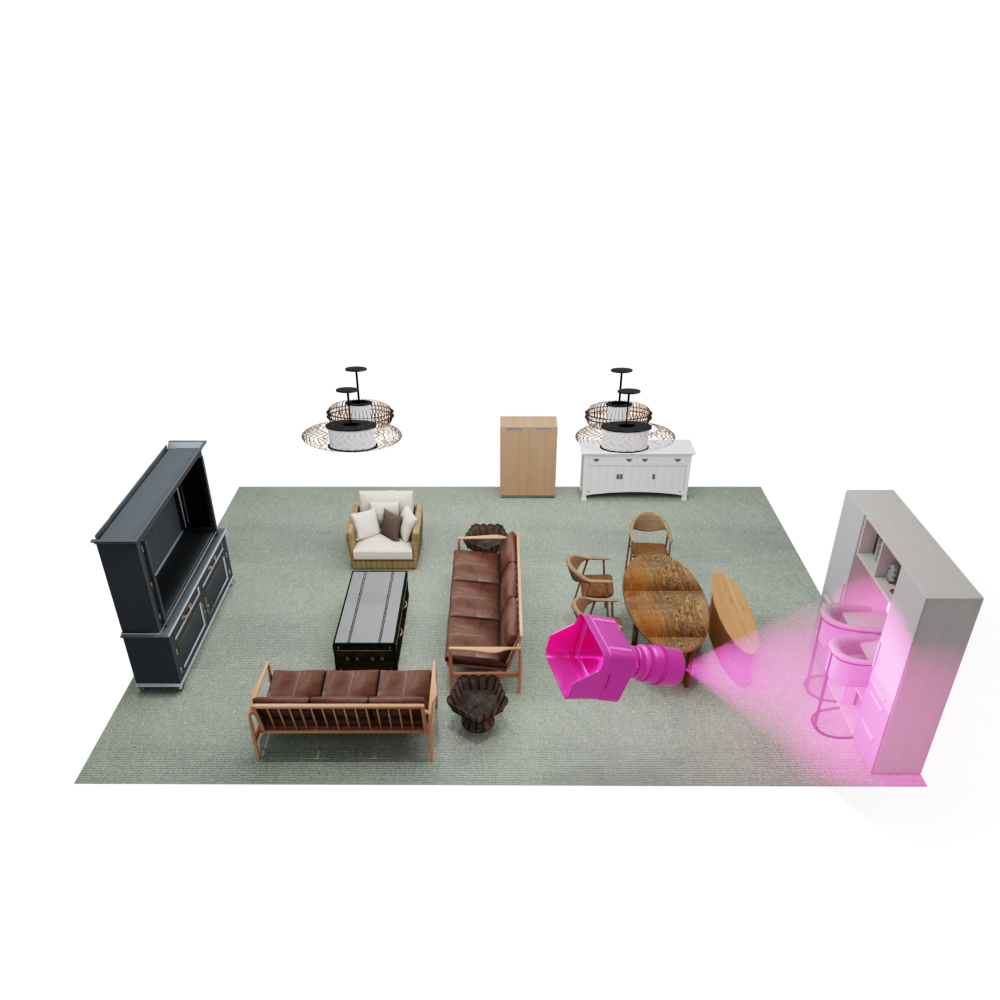}
            };
        \node[right = of x] (y){
         \includegraphics[width=0.45\linewidth, trim=100 200 100 350, clip]{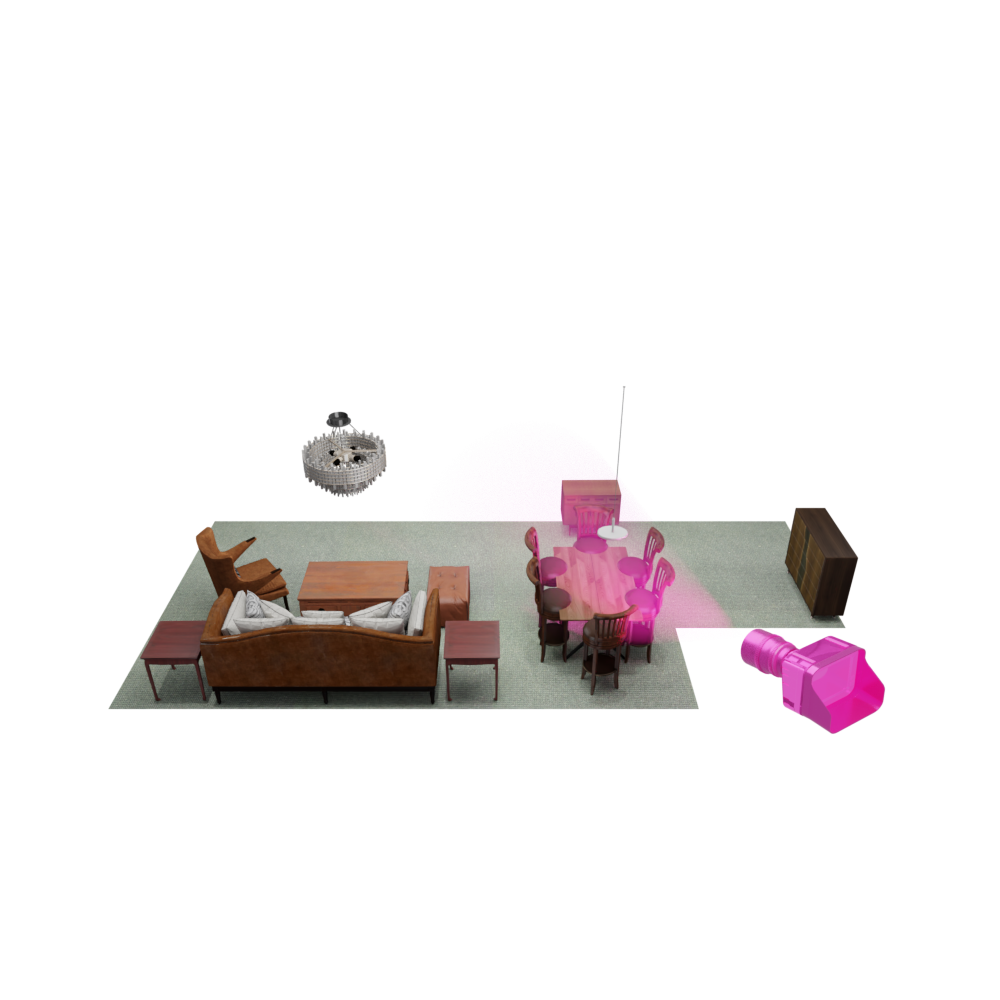}
         };
        \node[below= of x] (a)%
        {
           \includegraphics[width=0.45\linewidth, trim=50 0 20 350, clip]{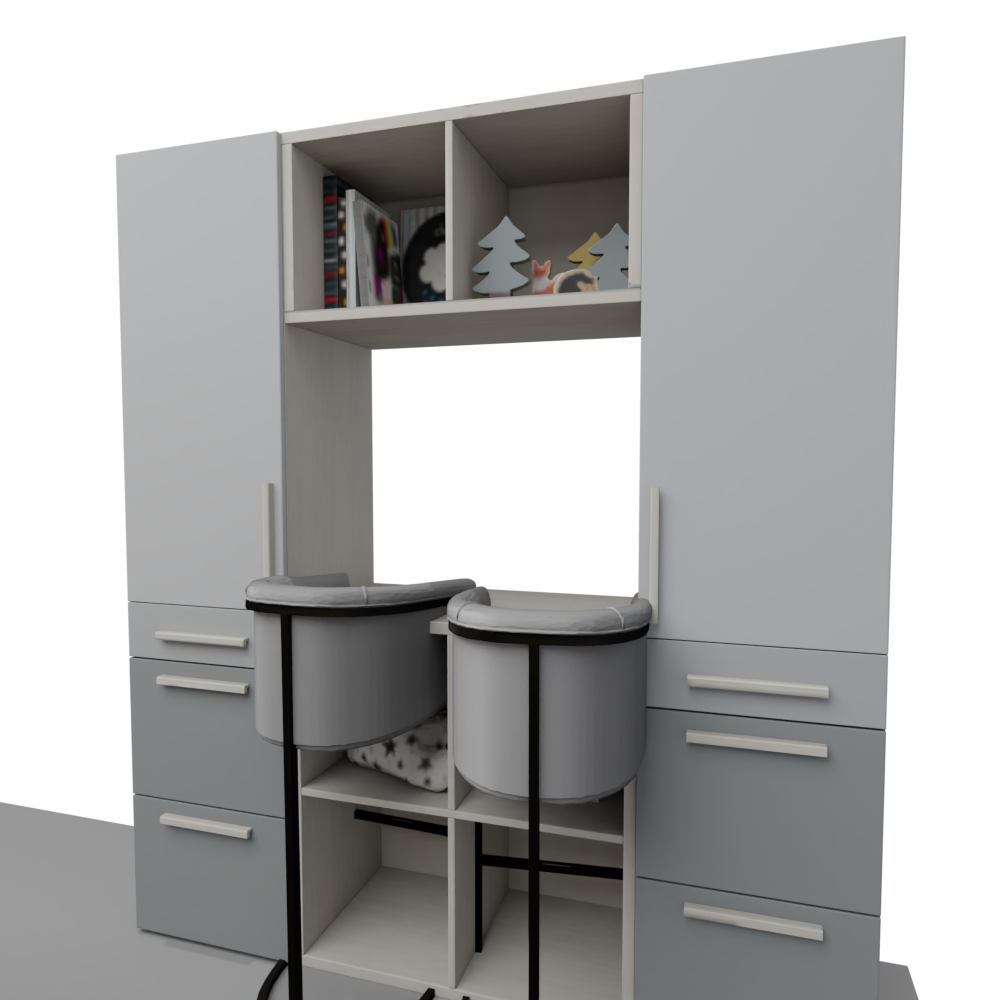}
        };
        \draw [pastel_pink, thin] (a.south west) rectangle (a.north east);
        \draw[pastel_pink, thin] (a.north west) -- ([xshift=-82pt, yshift=23pt]x.south east);
        \draw[pastel_pink, thin] (a.north east) -- ([xshift=-75pt, yshift=23pt]x.south east);
        
        \node[right= of a, below = of y] (b){
        %  100 200 100 350
        \includegraphics[width=0.45\linewidth, trim=50 0 20 350, clip]{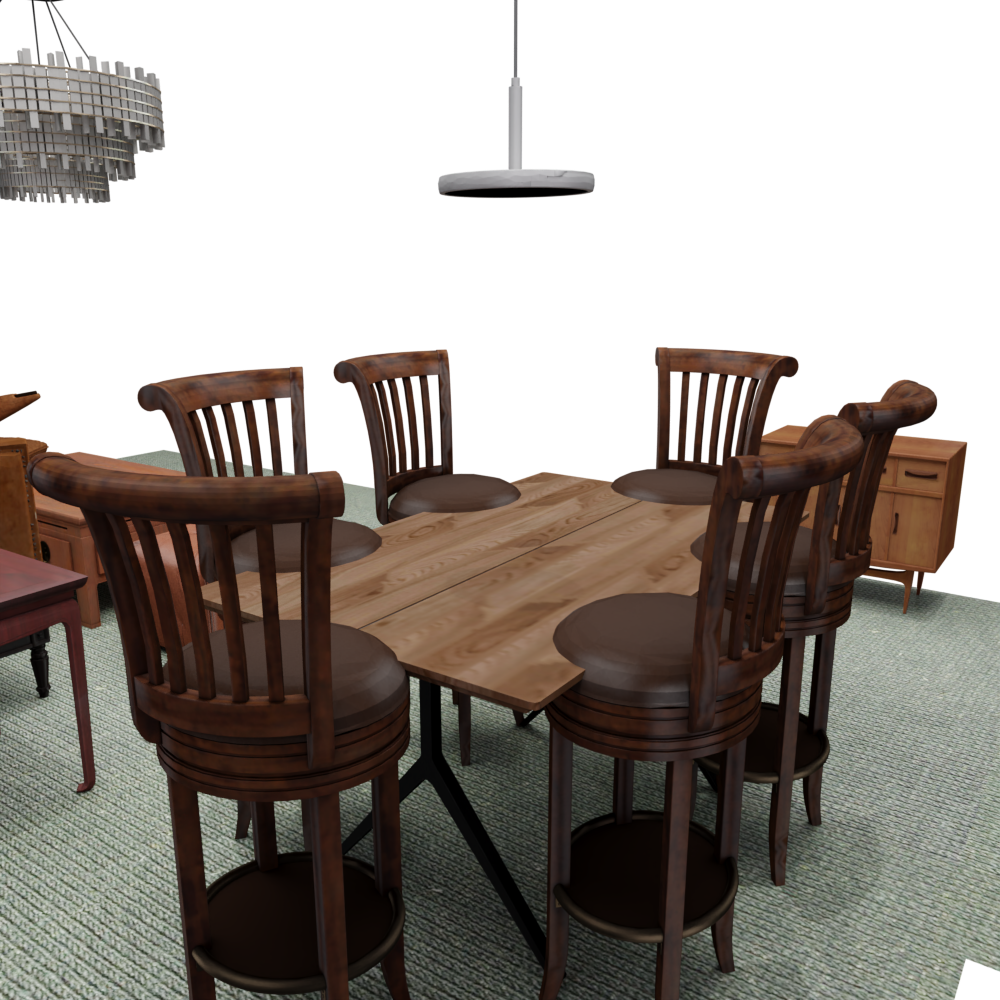}
        };
        \draw [pastel_pink, thin] (b.south west) rectangle (b.north east);
        \draw[pastel_pink, thin] (b.north west) -- ([xshift=-28pt, yshift=24pt]y.south east);
        \draw[pastel_pink, thin] (b.north east) -- ([xshift=-20pt, yshift=21pt]y.south east);
        \draw[black, thin] ([yshift=-3pt]a.south west) -- ([yshift=-3pt]b.south east);
        % \draw [black, thin] (a.south west) -- (b.south east);
    \end{tikzpicture}%

    % \vspace{-0.1}
    % \rule{\linewidth}{0.1mm}
    
    \includegraphics[width=0.22\linewidth]{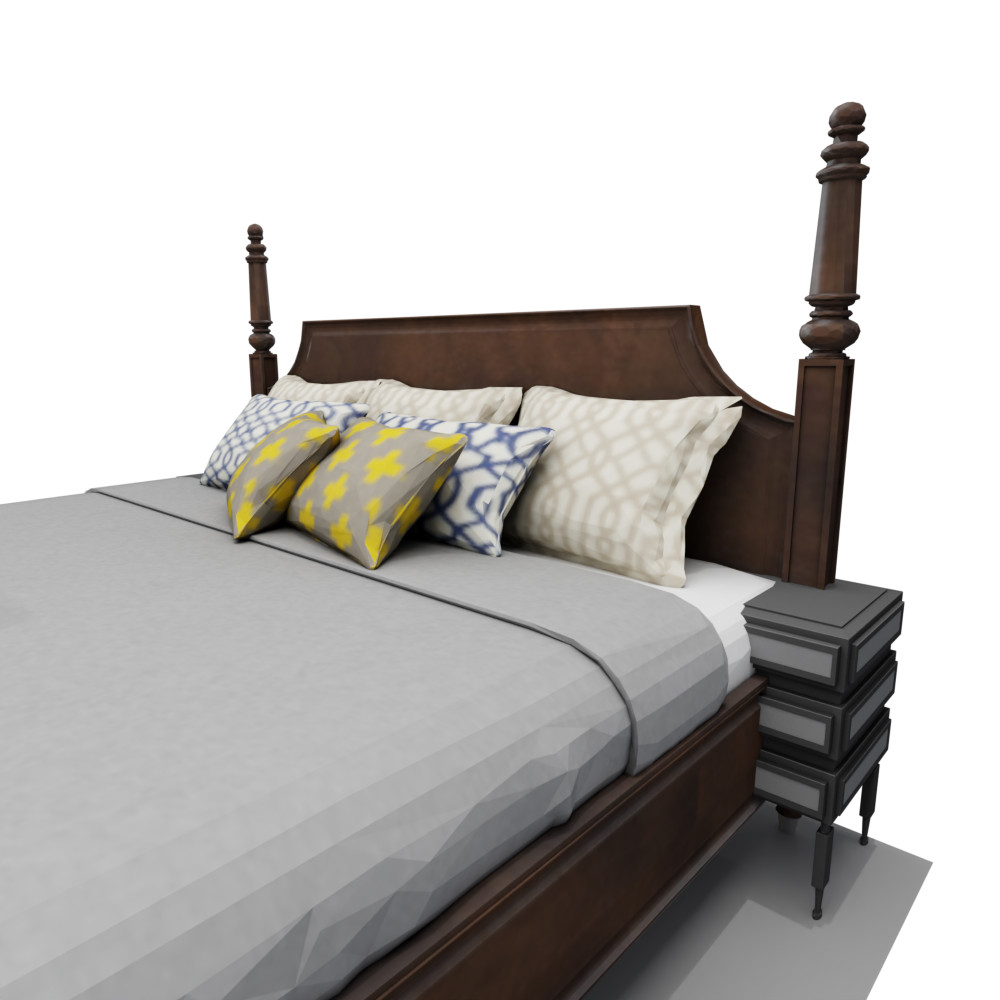}
    \includegraphics[width=0.22\linewidth]{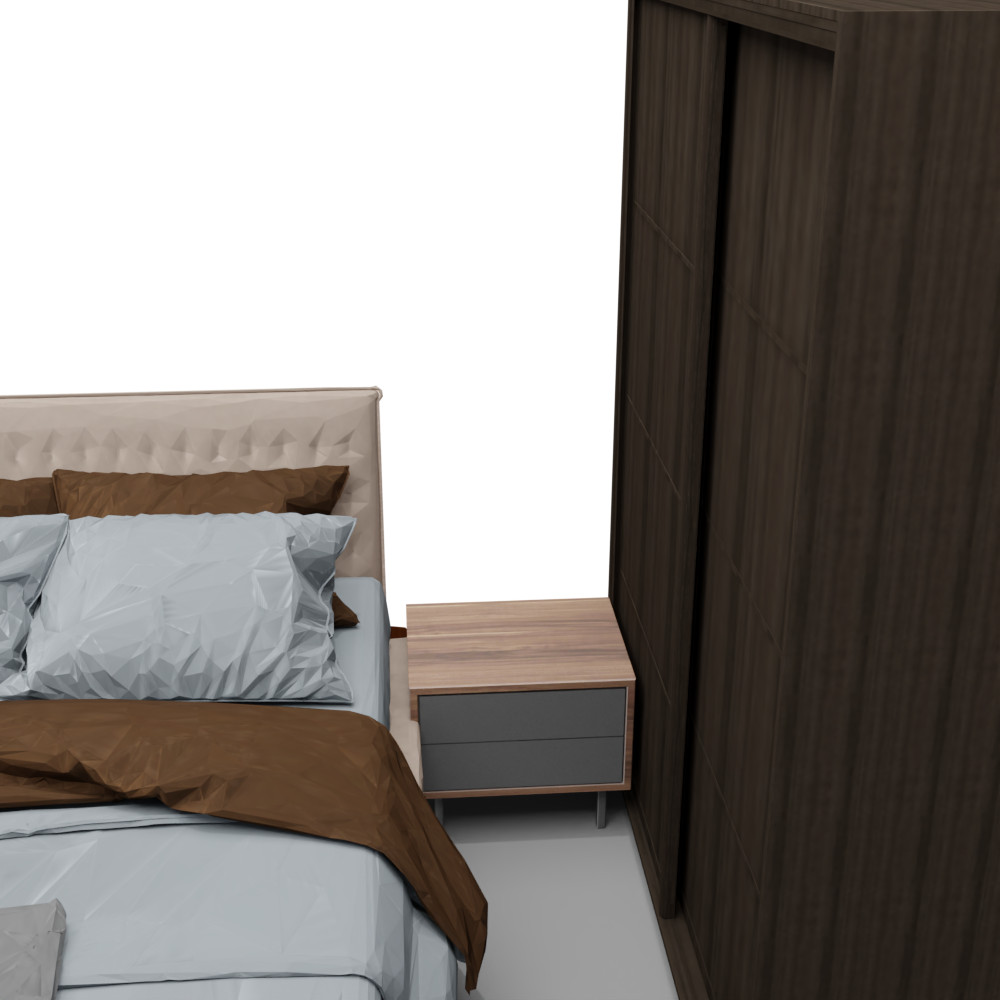}
    \includegraphics[width=0.22\linewidth]{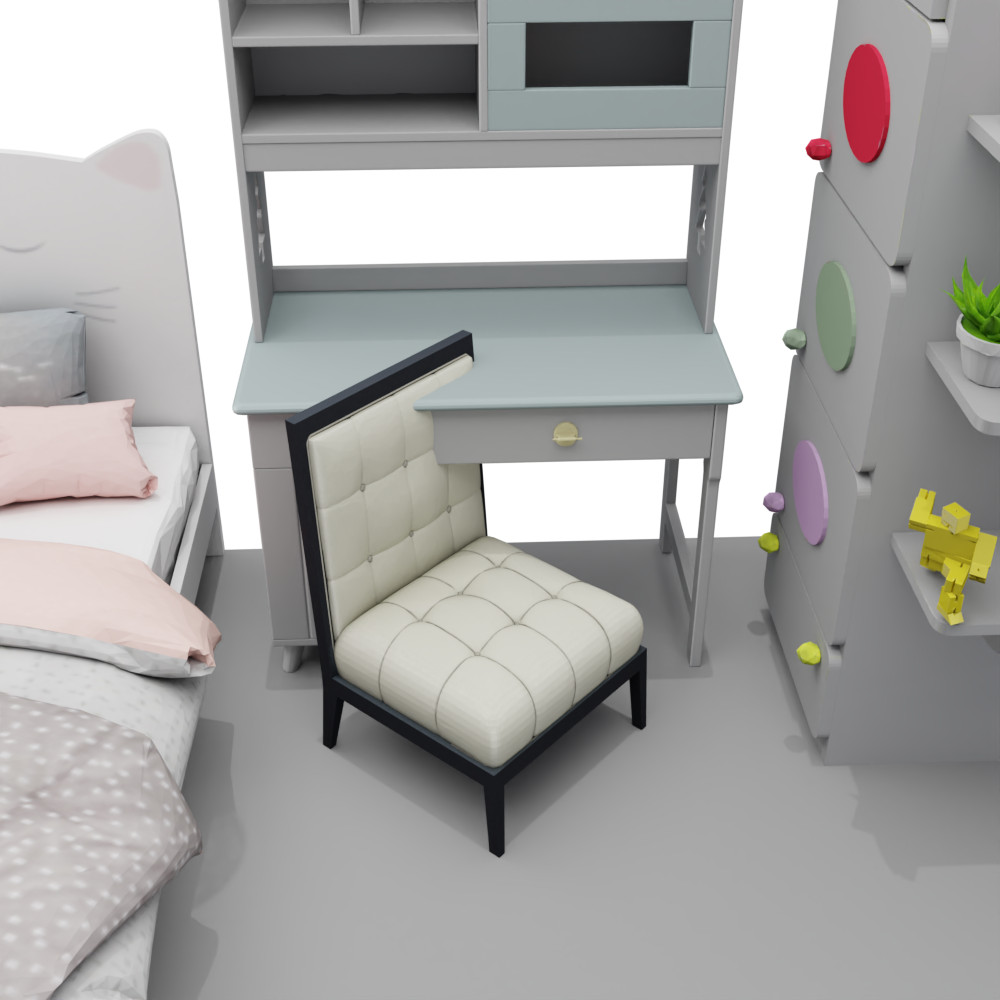}
    \includegraphics[width=0.22\linewidth]{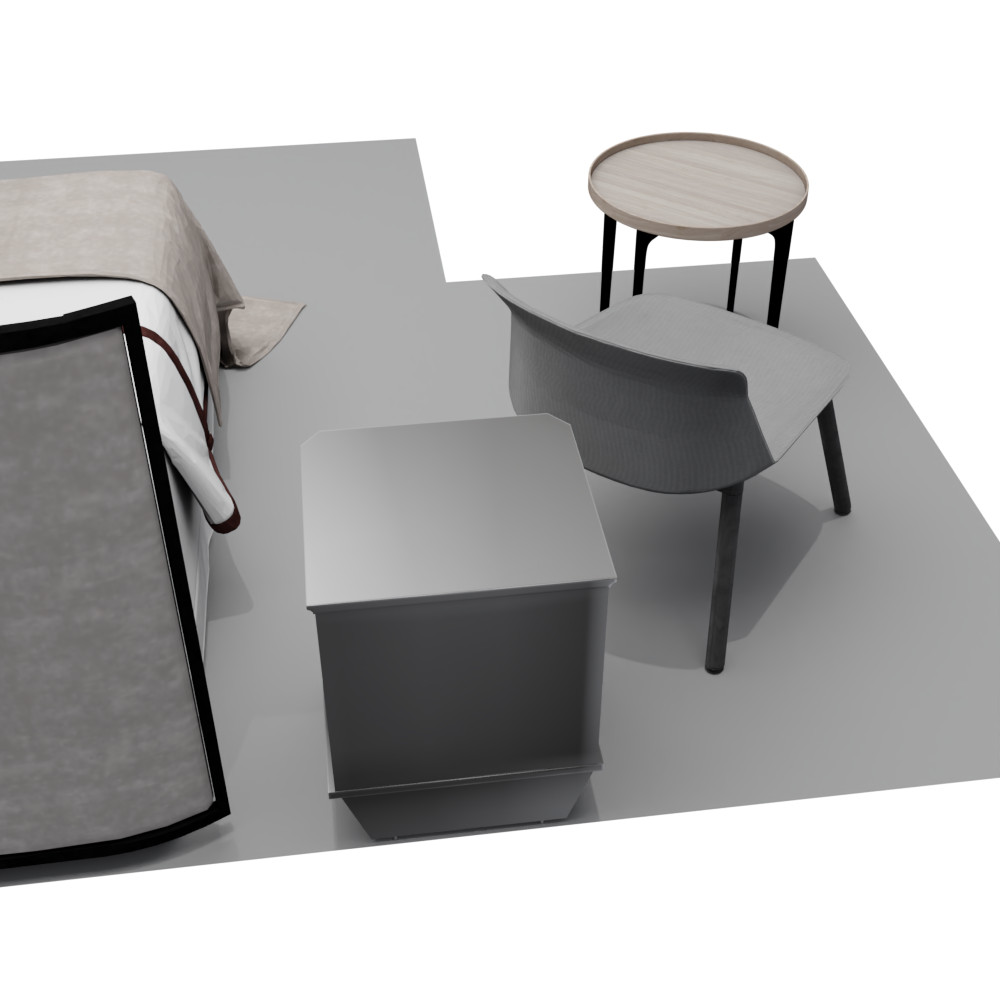}

    \caption{We show a few examples of inconsistencies in the 3D-FRONT dataset. \textbf{Top:} Camera placement in 3D-Front layouts. \textbf{Center:} The corresponding regions show errors in the ground-truth data. \textit{Left:} Chairs facing and intersecting a shelf. \textit{Right:} Chairs in the correct orientation, but intersecting with a table. \textbf{Bottom:} Some more ground-truth errors. \textit{(From Left to Right:)} Intersection. Blocking. Wrong Orientation and Intersection. Wrong Orientation.}
    
    \label{fig:threed_front}
\end{figure}

To the best of our knowledge, the 3D-Front~\cite{threed_front} dataset is the largest collection of indoor furniture layouts in the public domain. Its large scale is obtained, in part, by employing a semi-automatic pipeline, where a machine-learning system places the objects roughly, and an optimization step~\cite{scalable:weiss} refines the layouts further to conform to design standards. The only human involvement is verification that the layouts are valid - do not have object intersections, objects that block doors, etc. However, in our exploration, we find several inconsistencies still remain in the dataset. We mention a few - nightstands intersecting their nearest beds, nightstands obstructing wardrobes, chairs intersecting their closest tables, and chairs that face in the \textit{wrong} directions. We point out a few of these examples in Fig.~\ref{fig:threed_front}.

Our method, like other data-driven methods, learns the placement of objects from data. Thus, any errors in the ground-truth data itself would also show up in the sampled layouts. This is true, especially for \bed dataset, where the sampled nightstands often end up intersecting with beds.

\begin{figure}[h!]
    \centering
    \begin{subfigure}[t]{0.4\linewidth}
        \includegraphics[width=\linewidth]{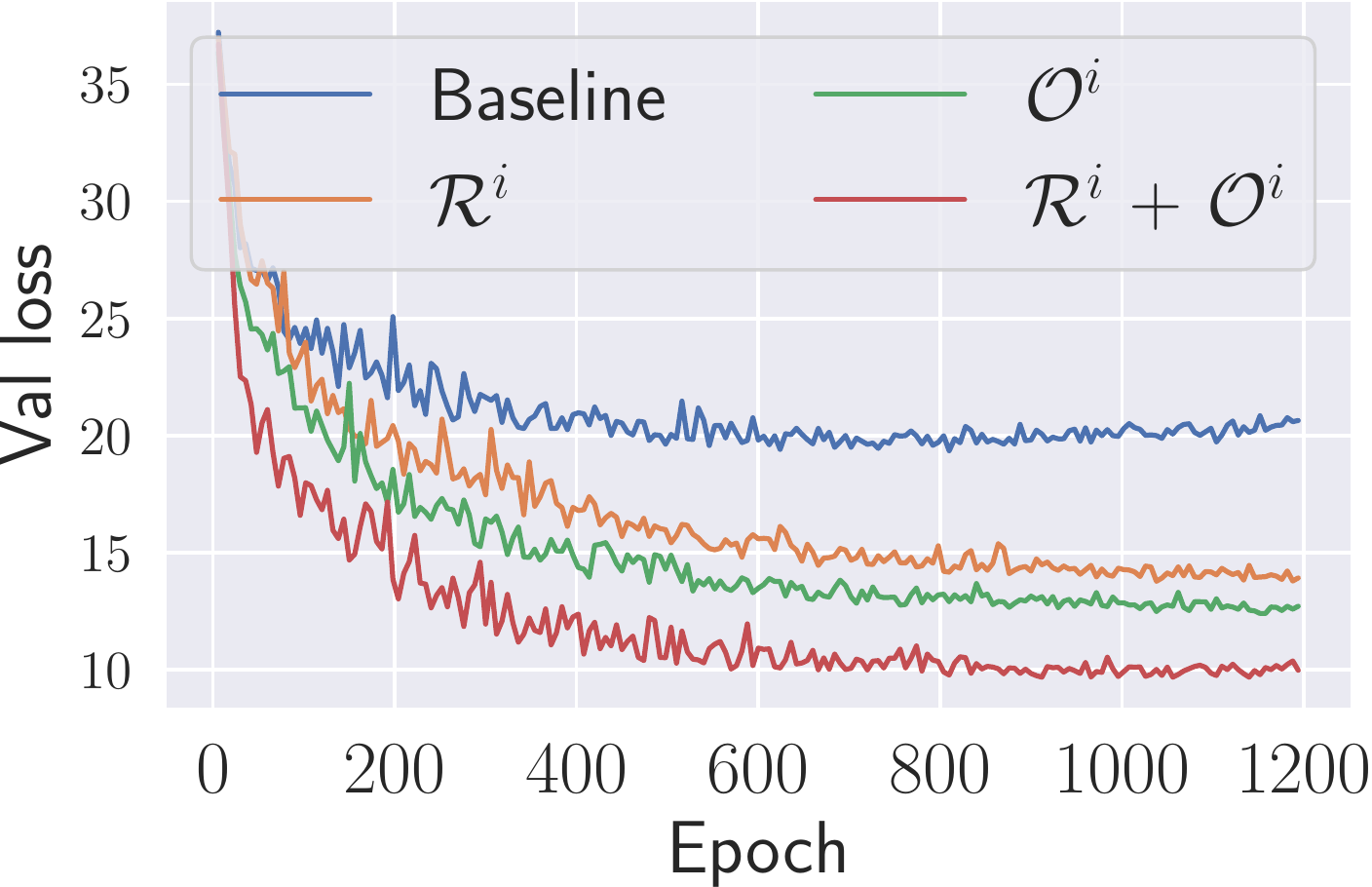}
        \subcaption[]{Adding additional tokens helps the decoder to better {align} the input and the output.}
        \label{fig:nll_pos}
    \end{subfigure}
    \hspace{0.1cm}
    \begin{subfigure}[t]{0.4\linewidth}
        \includegraphics[width=\linewidth]{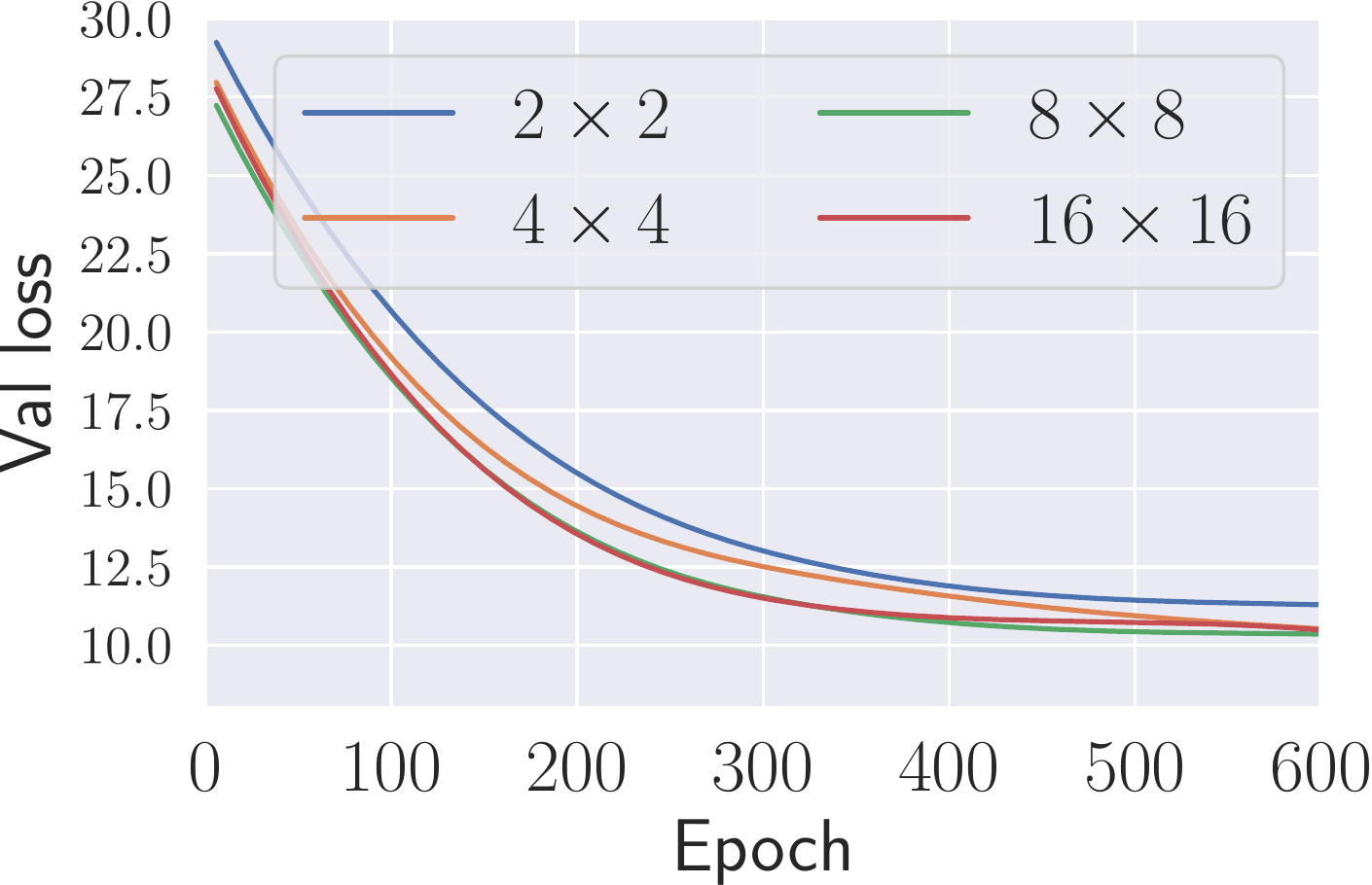}
        \subcaption[]{The smaller models perform worse, but adding more layers does not yield large correspondingly larger gains.}
        \label{fig:nll_size}
    \end{subfigure}
    % \rule{\linewidth}{0.1mm}
    
    \begin{subfigure}[b]{0.8\linewidth}
        \includegraphics[width=0.5\linewidth]{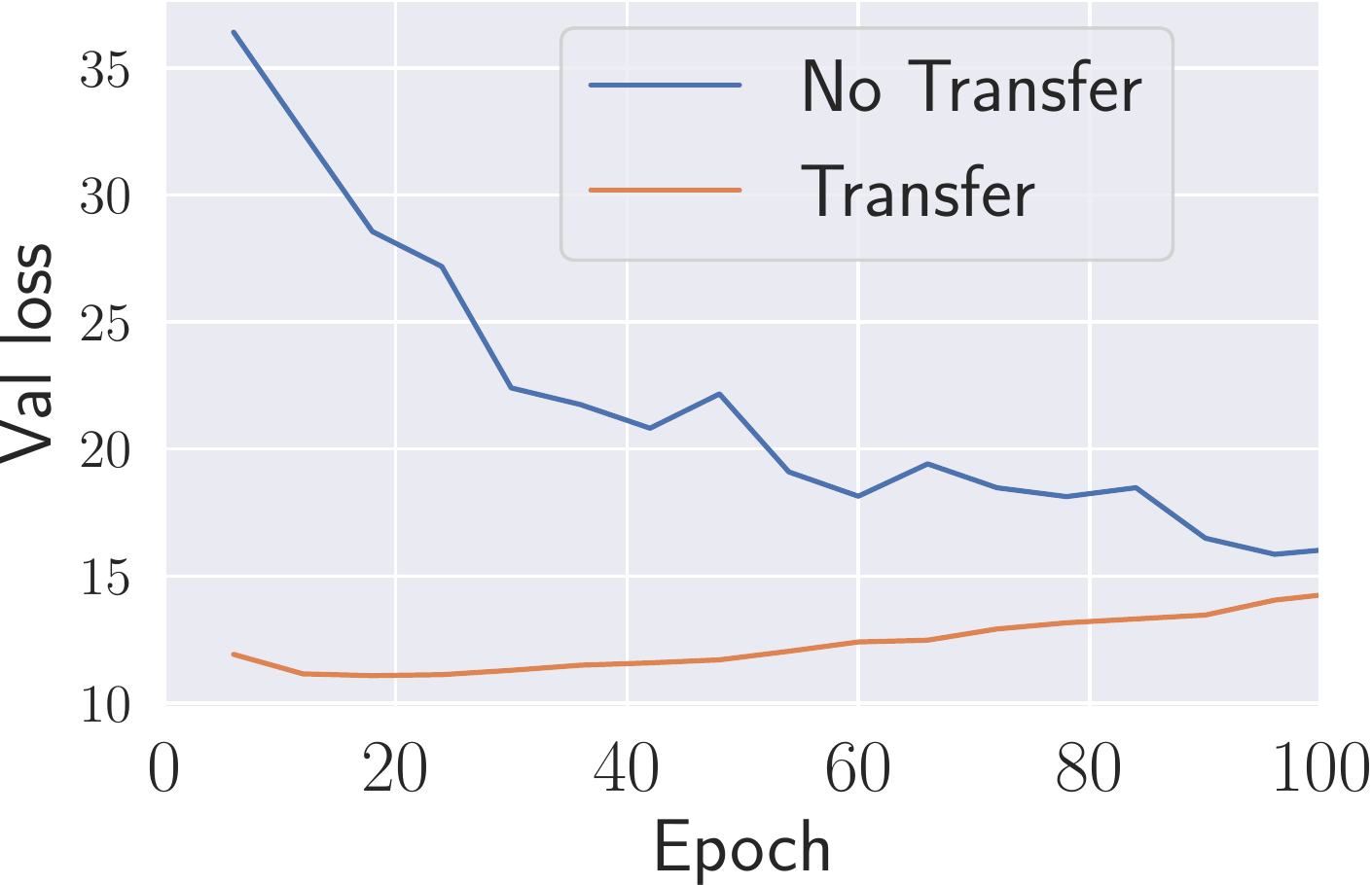}
        \hspace{0.1cm}
        \includegraphics[width=0.5\linewidth]{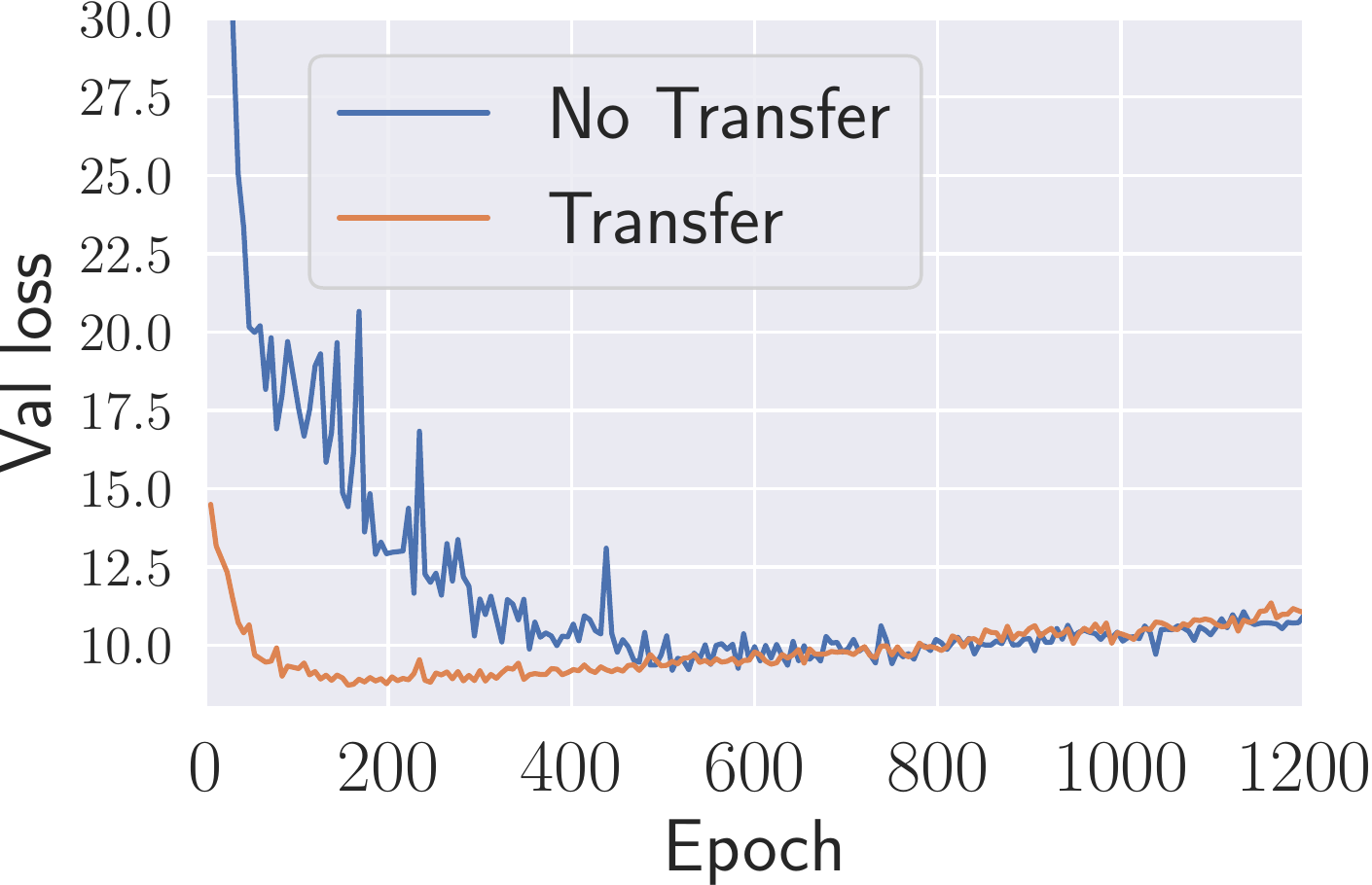}
        \subcaption[]{\textbf{Left:} Transfering weights on the \lib dataset. \textbf{Right:} Transferring weights on the \liv dataset.}
        \label{fig:nll_transfer}
    \end{subfigure}
    
    \begin{subfigure}[b]{0.8\linewidth}
        \includegraphics[width=0.5\linewidth]{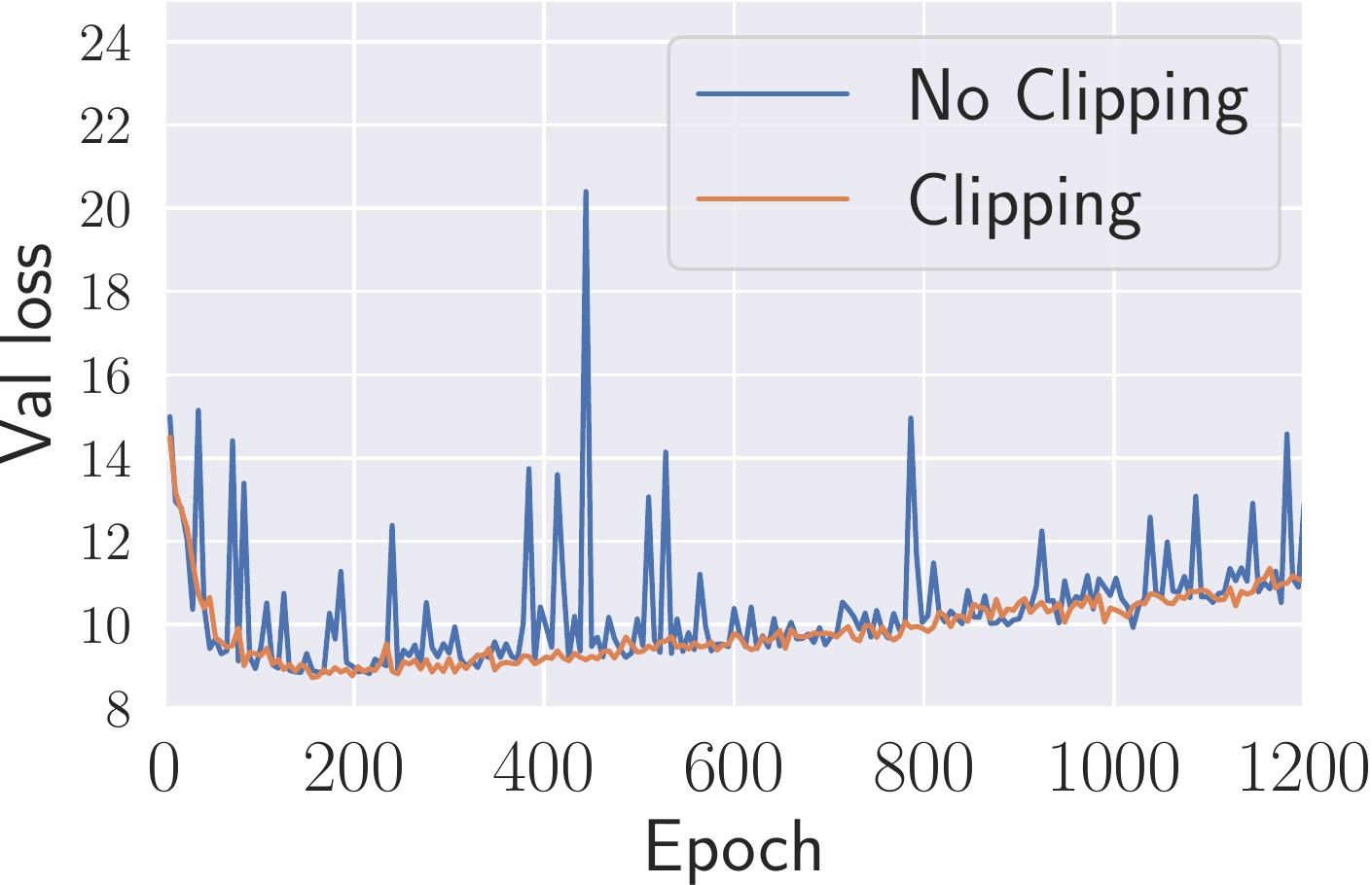}
        \hspace{0.1cm}
        \includegraphics[width=0.5\linewidth]{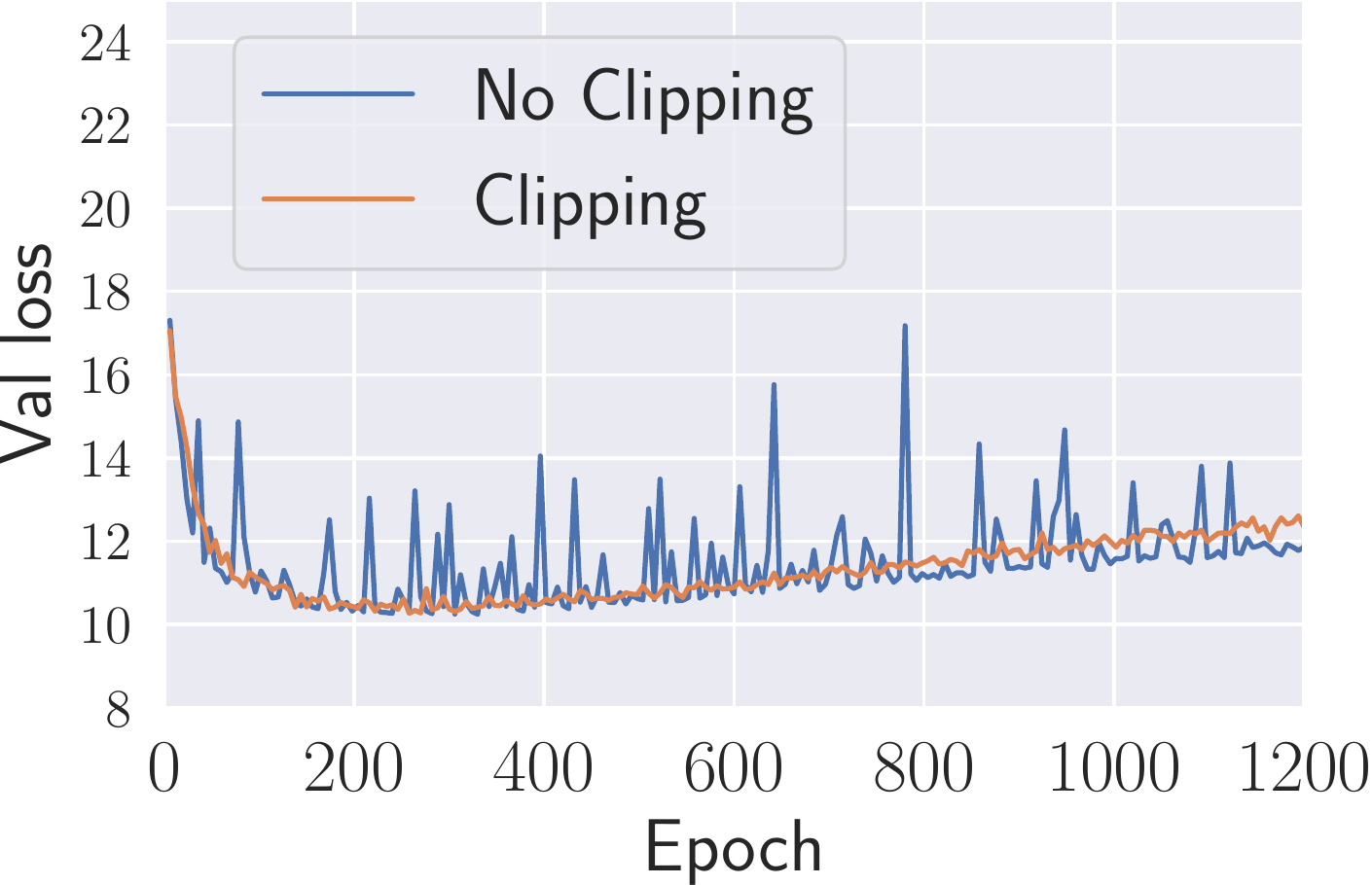}

        \subcaption[]{\textbf{Left:} Gradient clipping applied on the \din dataset. \textbf{Right:} Gradient clipping applied on the \liv dataset.}
        \label{fig:nll_clip}
    \end{subfigure}
    
    \begin{subfigure}[b]{0.8\linewidth}
        \centering
        \includegraphics[width=0.8\linewidth]{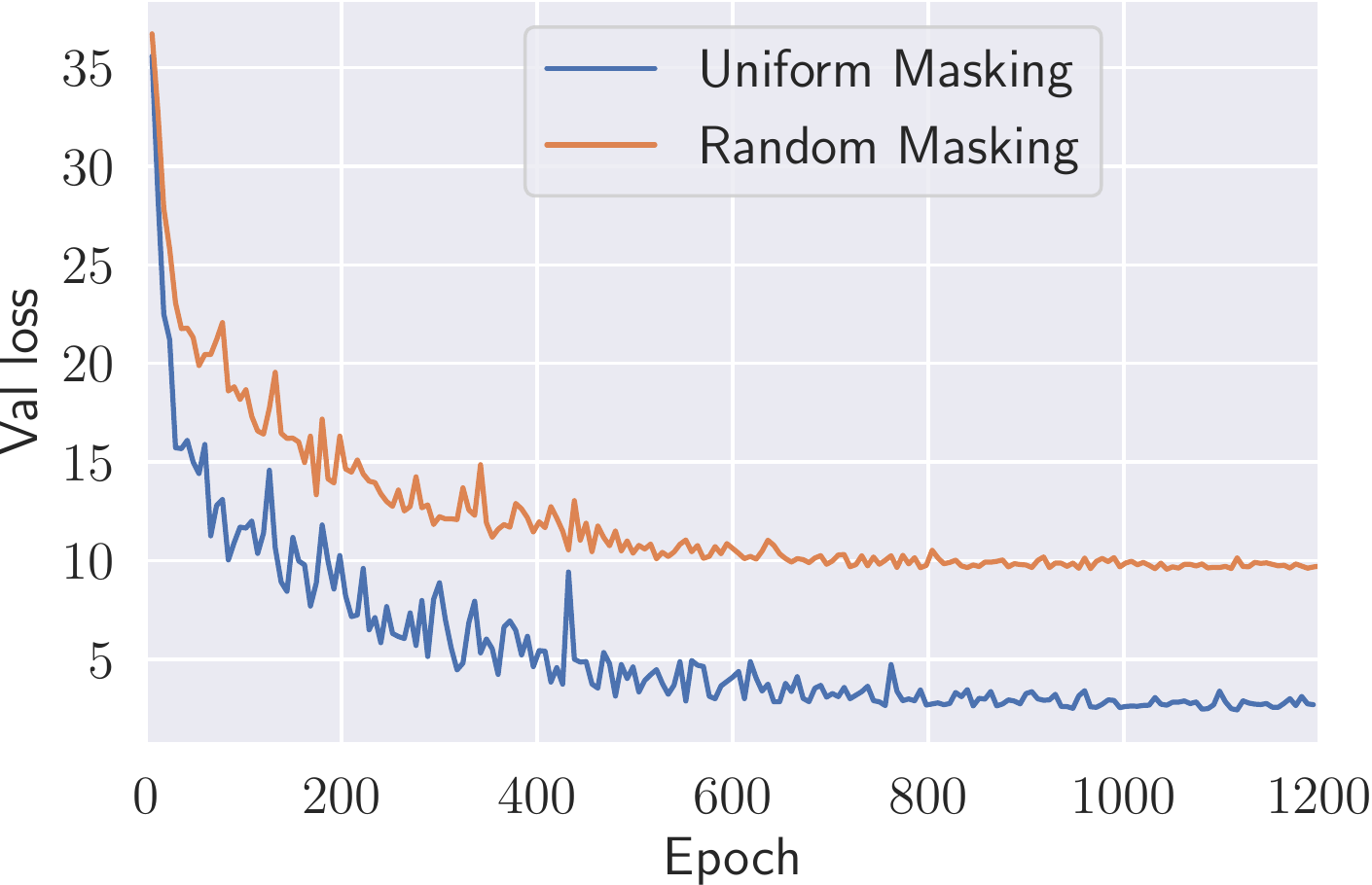}
        \subcaption[]{Using a uniform masking ratio of $0.15$ shows good performance in terms of NLL, but is unable to sample owing to large distribution shift between training and inference.}
        \label{fig:nll_mask}
    \end{subfigure}
    
    \caption{\textbf{Ablation Studies} We show the validation losses for the different architectural choices we make. }
    \label{fig:ablation}
\end{figure}

\section{Ablations}

In this section, we justify our design choices by conducting an ablation study. We train our model under different settings on the \bed dataset, unless specified otherwise, and use the validation loss, the Negative Log-Likelihood (NLL) as the metric to judge performance. This is because we empirically found the validation loss to correlate directly with sample quality. In particular, we ablate the choice of our position encodings, the number of layers and training with gradient clipping. We also include a discussion of the masking strategy and transfer learning.

\subsection{Position Encodings}
We consider the input conditioning to be a set. In contrast, the output is a sequence. Thus, the model needs additional information to \textit{align} the input and the output . We use \textit{object index tokens} $\mathcal{O}^i$ and the \textit{relative position tokens} $\mathcal{R}^i$ to provide this additional information.

During training, the objects themselves are permuted. The intuition is that $\mathcal{O}^i$ injects information about how \textit{early} or \textit{late} each object must appear in the output sequence. However, this information alone is not enough to disambiguate where each of the attributes of the object must appear. Hence, we also add $\mathcal{R}^i$ to the object attribute embeddings. Together, these embeddings localize the position of the attribute in the output sequence, given the current permutation.

In Fig.~\ref{fig:nll_pos}, we progressively add our embeddings to the \textsf{Baseline} model which is the model without any positional encodings on the encoder. It is clear that using our embeddings helps the model better \textit{align} the set-input and the sequence-output. While each of $\mathcal{O}^i$ and $\mathcal{R}^i$ roughly align the input and the output, it is only when using both the embeddings that the model can precisely locate the actual position of tokens in the output sequence.
\subsection{Number of Layers}
For all the experiments in the main paper, we used 4 transformer layers in both the encoder and the decoder. In Fig.~\ref{fig:nll_size}, we show how the model performance scales with scaling the number of layers. We see that the performance correlates strongly with the number of layers. However, the performance gains become marginal when going from 4 to 8 layers or 8 to 16 layers. 
These larger models take longer to train and sample from. We believe our 4 layer models provide a good compromise between performance and speed. 

Note that the values in Fig.~\ref{fig:nll_size} are smoothed by an interpolating spline to highlight the general trend.

\subsection{Gradient Clipping}
We found that the validation loss oscillated considerably during training. Upon further investigation, we noticed that the gradients norms tended to be unusually large, especially for the last layers in the parameter generating MLPs. Thus, we train the final networks with gradient clipping. Surprisingly, we found that even without gradient clipping, if we retain the model with the best NLL on the validation set, the performance is the same. However, with gradient clipping, we found the training curves to be much smoother (Fig.~\ref{fig:nll_clip}). Consequently, we were able to perform validation at less frequent intervals to select the best performing model, which sped up training.

\subsection{Masking Strategy}
MaskGIT~\cite{chang2022maskgit} find that using a robust masking strategy is important, as the usual $15\%$ masking leads to a distribution shift between training and sampling. We see in Fig.~\ref{fig:nll_mask} that masking with a uniform ratio of $15\%$ leads to better NLL as the network is more confident in it's predictions. But we found out that the we could not sample from such a trained network, as it would output a stop token after only generating a few objects, which intuitively makes sense, as the network would only see a few mask tokens during training.

\subsection{Transfer Learning:} We plot the validation loss in Fig.~\ref{fig:nll_transfer} on the \lib and \liv datasets under two configurations - \textsf{No Transfer}, where the models are trained from scratch and \textsf{Transfer}, where the model is first trained on the \bed dataset and these weights are used as initialization for training on the target dataset. 
We make a few observations: 1. The models begin to overfit fairly early. For the \bed dataset, the loss contiues to fall until epoch 1200, but in the \textsf{No Transfer} configuration for the \lib dataset, we see overfitting at epoch 150 and for the \liv dataset, at epoch 600. We hypothesize that this is due to the small size of these datasets compared to the \bed dataset. 2. The \textsf{No Transfer} configuration has a higher (worse) NLL as compared to the \textsf{Transfer} configuration, even when trained for longer.

These observations led us to use the \textsf{Transfer} configuration for the \lib, \liv and \din datasets.

\section{Sampling Details}
We highlight the difference between our sampling algorithm and the standard conditional sampling algorithm in this section. These differences are highlighted in {\color{blue} blue} in Alg. \ref{alg:decoding}. The primary difference is that in our sampling algorithm, a forward pass is made through the decoder every time a new token is sampled. This token then replaces the corresponding \mask token in both $C$ and $S$.

In addition, our algorithm runs for a fixed number of iterations (until all \mask tokens are replaced) compared to the standard algorithm which terminates when an \texttt{EOS} token is generated. This is both an advantage and a drawback - it is an advantage in the sense that a user can implicitly specify the number of objects by specifying the number of \mask tokens. It is a drawback in that the number of objects must be known before sampling can proceed.

\subsection{A Sampling Trick}
For our outlier detection examples, we use a simple trick - if there is only a single object to be sampled, we can create a permutation so that the \mask tokens of the object to be sampled are toward the end of the sequence in $C$ and $S$. With this permutation we only have to make forward passes beginning from the first masked token. All the tokens before the first masked token can simply be copied. This leads to faster sampling.

\hfill
{
\centering
\begin{minipage}[h]{0.48\linewidth}
    \small
    \begin{algorithm}[H]
      \caption{Standard Cond. Sampling}\label{alg:decoding}
      \begin{algorithmic}[1]
        \Require{$C = (c_i)_{i=1}^k$, $S = (\texttt{SOS}), s = \phi$}
        \State {${C^g = g_\phi([\cI, C])}$} \Comment{Only performed once}
        \While{$s \neq \texttt{EOS}$}% \Comment{Loop over semantic group}
            \State {$s = \textsf{SAMPLE}(f_\theta(S_{<i}, C^g))$}
            \State {$S\textsf{.append($s$)}$} \Comment{$C$ not updated}
        \EndWhile
        \State \textbf{return} $S$
      \end{algorithmic}
    \end{algorithm}
\end{minipage}
\begin{minipage}[h]{0.48\linewidth}
    \small
    \begin{algorithm}[H]
      \caption{Our Sampling}\label{alg:decoding}
      \begin{algorithmic}[1]
        \Require{$\cI$, $C = (\mask)_{i=1}^k$, $S = (\texttt{SOS})$}
        \For{$i \gets 1$ to $k$}% \Comment{Loop over semantic group}
            \State {${\color{blue} C^g = g_\phi([\cI, C])}$}
            \State {$s = \textsf{SAMPLE}(f_\theta(S_{<i}, C^g))$}
            \State {${\color{blue} C[i] = s}, \; S\textsf{.append($s$)}$}
        \EndFor
        \State \textbf{return} $S$
      \end{algorithmic}
    \end{algorithm}
\end{minipage}
}

In all our experiments, we set the number of objects to be sampled to be the same as the number of objects in the ground-truth layout associated with the particular floorplan boundary.

\section{Arbitrary Conditioning}
We first recap the sampling strategy of ATISS.
\begin{align}
    c_{\theta}: \mathbb{R}^{64} &\rightarrow \mathbb{R}^C &\hat{\bq} \mapsto \hat{\bc} \label{eq:att_ectractor_1} \\
    t_{\theta}: \mathbb{R}^{64}\times \mathbb{R}^{L_c} &\rightarrow \mathbb{R}^{3\times 3\times K}
        & (\hat{\bq}, \lambda(\bc)) \mapsto \hat{\bt} \\
    r_{\theta}: \mathbb{R}^{64}\times \mathbb{R}^{L_c} \times\mathbb{R}^{L_t} &\rightarrow \mathbb{R}^{1\times 3\times K}
    & (\hat{\bq}, \lambda(\bc), \gamma(\bt)) \mapsto \hat{\br} \\
    s_{\theta}: \mathbb{R}^{64}\times \mathbb{R}^{L_c} \times\mathbb{R}^{L_t} \times \mathbb{R}^{L_r} &\rightarrow \mathbb{R}^{3\times 3\times K}
    & (\hat{\bq}, \lambda(\bc), \gamma(\bt), \gamma(\br)) \mapsto \hat{\bs} \label{eq:att_ectractor_2}
\end{align}
These equations say the following: \textit{From a query vector} $\hat{\bq}$, \textit{the model predicts a class. From the query and class, the model predicts the translation. From the query, class, and translation, the model predicts a rotation, and so on.}
This means that in ATISS, \textit{future} attributes cannot affect the distribution of previous attributes. When conditioning, we can specify the class and then sample a translation, but we cannot specify a translation and let the model infer the most likely class for that given translation.

In contrast, COFS has bidirectional attention on the encoder side, enabling us to specify \textit{any} subset of object attributes. This is done by replacing the \mask token corresponding to the object attribute by its actual value in $C$. The copy-paste objective ensures that the same attribute will be sampled at the desired location by the decoder. The mask-predict objective trains the model to get the most-likely attributes for the unspecified tokens.

\begin{figure}[h!]
    \centering
    \begin{subfigure}[t]{0.28\linewidth}
        \includegraphics[width=\linewidth]{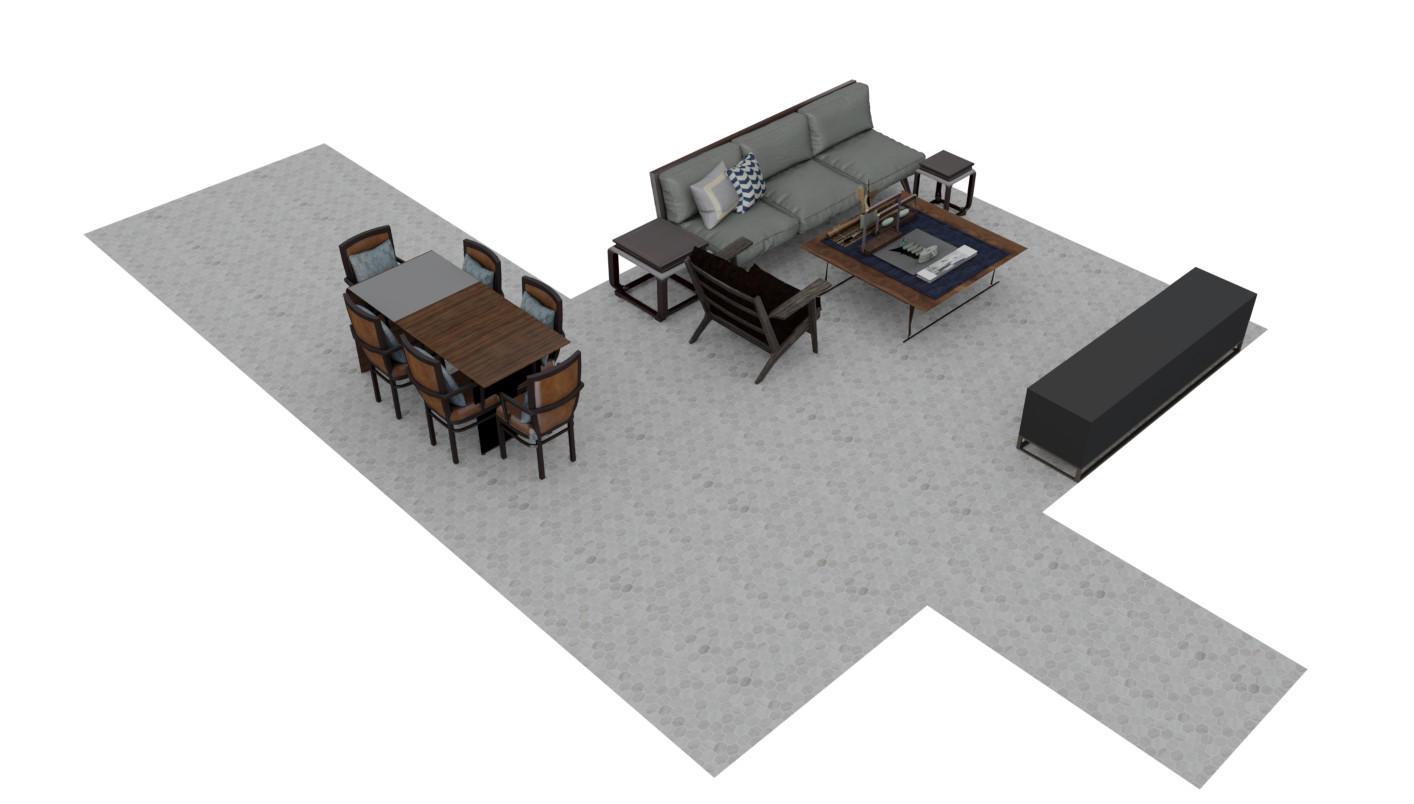}
        \subcaption[]{ Perspective view of the partial floorplan}
        \label{fig:arbit_persp}
    \end{subfigure}
    \hspace{0.1cm}
    \begin{subfigure}[t]{0.28\linewidth}
        \includegraphics[width=0.8\linewidth]{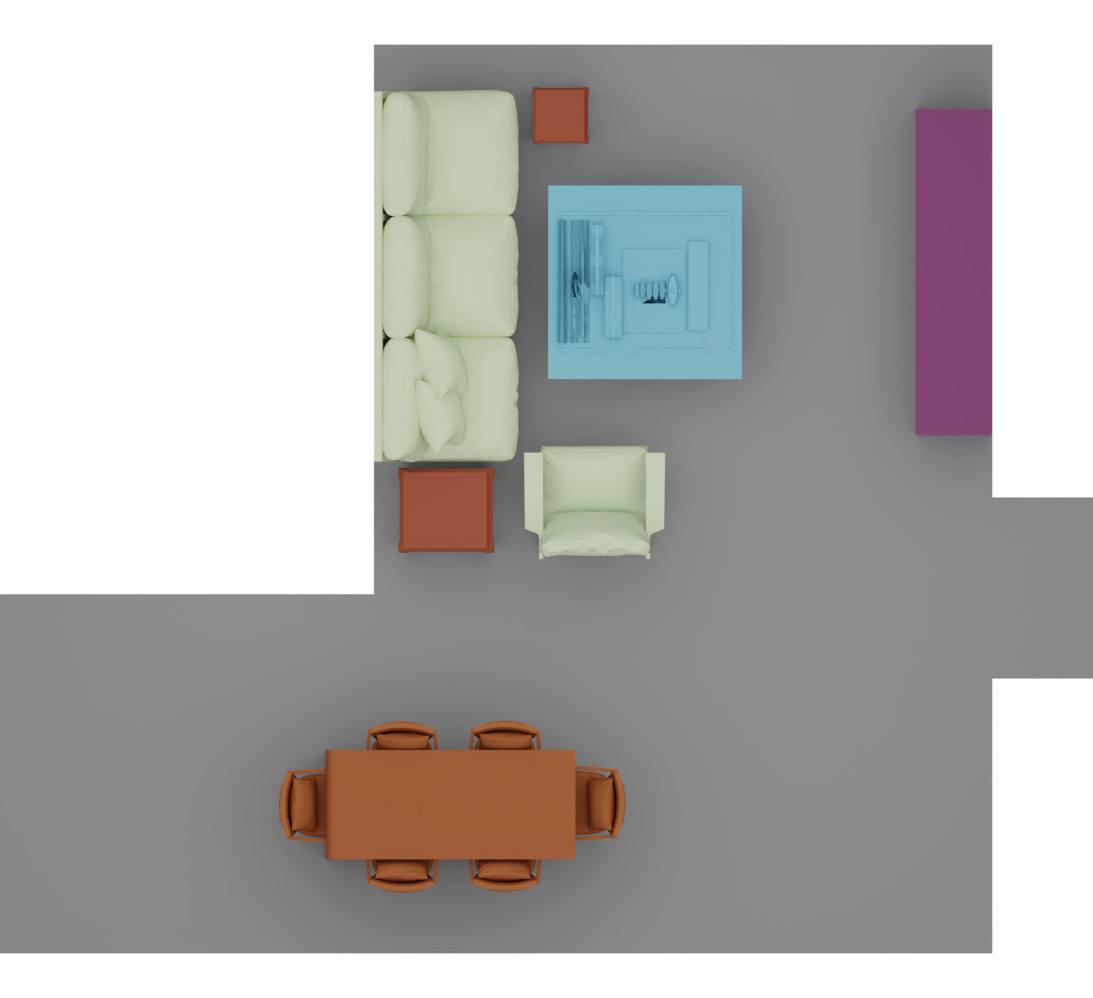}
        \subcaption[]{Simplified Orthographic view}
        \label{fig:arbit_ortho}
    \end{subfigure}
    \begin{subfigure}[t]{0.28\linewidth}
        \includegraphics[width=0.8\linewidth]{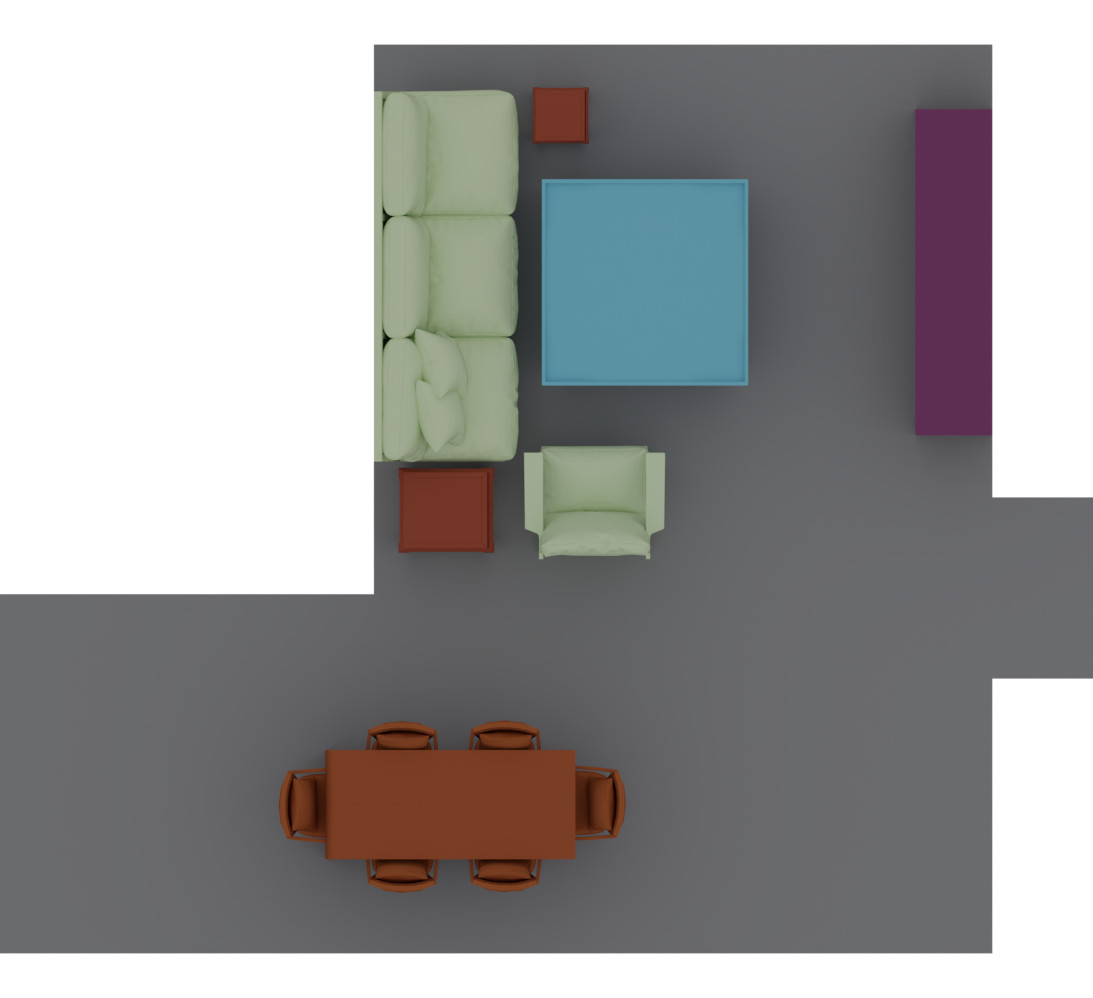}
        \subcaption[]{Unconditionally sampled table (cyan).}
        \label{fig:arbit_ortho_unc}
    \end{subfigure}
    \begin{subfigure}[t]{\linewidth}
        \includegraphics[width=\linewidth]{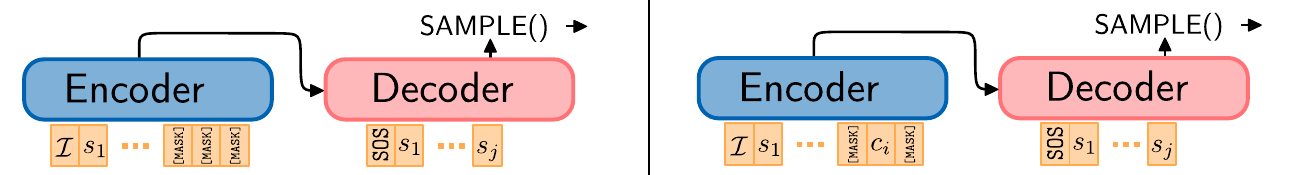}
        \subcaption[]{\textbf{Left:} Unconditional Sampling \textbf{Right:} Conditioning by specifying attributes in $C$.}
        \label{fig:arbit_c}
    \end{subfigure}
    \begin{subfigure}[t]{0.28\linewidth}
        \includegraphics[width=0.8\linewidth]{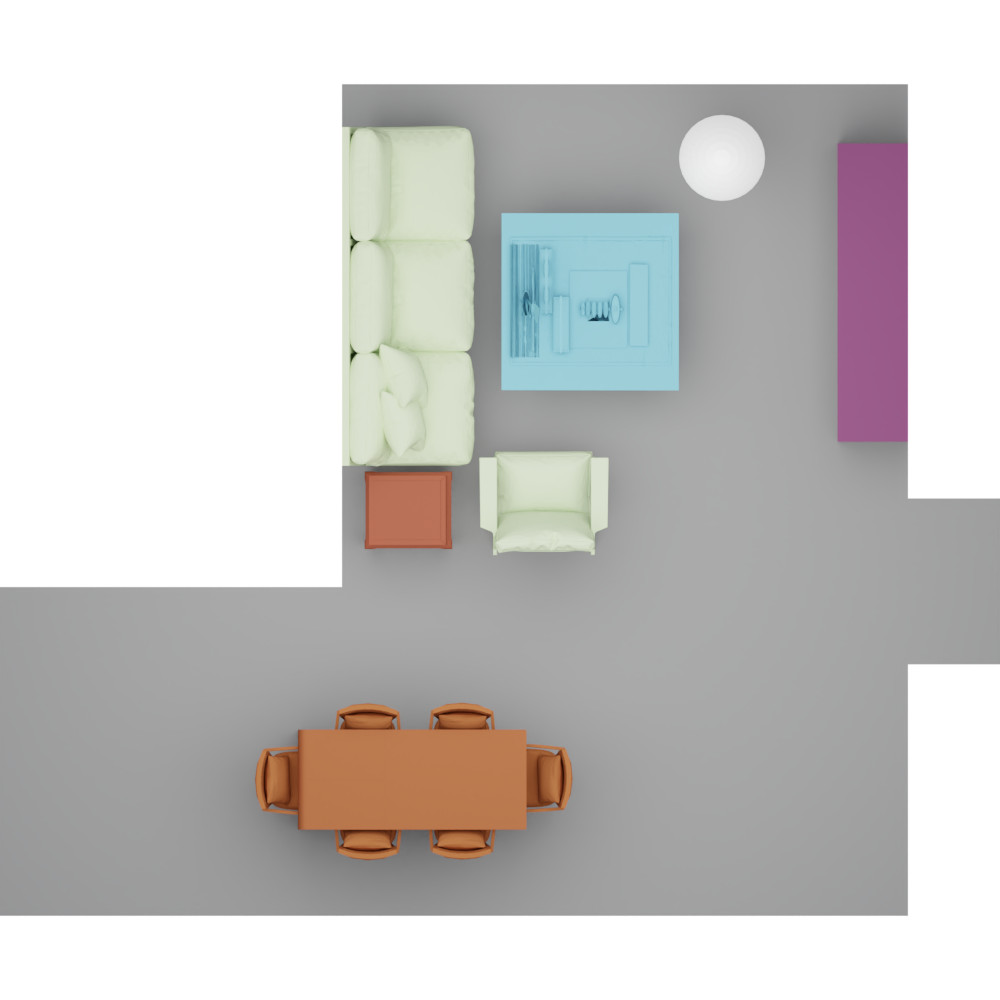}
        \subcaption[]{ We constrain the location of the object to be sampled. Shown by the white circle.}
        \label{fig:arbit_loc}
    \end{subfigure}
    \hspace{0.1cm}
    \begin{subfigure}[t]{0.28\linewidth}
        \includegraphics[width=0.8\linewidth]{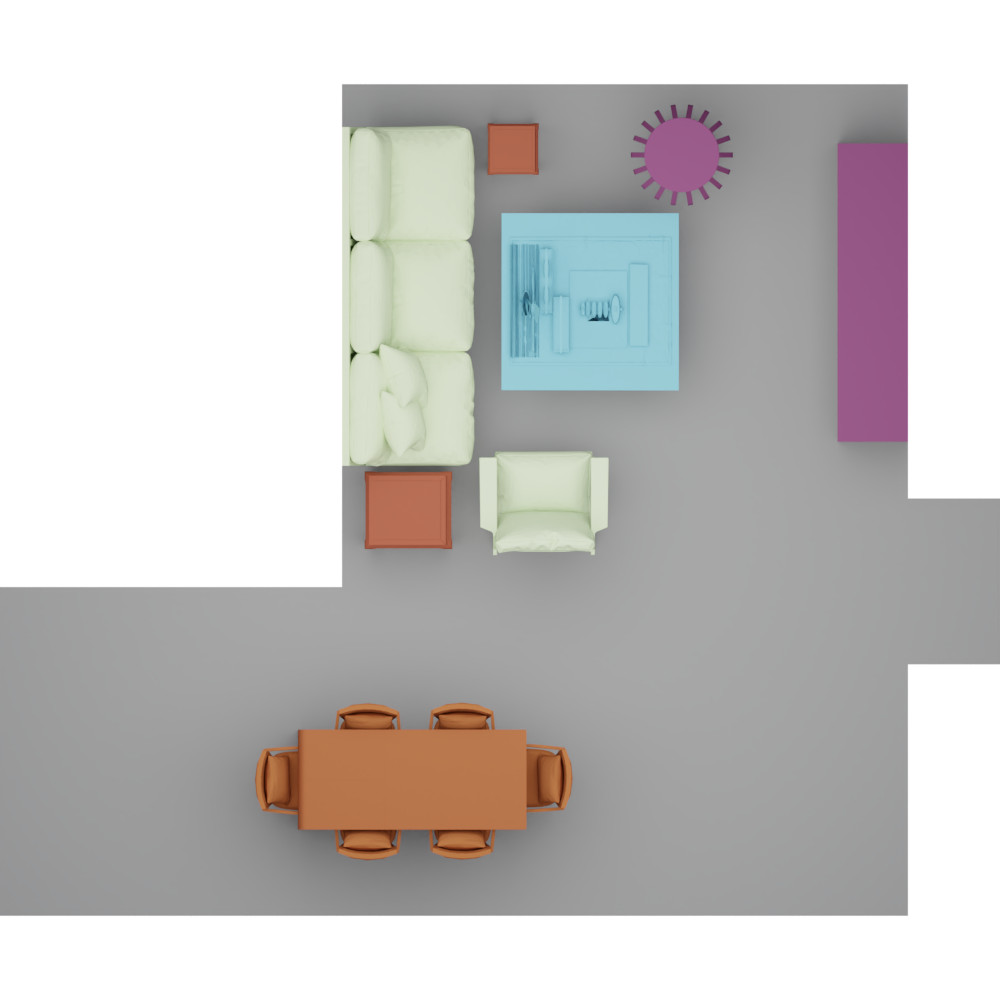}
        \subcaption[]{Conditionally generated sample. (Purple)}
        \label{fig:arbit_loc_sample}
    \end{subfigure}
        \hspace{0.1cm}
    \begin{subfigure}[t]{0.28\linewidth}
        \includegraphics[width=0.8\linewidth]{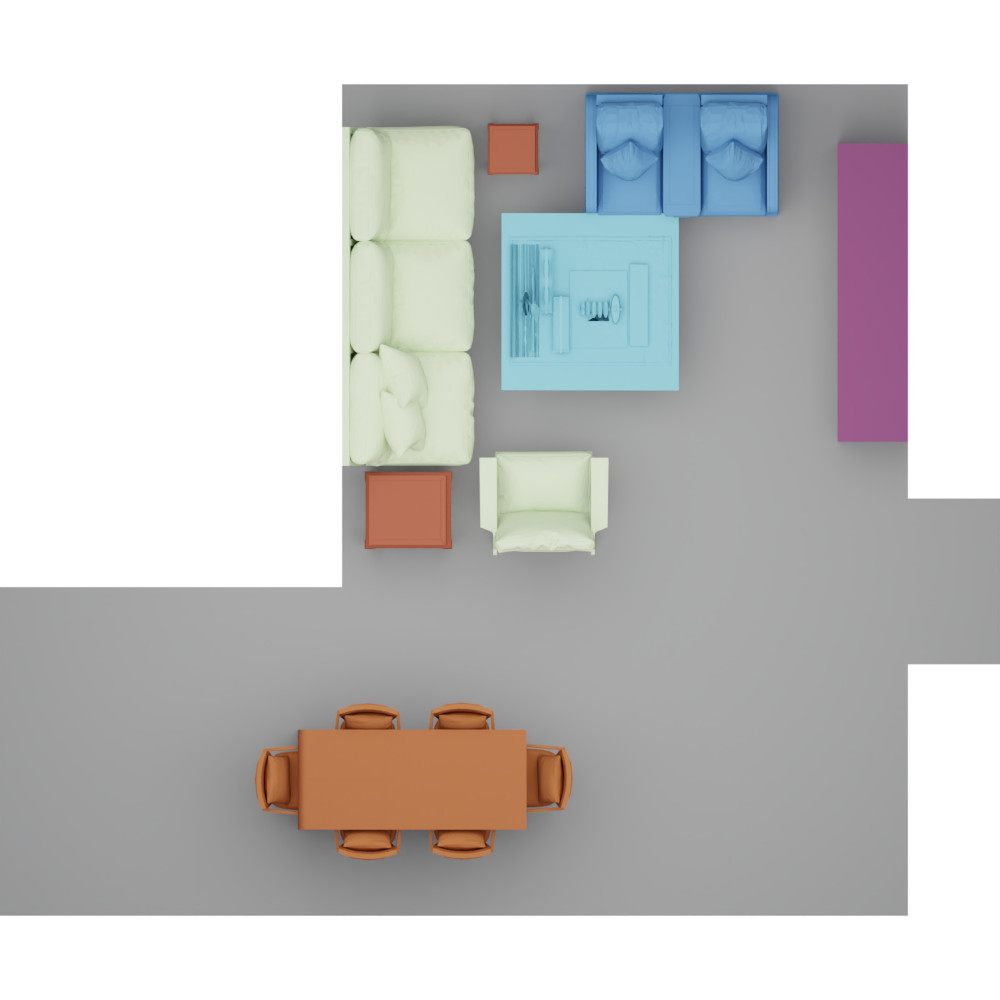}
        \subcaption[]{We now condition the size to be \textit{large}. The class automatically changes to satisfy the conditioning input.}
        \label{fig:arbit_loc_size_sample}
    \end{subfigure}
    \begin{subfigure}[t]{0.28\linewidth}
        \includegraphics[width=\linewidth]{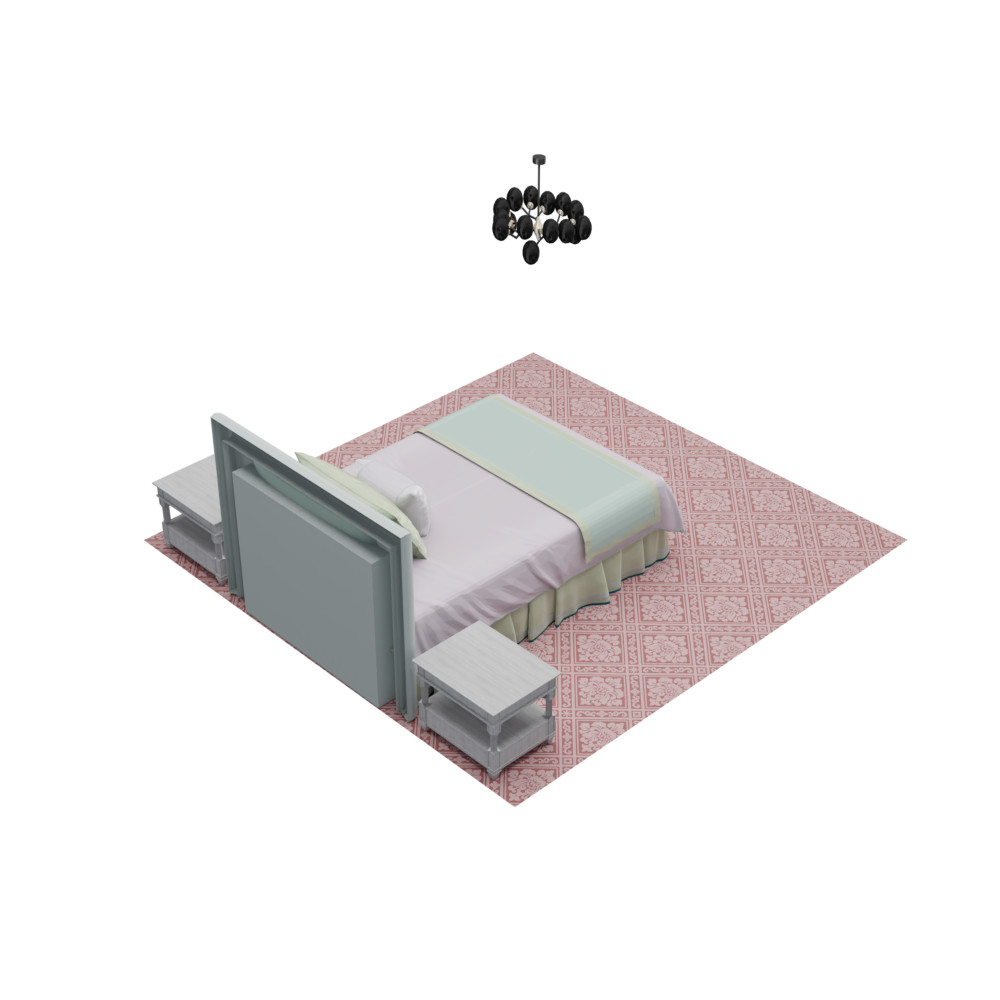}
        \subcaption[]{ Ground-truth Floorplan}
    \end{subfigure}
    \hspace{0.1cm}
    \begin{subfigure}[t]{0.28\linewidth}
        \includegraphics[width=\linewidth]{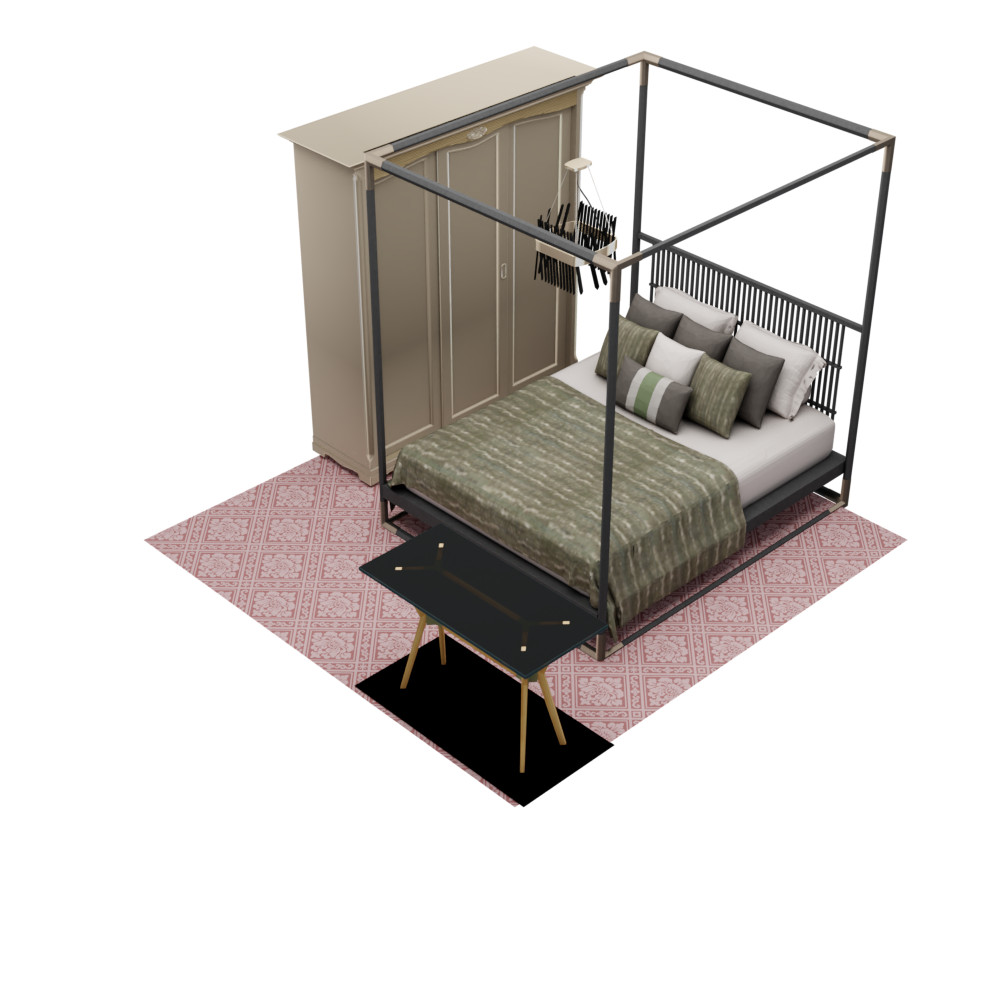}
        \subcaption[]{We fill the class tokens in $C$ to correspond to \texttt{lamp, bed, wardrobe, table}.}
    \end{subfigure}
        \hspace{0.1cm}
    \begin{subfigure}[t]{0.28\linewidth}
        \includegraphics[width=\linewidth]{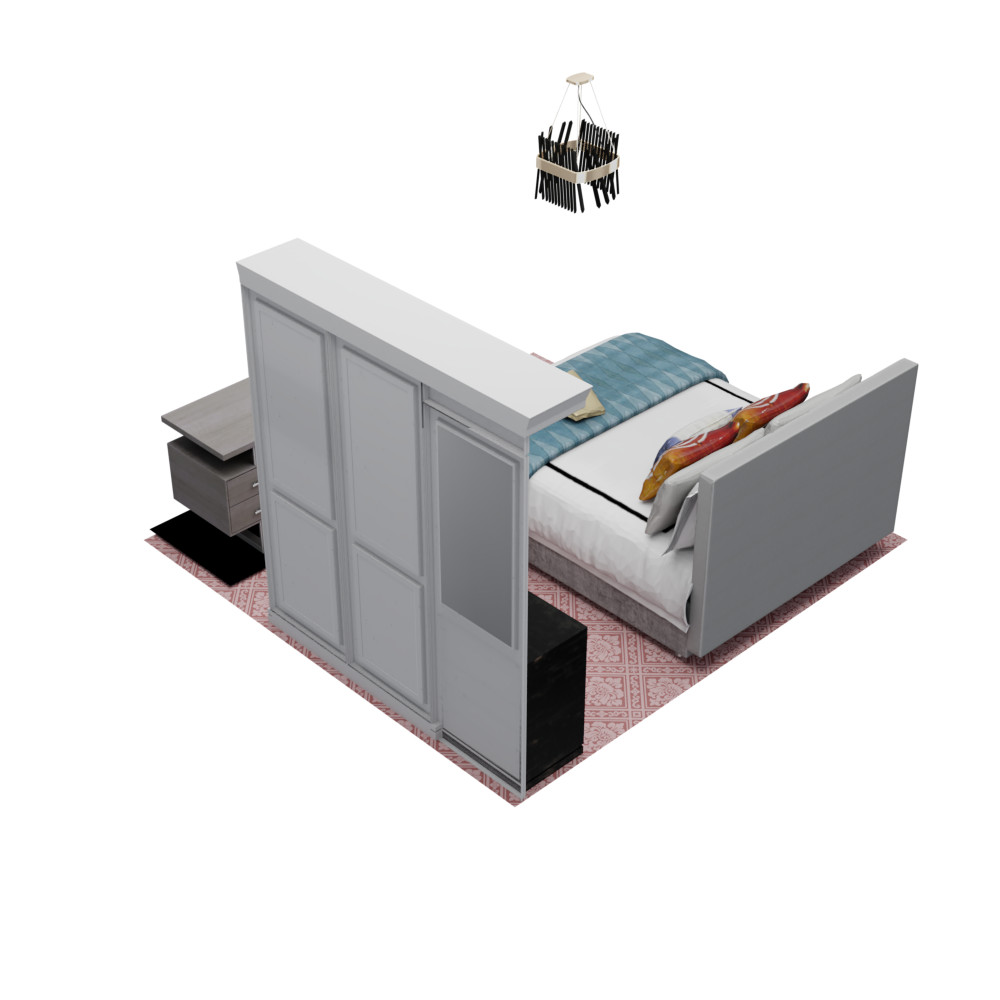}
        \subcaption[]{Same conditioning as before, but now we constrain the angle of the bed.}
    \end{subfigure}
    
    \begin{subfigure}[t]{0.28\linewidth}
        \includegraphics[width=\linewidth]{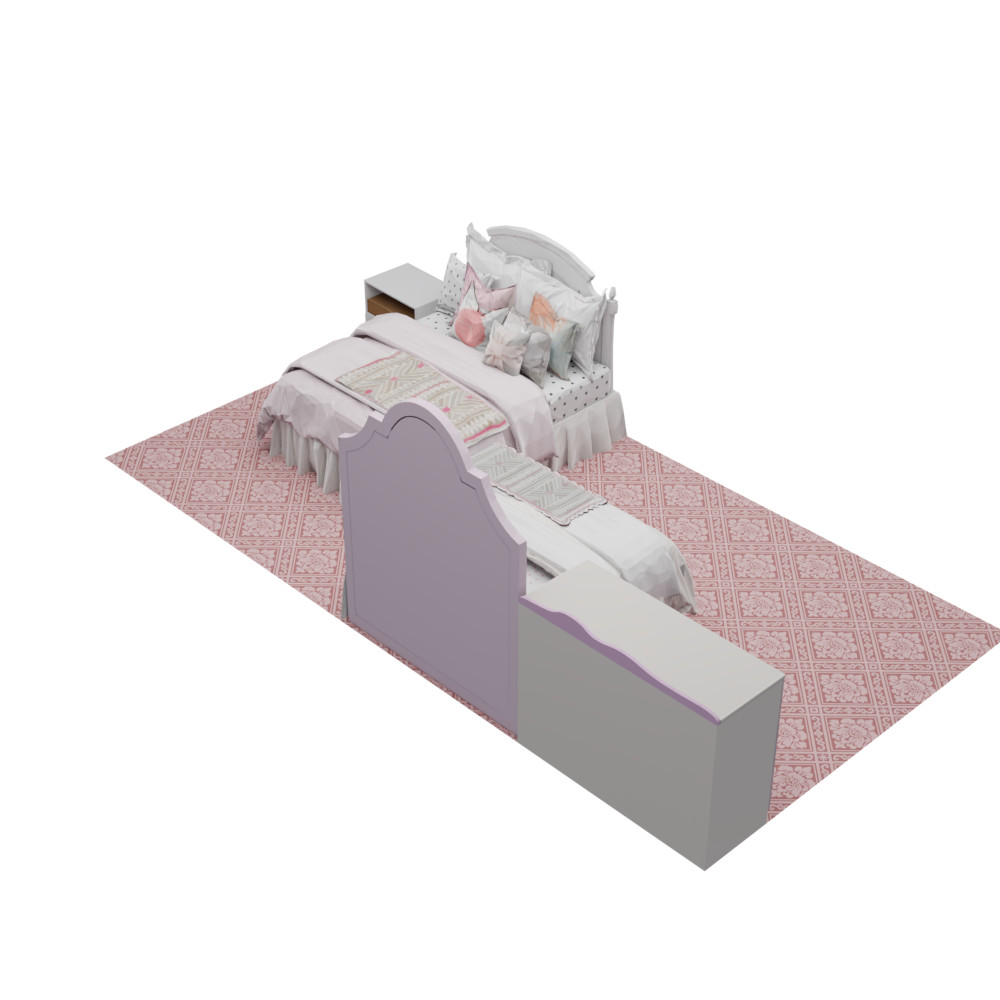}
        \subcaption[]{ Ground-Truth Floorplan}
    \end{subfigure}
    \hspace{0.1cm}
    \begin{subfigure}[t]{0.28\linewidth}
        \includegraphics[width=\linewidth]{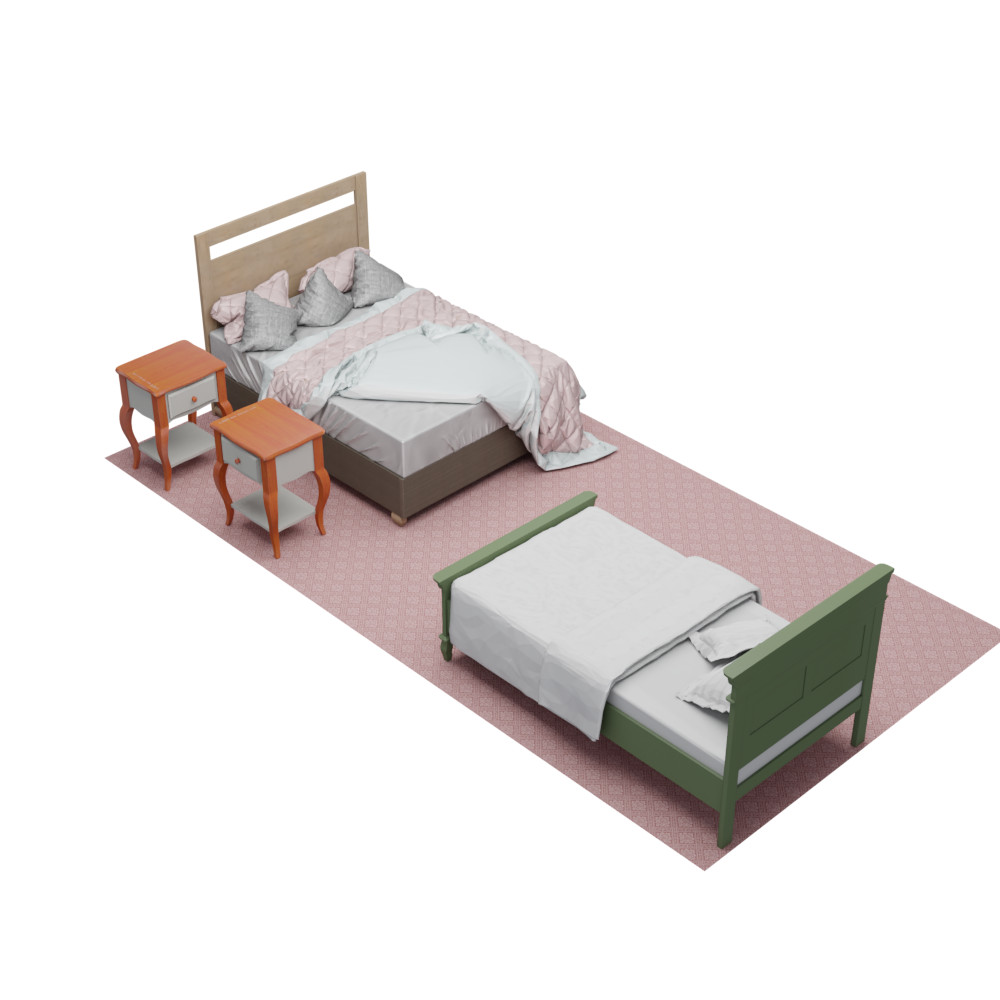}
         \subcaption[]{We condition to have two beds angled opposite each other, and two nightstands.}
    \end{subfigure}
        \hspace{0.1cm}
    \begin{subfigure}[t]{0.28\linewidth}
        \includegraphics[width=\linewidth]{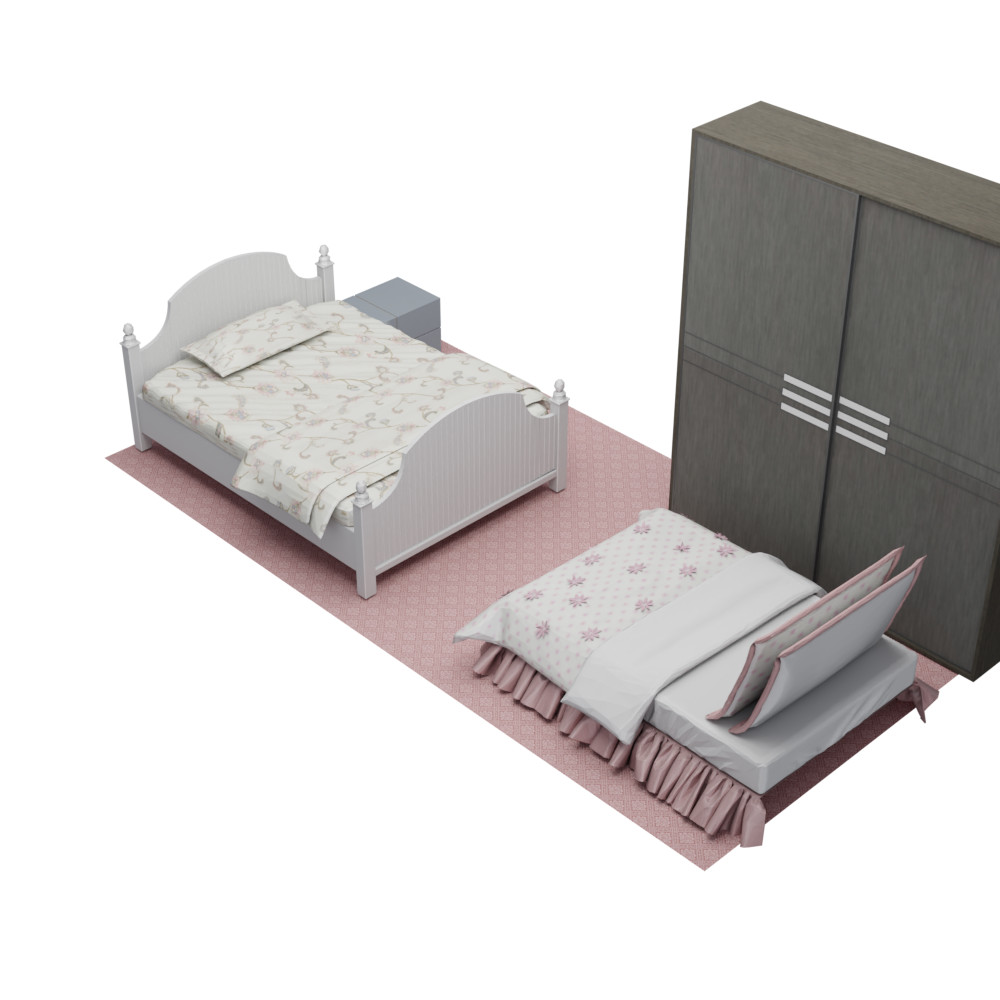}
        \subcaption[]{We change one nightstand to be a wardrobe.}
    \end{subfigure}
    \label{fig:unc_sampling}
\end{figure}

We describe the process using the following example:
We start out with a layout, shown in Fig.~\ref{fig:arbit_persp}. If we mask out the table in cyan (Fig.~\ref{fig:arbit_ortho}, and sample unconditionally, we get another similar table (Fig.~\ref{fig:arbit_ortho_unc}). We now wish to have some control over the generation process.

We now mask out a different object - stool in the upper left corner. We have masked out a single object, thus we have 8 \mask tokens. Our sequences $C$  and $S$ look like Fig.~\ref{fig:arbit_c}. If we want to specify the position of the next object, we simply set the token corresponding to position-attribute of the next object in $C$ - $c_i$ to the value we want. We show a few examples of this type of conditioning in Fig.~\ref{fig:arbit_loc_sample} and Fig.~\ref{fig:arbit_loc_size_sample}. In the rest of the figures, before beginning sampling, we set the class tokens. We see that the generated layouts follow the condition, while also generating plausible layouts, even if the classes of conditioning objects never occur together. As an example, there are only 5 examples of bedrooms with two beds, yet our model is able to reason about the placement of such challenging layouts in Row 5.

We further see that the model is able to place other objects in such a manner that the constrained objects can still satisfy their constraints. In Row 4, we see that when we constrain the angle of the bed, the other objects move in tandem to create a plausible layout.

\FloatBarrier

\section{Limitations and Discussion}

\begin{figure}[hb!]
    \centering
    \includegraphics[width=0.28\linewidth]{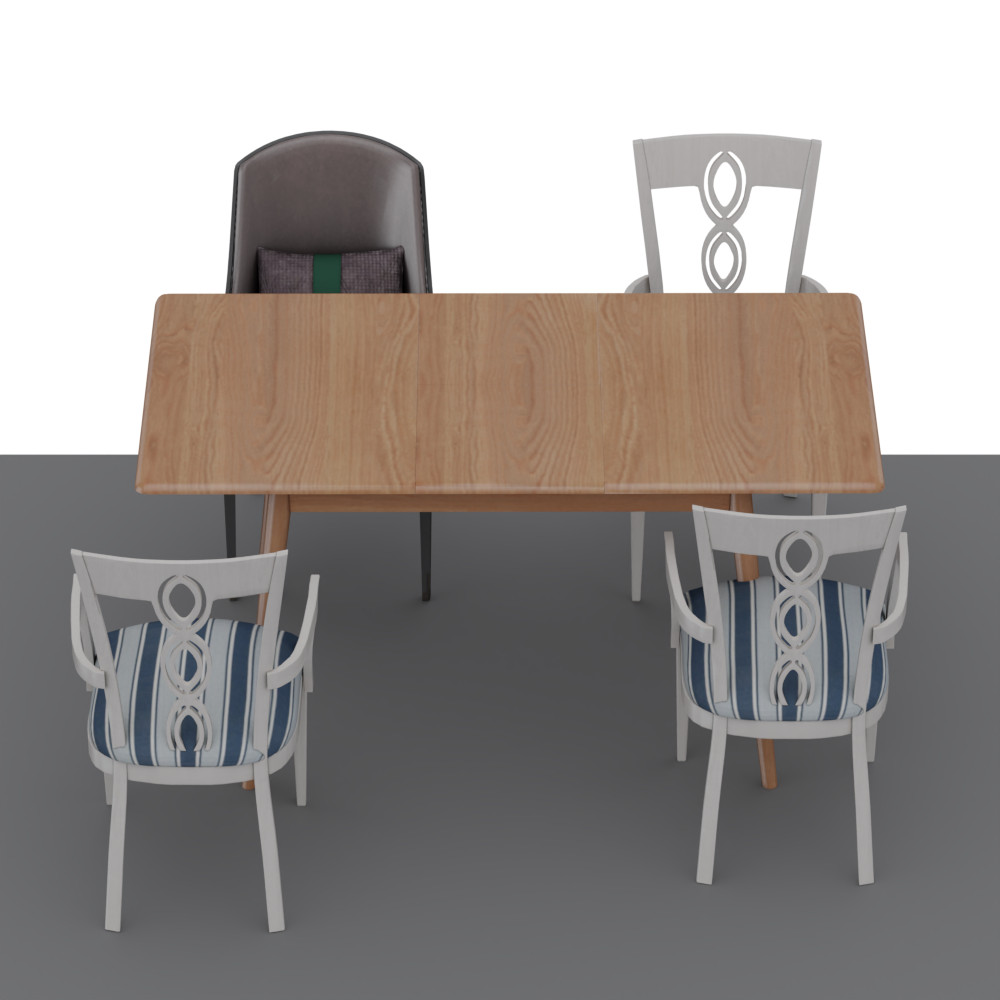}
    \includegraphics[width=0.28\linewidth]{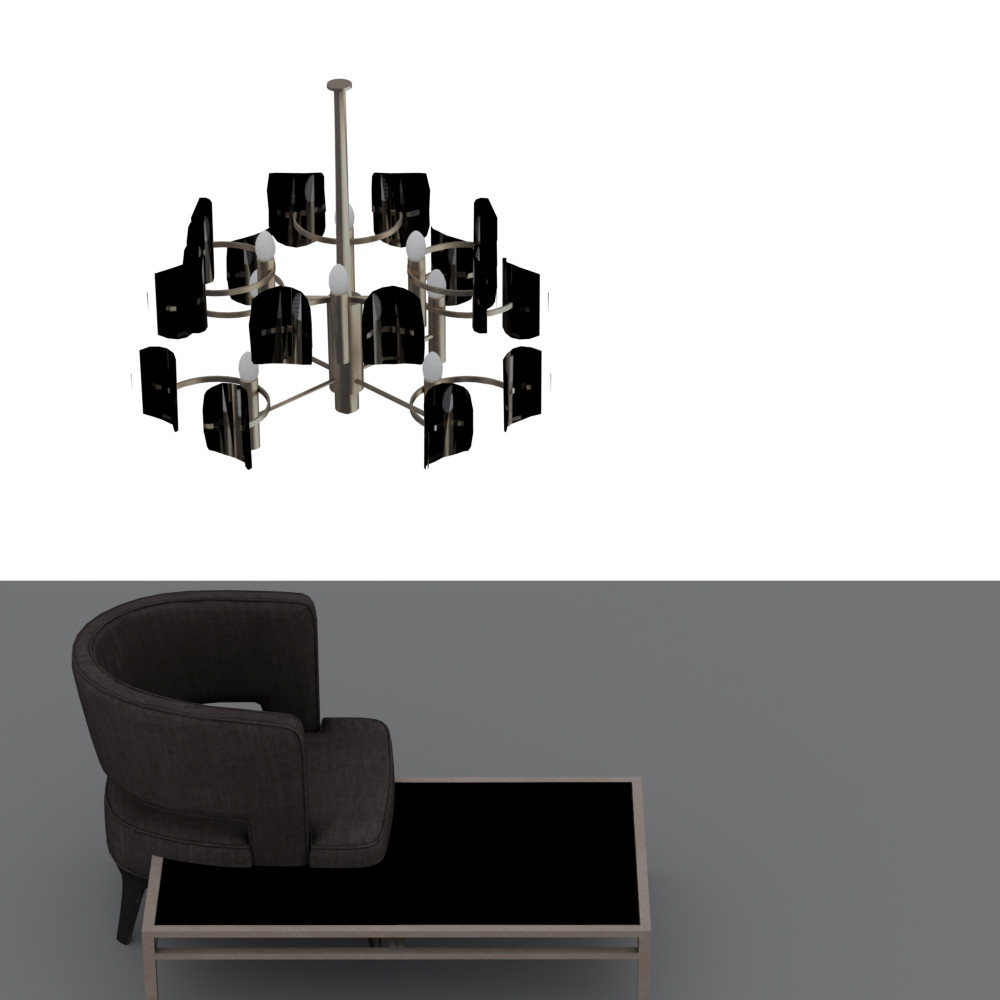}
    \includegraphics[width=0.28\linewidth]{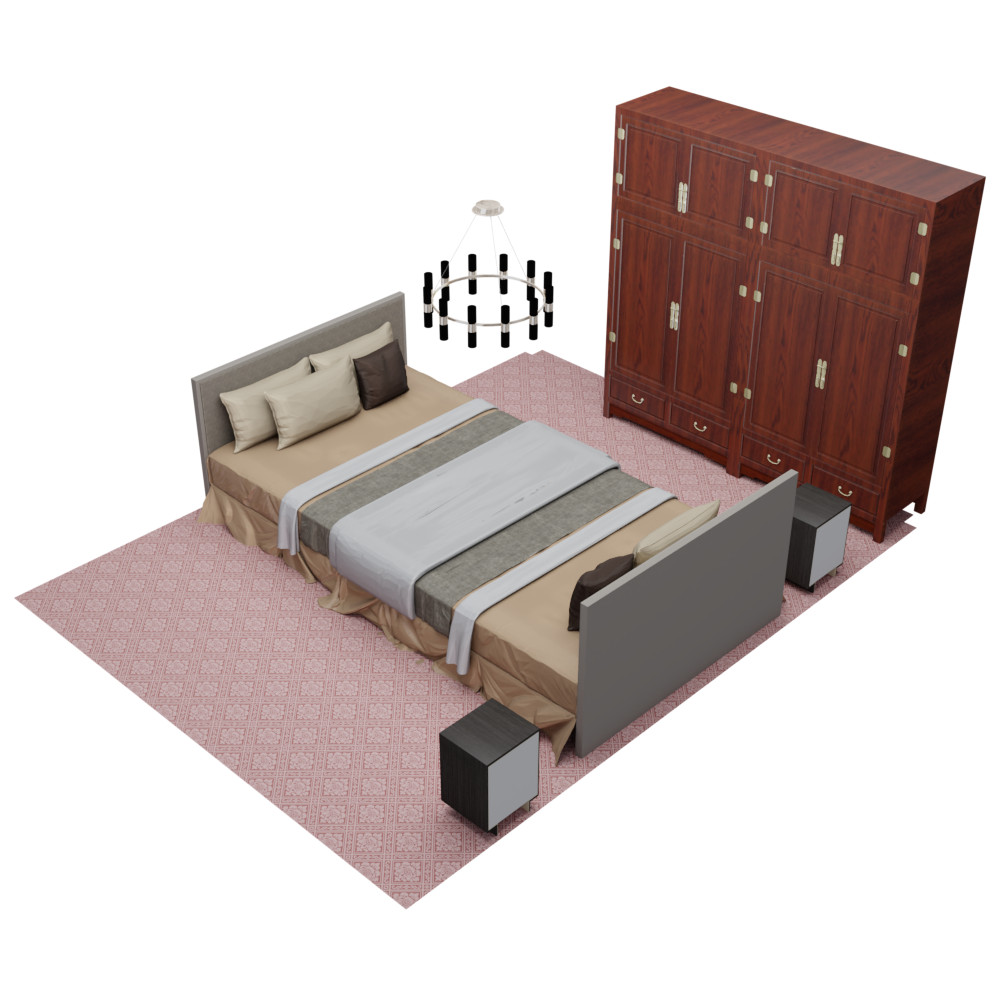}
    \caption{We show failure cases in the samples generated by our model.}
    \label{fig:limitations}
\end{figure}

We now discuss limitations of our model. The first is related to our simple object retrieval scheme based only on bounding box sizes. This often leads to stylistically different objects in close proximity even if the bounding box dimensions are slightly different. We show such an example in Fig.~\ref{fig:limitations}(left). The second is related to the training objective of the model - we only consider the cross entropy/NLL. Thus, the network does not have explicit knowledge of design principles such as non-intersection, or object co-occurrence. This means that the model completely relies on the data being high-quality to ensure such output. We highlighted the fact that certain scenes in the dataset have problematic layouts, and our method cannot filter them out. We show an example of intersections in Fig.~\ref{fig:limitations}(center). Thirdly, the the performance on the \liv and \din datasets is not as good as the other classes, which is clear from the CAS scores. This is in part because the datasets are small but also have significantly more objects than \bed or \lib. This leads to accumulated errors. We would like to explore novel sampling strategies to mitigate such errors. Lastly, while the conditioning works well, it is not guaranteed to generate a good layout. For example, in Fig.~\ref{fig:limitations}(right) we set the condition to be two beds opposite each other, but the network is unable to place them in valid locations. Adding explicit design knowledge would help mitigate such arrangements, but we leave that extension to future work.

\section{Additional Results}

We show additional results on unconditional sampling from our model in the concluding figures. Our synthesised layouts are novel and do not merely copy the ground-truth layout. In addition, we see that our layouts respect the floorplan boundary and mimic the underlying style of the datasets, in terms of object-object co-occurrence.

%%%%%%%%%%%%%%%%%%%%%%%%%%%%%%%%%%%%%%%%%%%%%%%%%%%%%%%%%%%%%%%%%%%%%%%%%%%%%%%%%%%%%%%%%%%%
%%%%%%%%%%%%%%%%%%%%%%%%%%%%%%%%%%%%%%%%%%%%%%%%%%%%%%%%%%%%%%%%%%%%%%%%%%%%%%%%%%%%%%%%%%%%%%%5
%%%%%%%%%%%%%%%%%%%%%%%%%%%%%%%%%55
%% LIBRARIES
\begin{figure*}[t!]

    \centering
    \vspace{-1.5em}
    % \hfill
    \begin{subfigure}[b]{0.22\linewidth}
        \centering
	    \small Boundary
    \end{subfigure}%
    \begin{subfigure}[b]{0.22\linewidth}
        \centering
        \small GT
    \end{subfigure}%
    \begin{subfigure}[b]{0.22\linewidth}
        \centering
        \small ATISS
    \end{subfigure}%
    \begin{subfigure}[b]{0.22\linewidth}
        \centering
        \small Ours
    \end{subfigure}%
    % \hfill%
    \vskip\baselineskip%
    \vspace{-0.75em}
    %%%%%%%%%%%%%%%%%%%%%%%%%%%%%%%%%%%%
    % \hfill
    \begin{subfigure}[b]{0.22\linewidth}
        \centering
	    \includegraphics[width=\textwidth,  clip]{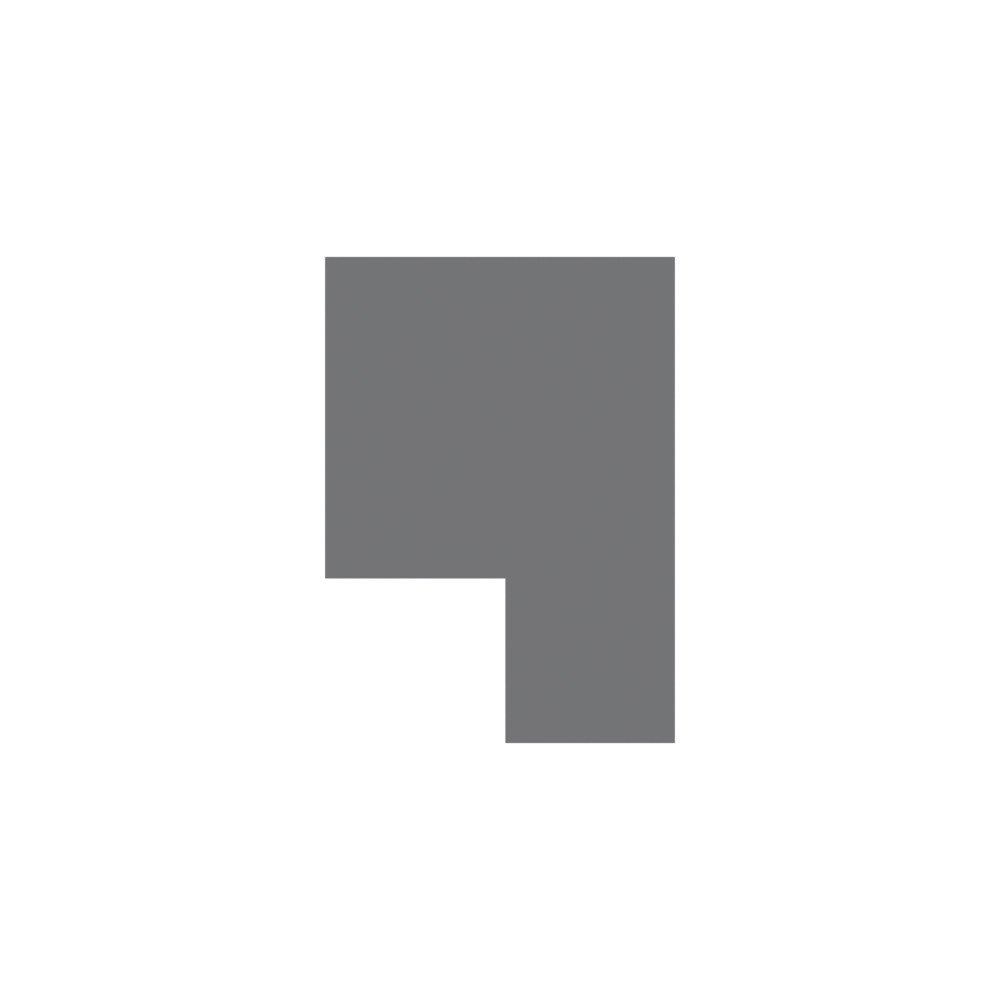}
    \end{subfigure}%
    \begin{subfigure}[b]{0.22\linewidth}
        \centering
            \begin{overpic}[width=\textwidth,  clip]{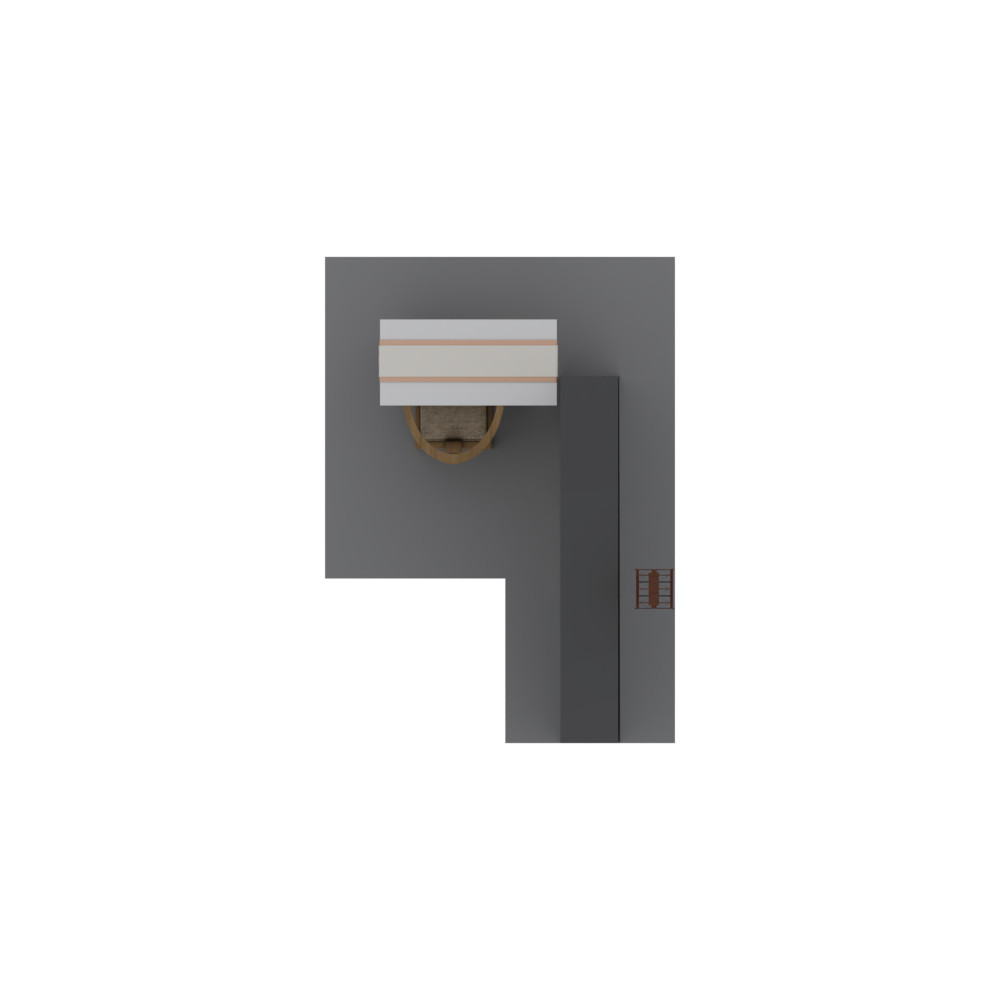}
        \end{overpic}
    \end{subfigure}%
    \begin{subfigure}[b]{0.22\linewidth}
        \centering
        \begin{overpic}[width=\textwidth, clip]{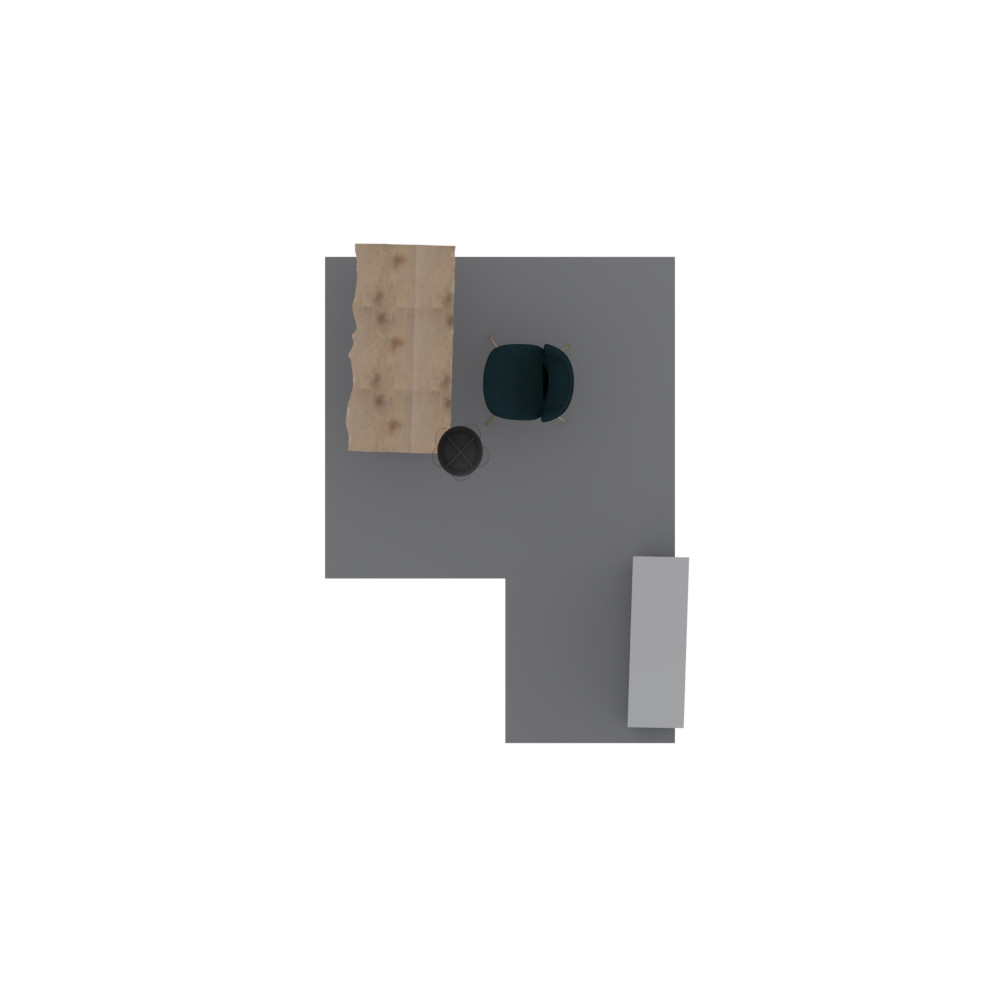}
	    \end{overpic}
    \end{subfigure}%
    \begin{subfigure}[b]{0.22\linewidth}
        \centering
        \begin{overpic}[width=\textwidth, clip]{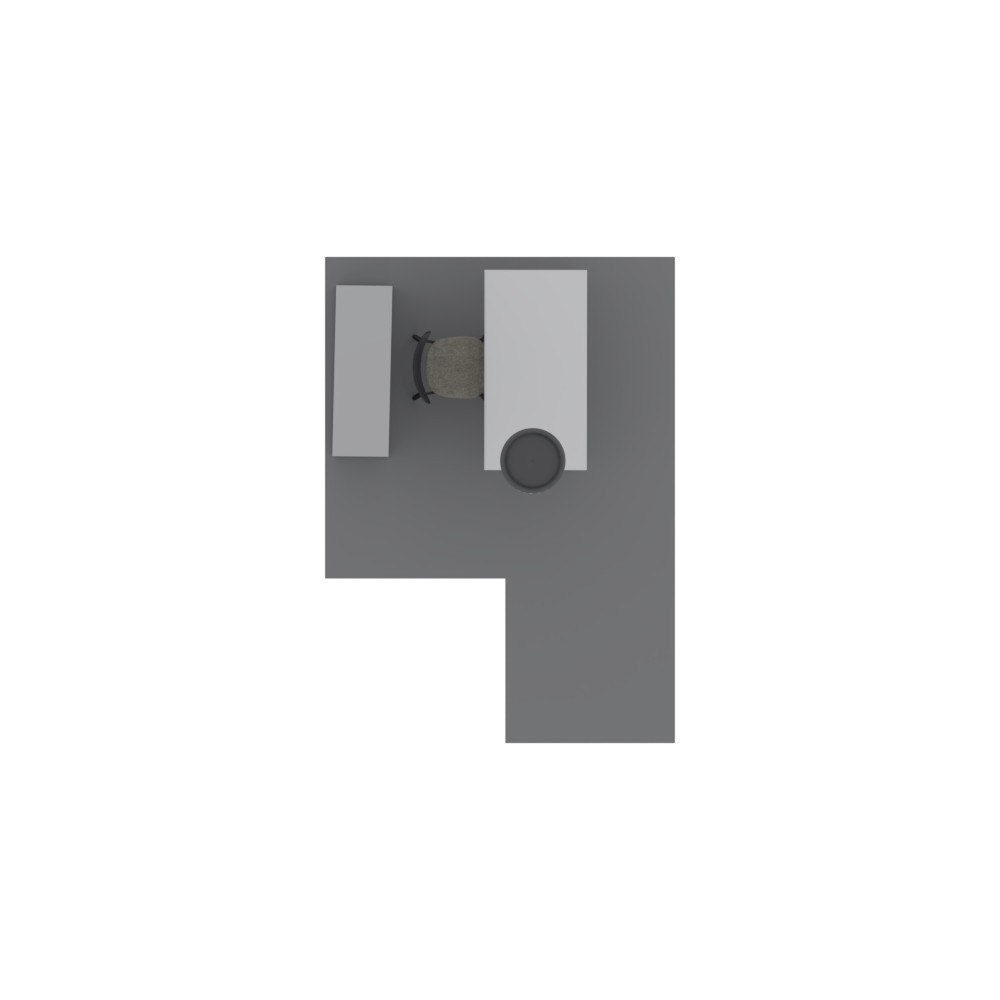}
        \end{overpic}
    \end{subfigure}%
    % \hfill%
    \vskip\baselineskip%
    \vspace{-0.75em}
    % \hfill
    %%%%%%%%%%%%%%%%%%%%%%%%%%%%%%%%%%%%%%%%%%%%%%%%%%%%%%%%%%%%%%%%%%%%%
    \begin{subfigure}[b]{0.22\linewidth}
        \centering
	    \includegraphics[width=\textwidth, clip]{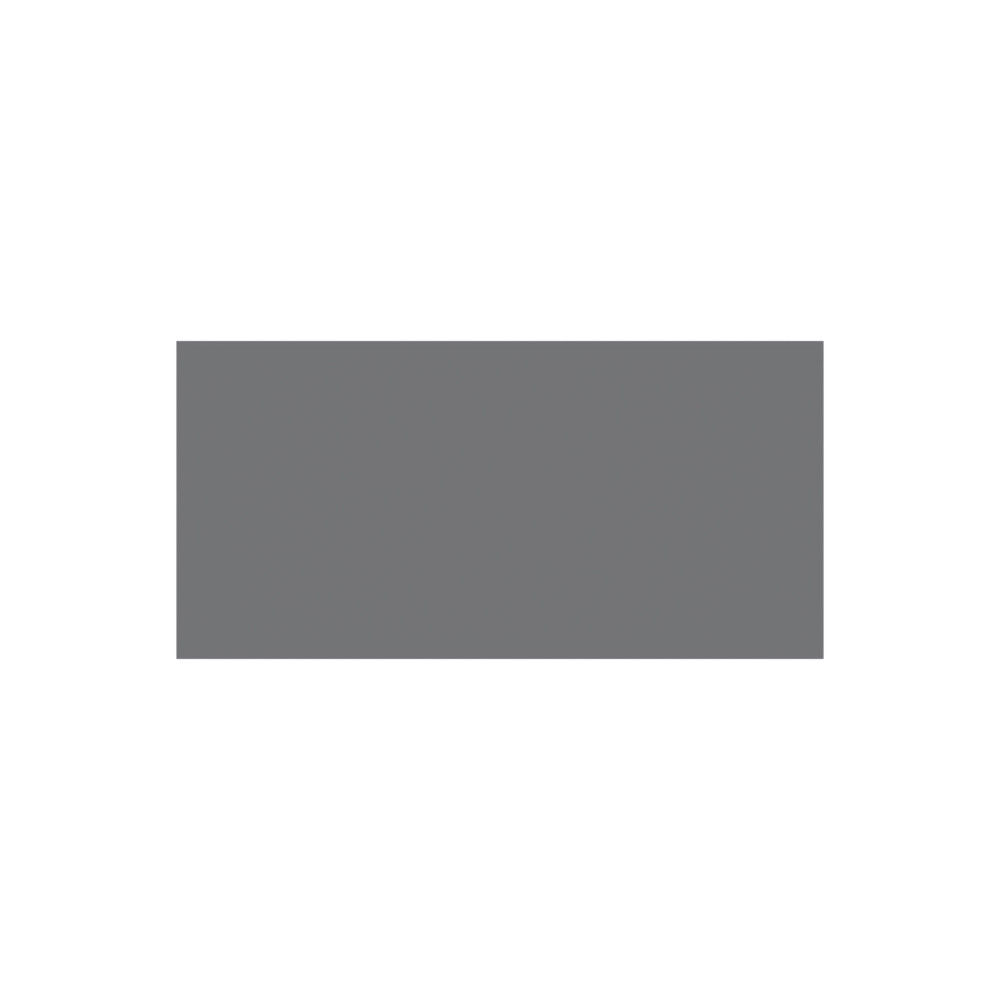}
    \end{subfigure}%
    \begin{subfigure}[b]{0.22\linewidth}
        \centering
        \begin{overpic}[width=\textwidth,  clip]{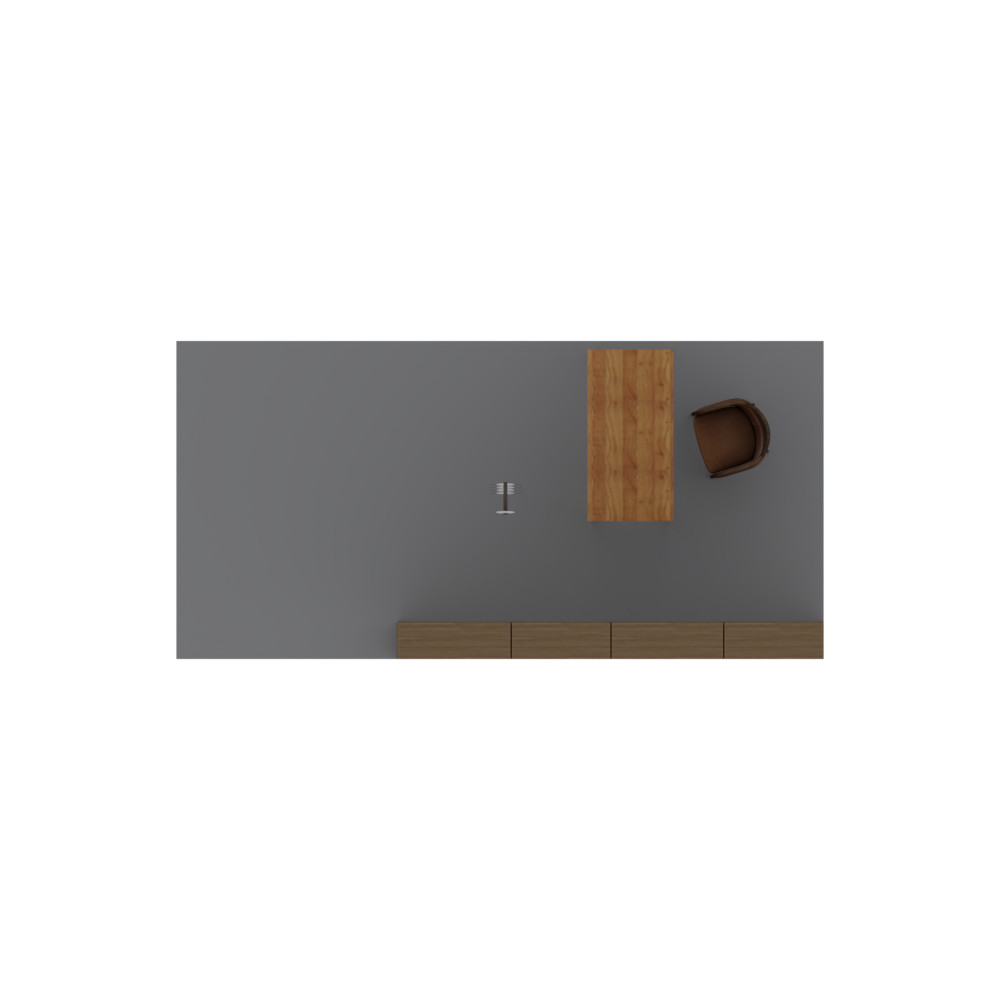}
        \end{overpic}
    \end{subfigure}%
    \begin{subfigure}[b]{0.22\linewidth}
        \centering
        \begin{overpic}[width=\textwidth, clip]{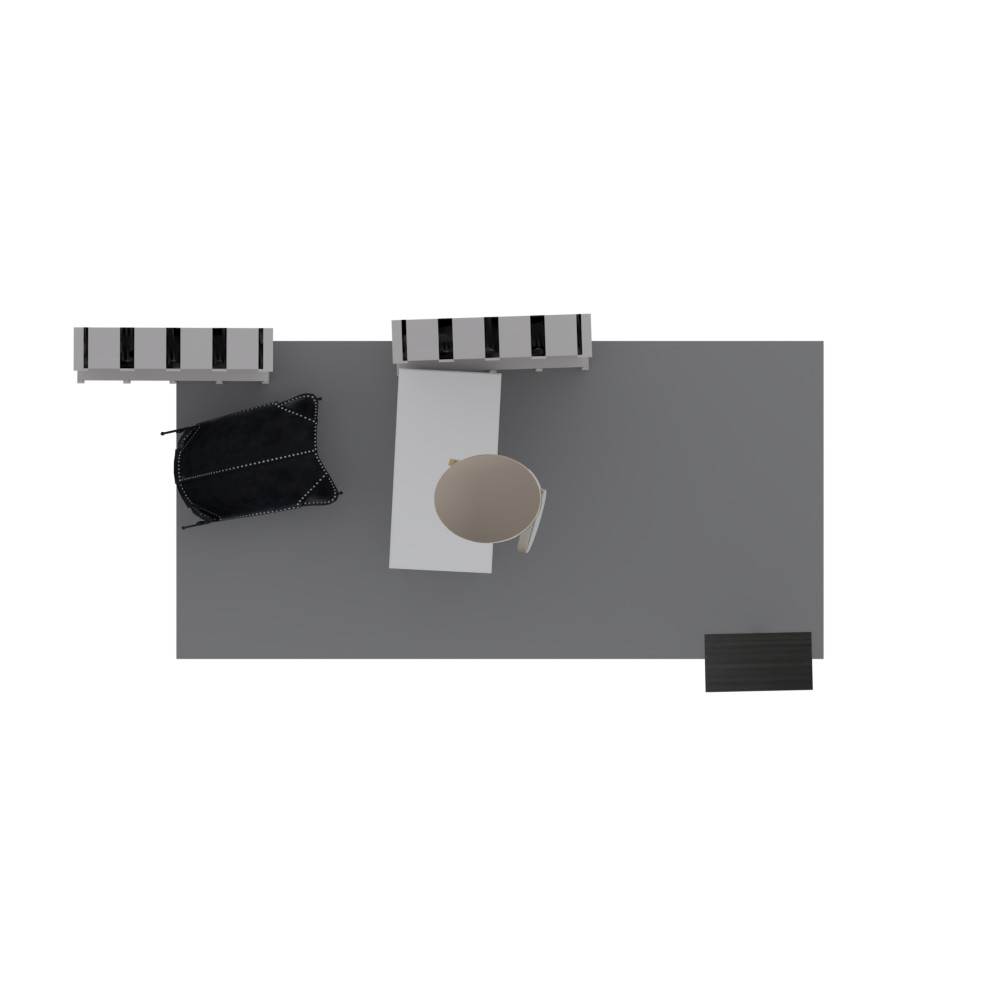}
	    \end{overpic}
    \end{subfigure}%
    \begin{subfigure}[b]{0.22\linewidth}
        \begin{overpic}[width=\textwidth,  clip]{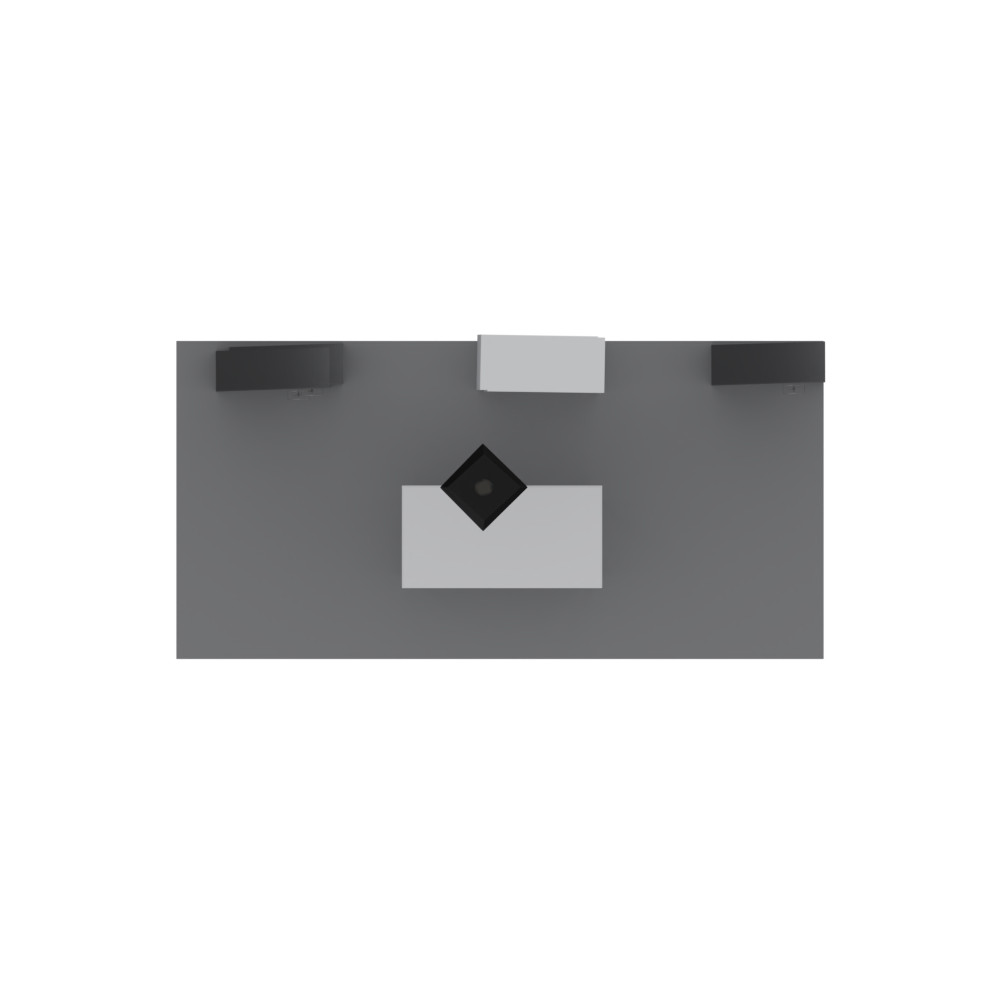}
        \end{overpic}
    \end{subfigure}%
    % \hfill%
    \vskip\baselineskip%
    \vspace{-0.75em}
    % \hfill
    %%%%%%%%%%%%%%%%%%%%%%%%%%%%%%%%%%%%%%%%%%%%%%%%%%%%%%%%%%%%%%%%%%%%%%%%%%%%%%%%%%%%%%%%%%%%%%%%%%%5
    \begin{subfigure}[b]{0.22\linewidth}
        \centering
	    \includegraphics[width=\textwidth, clip]{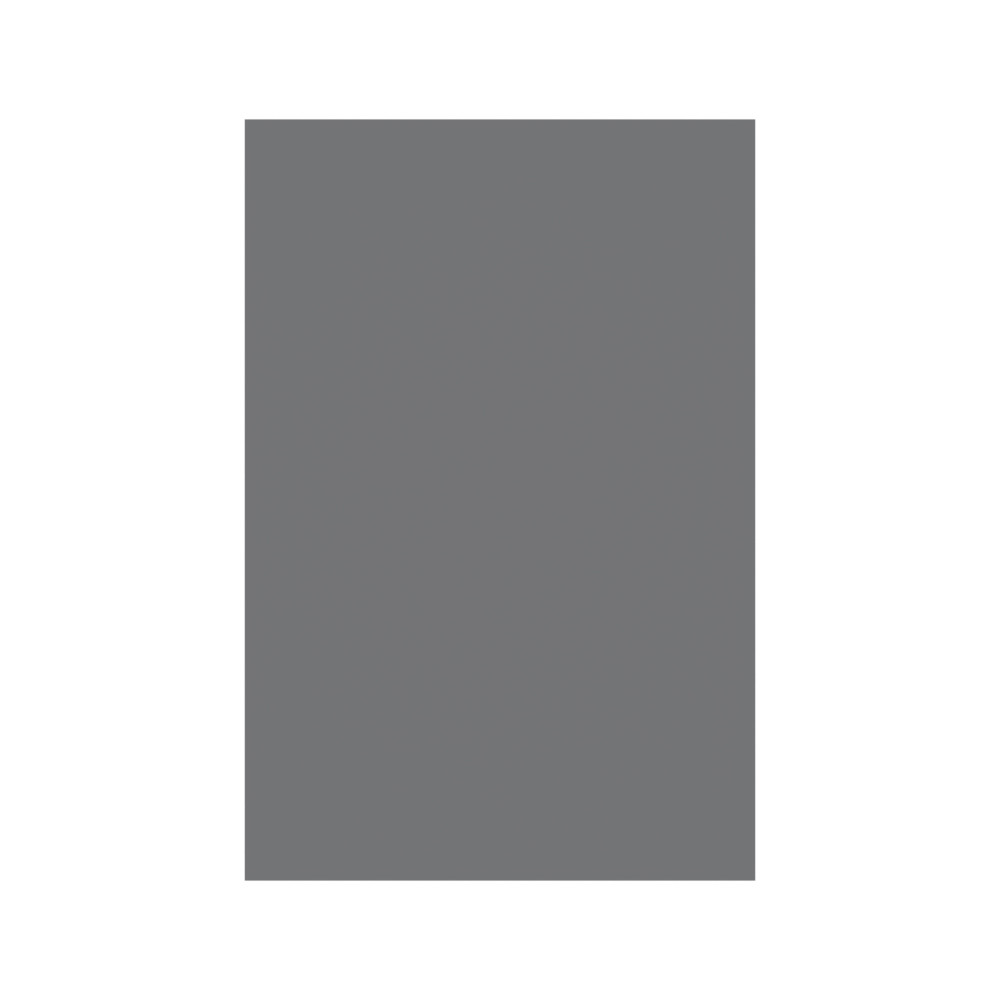}
    \end{subfigure}%
    \begin{subfigure}[b]{0.22\linewidth}
        \centering
        \begin{overpic}[width=\textwidth,  clip]{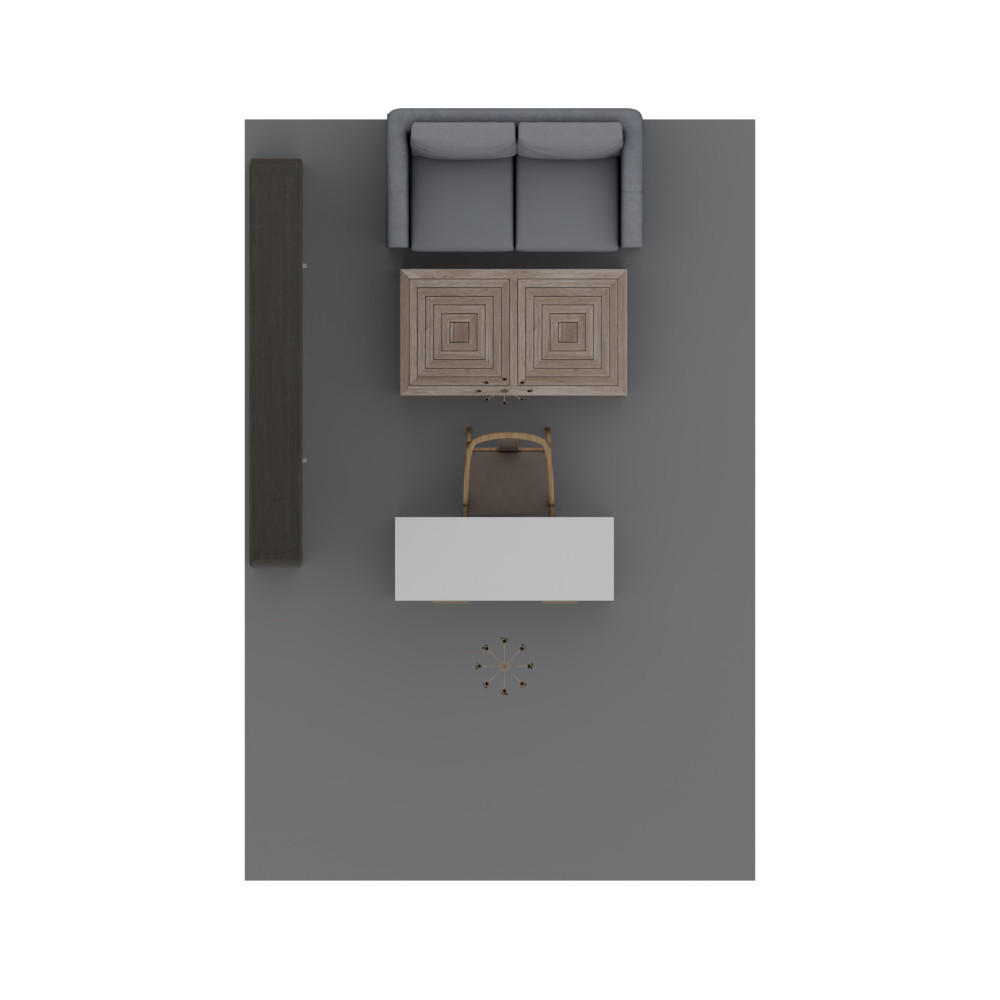}
        \end{overpic}
    \end{subfigure}%
    \begin{subfigure}[b]{0.22\linewidth}
        \centering
        \begin{overpic}[width=\textwidth, clip]{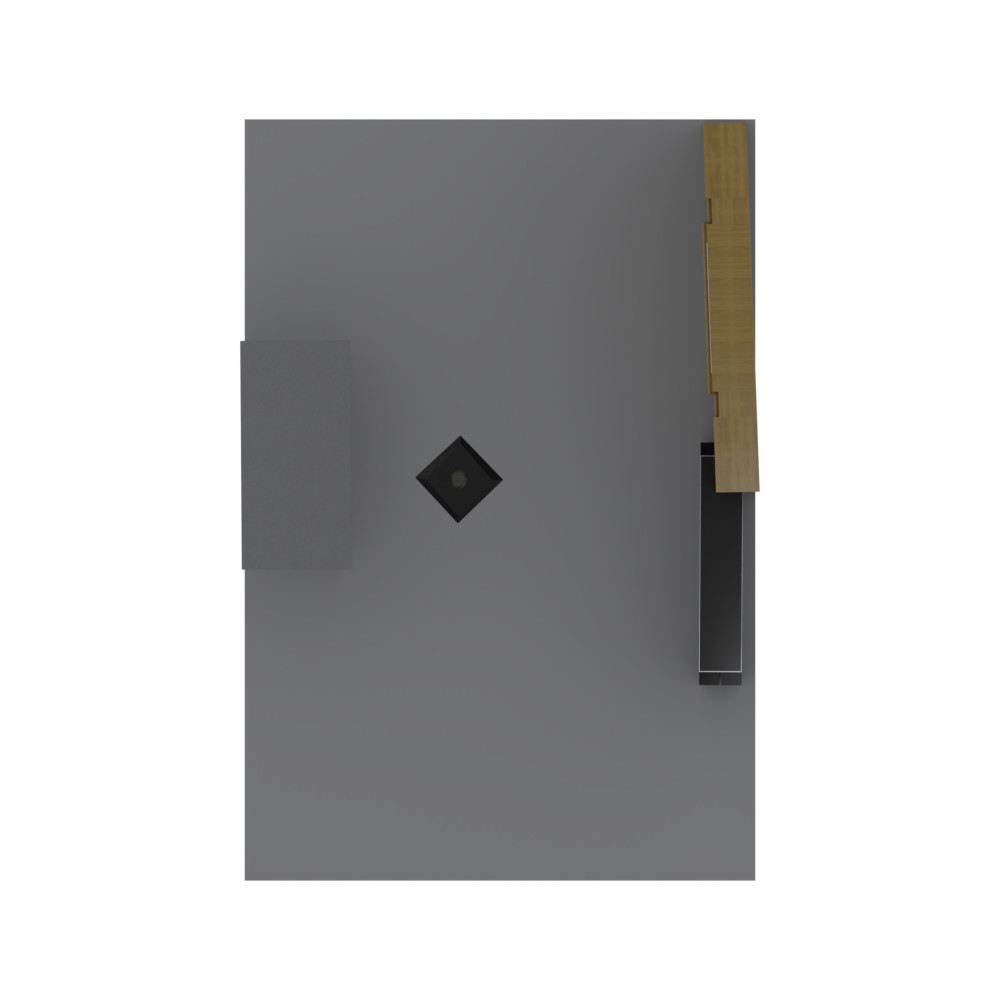}
	    \end{overpic}
    \end{subfigure}%
    \begin{subfigure}[b]{0.22\linewidth}
        \begin{overpic}[width=\textwidth,  clip]{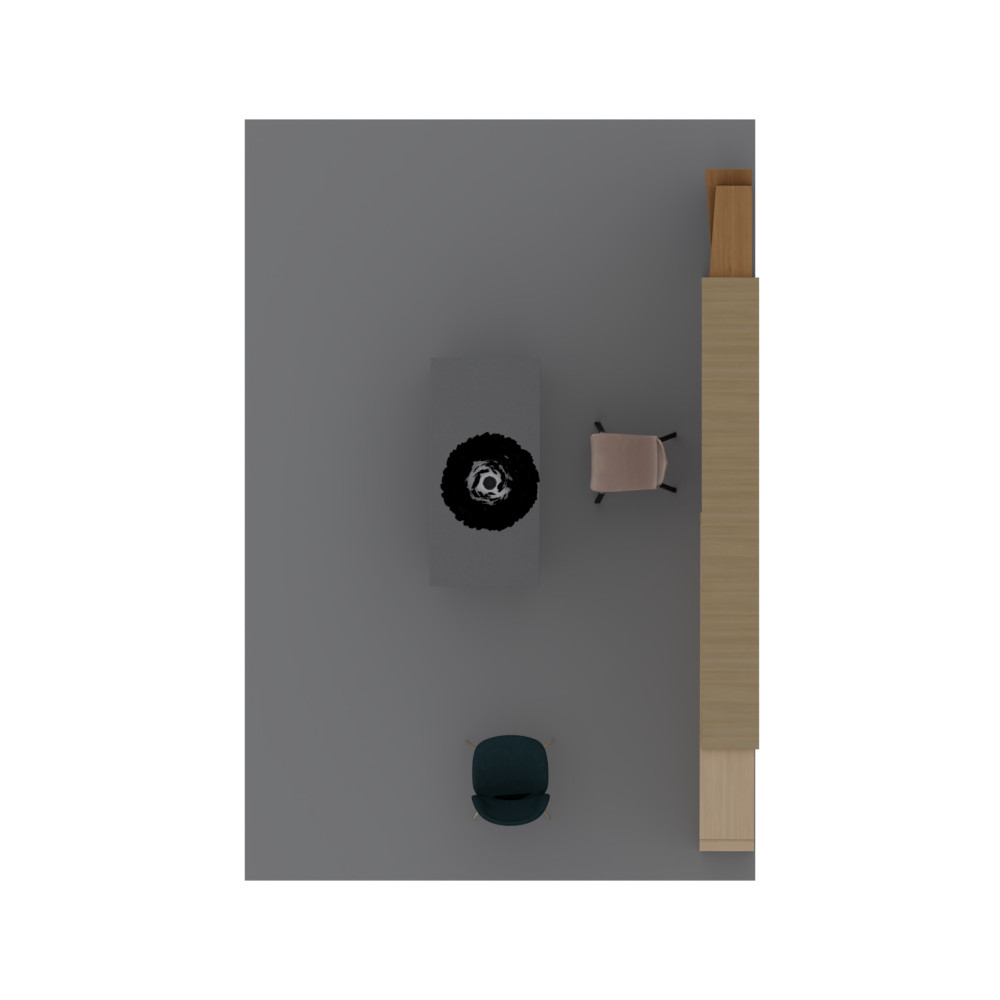}
        \end{overpic}
    \end{subfigure}%
    % \hfill%
    \vskip\baselineskip%
    \vspace{-0.75em}
    \begin{subfigure}[b]{0.22\linewidth}
        \centering
	    \includegraphics[width=\textwidth, clip]{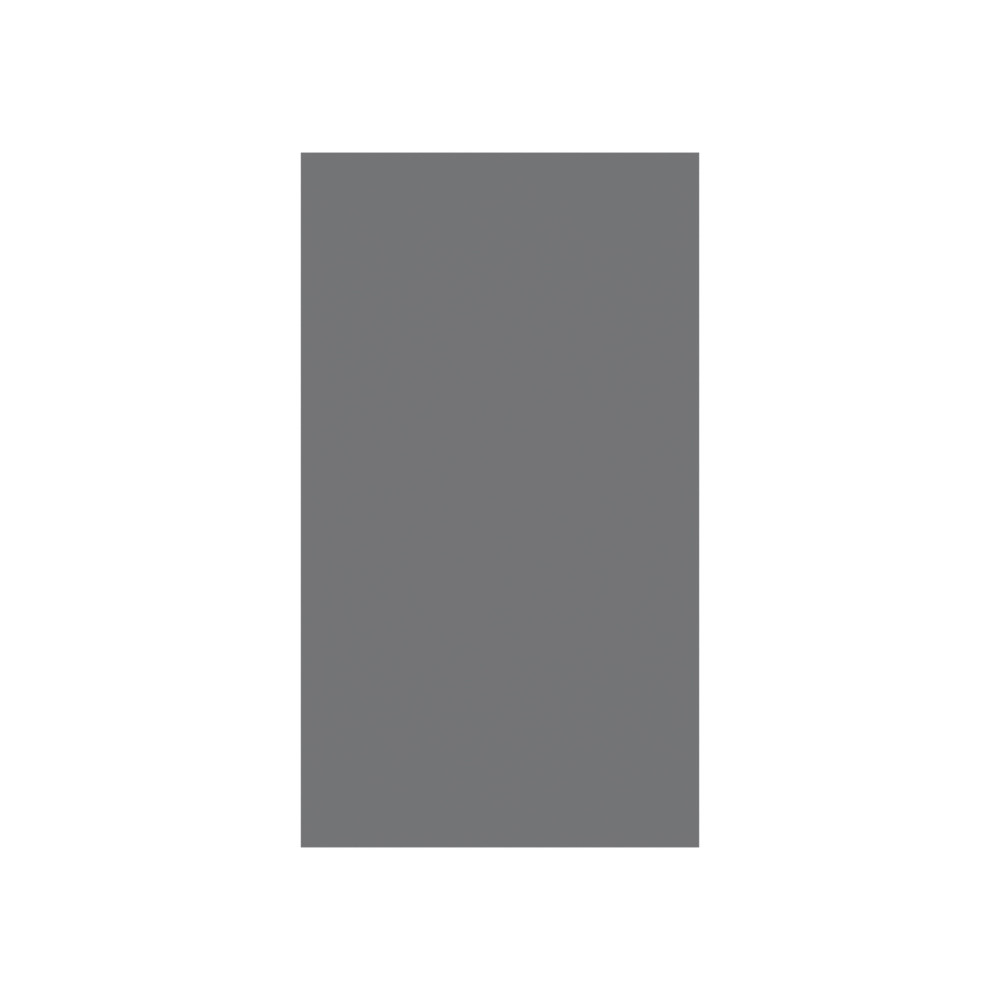}
    \end{subfigure}%
    \begin{subfigure}[b]{0.22\linewidth}
        \centering
        \begin{overpic}[width=\textwidth,  clip]{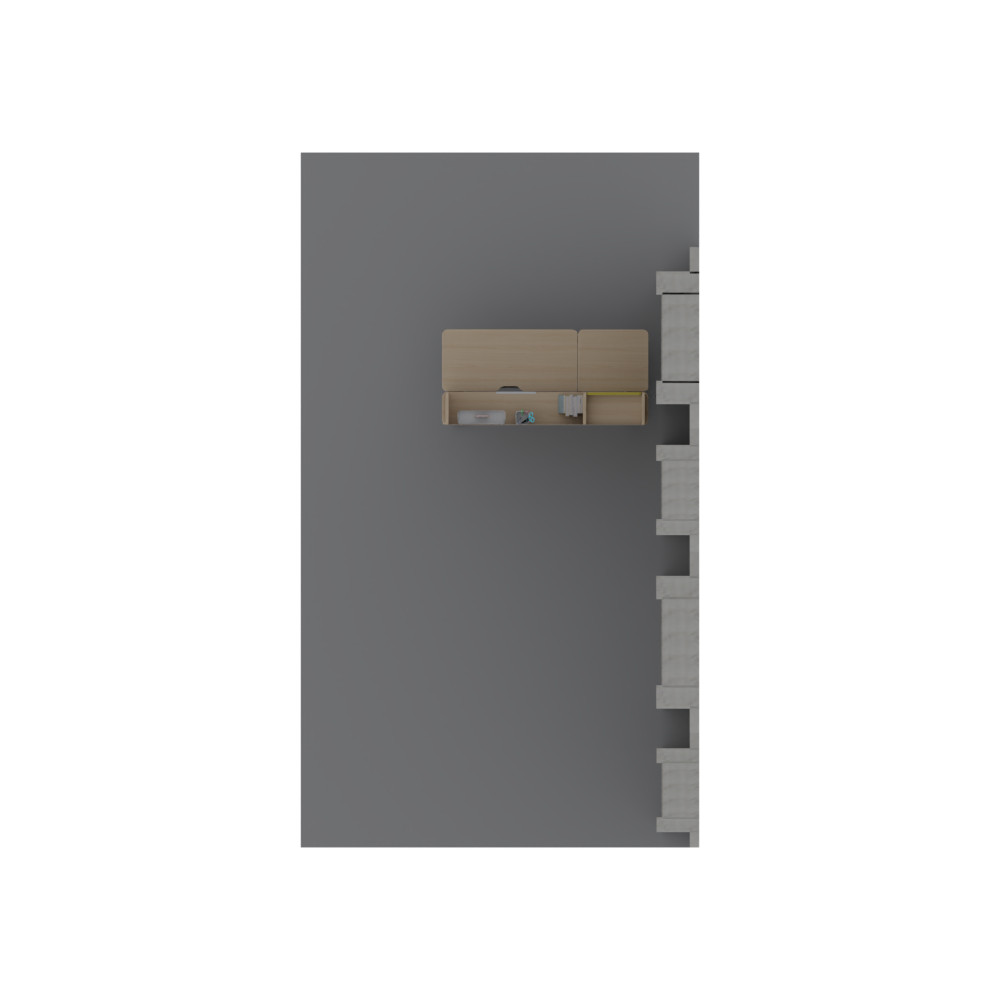}
        \end{overpic}
    \end{subfigure}%
    \begin{subfigure}[b]{0.22\linewidth}
        \centering
        \begin{overpic}[width=\textwidth,  clip]{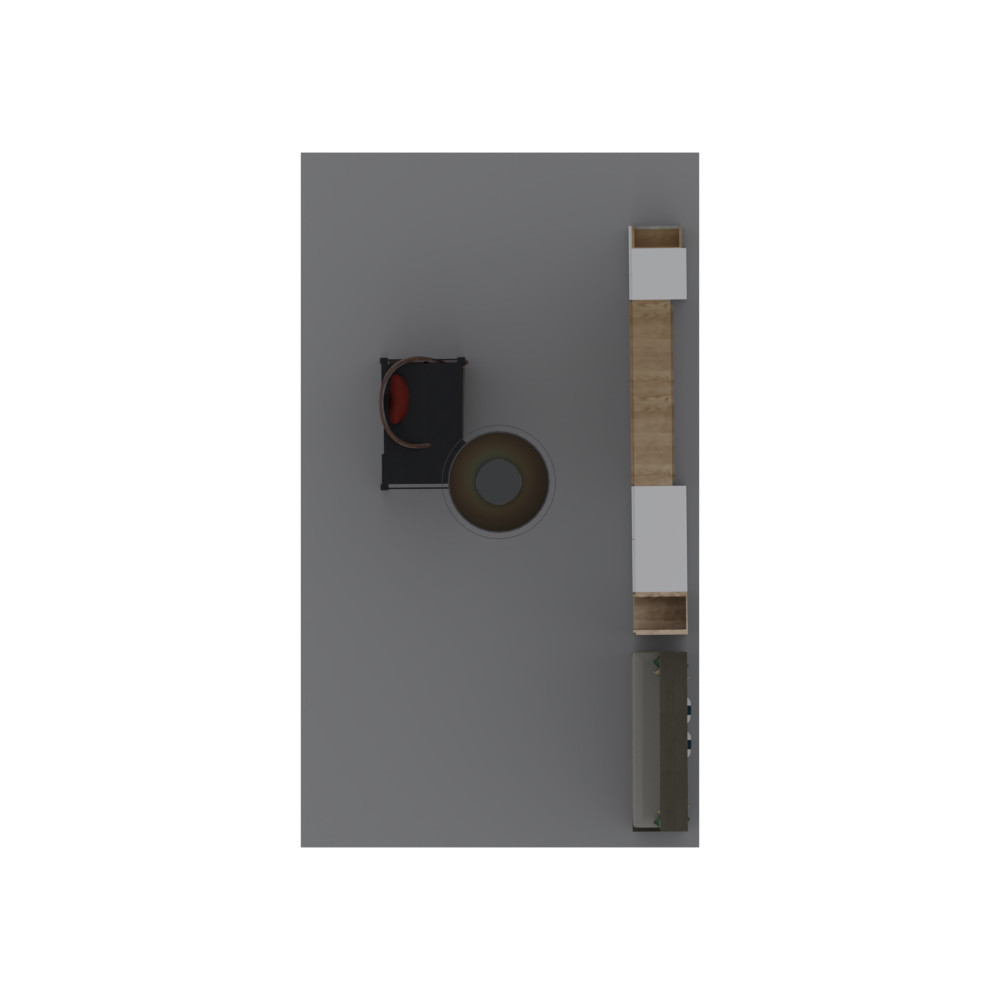}
	    \end{overpic}
    \end{subfigure}%
    \begin{subfigure}[b]{0.22\linewidth}
        \begin{overpic}[width=\textwidth,  clip]{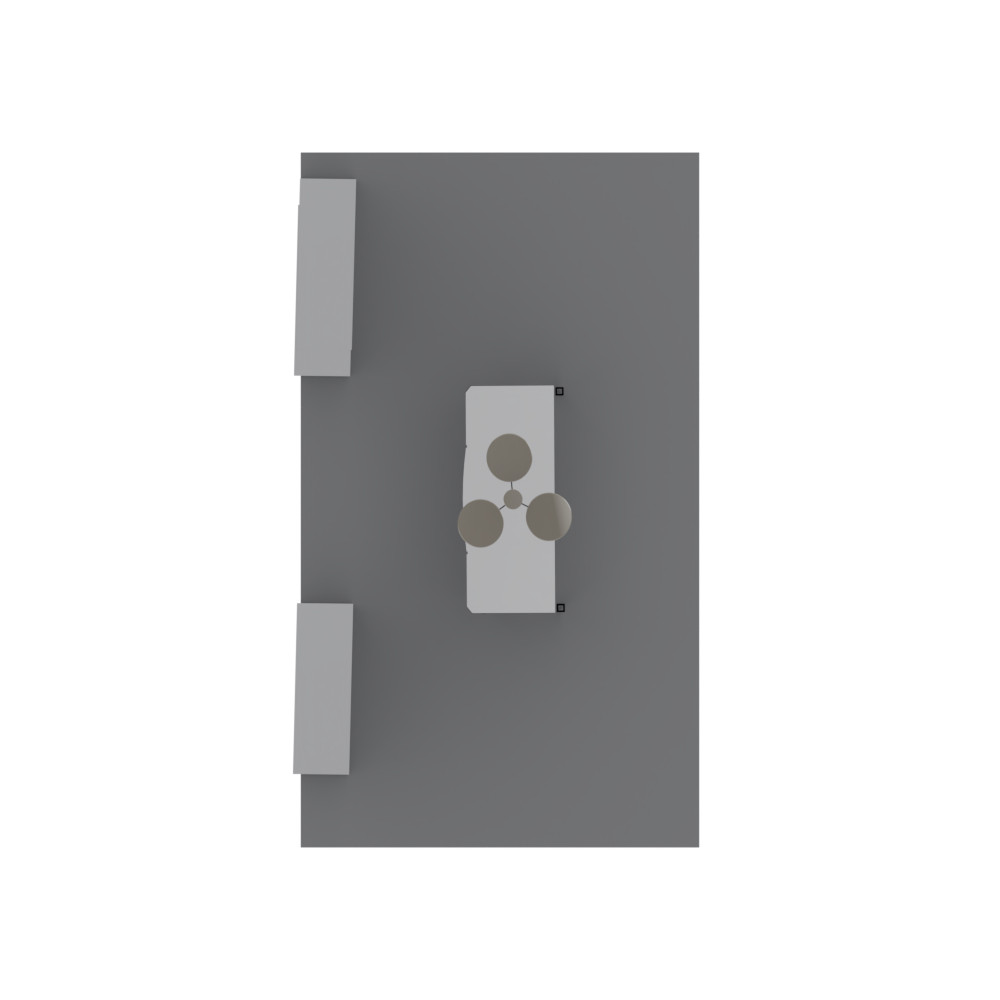}
        \end{overpic}
    \end{subfigure}%
    % \hfill%
    \vskip\baselineskip%
    \vspace{-0.75em}
    \begin{subfigure}[b]{0.22\linewidth}
        \centering
	    \includegraphics[width=\textwidth, clip]{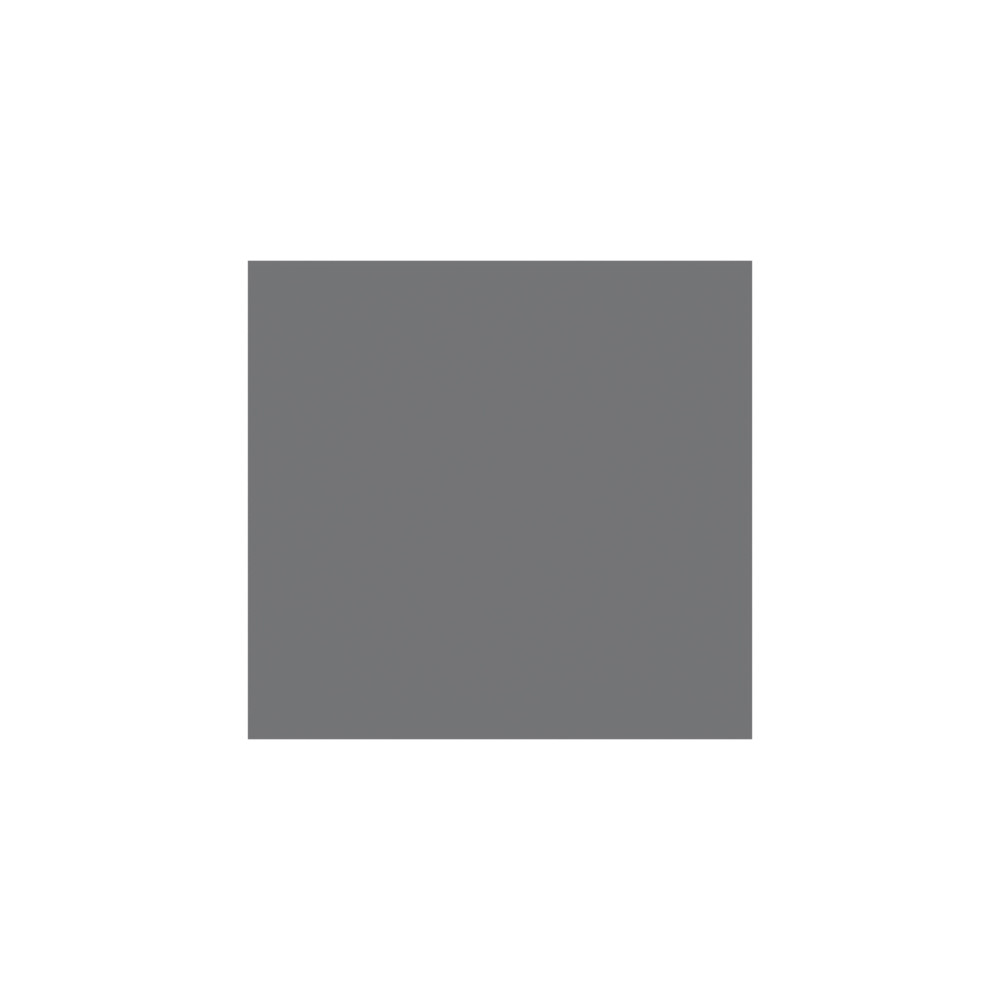}
    \end{subfigure}%
    \begin{subfigure}[b]{0.22\linewidth}
        \centering
        \begin{overpic}[width=\textwidth,  clip]{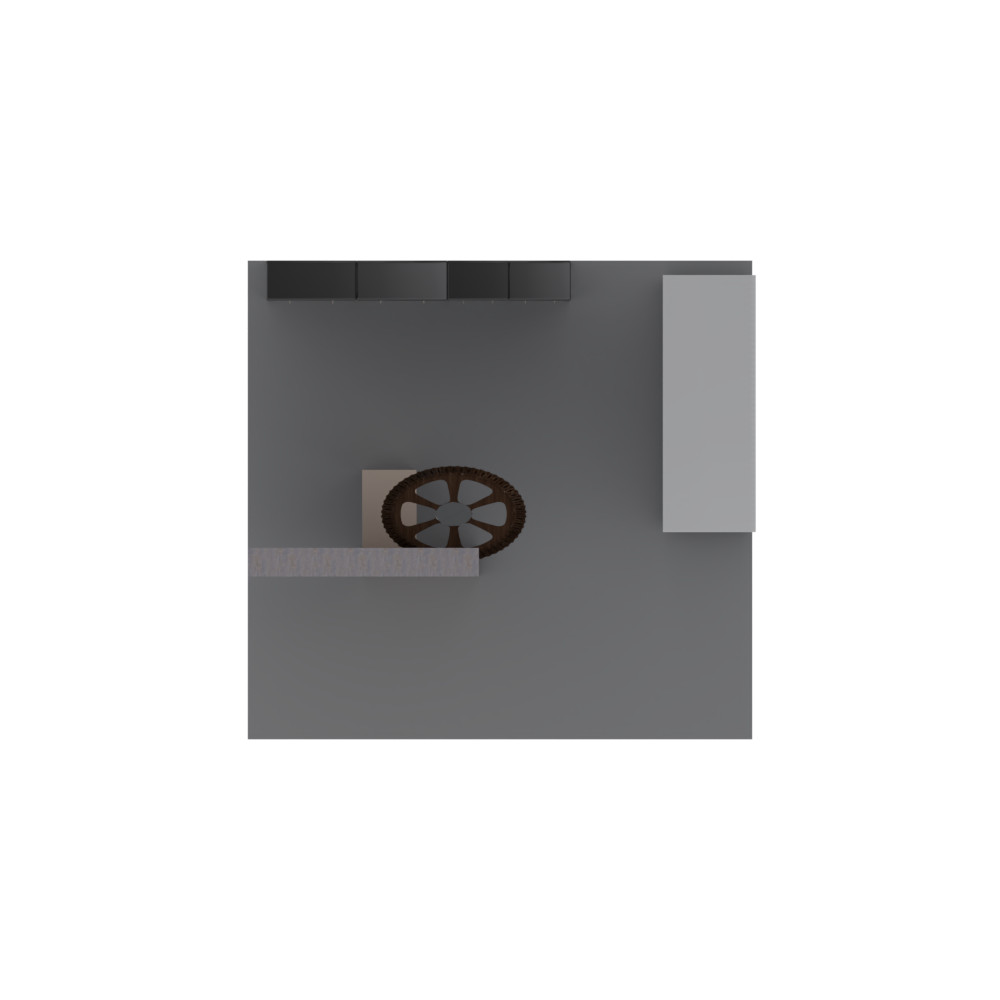}
        \end{overpic}
    \end{subfigure}%
    \begin{subfigure}[b]{0.22\linewidth}
        \centering
        \begin{overpic}[width=\textwidth,  clip]{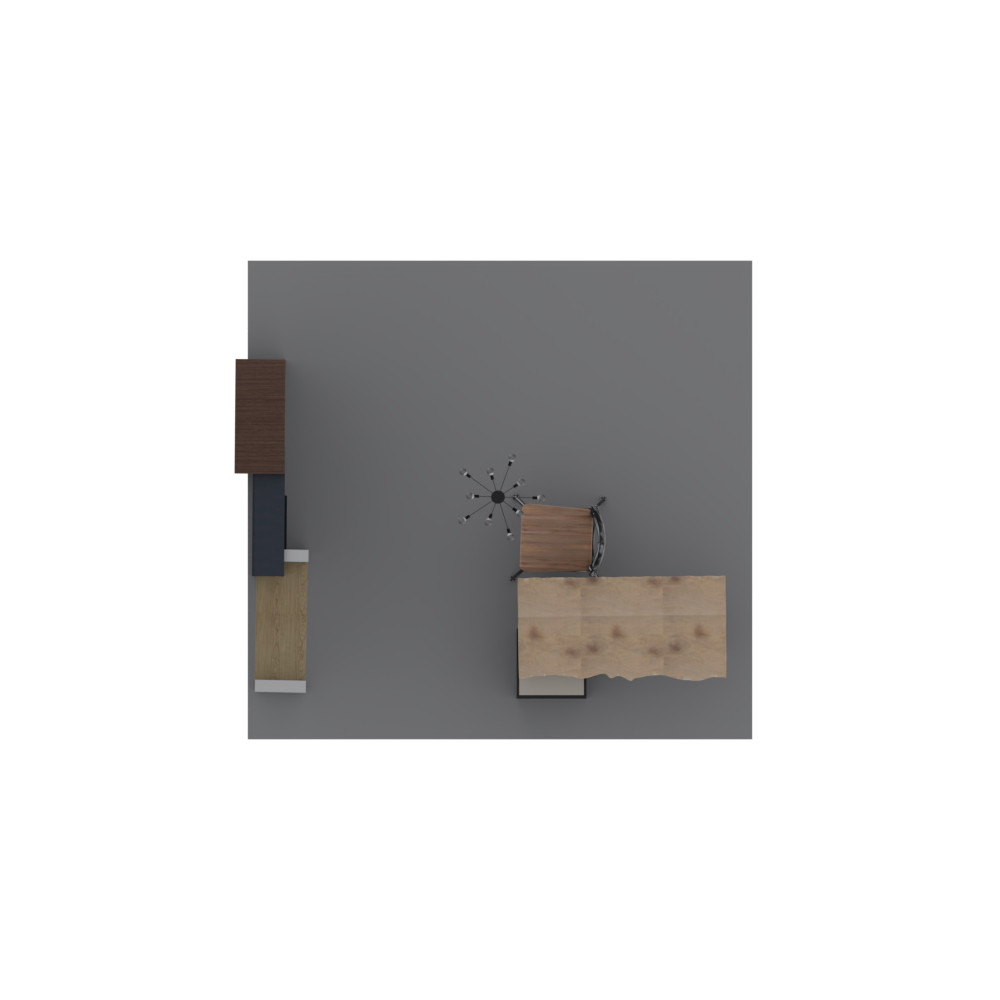}
	    \end{overpic}
    \end{subfigure}%
    \begin{subfigure}[b]{0.22\linewidth}
        \begin{overpic}[width=\textwidth,  clip]{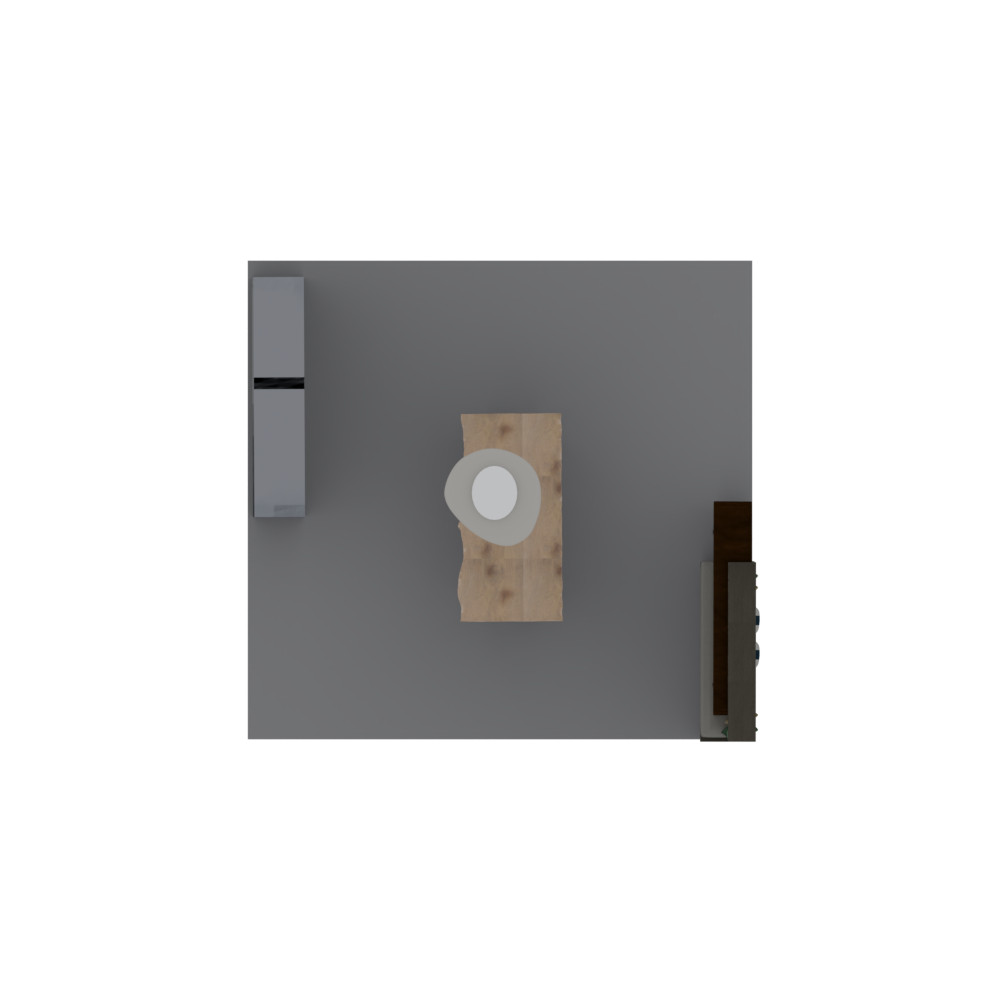}
        \end{overpic}
    \end{subfigure}%
    % \hfill%
    \vskip\baselineskip%
    \vspace{-0.75em}
    \begin{subfigure}[b]{0.22\linewidth}
        \centering
	    \includegraphics[width=\textwidth, clip]{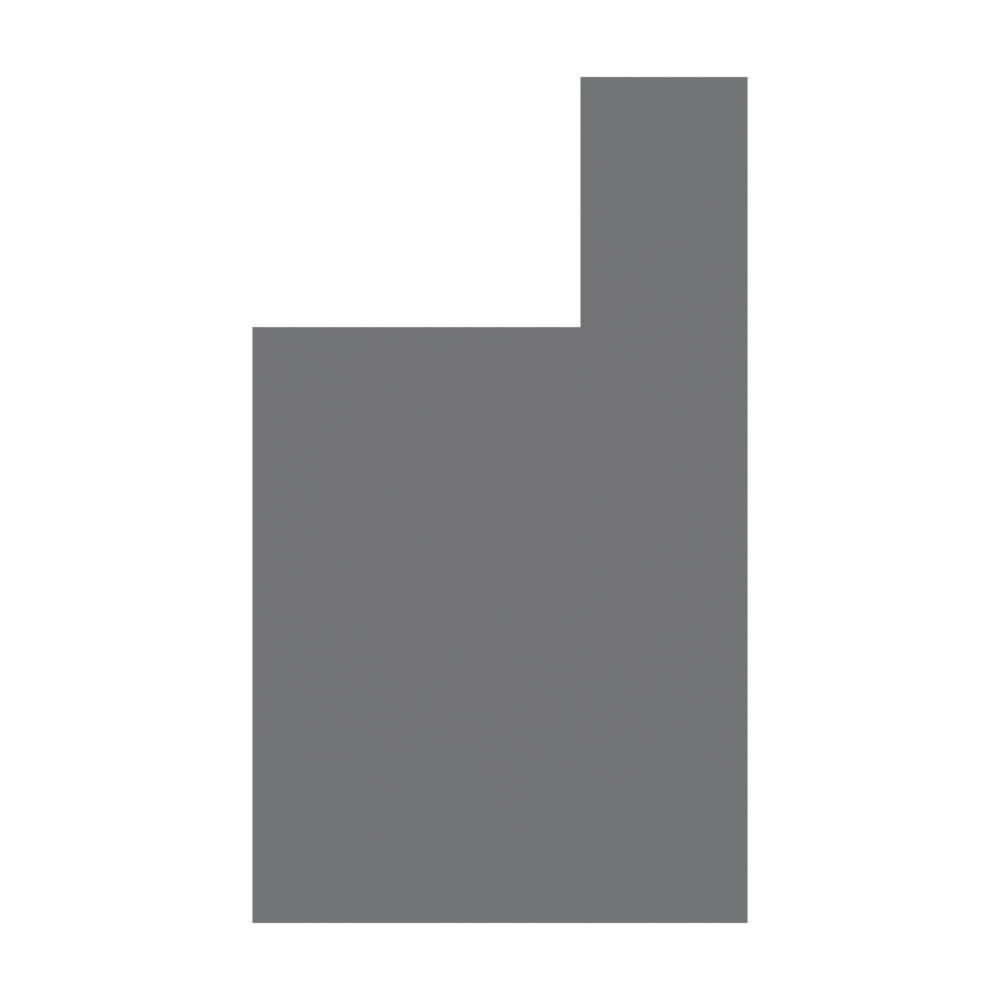}
    \end{subfigure}%
    \begin{subfigure}[b]{0.22\linewidth}
        \centering
        \begin{overpic}[width=\textwidth,  clip]{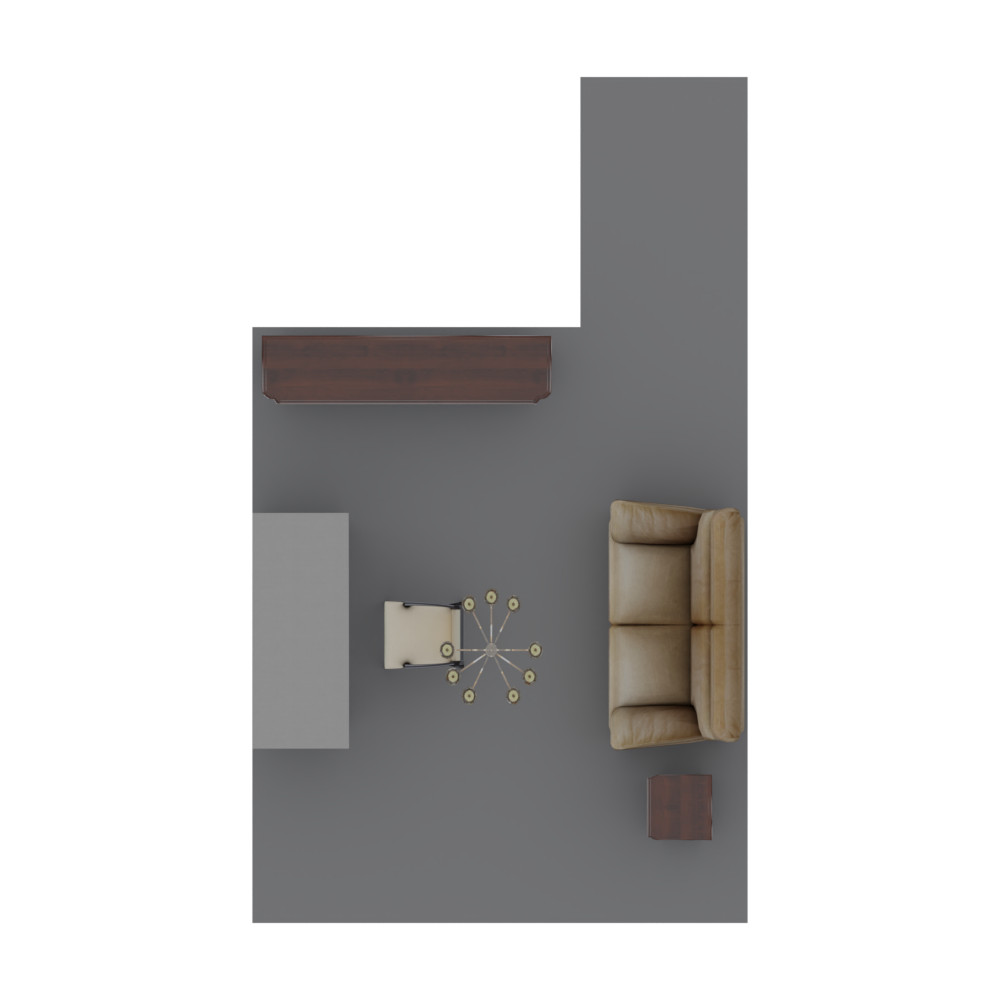}
        \end{overpic}
    \end{subfigure}%
    \begin{subfigure}[b]{0.22\linewidth}
        \centering
        \begin{overpic}[width=\textwidth,  clip]{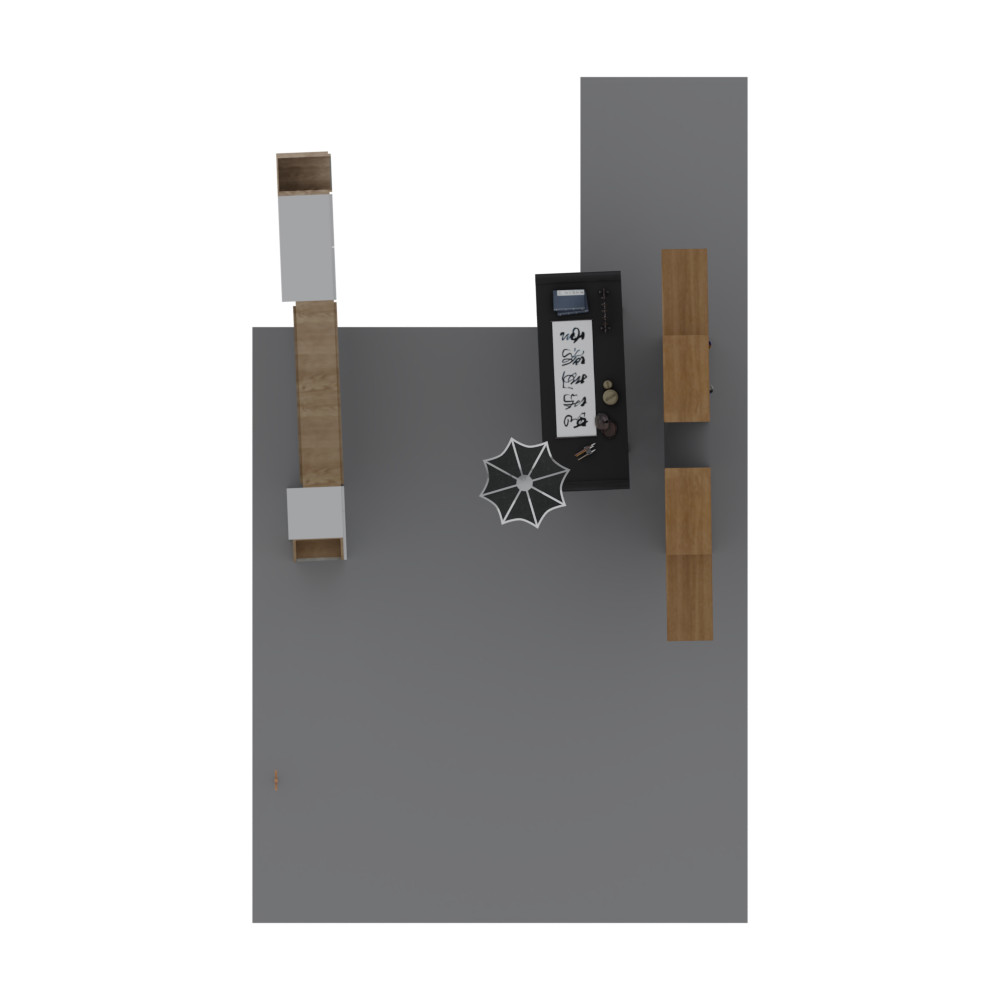}
	    \end{overpic}
    \end{subfigure}%
    \begin{subfigure}[b]{0.22\linewidth}
        \begin{overpic}[width=\textwidth, clip]{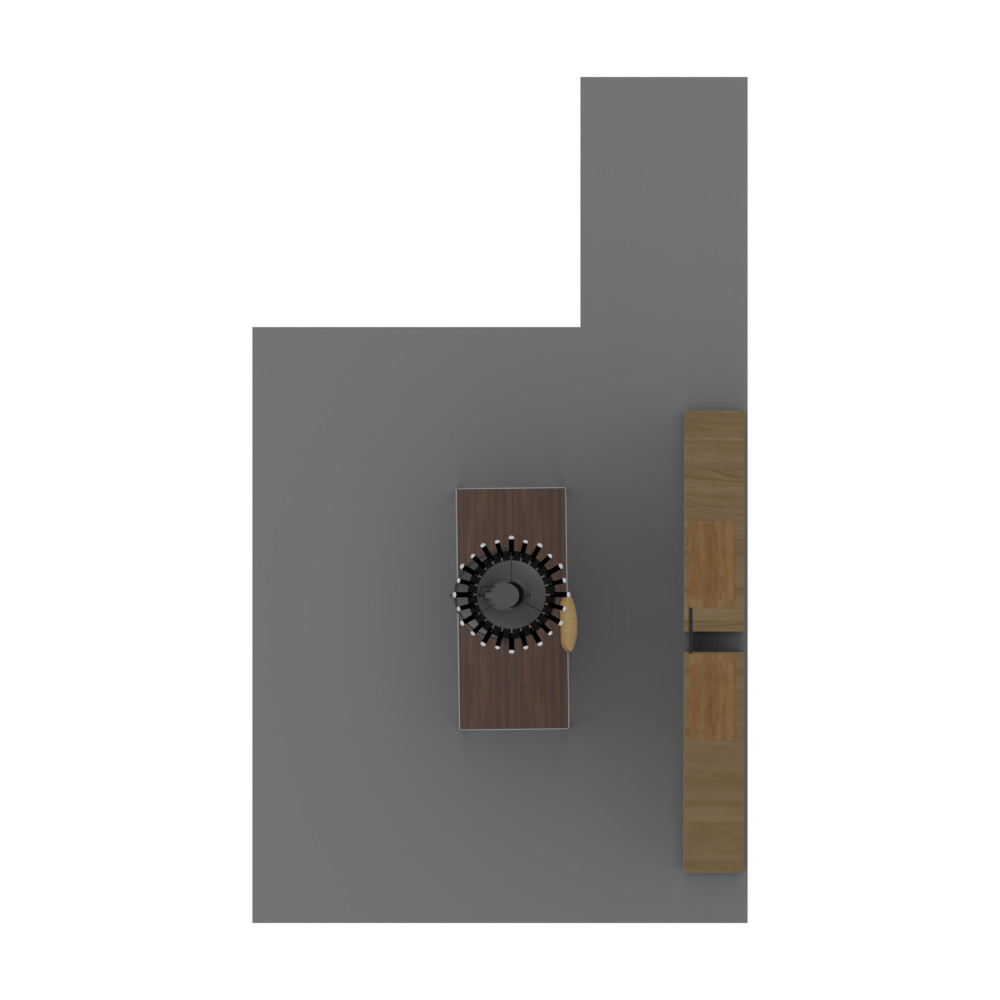}
        \end{overpic}
    \end{subfigure}%
    
    \caption{\textbf{Scene generation from scratch:} We compare generated scenes from GT, ATISS, and our model on \lib class.}
    \label{fig:addl_lib}
\vspace{-0.75em}
\end{figure*}

%%%%%%%%%%%%%%%%%%%%%%%%%%%%%%%%%%%%%%%%%%%%%%%%%%%%%%%%%%%%%%%%%%
%%%%%%%%%%%%%%%%%%%%%%%%%%%%%%%%%%%%%%%%%%%%%%%%%%%%%%%%%%%%%%%%%%%5
%%%%%%%%%%%%%%%%%%%%%%%%%%%%%%%%%%%%%%%%%%%%%%%%%%%%%%%%%%%%%%%%%%%%5
%% BEDROOMS

\begin{figure*}[t!]

    \centering
    \vspace{-1.5em}
    % \hfill
    \begin{subfigure}[b]{0.22\linewidth}
        \centering
	    \small Boundary
    \end{subfigure}%
    \begin{subfigure}[b]{0.22\linewidth}
        \centering
        \small GT
    \end{subfigure}%
    \begin{subfigure}[b]{0.22\linewidth}
        \centering
        \small ATISS
    \end{subfigure}%
    \begin{subfigure}[b]{0.22\linewidth}
        \centering
        \small Ours
    \end{subfigure}%
    % \hfill%
    \vskip\baselineskip%
    \vspace{-0.75em}
    %%%%%%%%%%%%%%%%%%%%%%%%%%%%%%%%%%%%
    % \hfill
    \begin{subfigure}[b]{0.22\linewidth}
        \centering
	    \includegraphics[width=\textwidth,  clip]{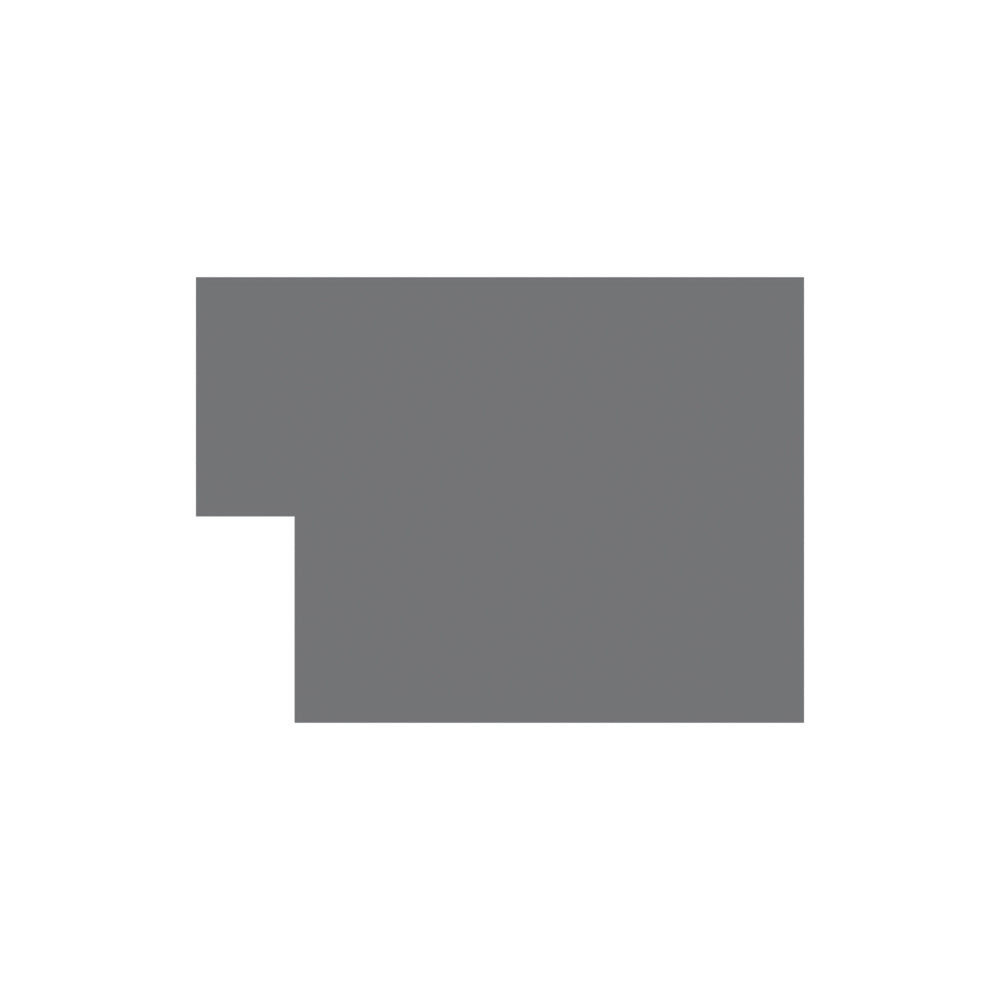}
    \end{subfigure}%
    \begin{subfigure}[b]{0.22\linewidth}
        \centering
            \begin{overpic}[width=\textwidth,  clip]{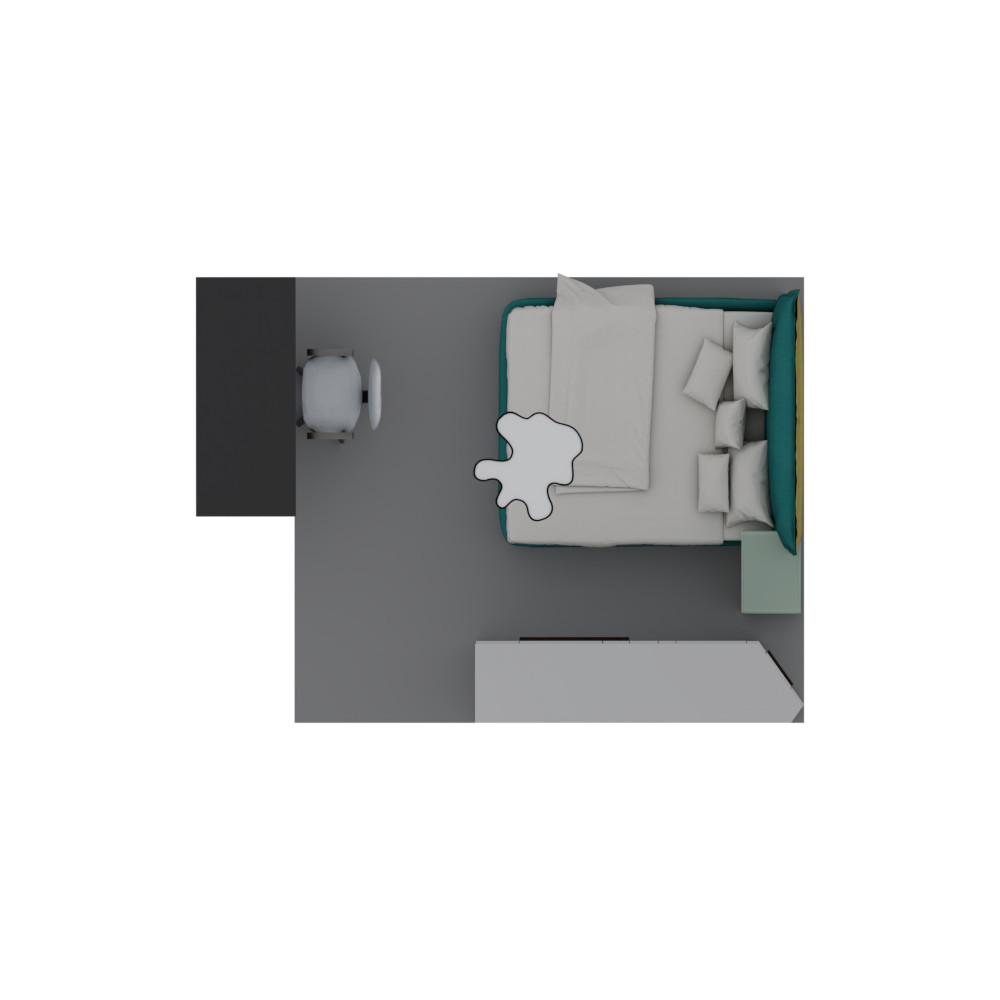}
        \end{overpic}
    \end{subfigure}%
    \begin{subfigure}[b]{0.22\linewidth}
        \centering
        \begin{overpic}[width=\textwidth, clip]{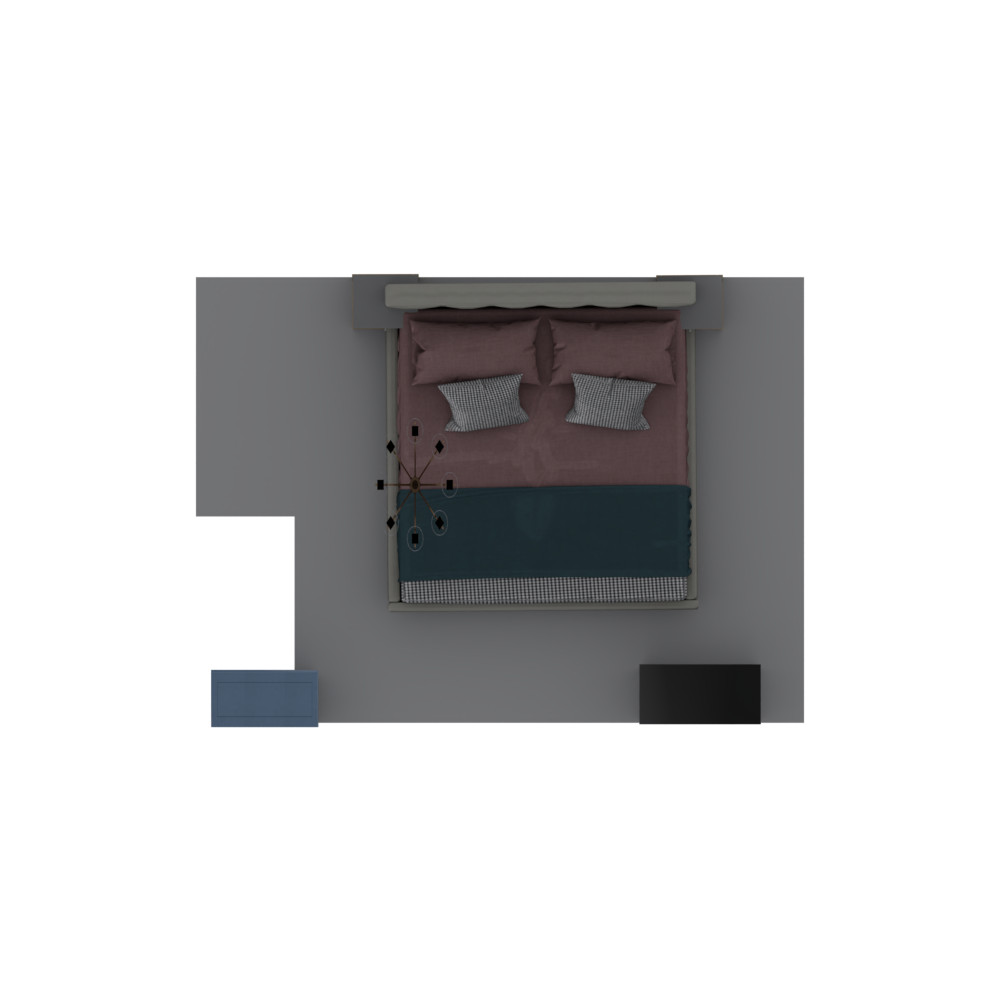}
	    \end{overpic}
    \end{subfigure}%
    \begin{subfigure}[b]{0.22\linewidth}
        \centering
        \begin{overpic}[width=\textwidth, clip]{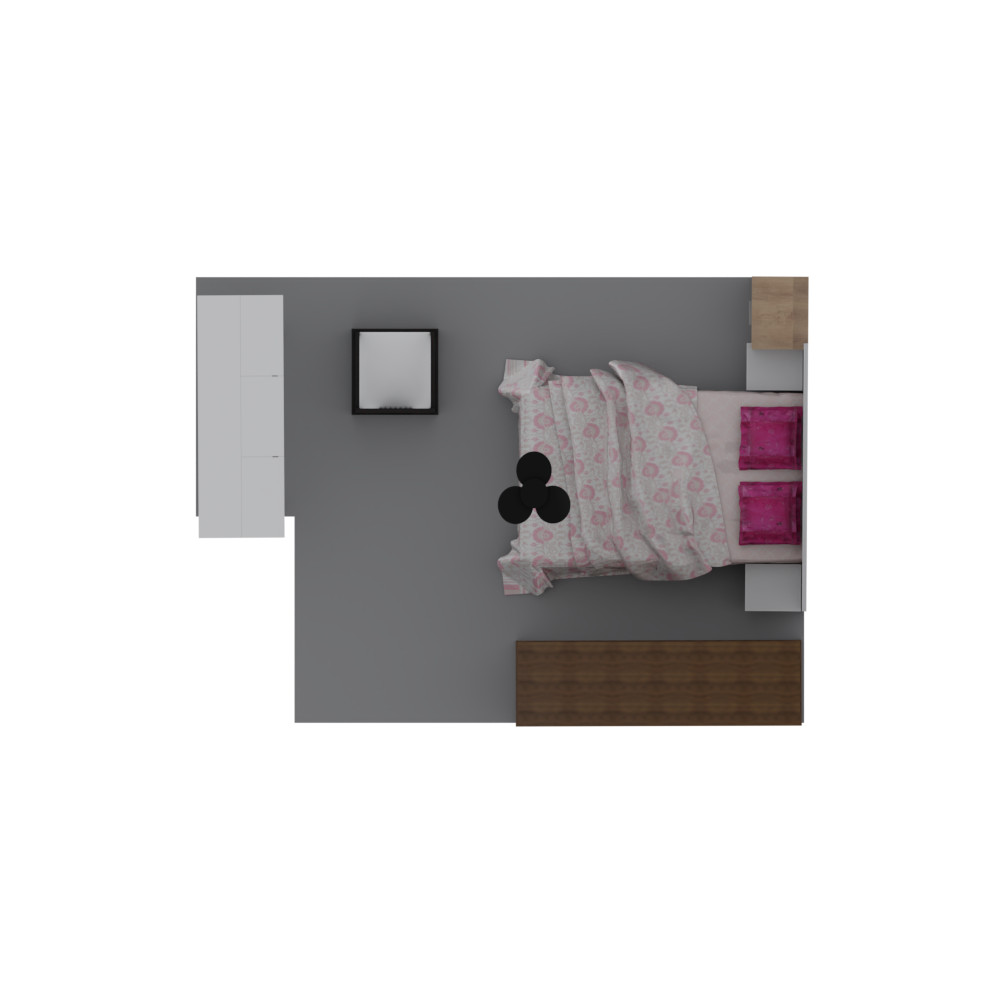}
        \end{overpic}
    \end{subfigure}%
    % \hfill%
    \vskip\baselineskip%
    \vspace{-0.75em}
    % \hfill
    %%%%%%%%%%%%%%%%%%%%%%%%%%%%%%%%%%%%%%%%%%%%%%%%%%%%%%%%%%%%%%%%%%%%%
    \begin{subfigure}[b]{0.22\linewidth}
        \centering
	    \includegraphics[width=\textwidth, clip]{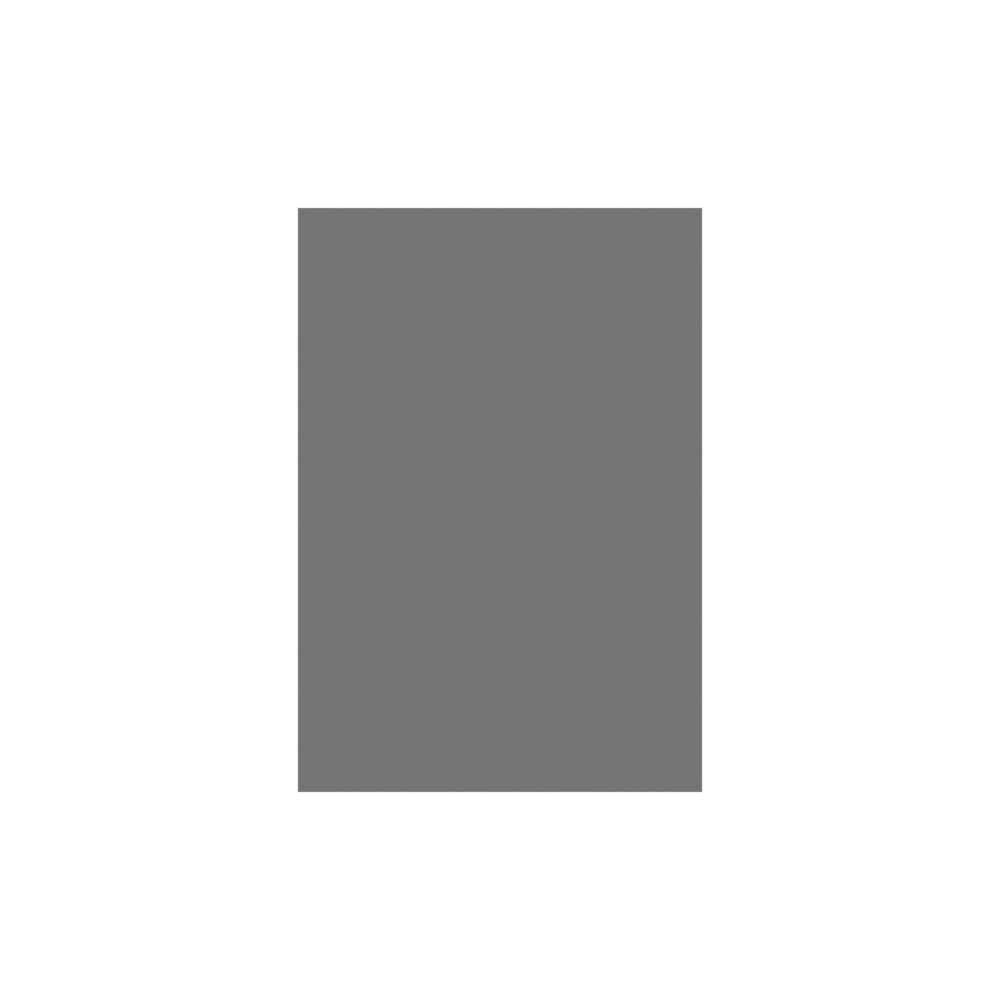}
    \end{subfigure}%
    \begin{subfigure}[b]{0.22\linewidth}
        \centering
        \begin{overpic}[width=\textwidth,  clip]{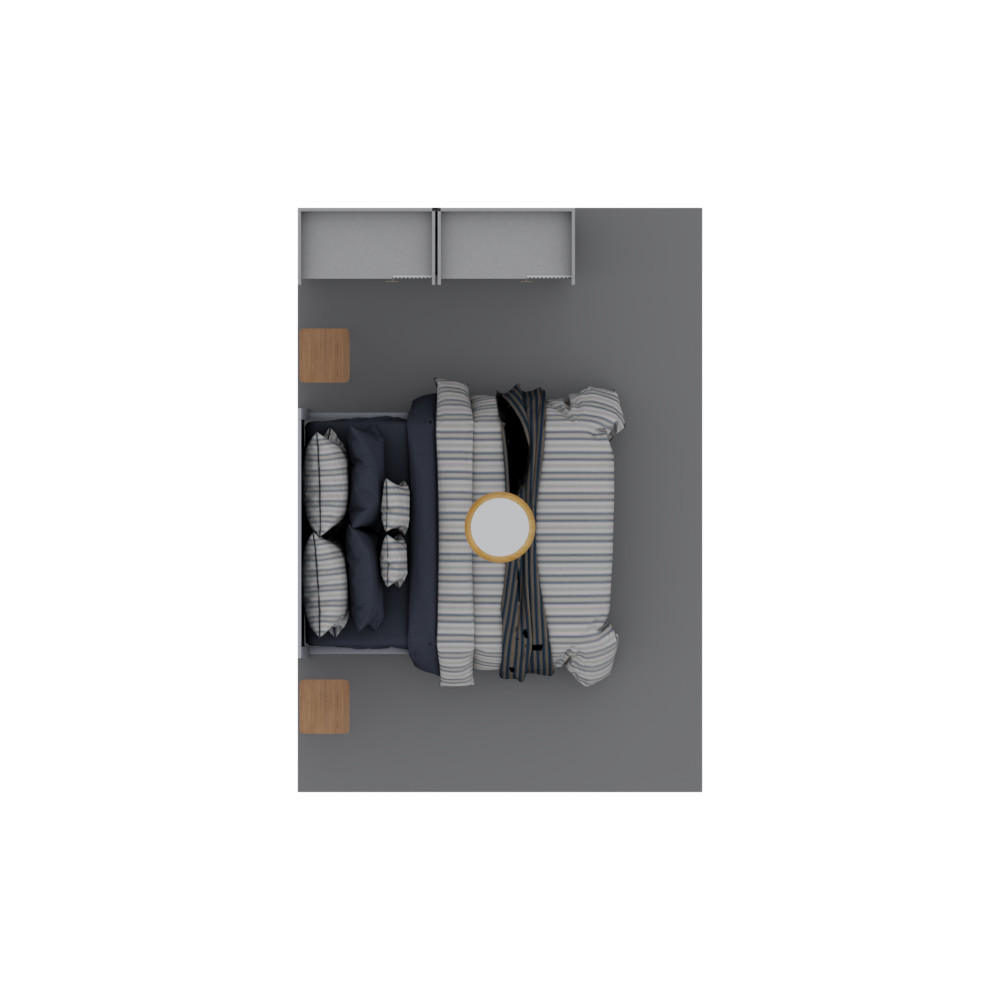}
        \end{overpic}
    \end{subfigure}%
    \begin{subfigure}[b]{0.22\linewidth}
        \centering
        \begin{overpic}[width=\textwidth, clip]{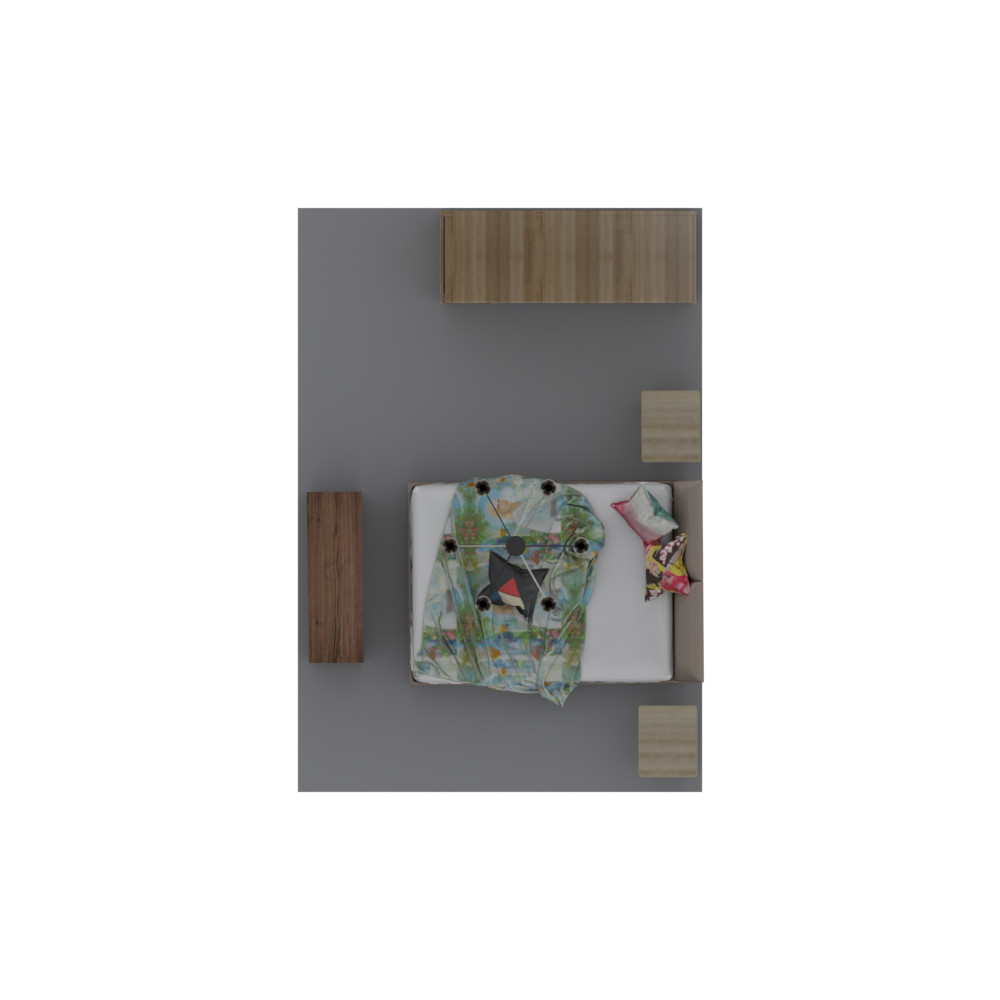}
	    \end{overpic}
    \end{subfigure}%
    \begin{subfigure}[b]{0.22\linewidth}
        \begin{overpic}[width=\textwidth,  clip]{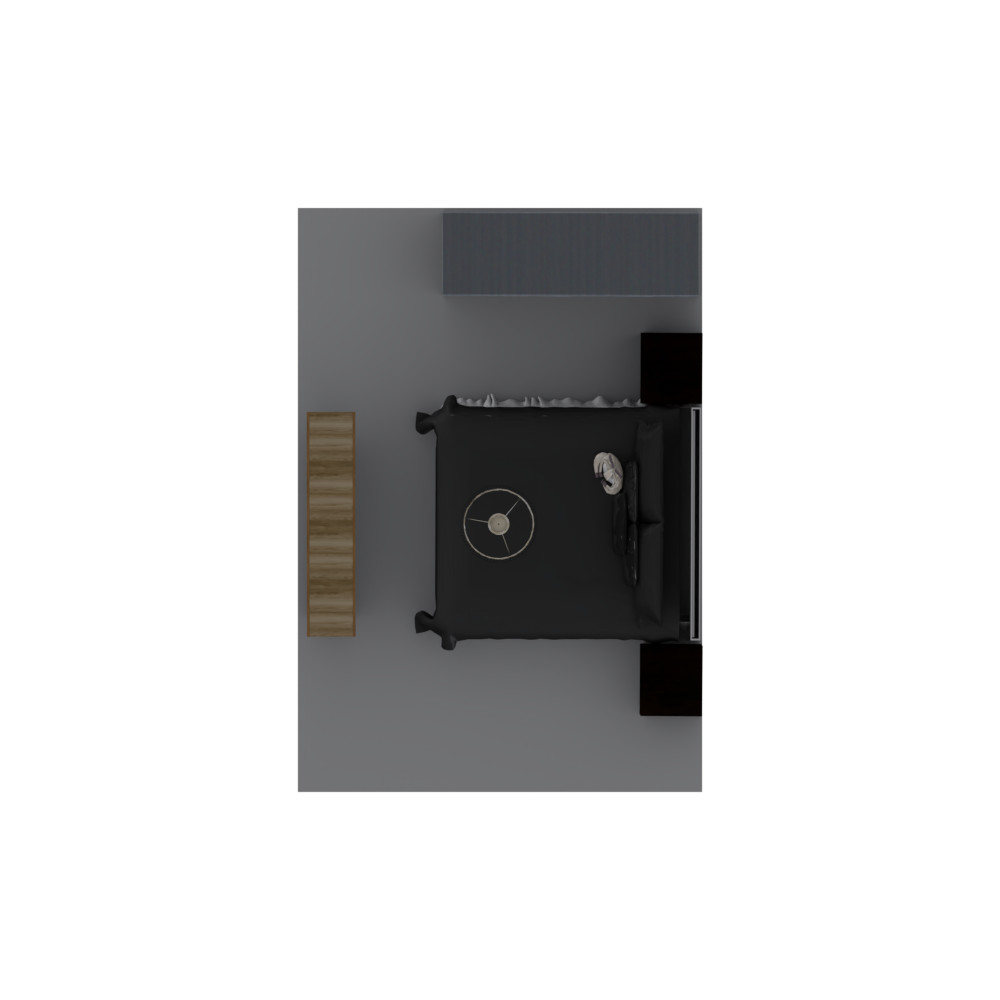}
        \end{overpic}
    \end{subfigure}%
    % \hfill%
    \vskip\baselineskip%
    \vspace{-0.75em}
    % \hfill
    %%%%%%%%%%%%%%%%%%%%%%%%%%%%%%%%%%%%%%%%%%%%%%%%%%%%%%%%%%%%%%%%%%%%%%%%%%%%%%%%%%%%%%%%%%%%%%%%%%%5
    \begin{subfigure}[b]{0.22\linewidth}
        \centering
	    \includegraphics[width=\textwidth, clip]{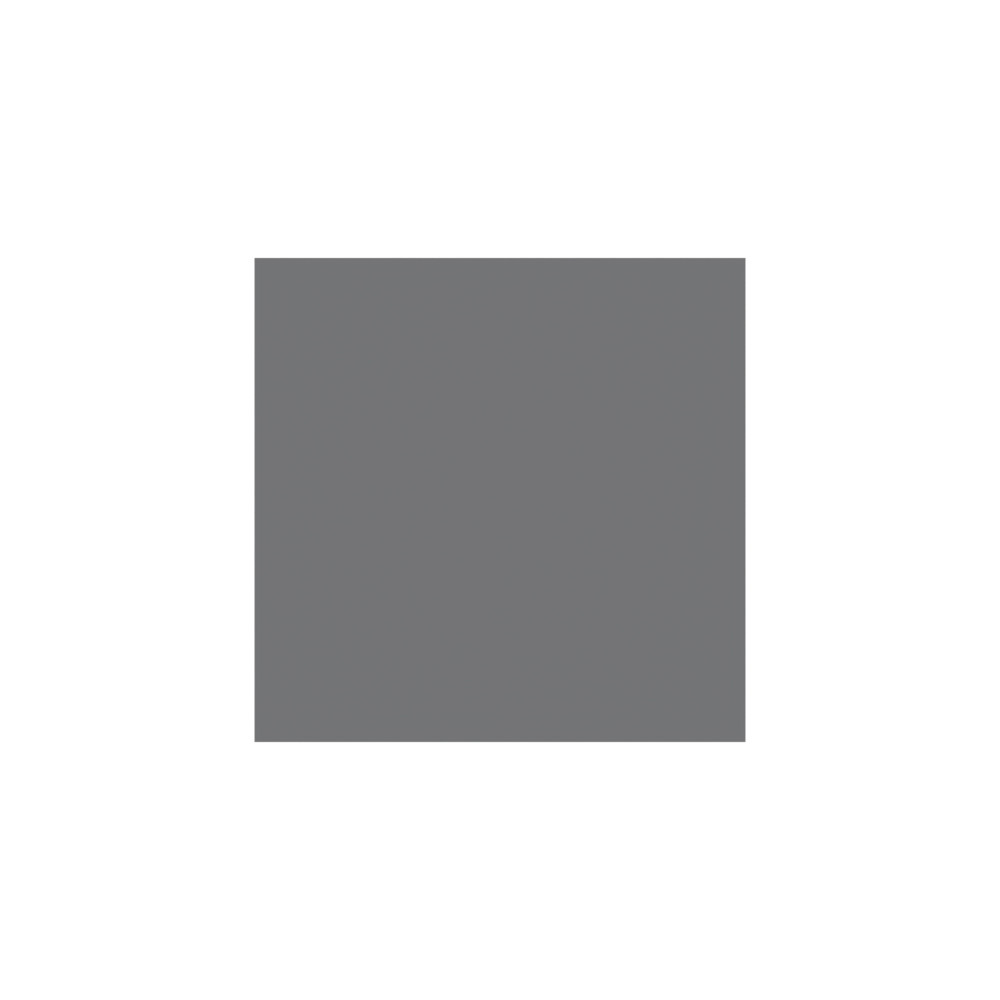}
    \end{subfigure}%
    \begin{subfigure}[b]{0.22\linewidth}
        \centering
        \begin{overpic}[width=\textwidth,  clip]{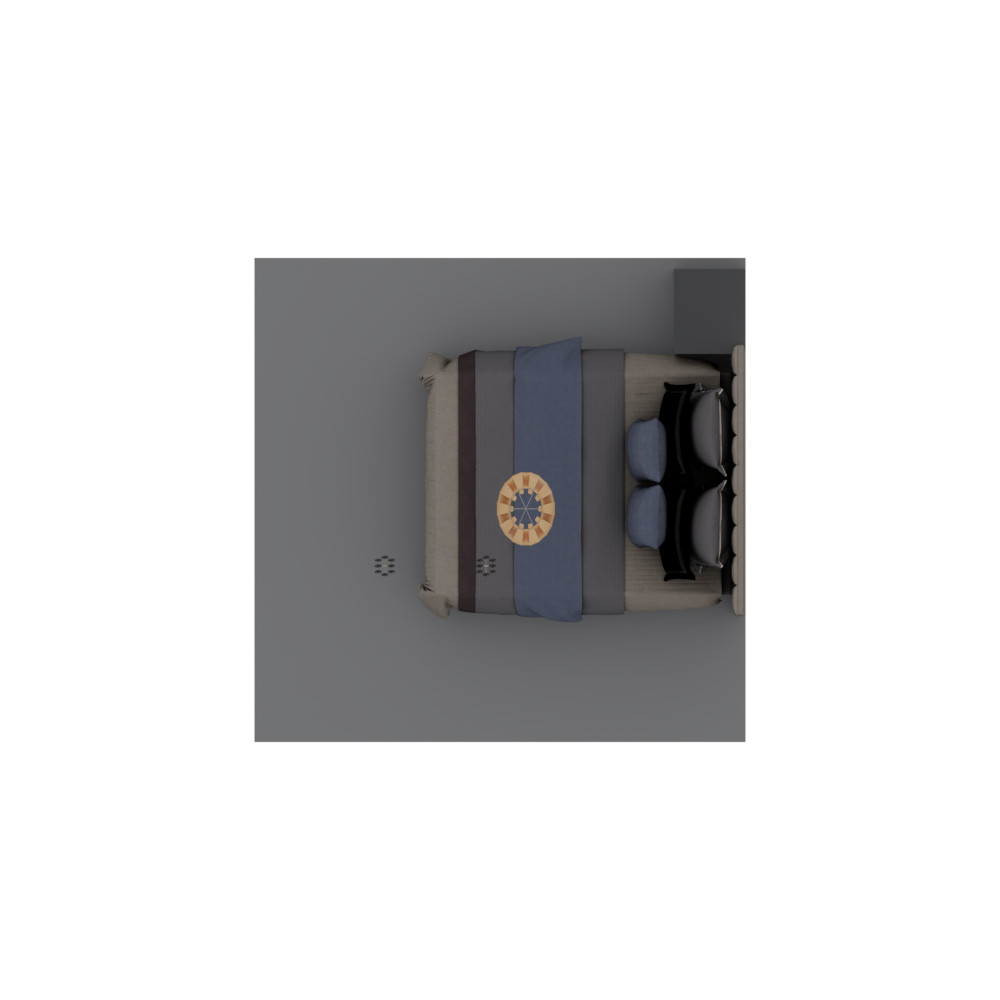}
        \end{overpic}
    \end{subfigure}%
    \begin{subfigure}[b]{0.22\linewidth}
        \centering
        \begin{overpic}[width=\textwidth, clip]{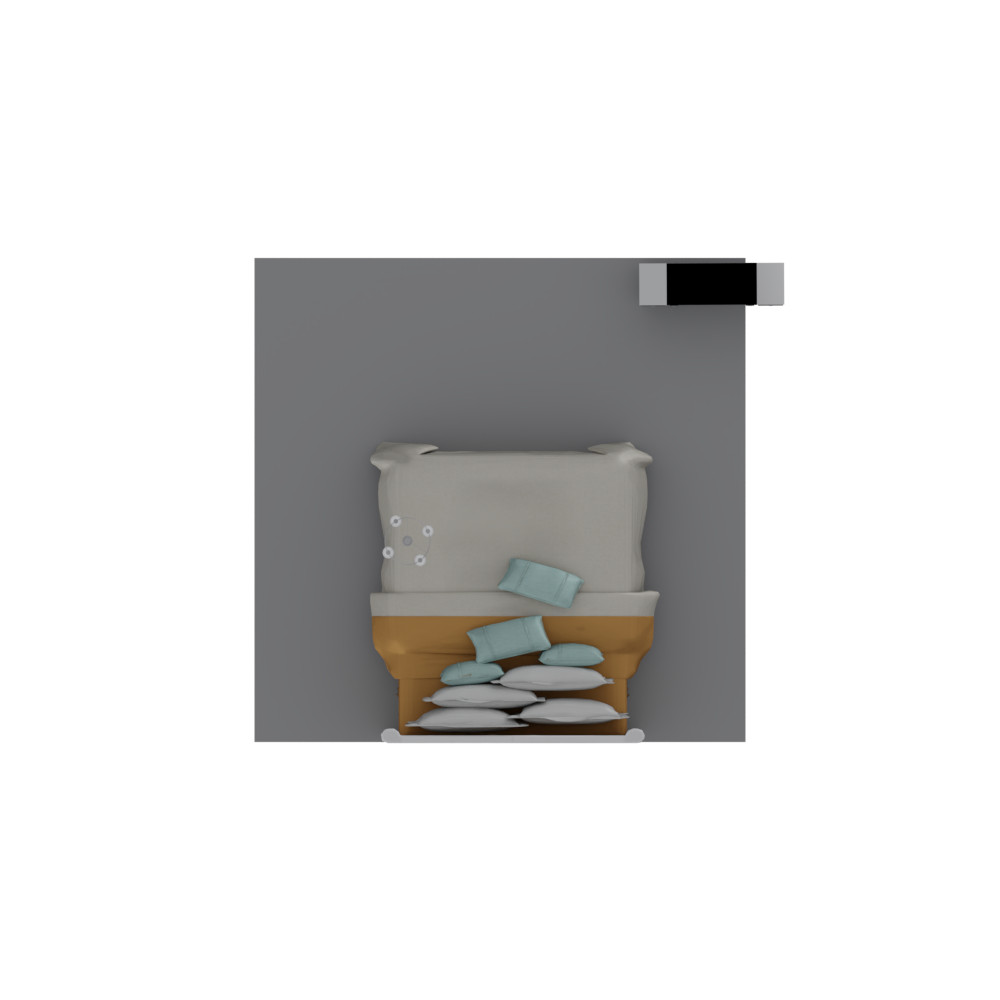}
	    \end{overpic}
    \end{subfigure}%
    \begin{subfigure}[b]{0.22\linewidth}
        \begin{overpic}[width=\textwidth,  clip]{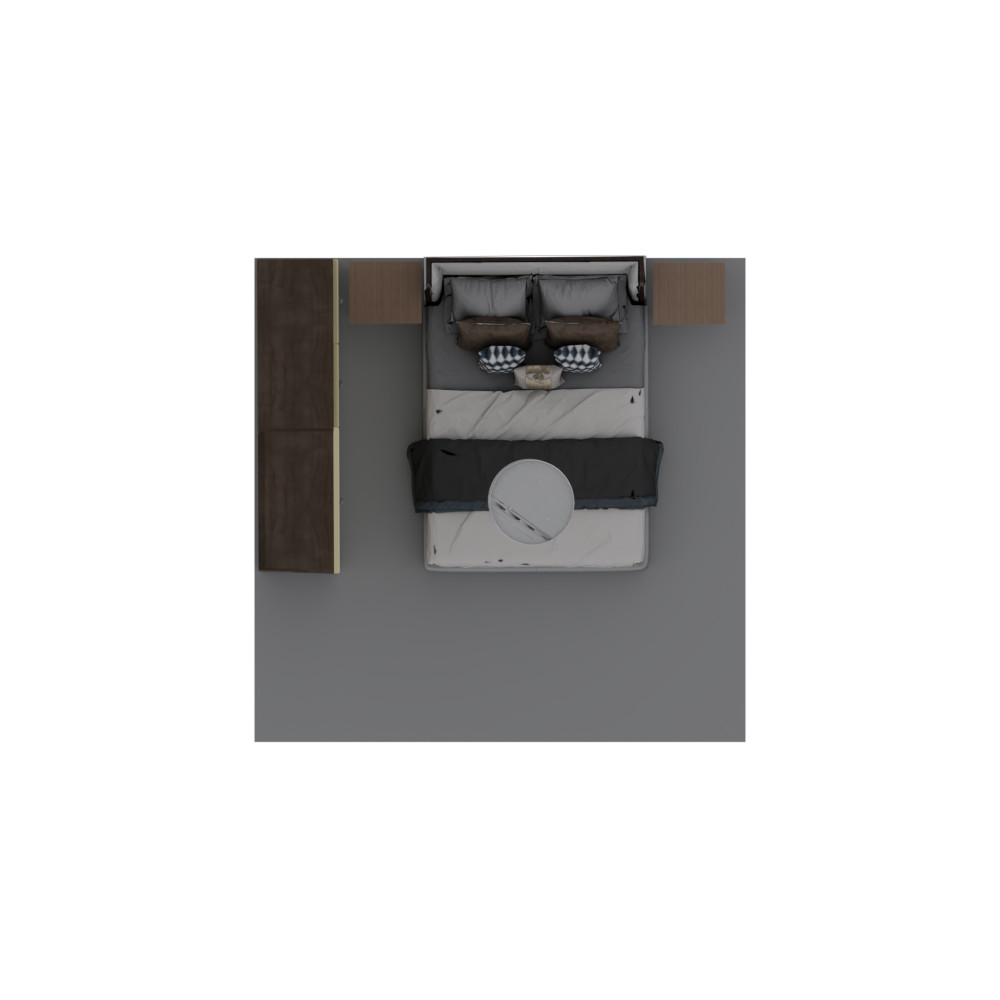}
        \end{overpic}
    \end{subfigure}%
    % \hfill%
    \vskip\baselineskip%
    \vspace{-0.75em}
    \begin{subfigure}[b]{0.22\linewidth}
        \centering
	    \includegraphics[width=\textwidth, clip]{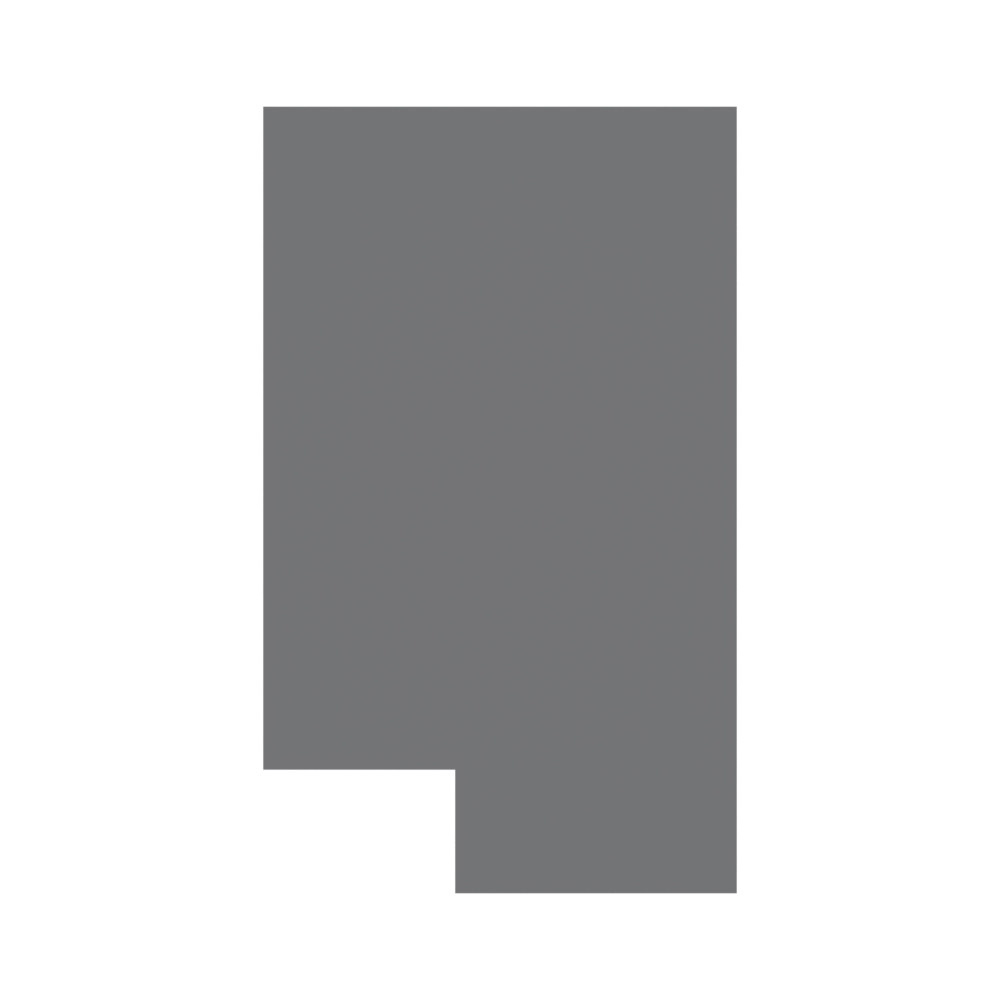}
    \end{subfigure}%
    \begin{subfigure}[b]{0.22\linewidth}
        \centering
        \begin{overpic}[width=\textwidth,  clip]{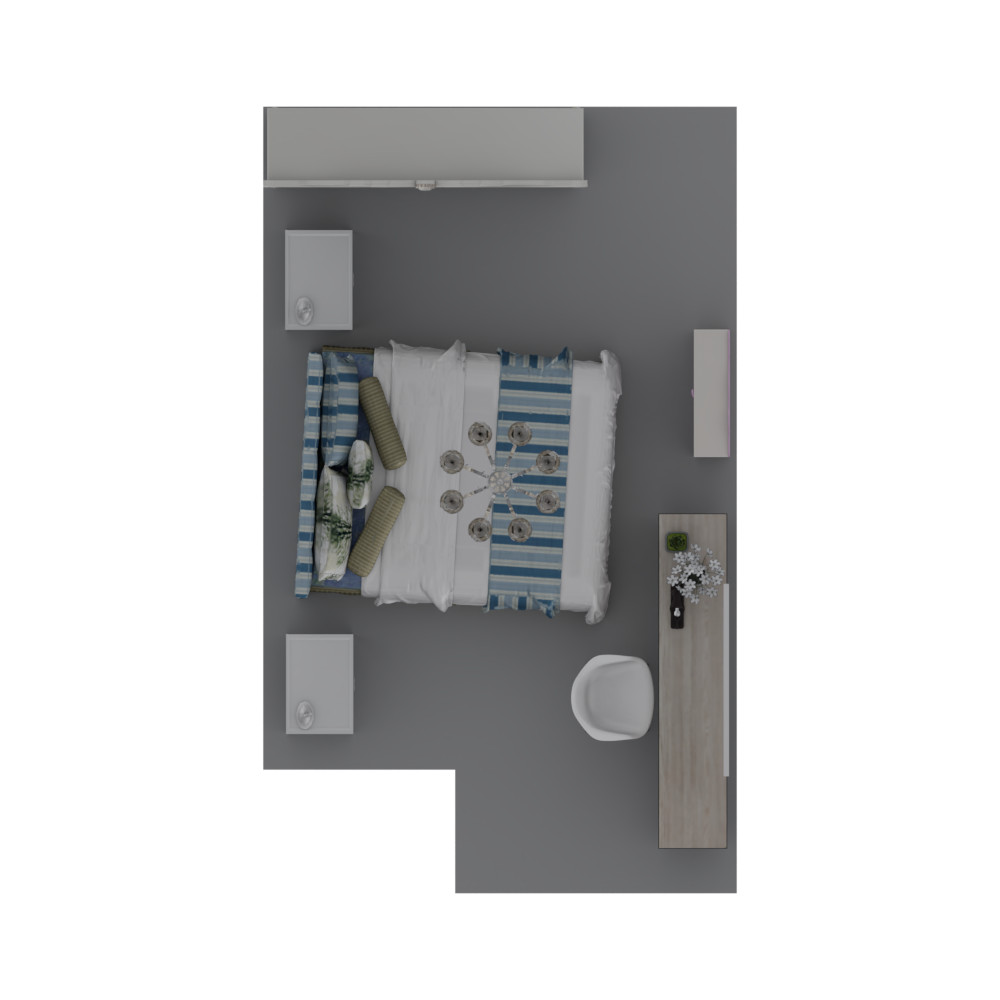}
        \end{overpic}
    \end{subfigure}%
    \begin{subfigure}[b]{0.22\linewidth}
        \centering
        \begin{overpic}[width=\textwidth,  clip]{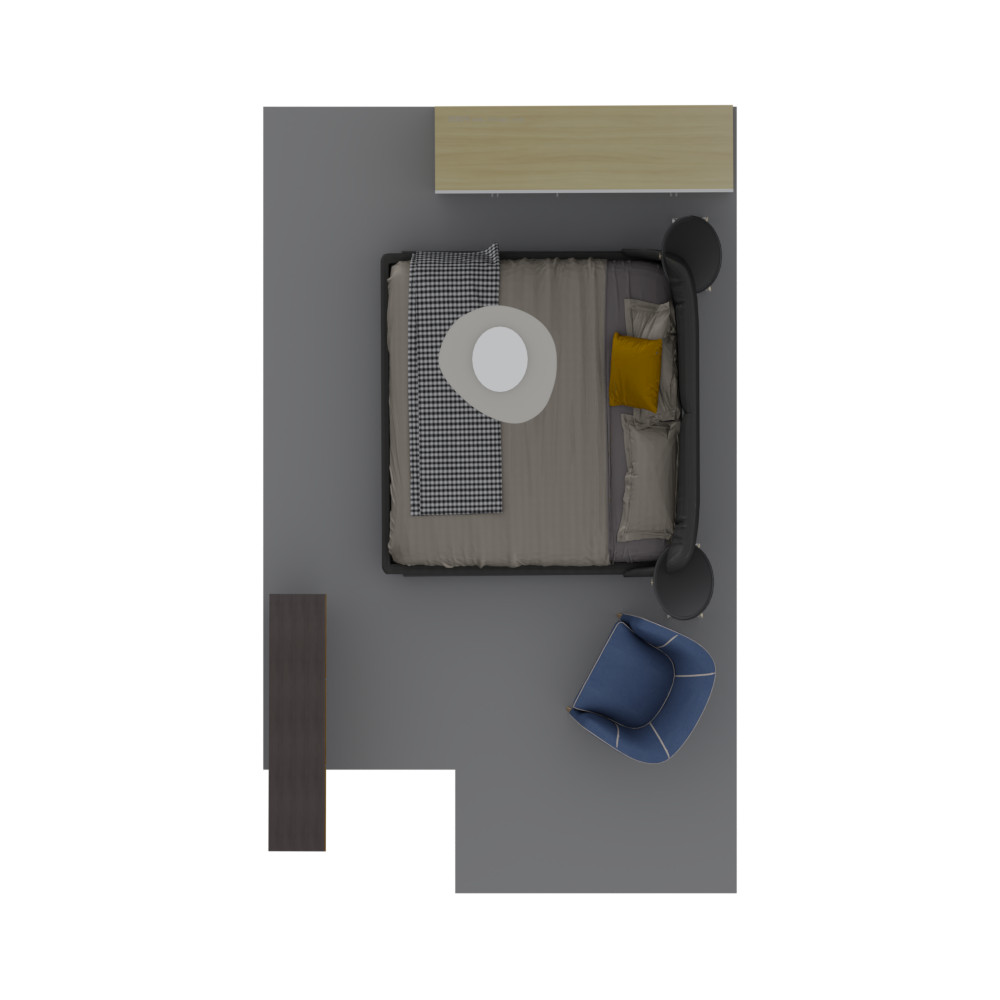}
	    \end{overpic}
    \end{subfigure}%
    \begin{subfigure}[b]{0.22\linewidth}
        \begin{overpic}[width=\textwidth,  clip]{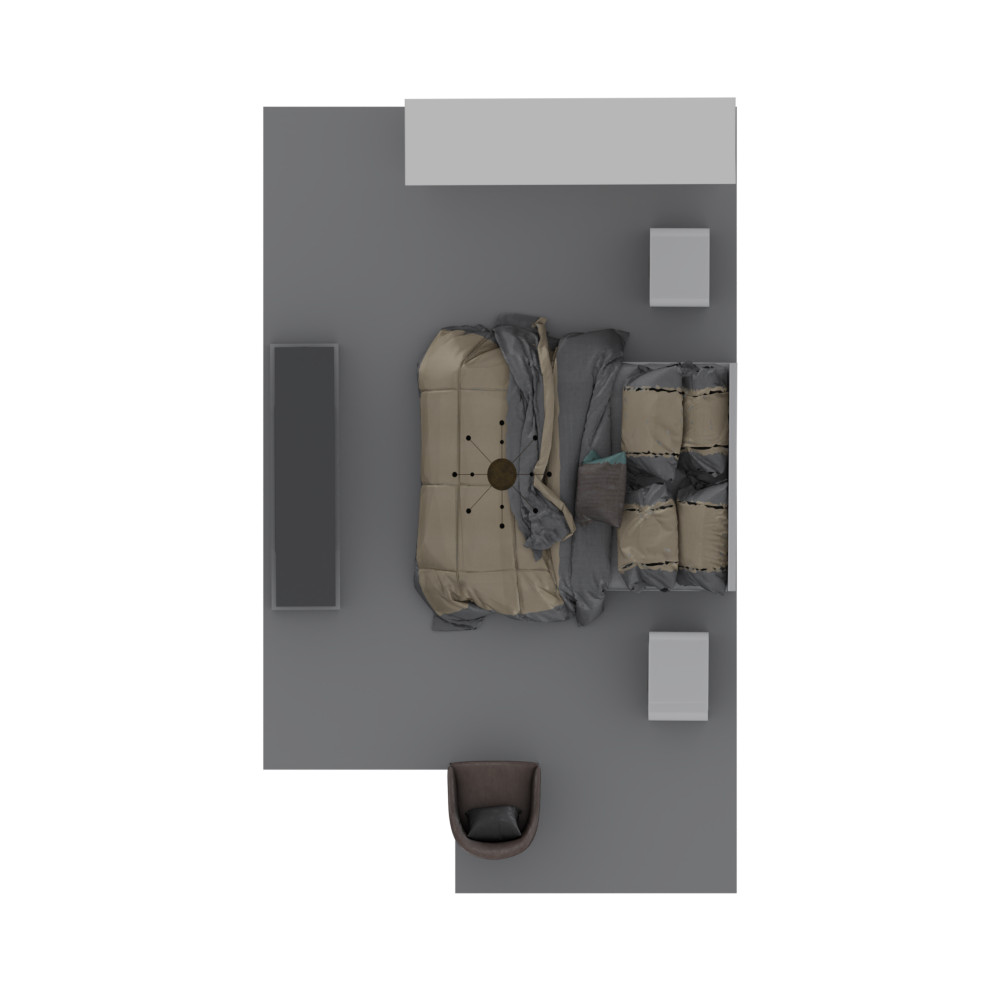}
        \end{overpic}
    \end{subfigure}%
    % \hfill%
    \vskip\baselineskip%
    \vspace{-0.75em}
    \begin{subfigure}[b]{0.22\linewidth}
        \centering
	    \includegraphics[width=\textwidth, clip]{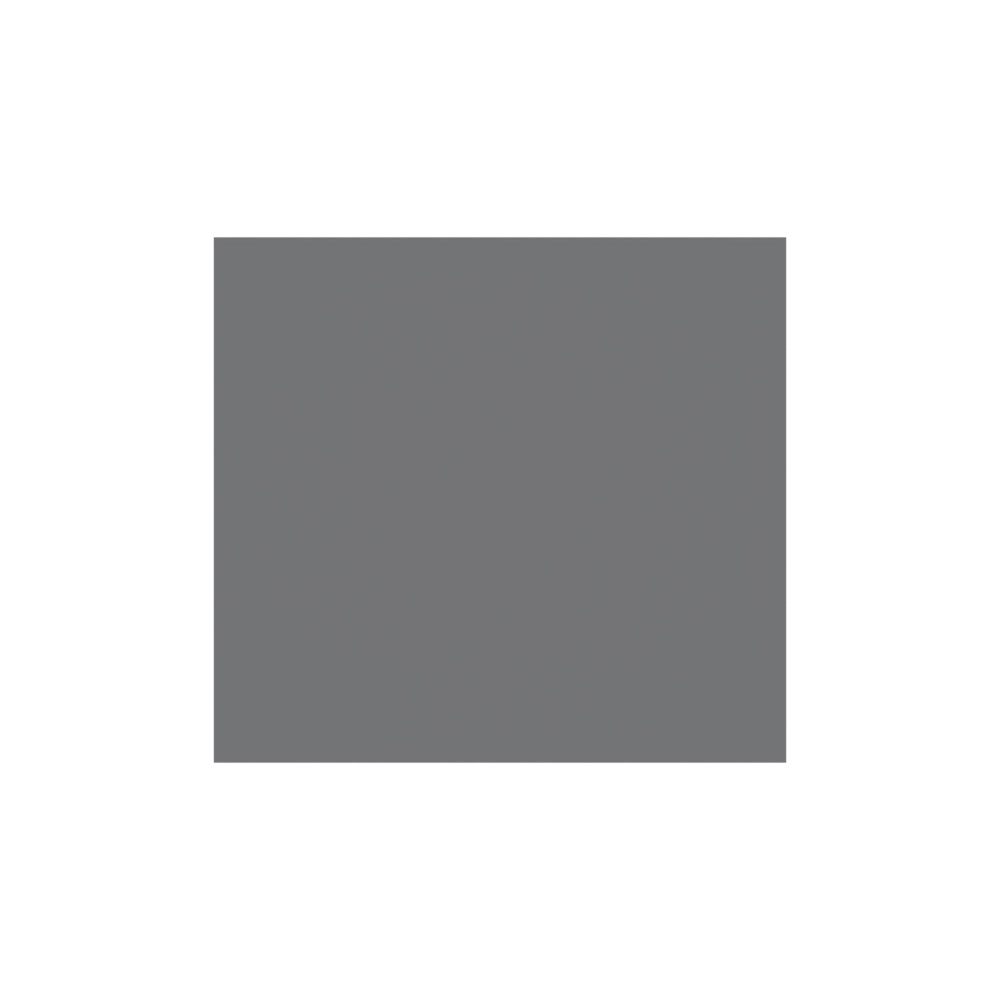}
    \end{subfigure}%
    \begin{subfigure}[b]{0.22\linewidth}
        \centering
        \begin{overpic}[width=\textwidth,  clip]{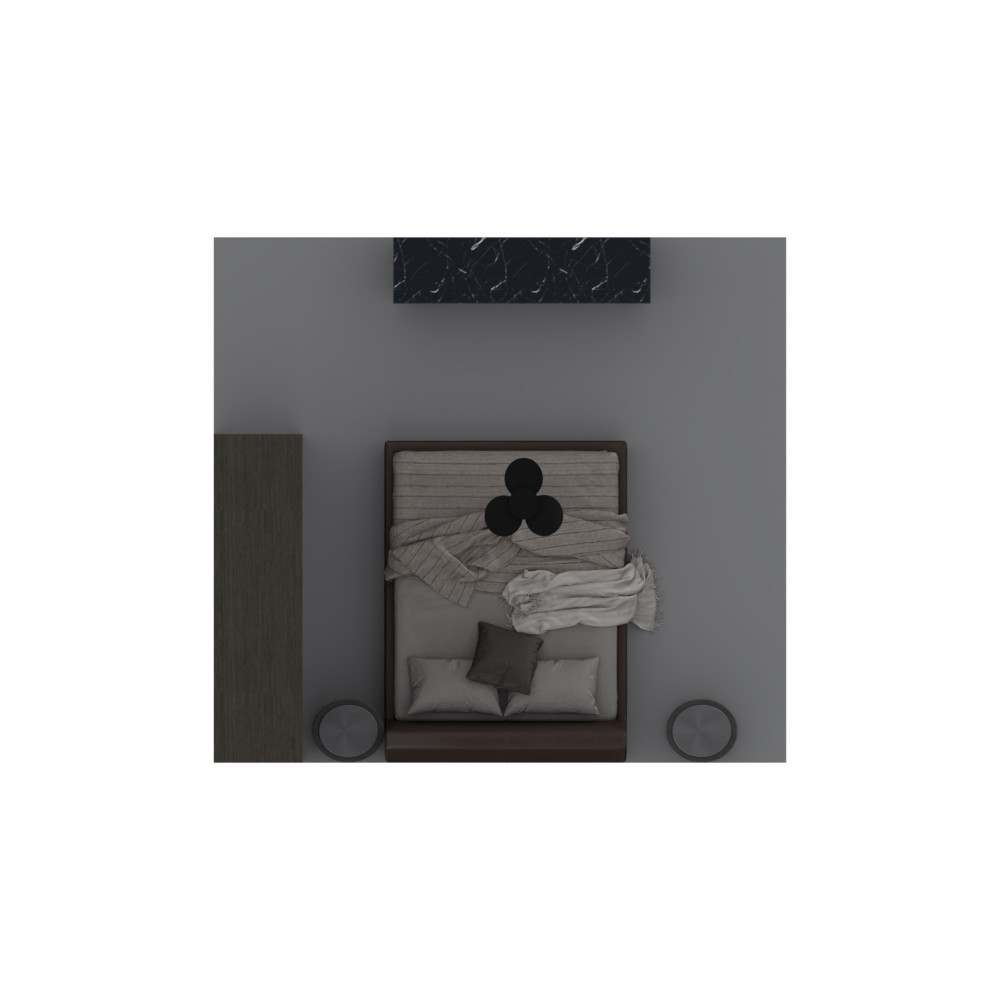}
        \end{overpic}
    \end{subfigure}%
    \begin{subfigure}[b]{0.22\linewidth}
        \centering
        \begin{overpic}[width=\textwidth,  clip]{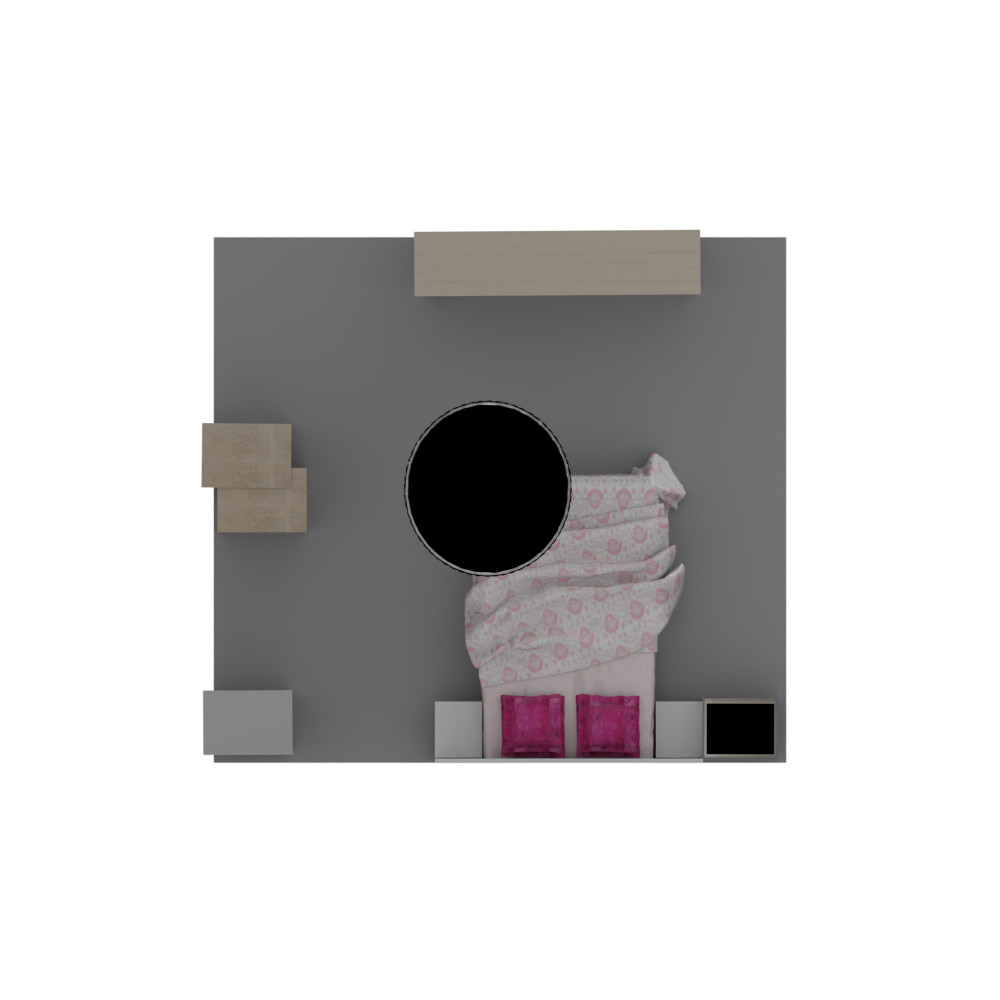}
	    \end{overpic}
    \end{subfigure}%
    \begin{subfigure}[b]{0.22\linewidth}
        \begin{overpic}[width=\textwidth,  clip]{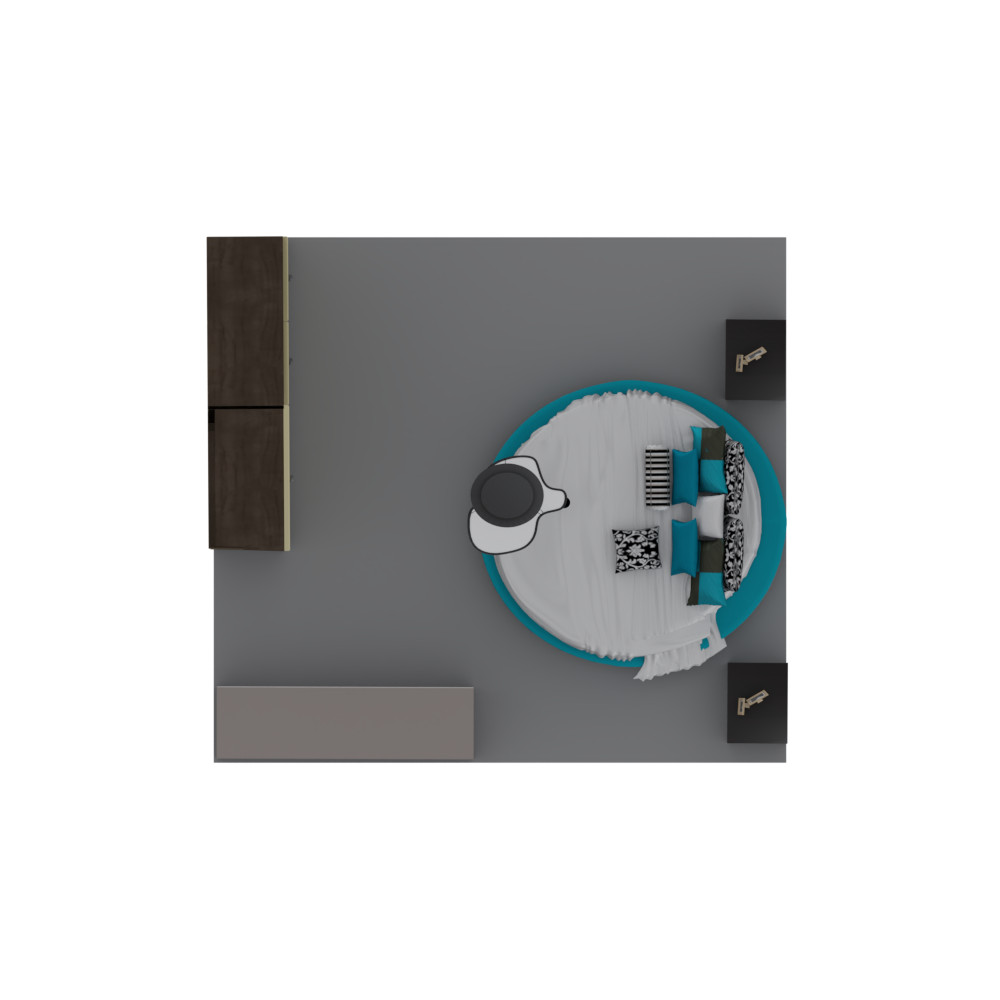}
        \end{overpic}
    \end{subfigure}%
    % \hfill%
    \vskip\baselineskip%
    \vspace{-0.75em}
    \begin{subfigure}[b]{0.22\linewidth}
        \centering
	    \includegraphics[width=\textwidth, clip]{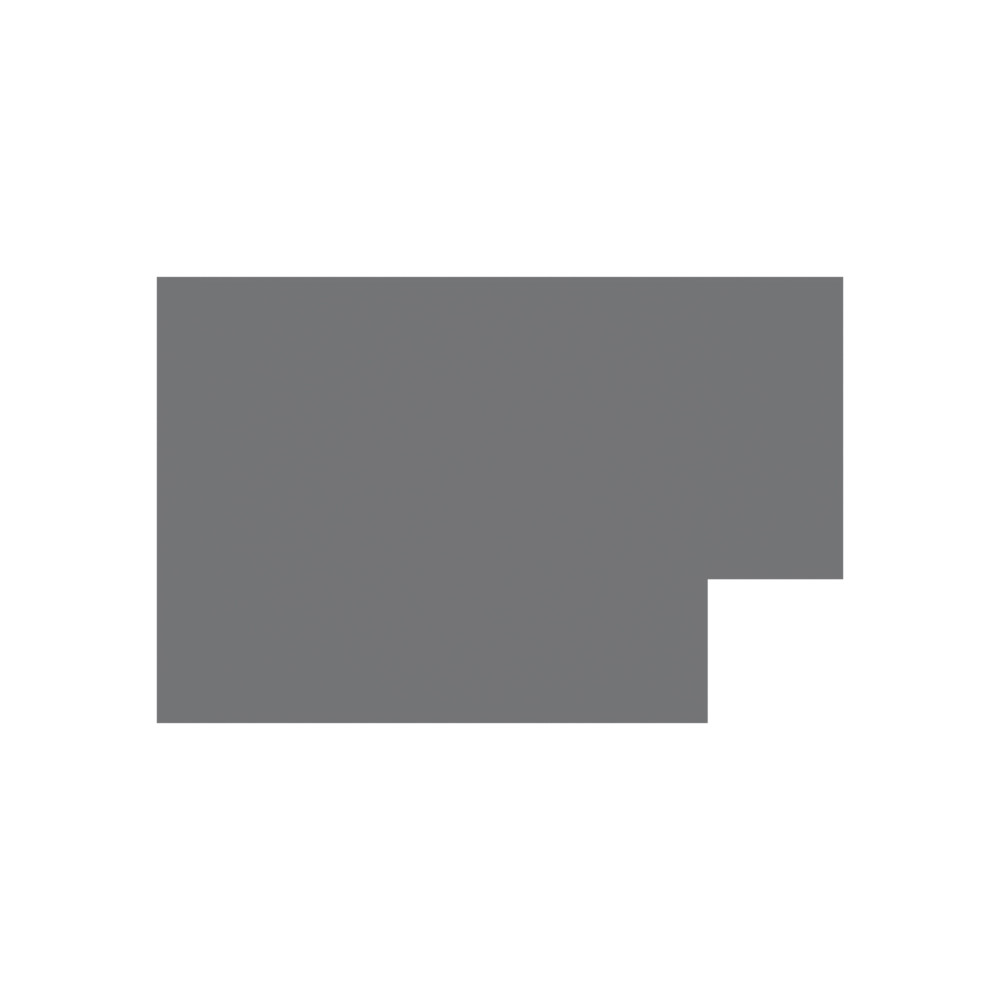}
    \end{subfigure}%
    \begin{subfigure}[b]{0.22\linewidth}
        \centering
        \begin{overpic}[width=\textwidth,  clip]{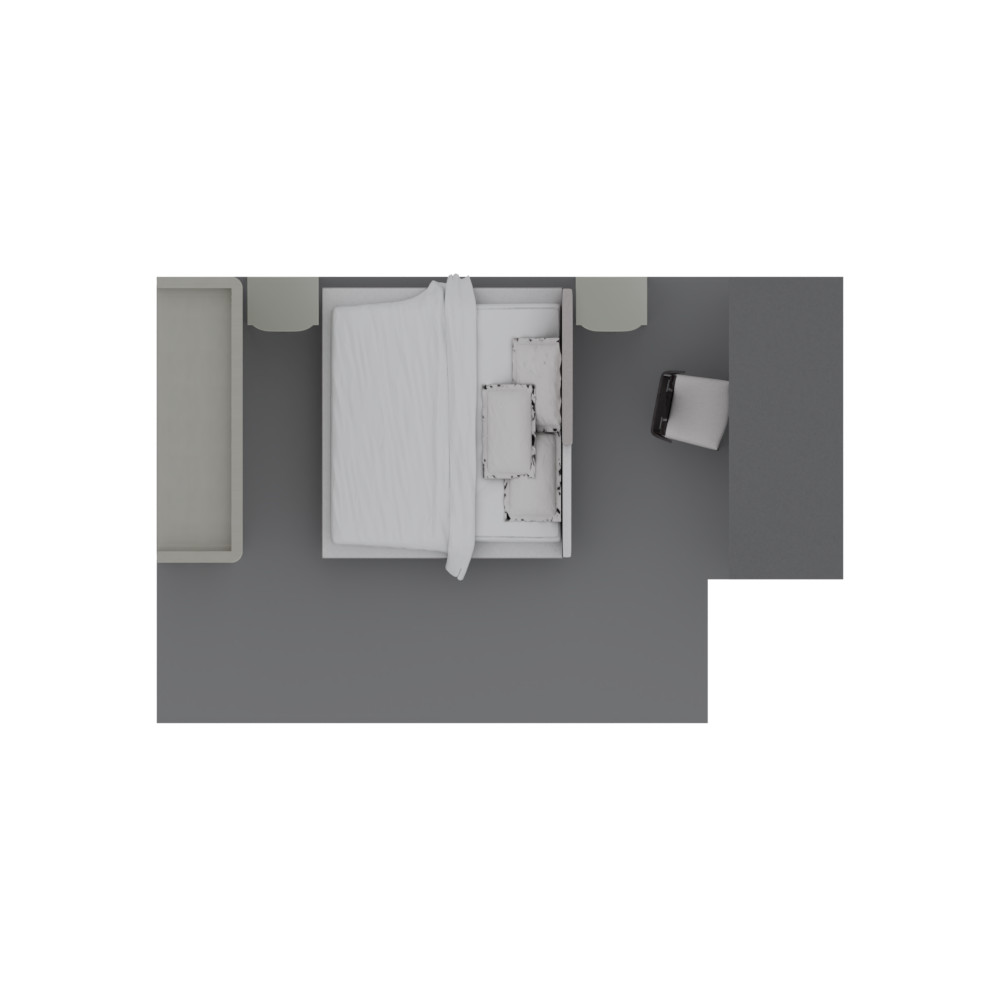}
        \end{overpic}
    \end{subfigure}%
    \begin{subfigure}[b]{0.22\linewidth}
        \centering
        \begin{overpic}[width=\textwidth,  clip]{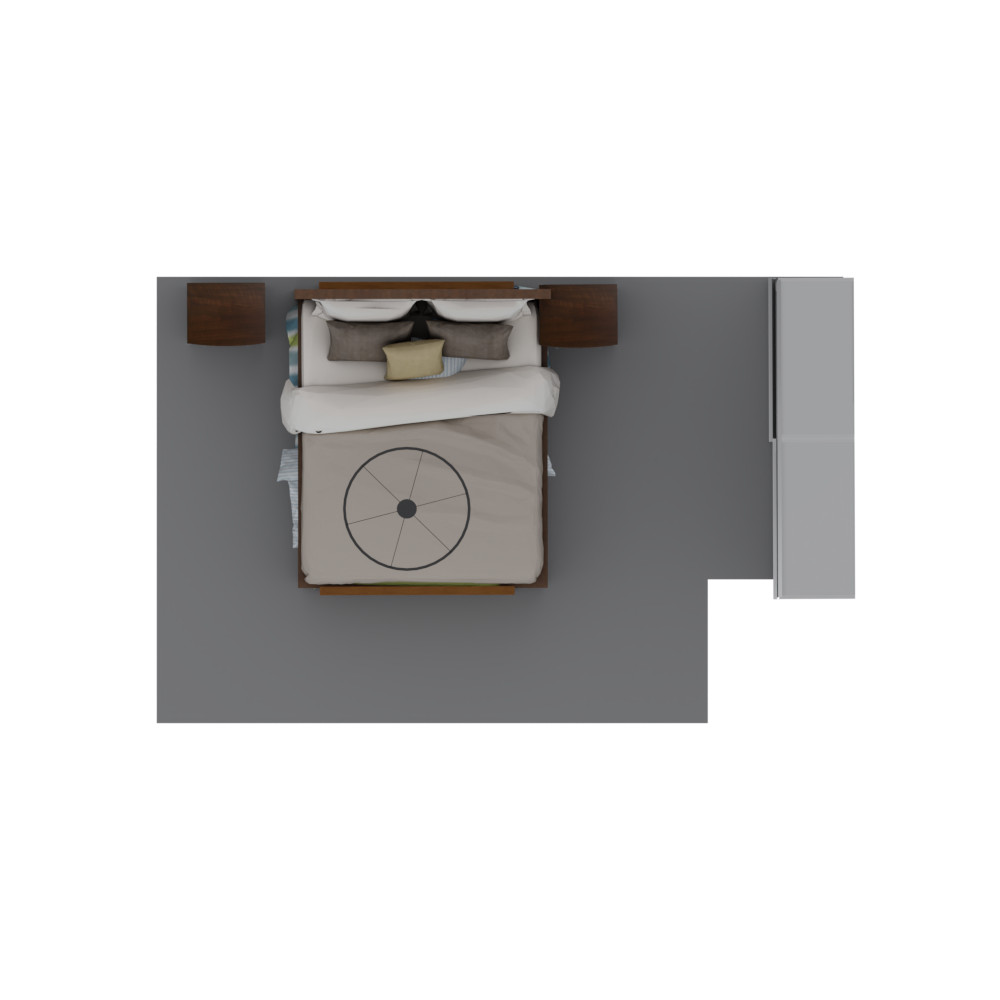}
	    \end{overpic}
    \end{subfigure}%
    \begin{subfigure}[b]{0.22\linewidth}
        \begin{overpic}[width=\textwidth, clip]{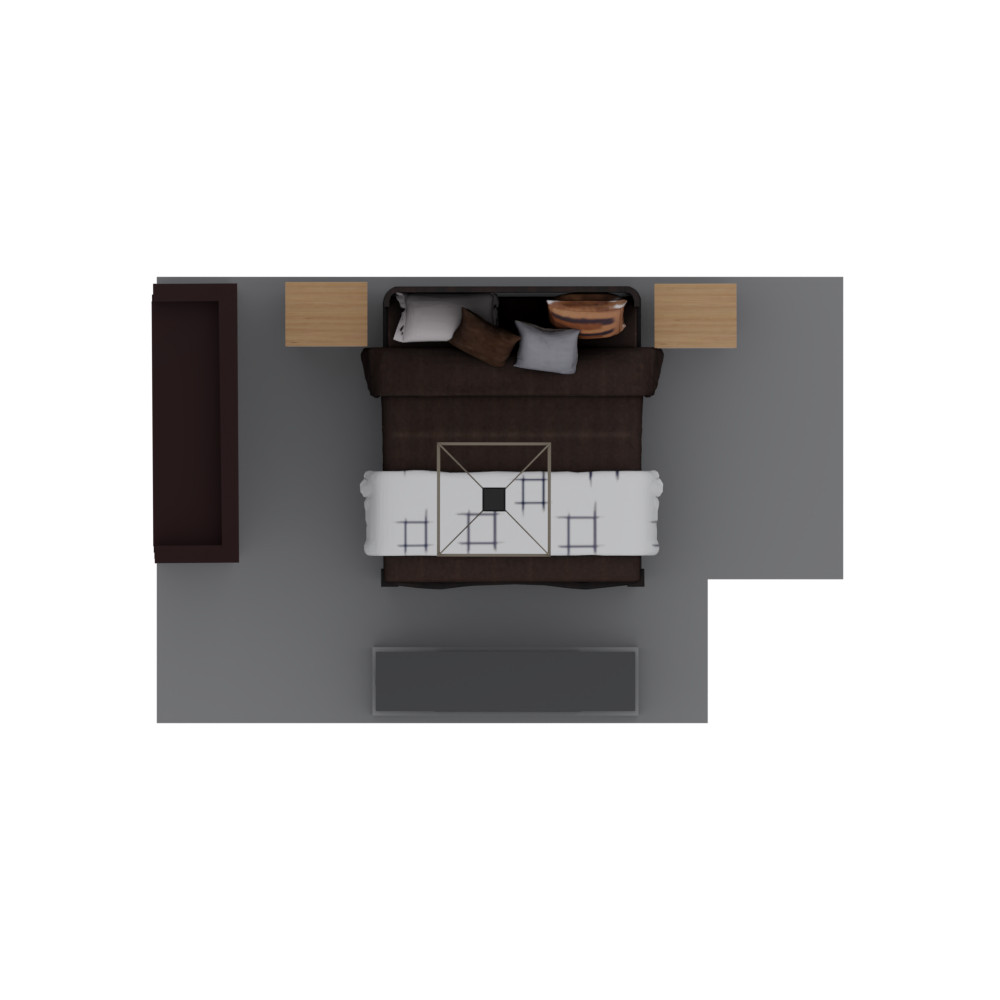}
        \end{overpic}
    \end{subfigure}%
    
    \caption{\textbf{Scene generation from scratch:} We compare generated scenes from GT, ATISS, and our model on \bed class.}
    \label{fig:addl_lib}
\vspace{-0.75em}
\end{figure*}

%%%%%%%%%%%%%%%%%%%%%%%%%%%%%%%%%%%%%%%%%%%%%%%%%%%%%%%%%%%%%%%%%%
%%%%%%%%%%%%%%%%%%%%%%%%%%%%%%%%%%%%%%%%%%%%%%%%%%%%%%%%%%%%%%%%%%%5
%%%%%%%%%%%%%%%%%%%%%%%%%%%%%%%%%%%%%%%%%%%%%%%%%%%%%%%%%%%%%%%%%%%%5
%% DINING

\begin{figure*}[t!]

    \centering
    \vspace{-1.5em}
    % \hfill
    \begin{subfigure}[b]{0.22\linewidth}
        \centering
	    \small Boundary
    \end{subfigure}%
    \begin{subfigure}[b]{0.22\linewidth}
        \centering
        \small GT
    \end{subfigure}%
    \begin{subfigure}[b]{0.22\linewidth}
        \centering
        \small ATISS
    \end{subfigure}%
    \begin{subfigure}[b]{0.22\linewidth}
        \centering
        \small Ours
    \end{subfigure}%
    % \hfill%
    \vskip\baselineskip%
    \vspace{-0.75em}
    %%%%%%%%%%%%%%%%%%%%%%%%%%%%%%%%%%%%
    % \hfill
    \begin{subfigure}[b]{0.22\linewidth}
        \centering
	    \includegraphics[width=0.8\textwidth,  clip]{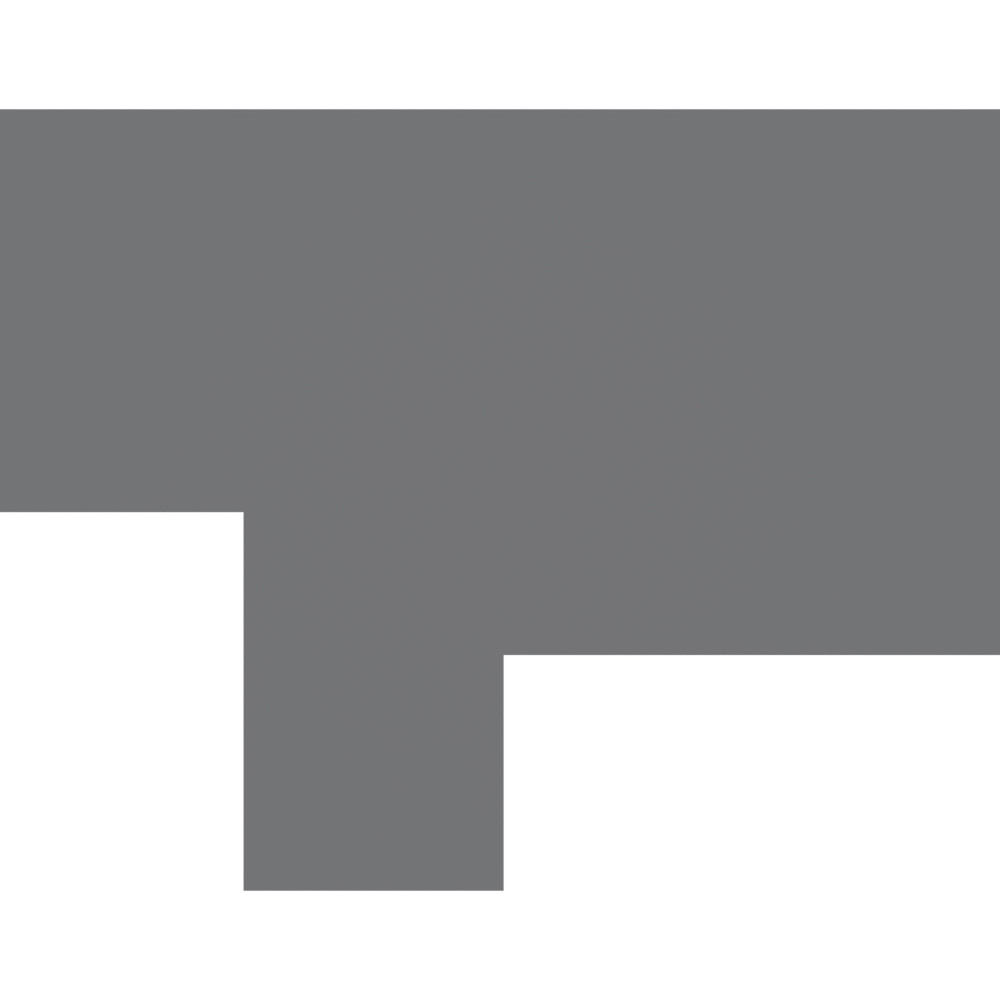}
    \end{subfigure}%
    \begin{subfigure}[b]{0.22\linewidth}
        \centering
            \begin{overpic}[width=0.8\textwidth,  clip]{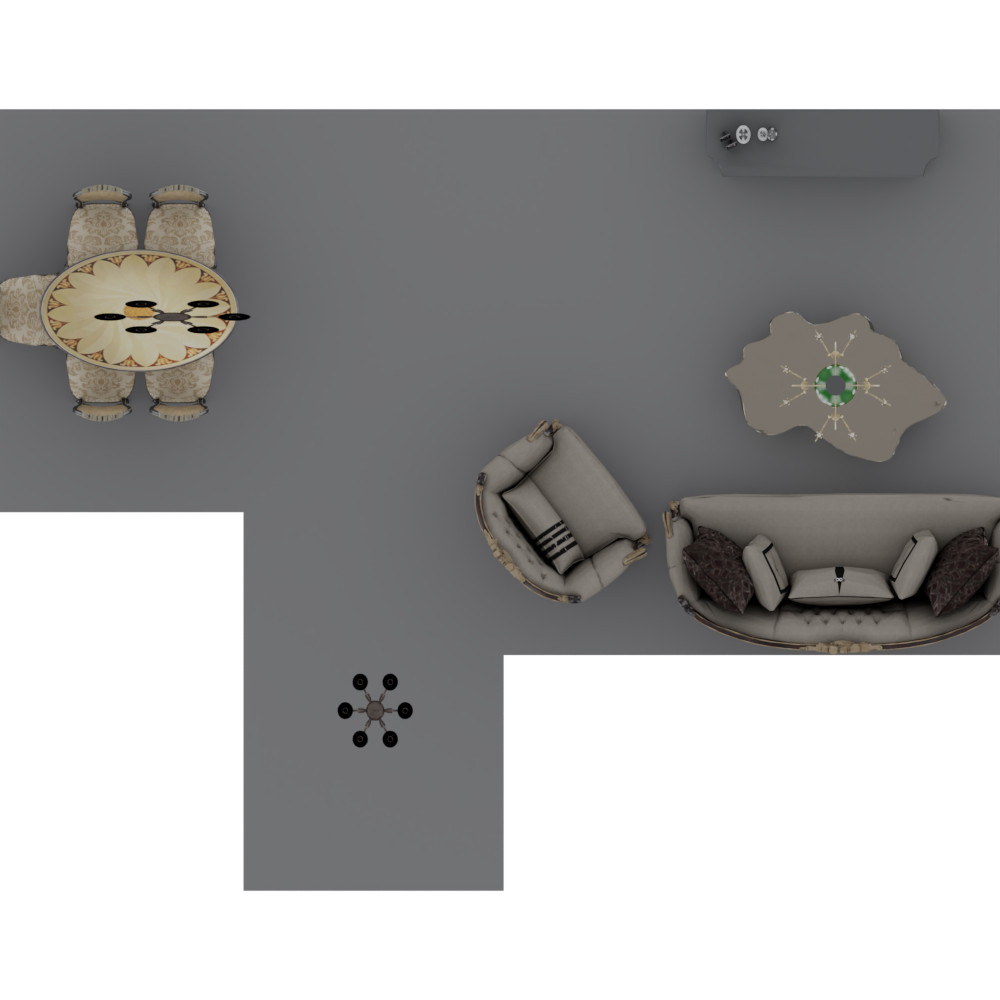}
        \end{overpic}
    \end{subfigure}%
    \begin{subfigure}[b]{0.22\linewidth}
        \centering
        \begin{overpic}[width=0.8\textwidth, clip]{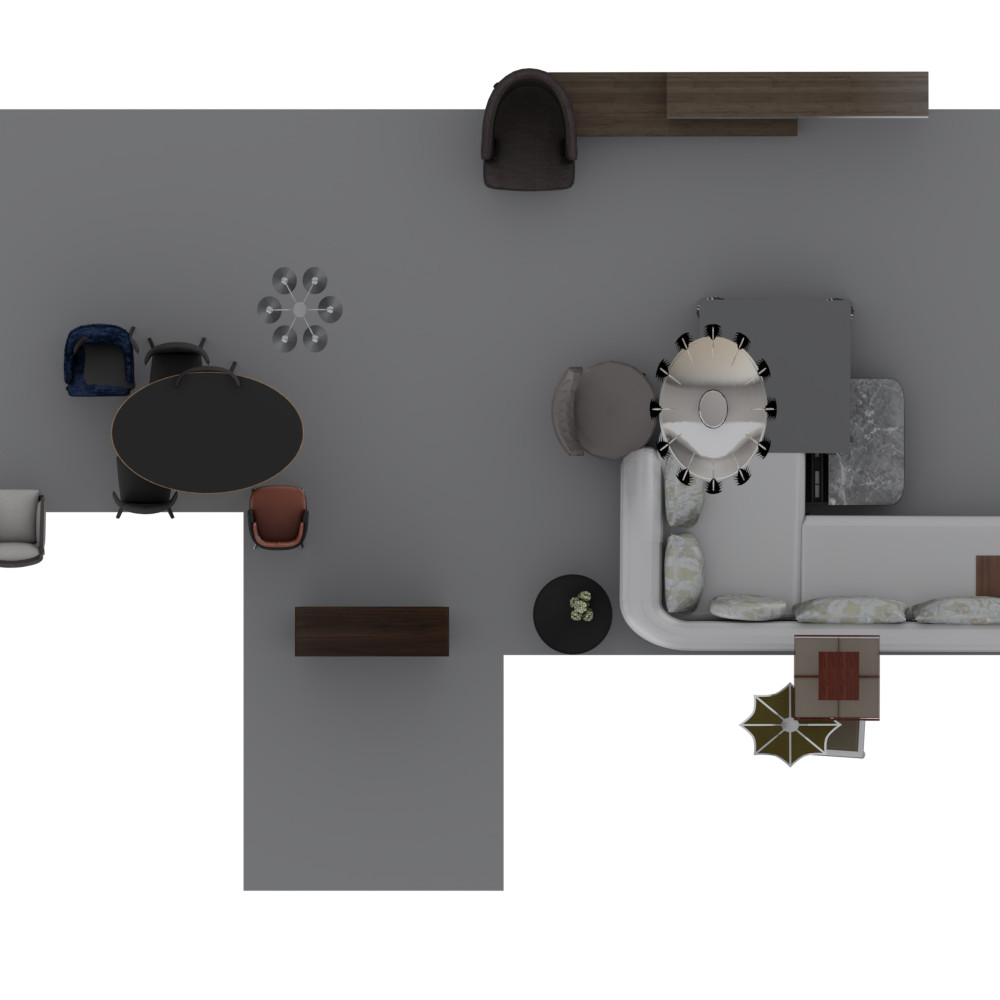}
	    \end{overpic}
    \end{subfigure}%
    \begin{subfigure}[b]{0.22\linewidth}
        \centering
        \begin{overpic}[width=0.8\textwidth, clip]{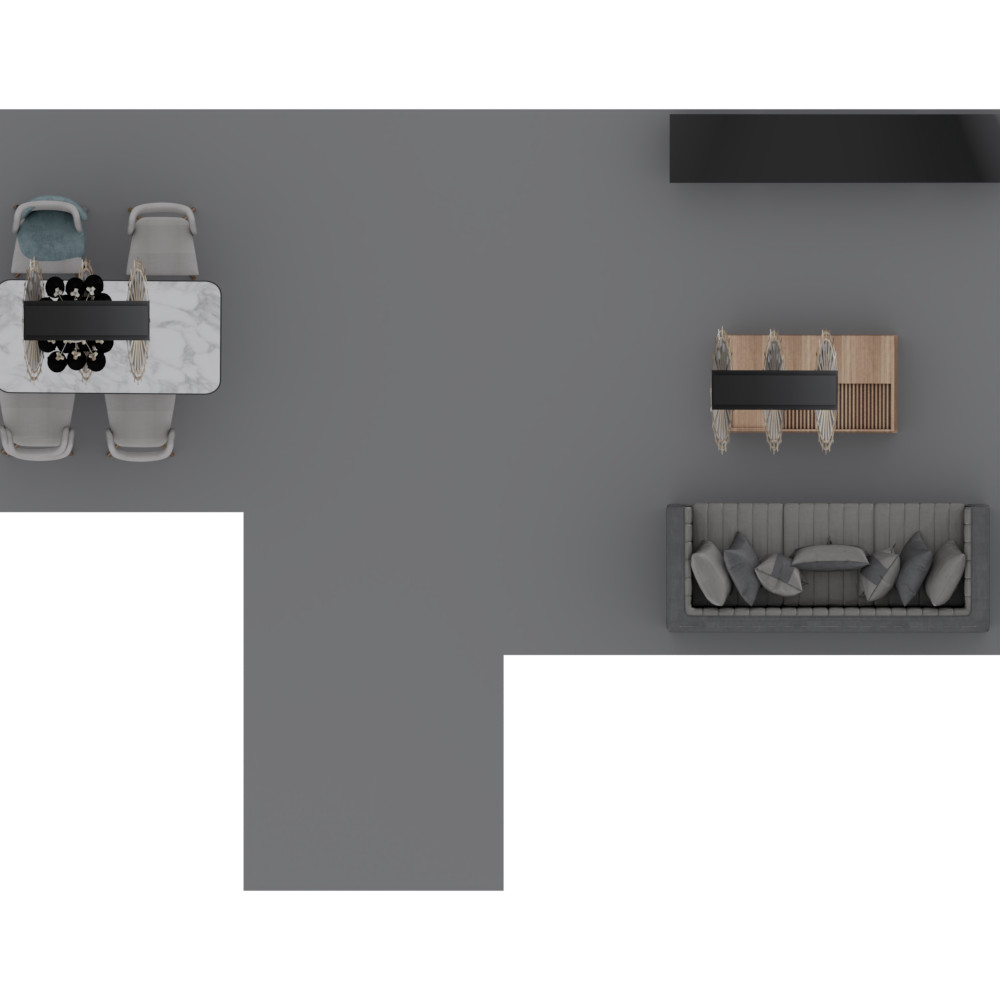}
        \end{overpic}
    \end{subfigure}%
    % \hfill%
    \vskip\baselineskip%
    \vspace{-0.75em}
    % \hfill
    %%%%%%%%%%%%%%%%%%%%%%%%%%%%%%%%%%%%%%%%%%%%%%%%%%%%%%%%%%%%%%%%%%%%%
    \begin{subfigure}[b]{0.22\linewidth}
        \centering
	    \includegraphics[width=\textwidth, clip]{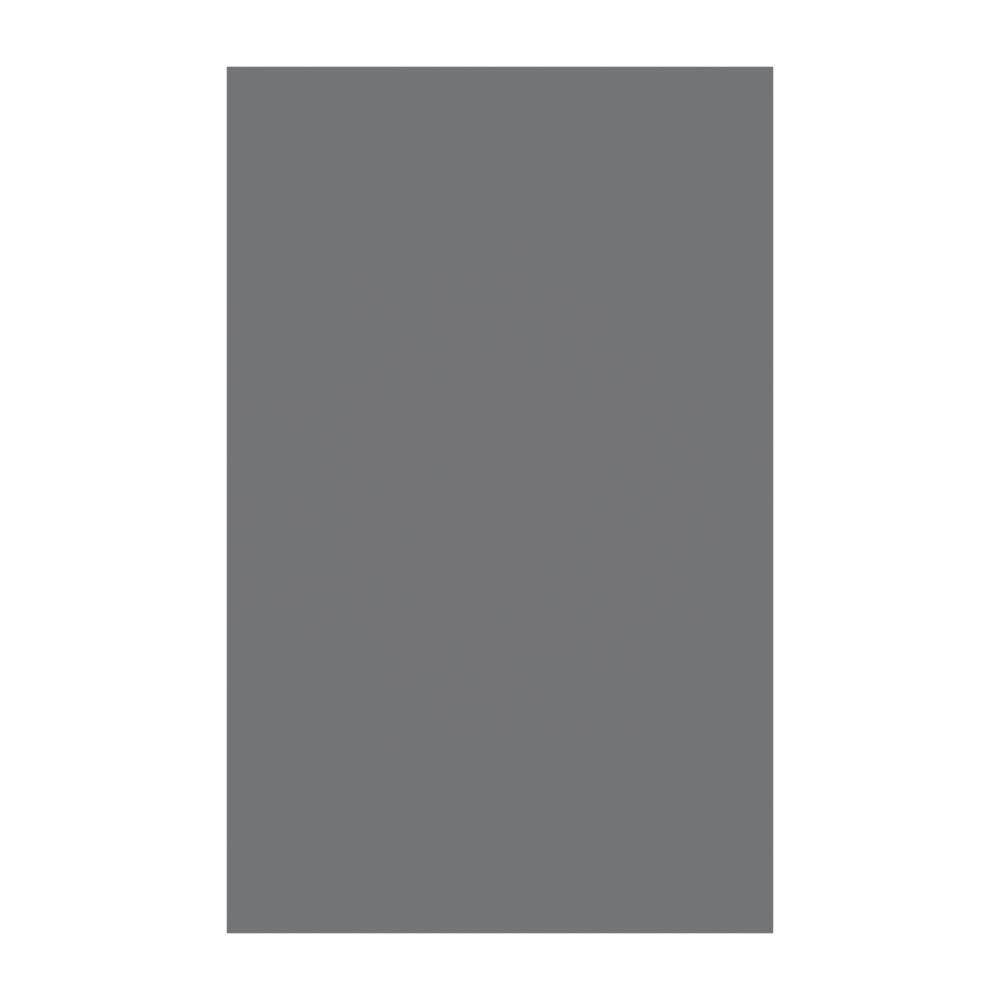}
    \end{subfigure}%
    \begin{subfigure}[b]{0.22\linewidth}
        \centering
        \begin{overpic}[width=\textwidth,  clip]{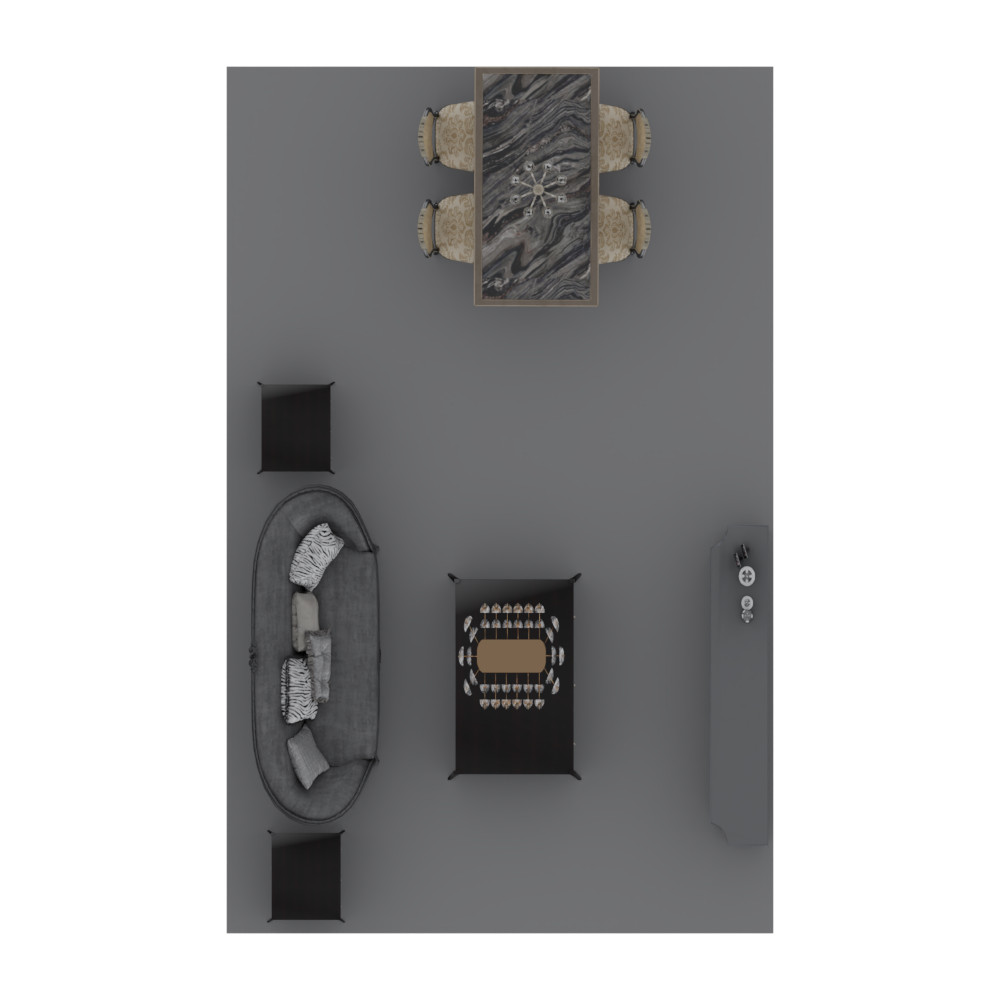}
        \end{overpic}
    \end{subfigure}%
    \begin{subfigure}[b]{0.22\linewidth}
        \centering
        \begin{overpic}[width=\textwidth, clip]{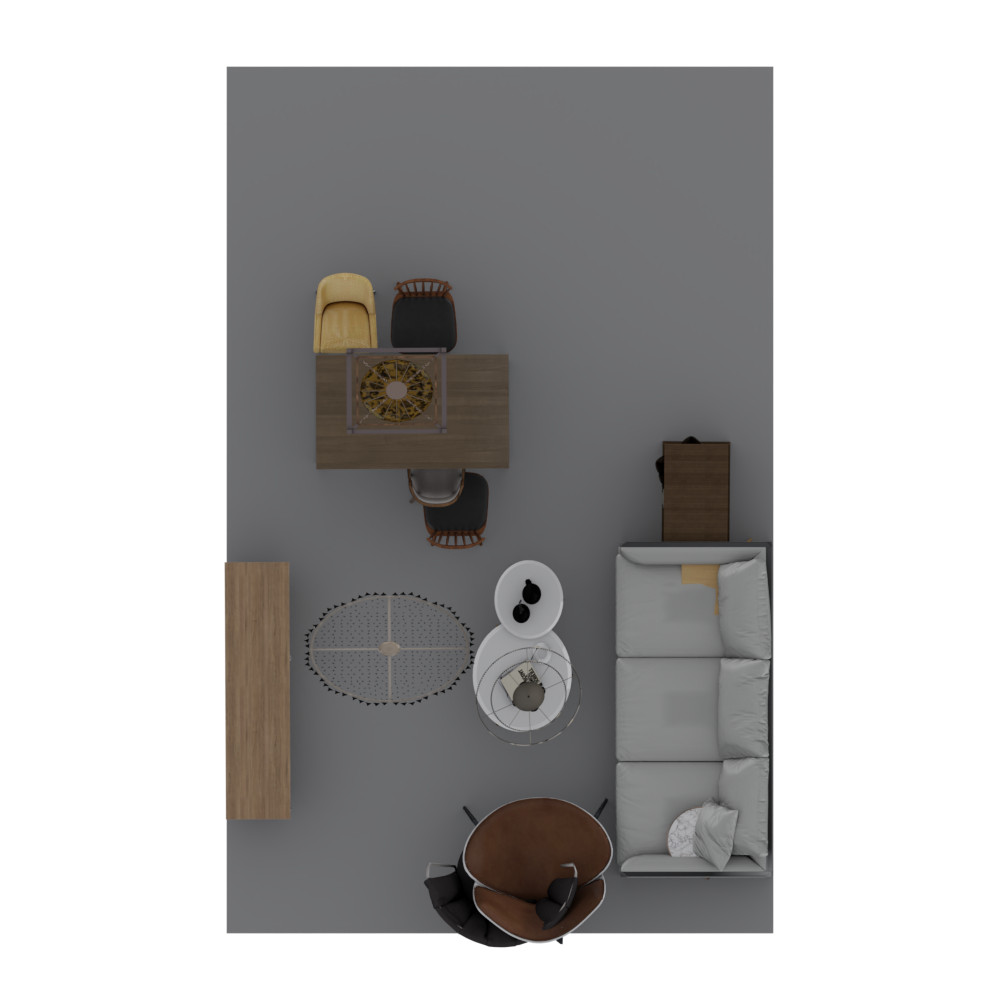}
	    \end{overpic}
    \end{subfigure}%
    \begin{subfigure}[b]{0.22\linewidth}
        \begin{overpic}[width=\textwidth,  clip]{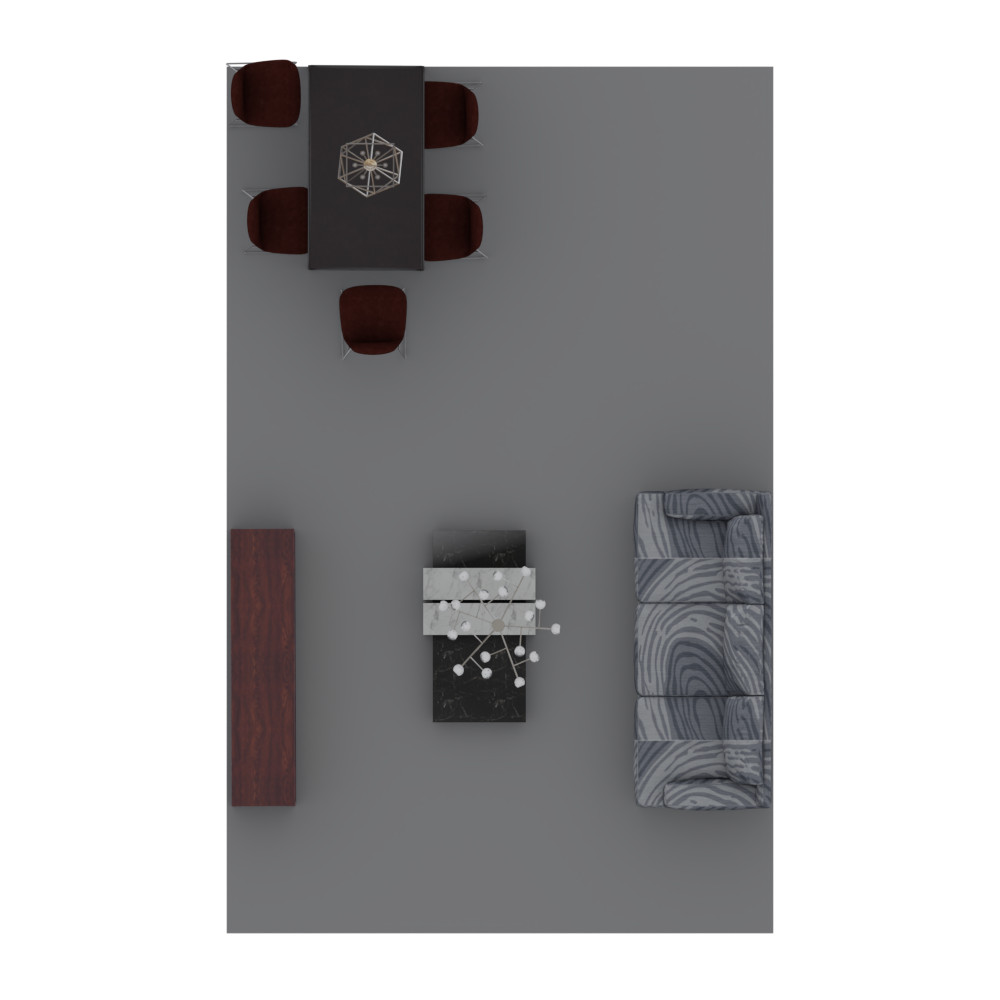}
        \end{overpic}
    \end{subfigure}%
    % \hfill%
    \vskip\baselineskip%
    \vspace{-0.75em}
    % \hfill
    %%%%%%%%%%%%%%%%%%%%%%%%%%%%%%%%%%%%%%%%%%%%%%%%%%%%%%%%%%%%%%%%%%%%%%%%%%%%%%%%%%%%%%%%%%%%%%%%%%%5
    \begin{subfigure}[b]{0.22\linewidth}
        \centering
	    \includegraphics[width=\textwidth, clip]{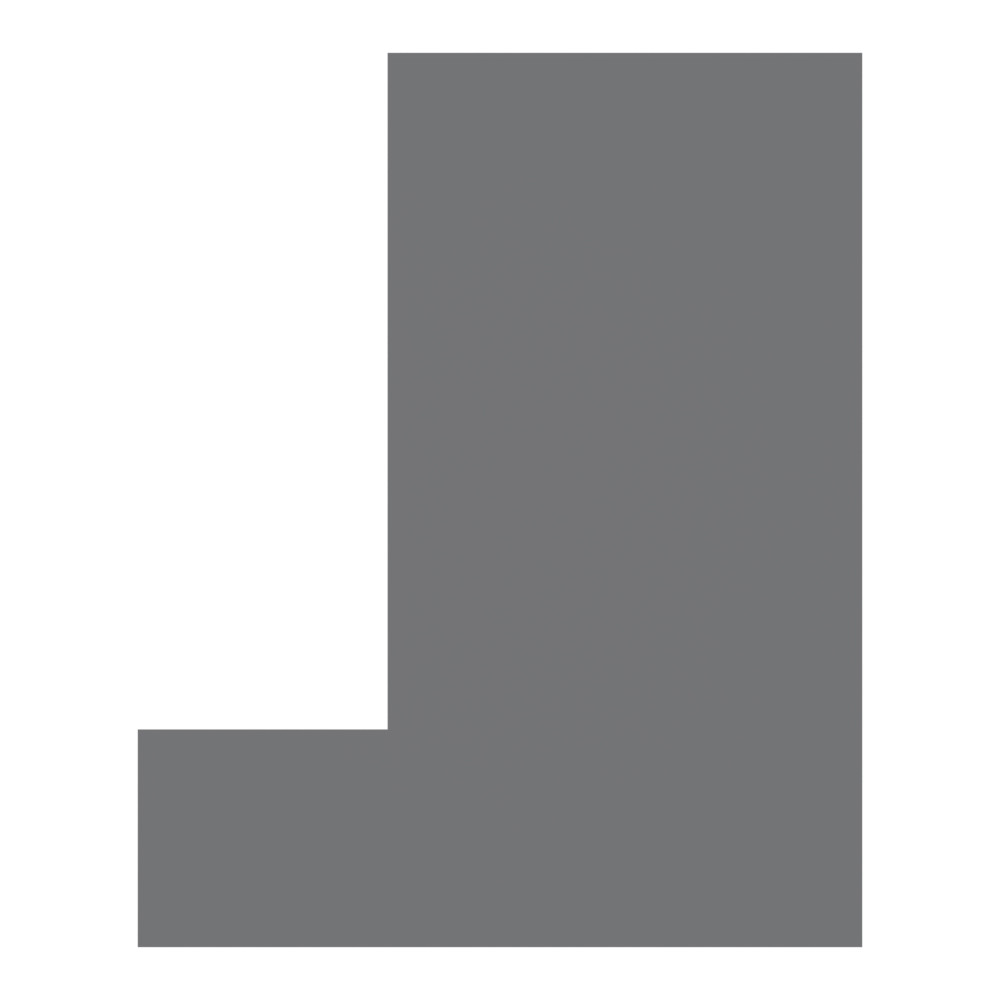}
    \end{subfigure}%
    \begin{subfigure}[b]{0.22\linewidth}
        \centering
        \begin{overpic}[width=\textwidth,  clip]{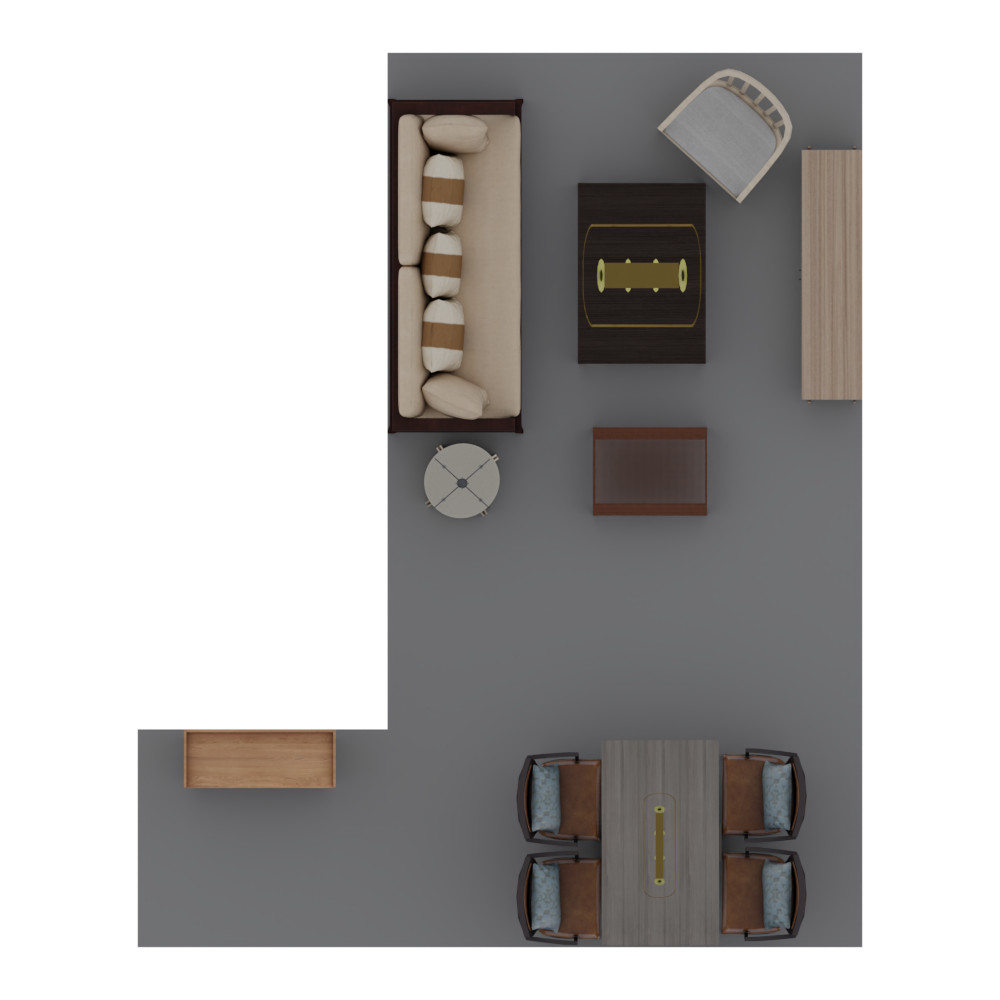}
        \end{overpic}
    \end{subfigure}%
    \begin{subfigure}[b]{0.22\linewidth}
        \centering
        \begin{overpic}[width=\textwidth, clip]{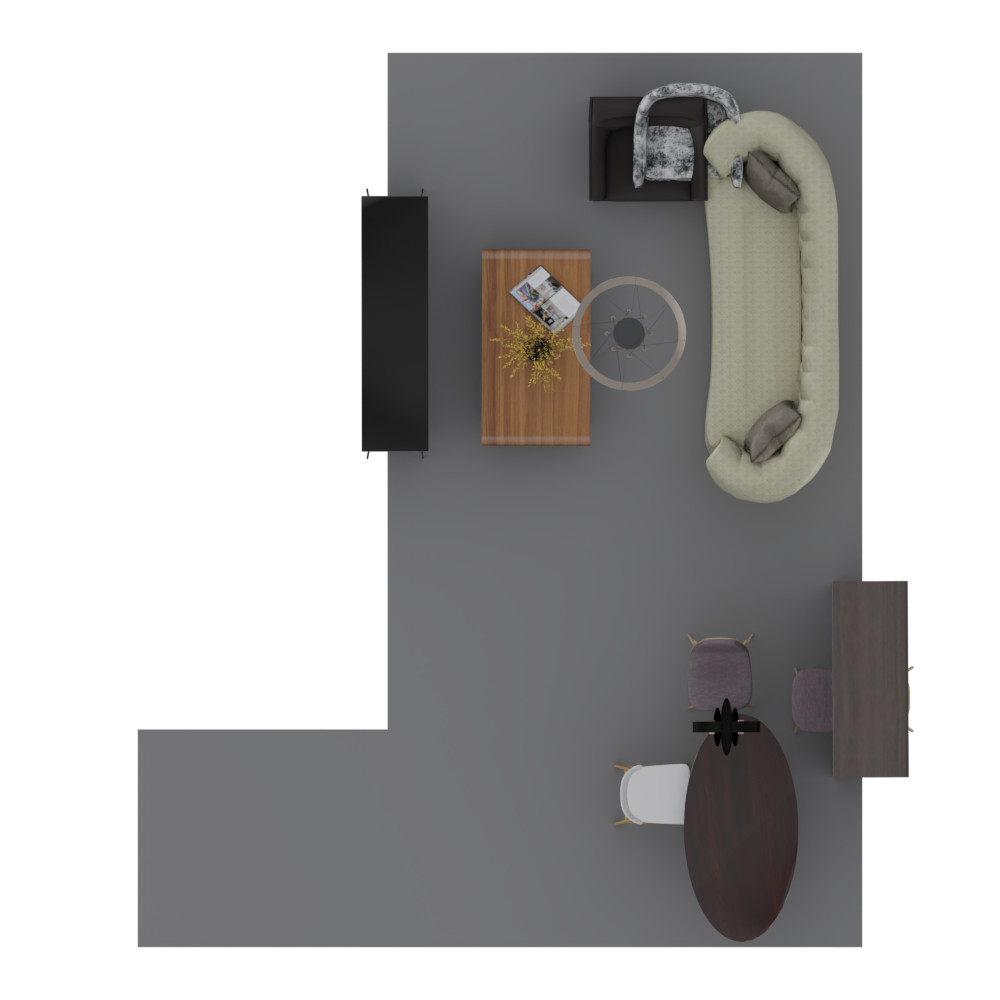}
	    \end{overpic}
    \end{subfigure}%
    \begin{subfigure}[b]{0.22\linewidth}
        \begin{overpic}[width=\textwidth,  clip]{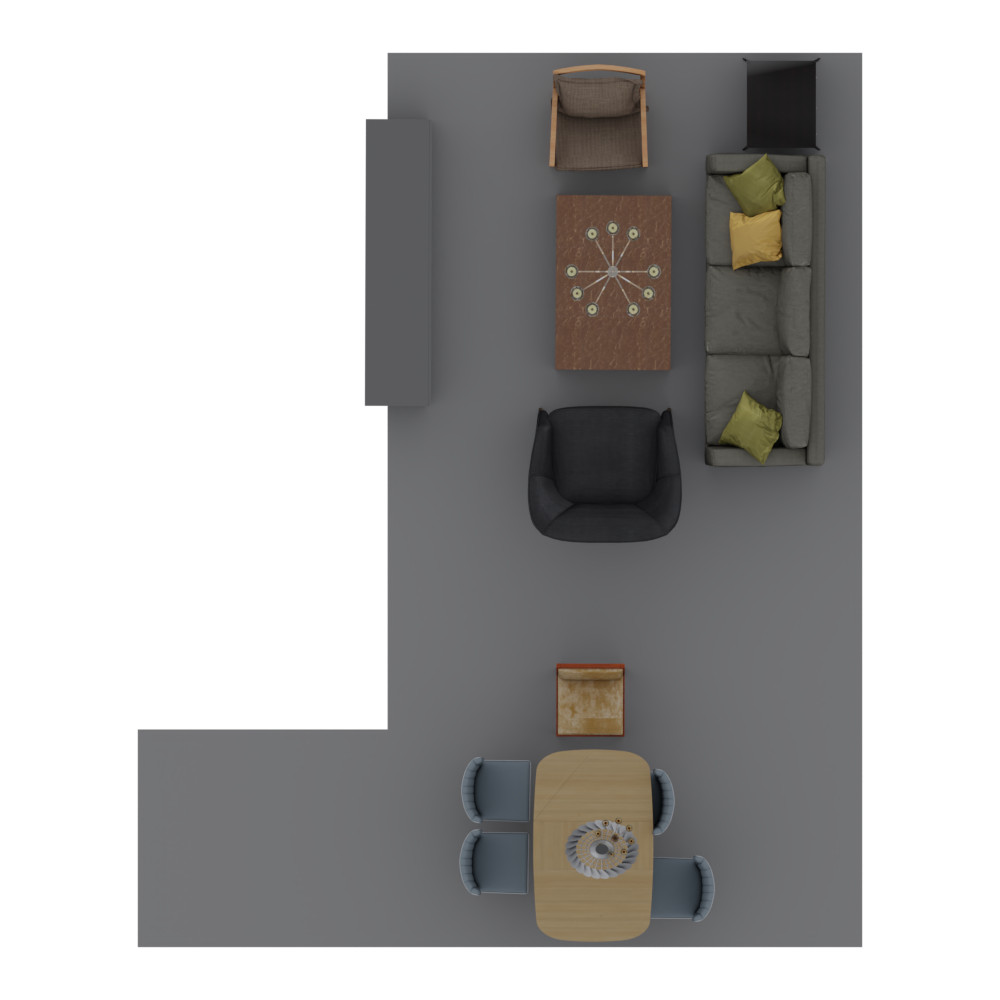}
        \end{overpic}
    \end{subfigure}%
    % \hfill%
    \vskip\baselineskip%
    \vspace{-0.75em}
    \begin{subfigure}[b]{0.22\linewidth}
        \centering
	    \includegraphics[width=\textwidth, clip]{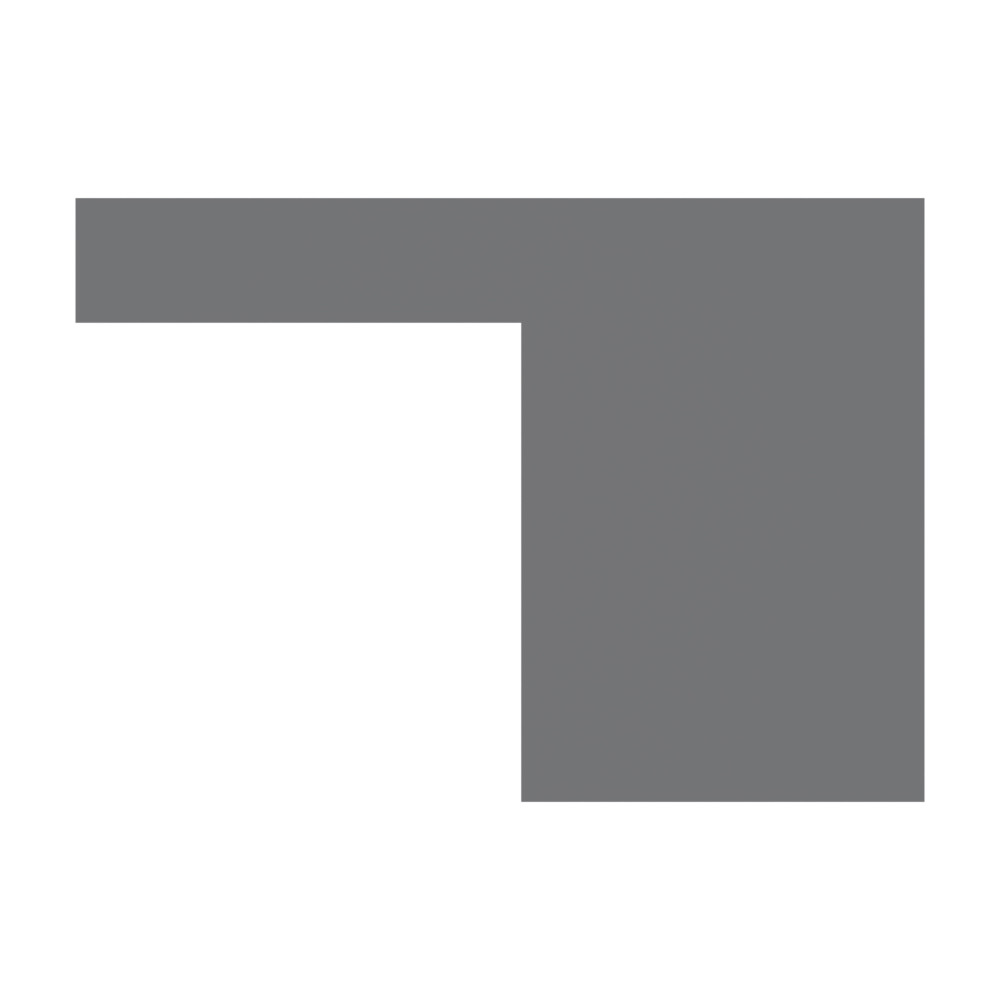}
    \end{subfigure}%
    \begin{subfigure}[b]{0.22\linewidth}
        \centering
        \begin{overpic}[width=\textwidth,  clip]{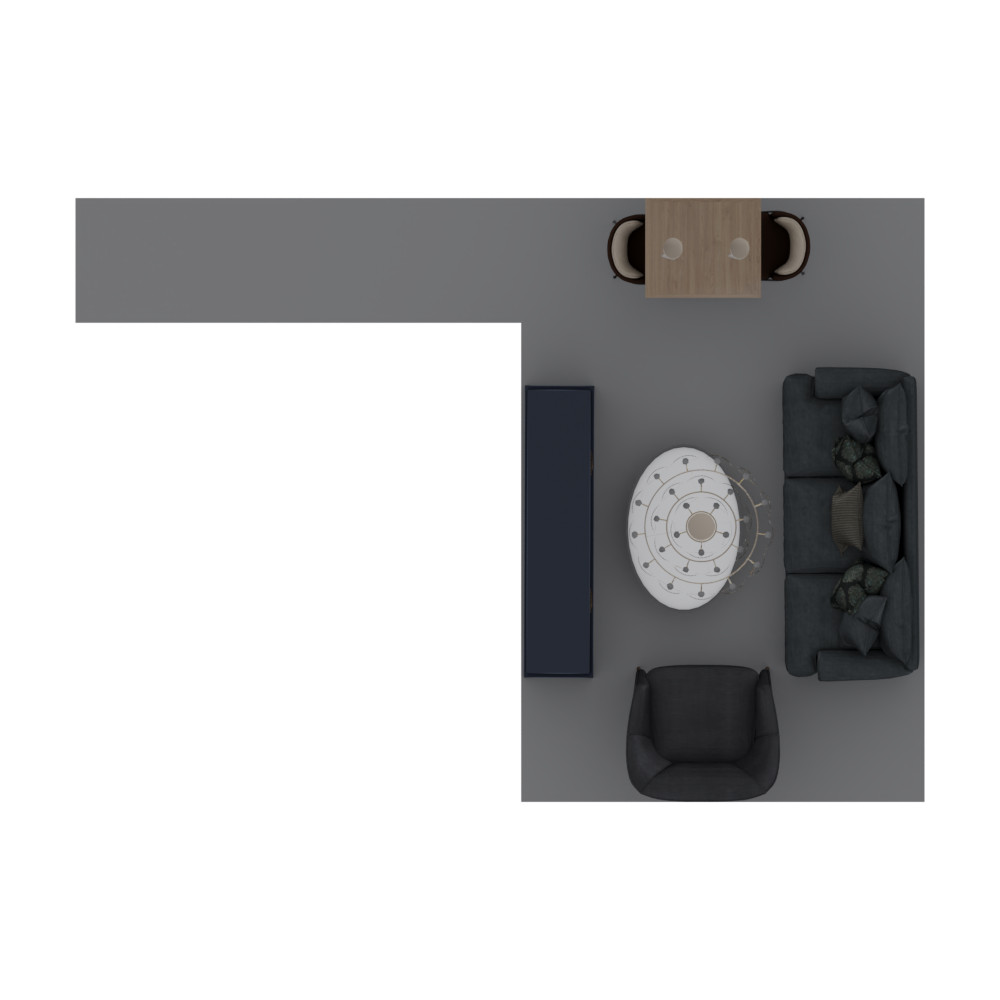}
        \end{overpic}
    \end{subfigure}%
    \begin{subfigure}[b]{0.22\linewidth}
        \centering
        \begin{overpic}[width=\textwidth,  clip]{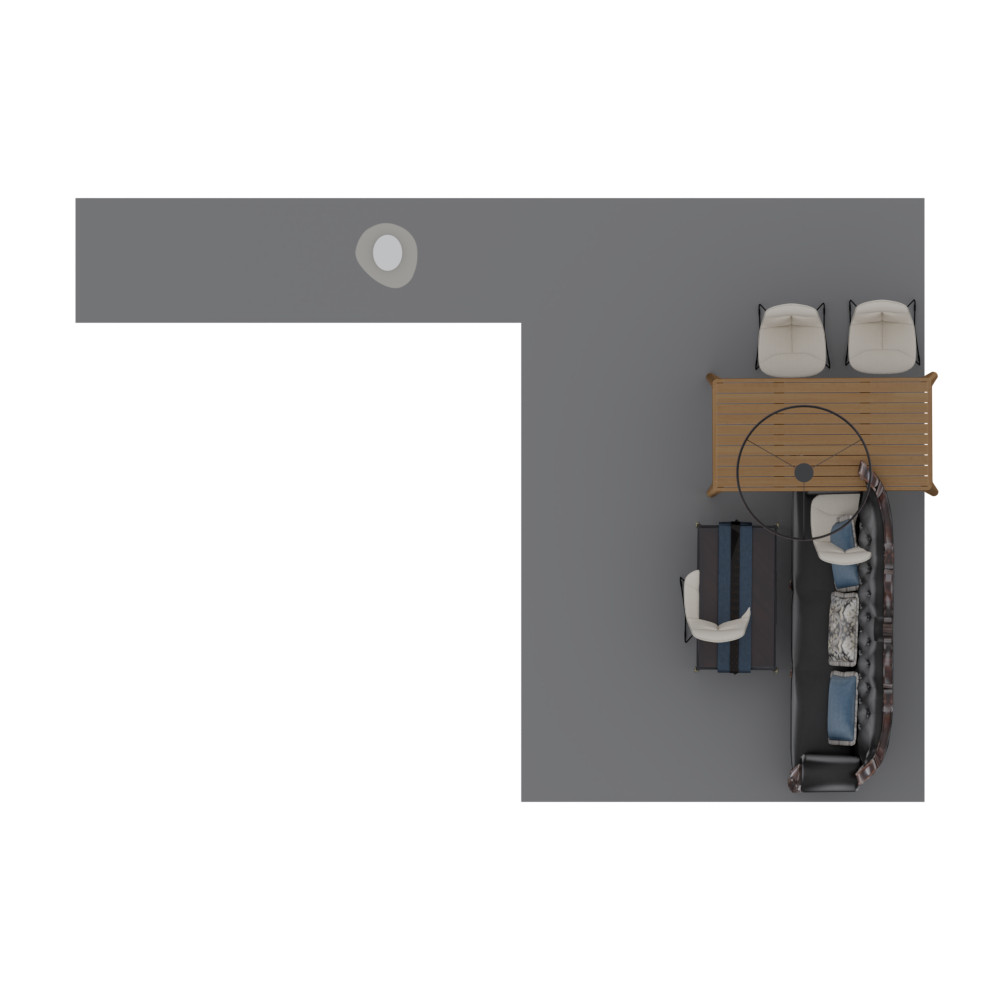}
	    \end{overpic}
    \end{subfigure}%
    \begin{subfigure}[b]{0.22\linewidth}
        \begin{overpic}[width=\textwidth,  clip]{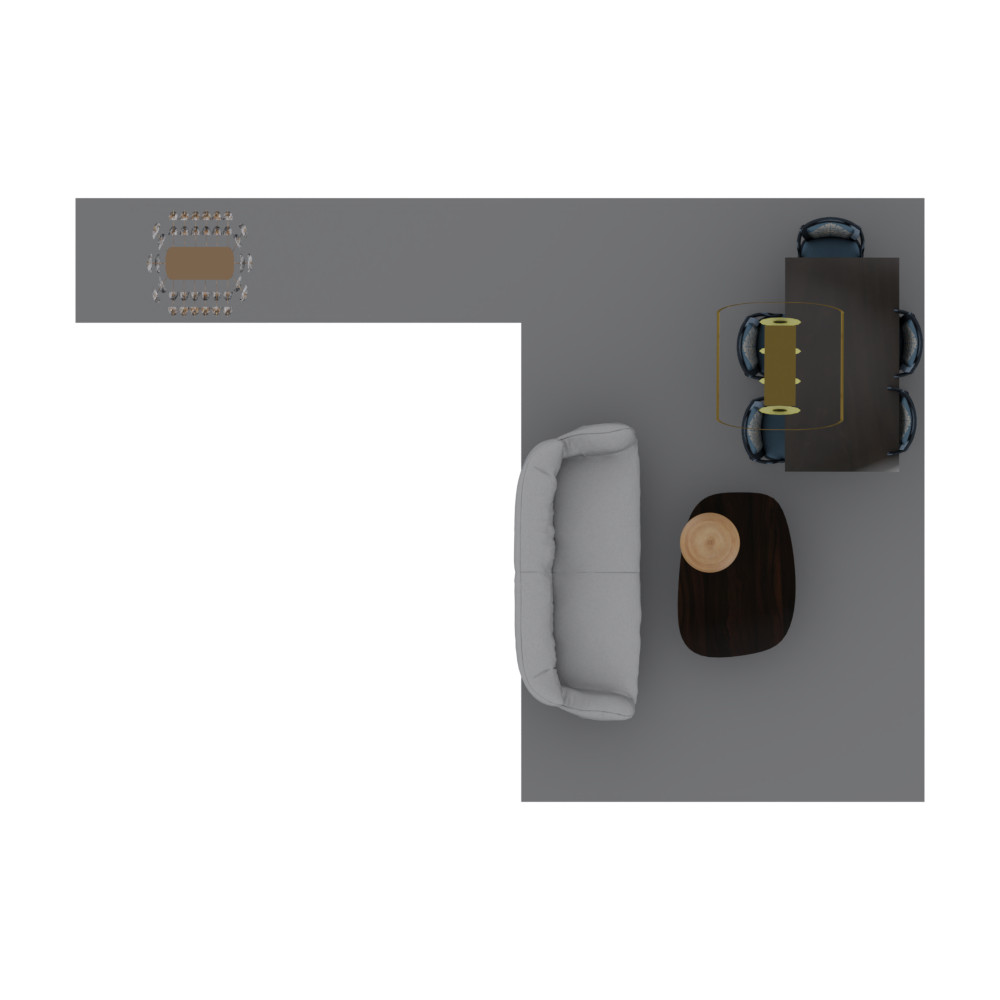}
        \end{overpic}
    \end{subfigure}%
    % \hfill%
    \vskip\baselineskip%
    \vspace{-0.75em}
    \begin{subfigure}[b]{0.22\linewidth}
        \centering
	    \includegraphics[width=\textwidth, clip]{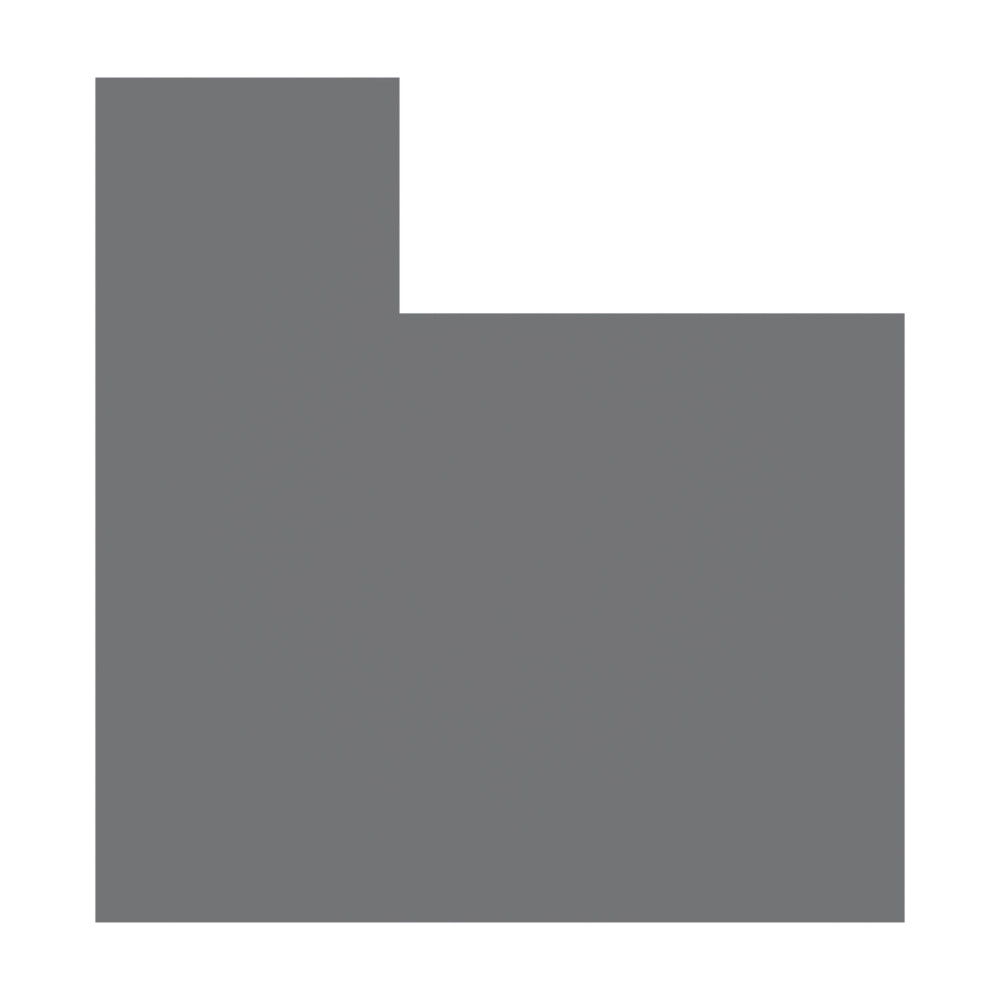}
    \end{subfigure}%
    \begin{subfigure}[b]{0.22\linewidth}
        \centering
        \begin{overpic}[width=\textwidth,  clip]{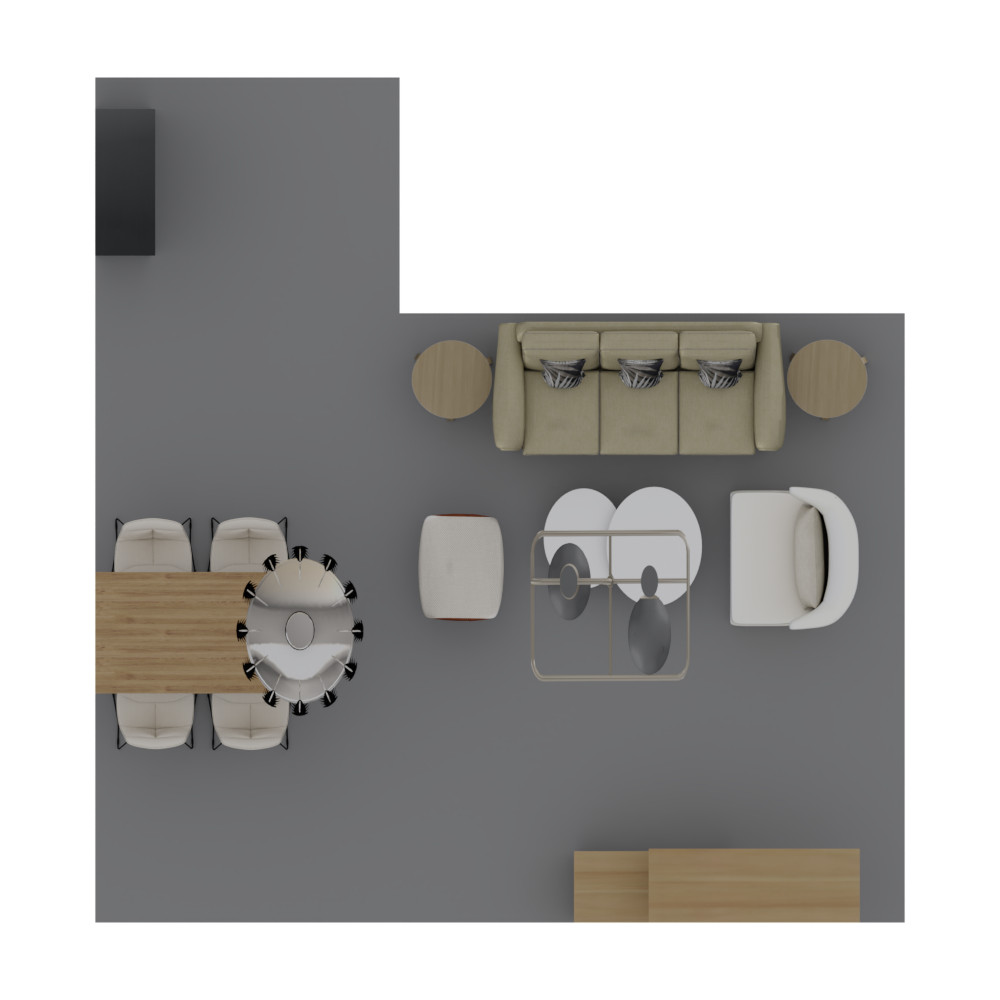}
        \end{overpic}
    \end{subfigure}%
    \begin{subfigure}[b]{0.22\linewidth}
        \centering
        \begin{overpic}[width=\textwidth,  clip]{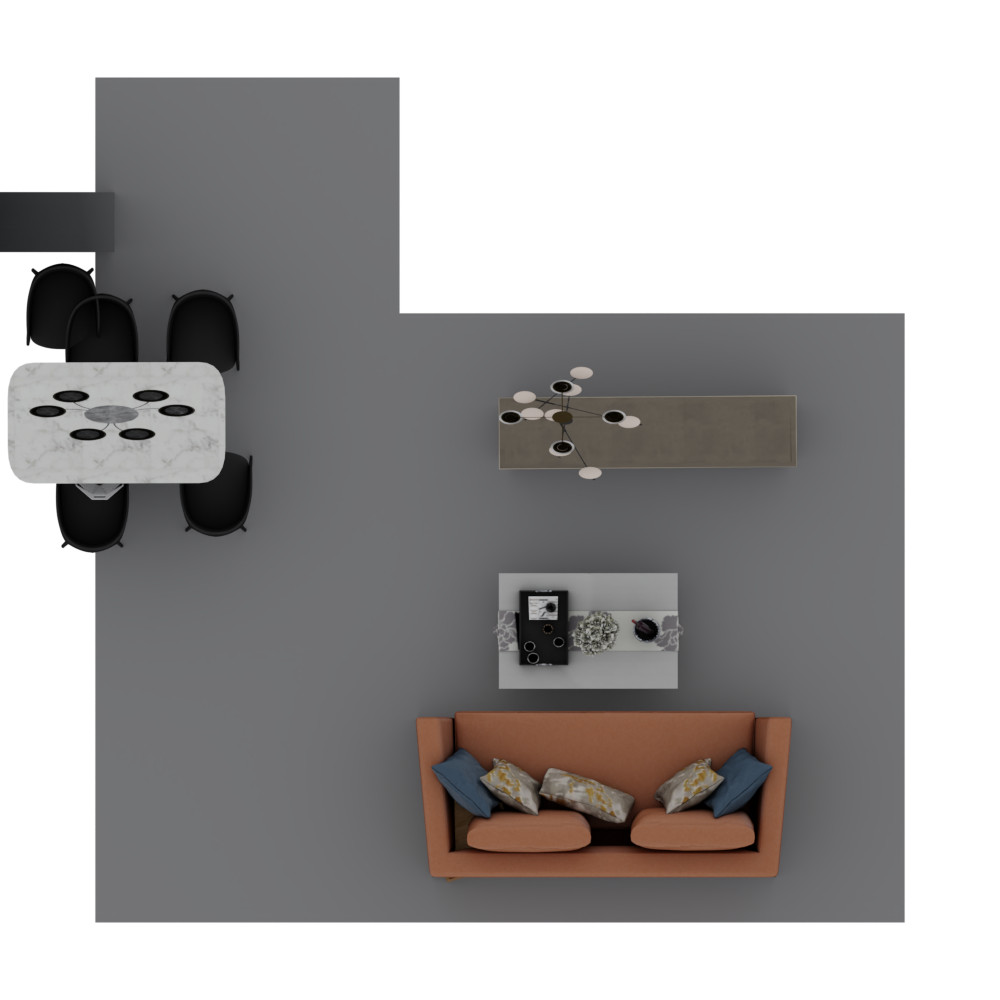}
	    \end{overpic}
    \end{subfigure}%
    \begin{subfigure}[b]{0.22\linewidth}
        \begin{overpic}[width=\textwidth,  clip]{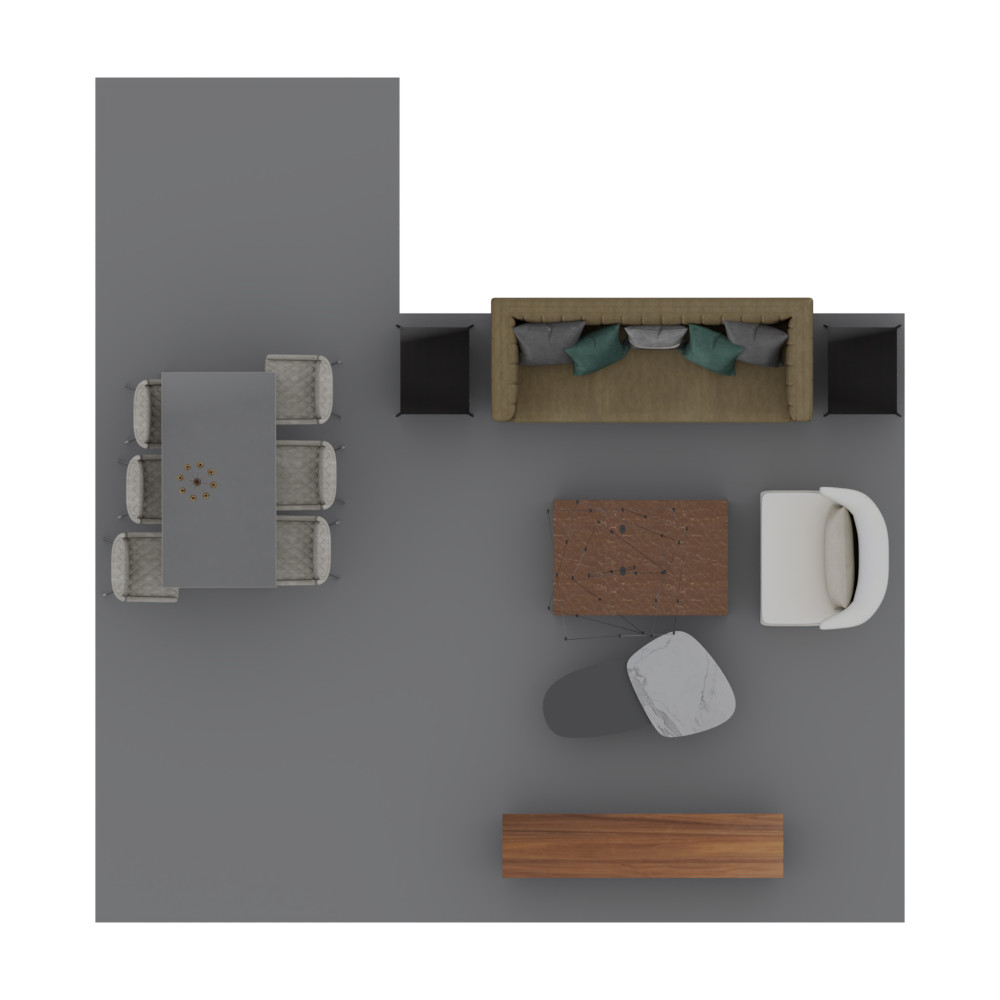}
        \end{overpic}
    \end{subfigure}%
    % \hfill%
    \vskip\baselineskip%
    \vspace{-0.75em}
    \begin{subfigure}[b]{0.22\linewidth}
        \centering
	    \includegraphics[width=\textwidth, clip]{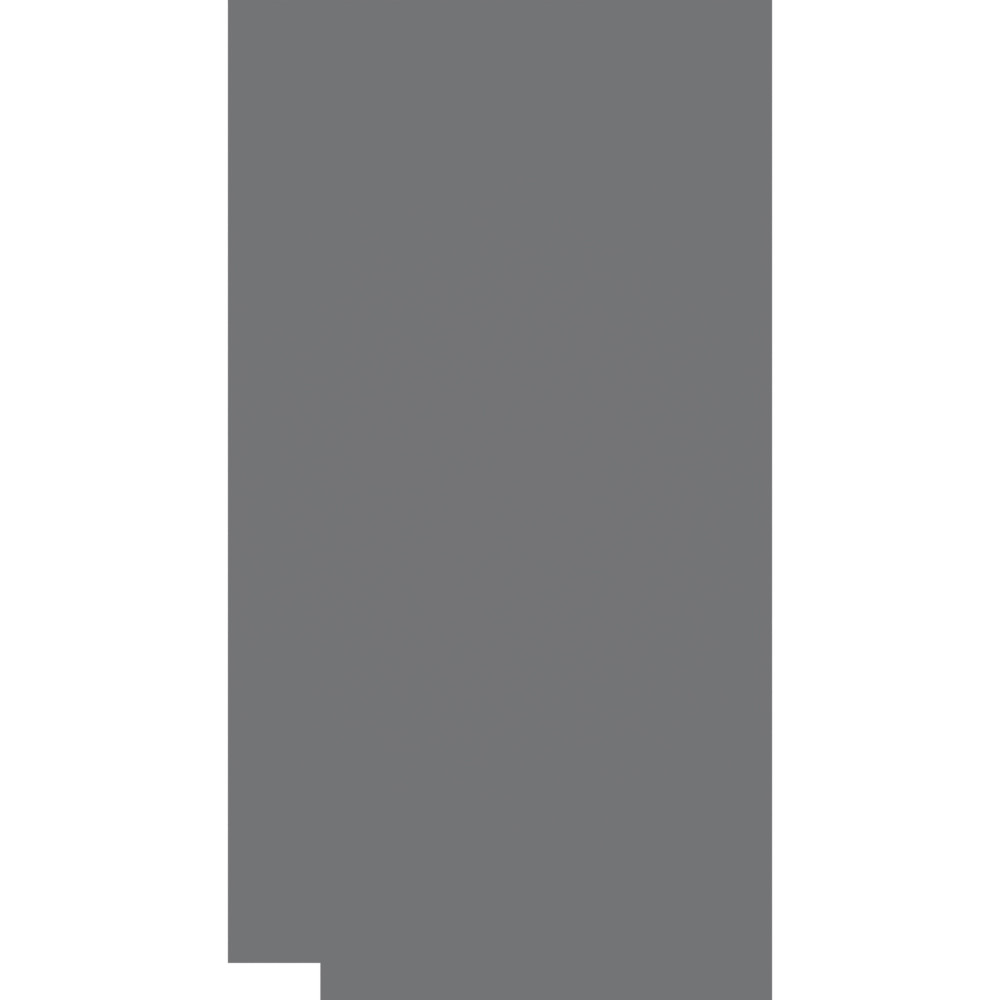}
    \end{subfigure}%
    \begin{subfigure}[b]{0.22\linewidth}
        \centering
        \begin{overpic}[width=\textwidth,  clip]{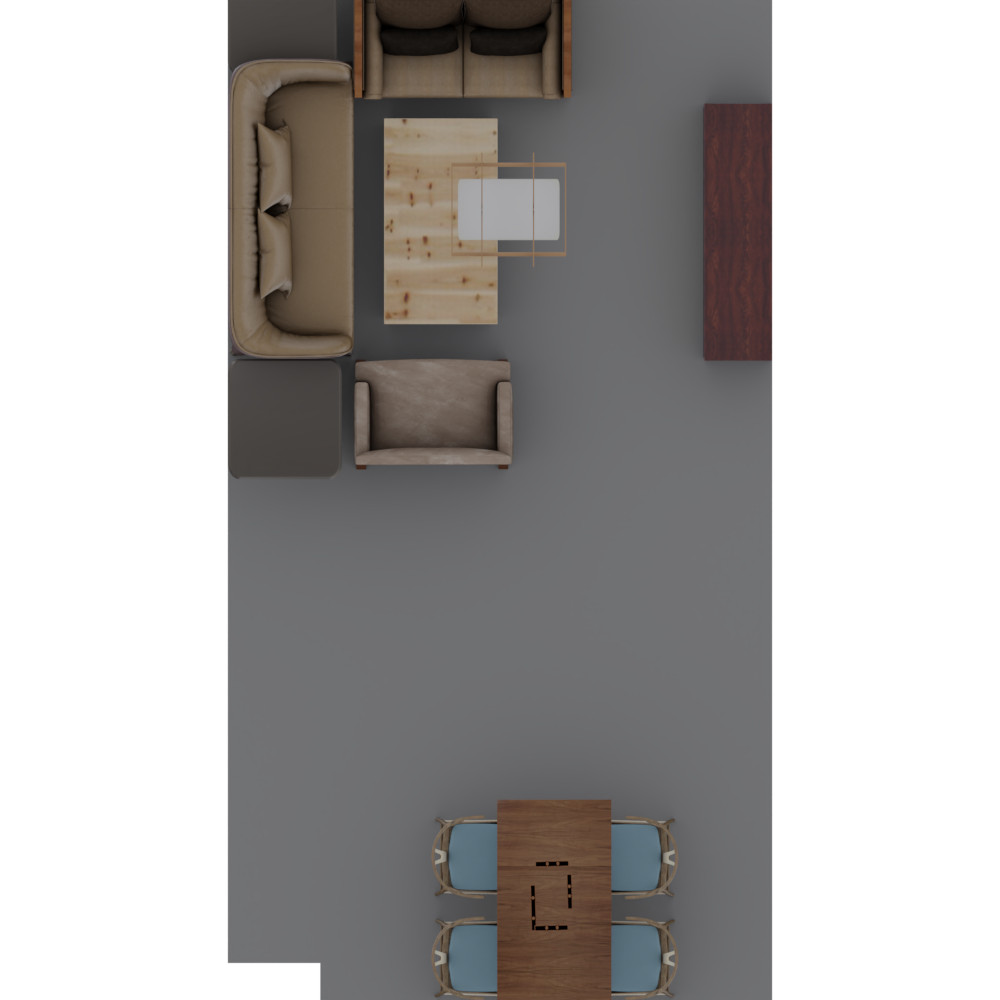}
        \end{overpic}
    \end{subfigure}%
    \begin{subfigure}[b]{0.22\linewidth}
        \centering
        \begin{overpic}[width=\textwidth,  clip]{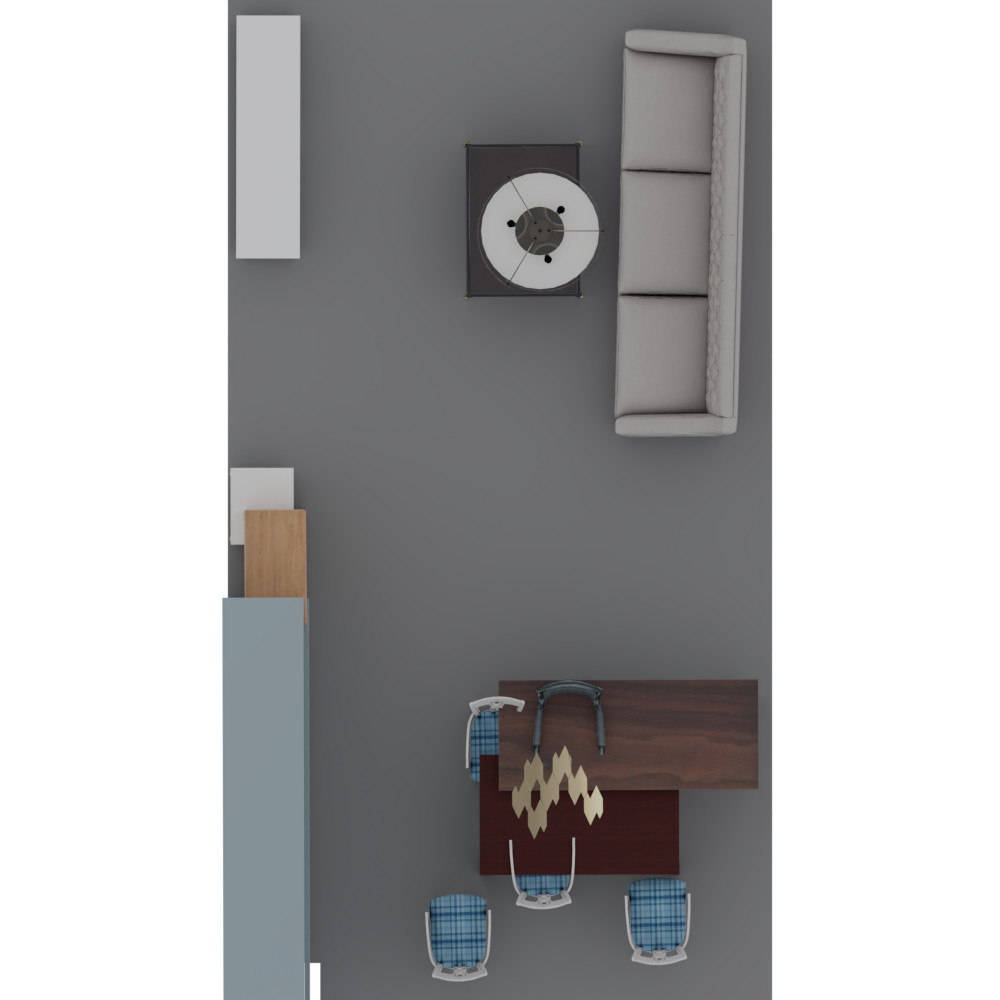}
	    \end{overpic}
    \end{subfigure}%
    \begin{subfigure}[b]{0.22\linewidth}
        \begin{overpic}[width=\textwidth, clip]{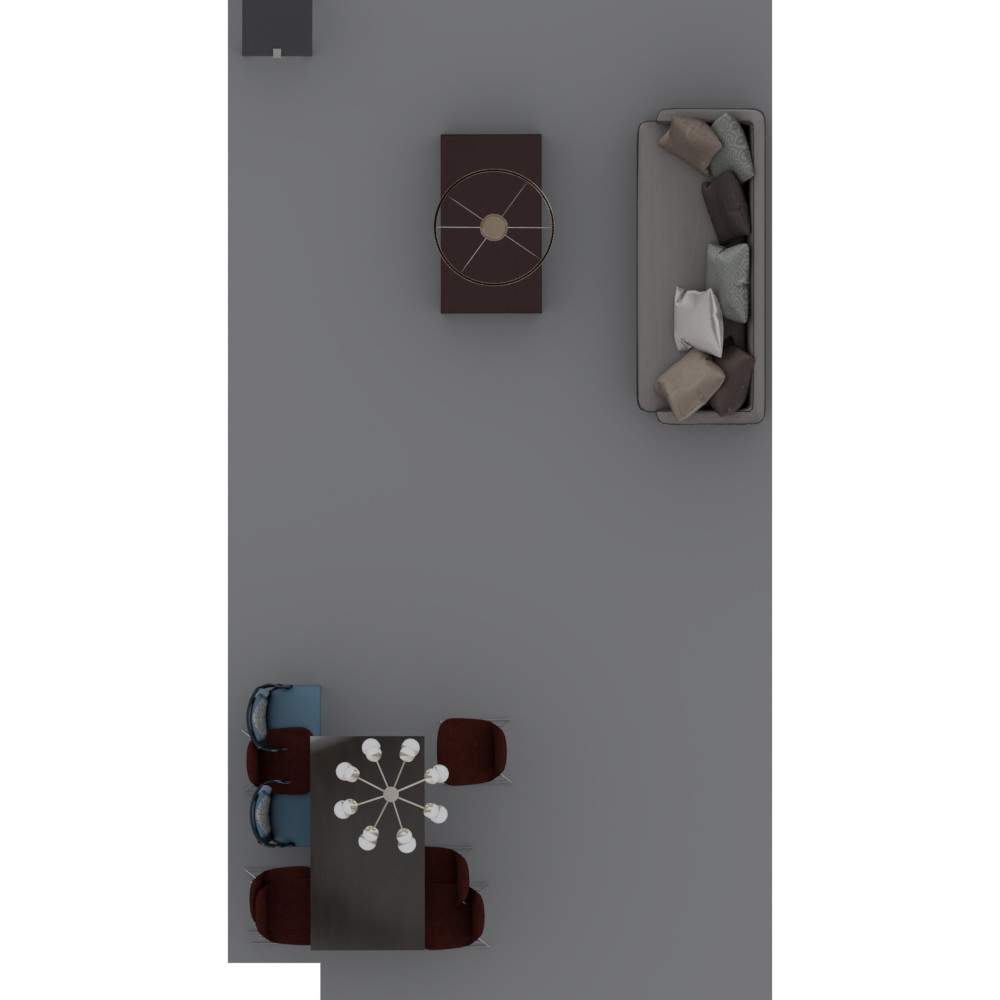}
        \end{overpic}
    \end{subfigure}%
    
    \caption{\textbf{Scene generation from scratch:} We compare generated scenes from GT, ATISS, and our model on \din class.}
    \label{fig:addl_din}
\vspace{-0.75em}
\end{figure*}

%%%%%%%%%%%%%%%%%%%%%%%%%%%%%%%%%%%%%%%%%%%%%%%%%%%%%%%%%%%%%%%%%%
%%%%%%%%%%%%%%%%%%%%%%%%%%%%%%%%%%%%%%%%%%%%%%%%%%%%%%%%%%%%%%%%%%%5
%%%%%%%%%%%%%%%%%%%%%%%%%%%%%%%%%%%%%%%%%%%%%%%%%%%%%%%%%%%%%%%%%%%%5
%% LIVING

\begin{figure*}[t!]

    \centering
    \vspace{-1.5em}
    % \hfill
    \begin{subfigure}[b]{0.22\linewidth}
        \centering
	    \small Boundary
    \end{subfigure}%
    \begin{subfigure}[b]{0.22\linewidth}
        \centering
        \small GT
    \end{subfigure}%
    \begin{subfigure}[b]{0.22\linewidth}
        \centering
        \small ATISS
    \end{subfigure}%
    \begin{subfigure}[b]{0.22\linewidth}
        \centering
        \small Ours
    \end{subfigure}%
    % \hfill%
    \vskip\baselineskip%
    \vspace{-0.75em}
    %%%%%%%%%%%%%%%%%%%%%%%%%%%%%%%%%%%%
    % \hfill
    \begin{subfigure}[b]{0.22\linewidth}
        \centering
	    \includegraphics[width=0.8\textwidth,  clip]{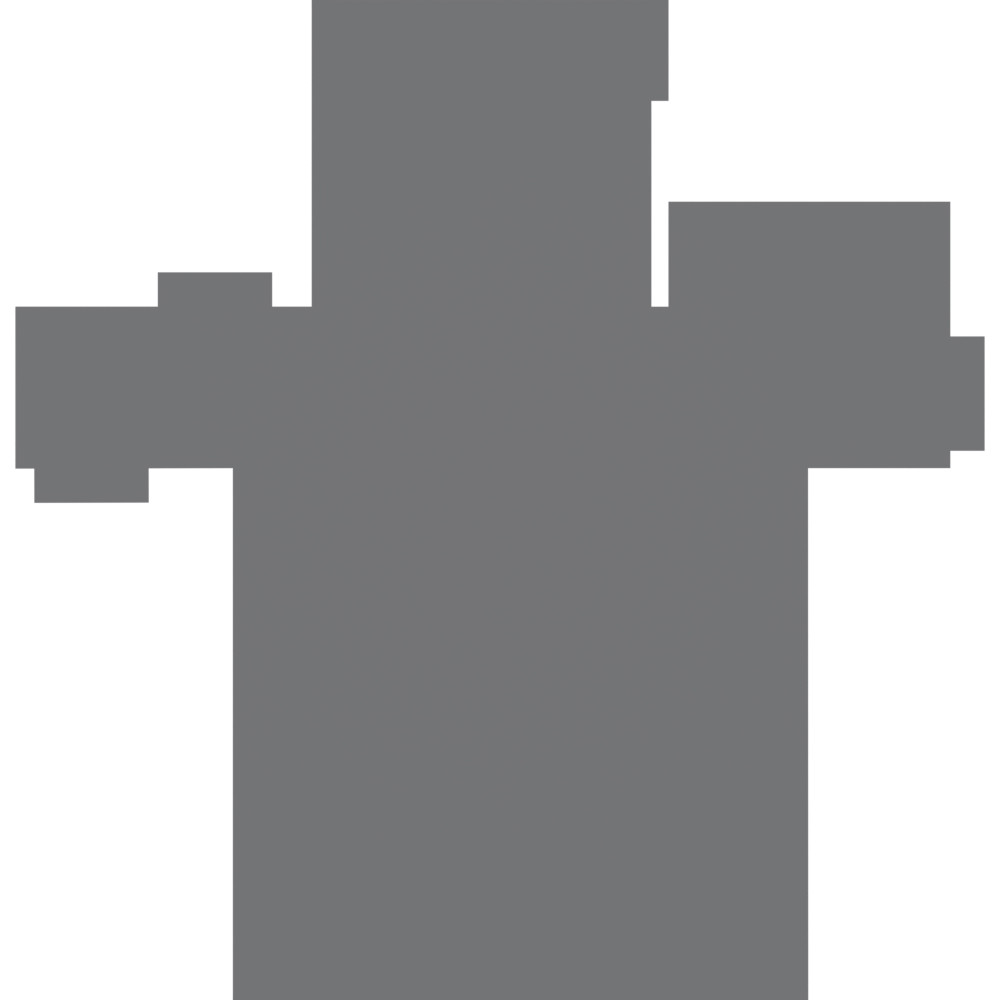}
    \end{subfigure}%
    \begin{subfigure}[b]{0.22\linewidth}
        \centering
            \begin{overpic}[width=0.8\textwidth,  clip]{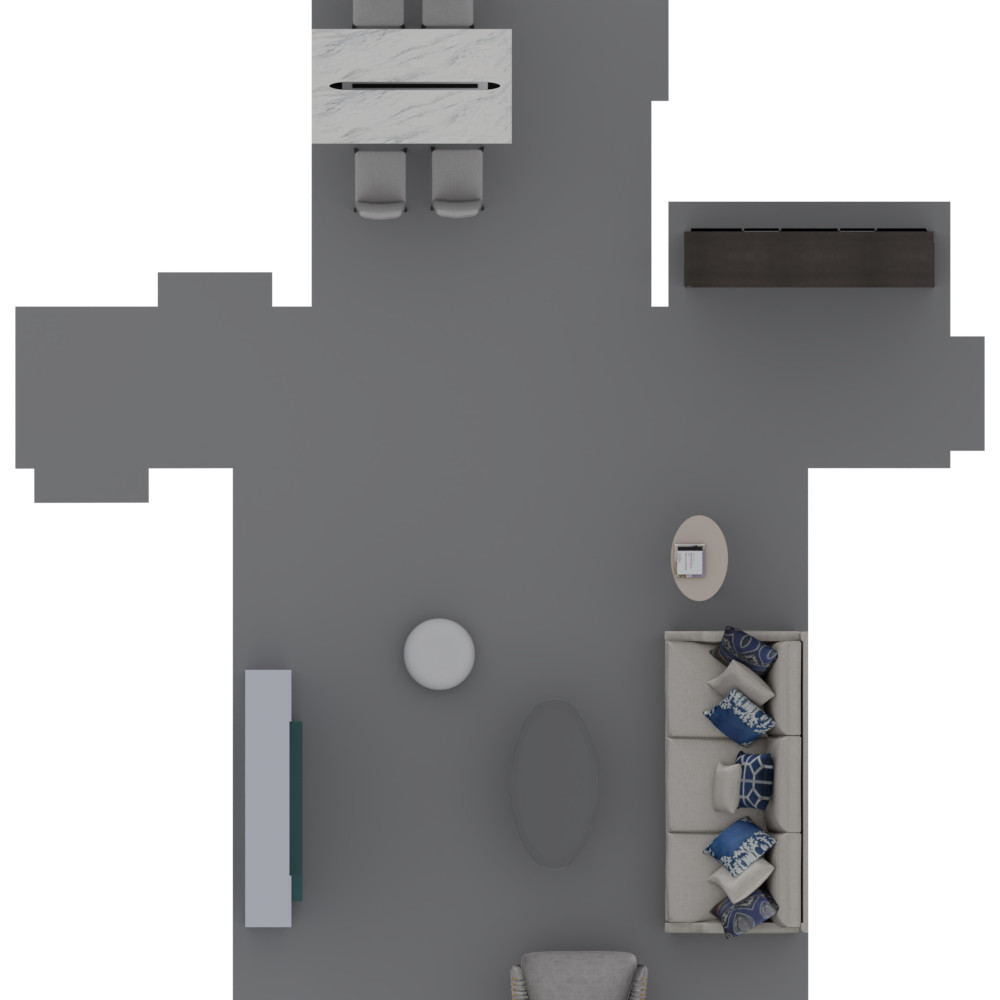}
        \end{overpic}
    \end{subfigure}%
    \begin{subfigure}[b]{0.22\linewidth}
        \centering
        \begin{overpic}[width=0.8\textwidth, clip]{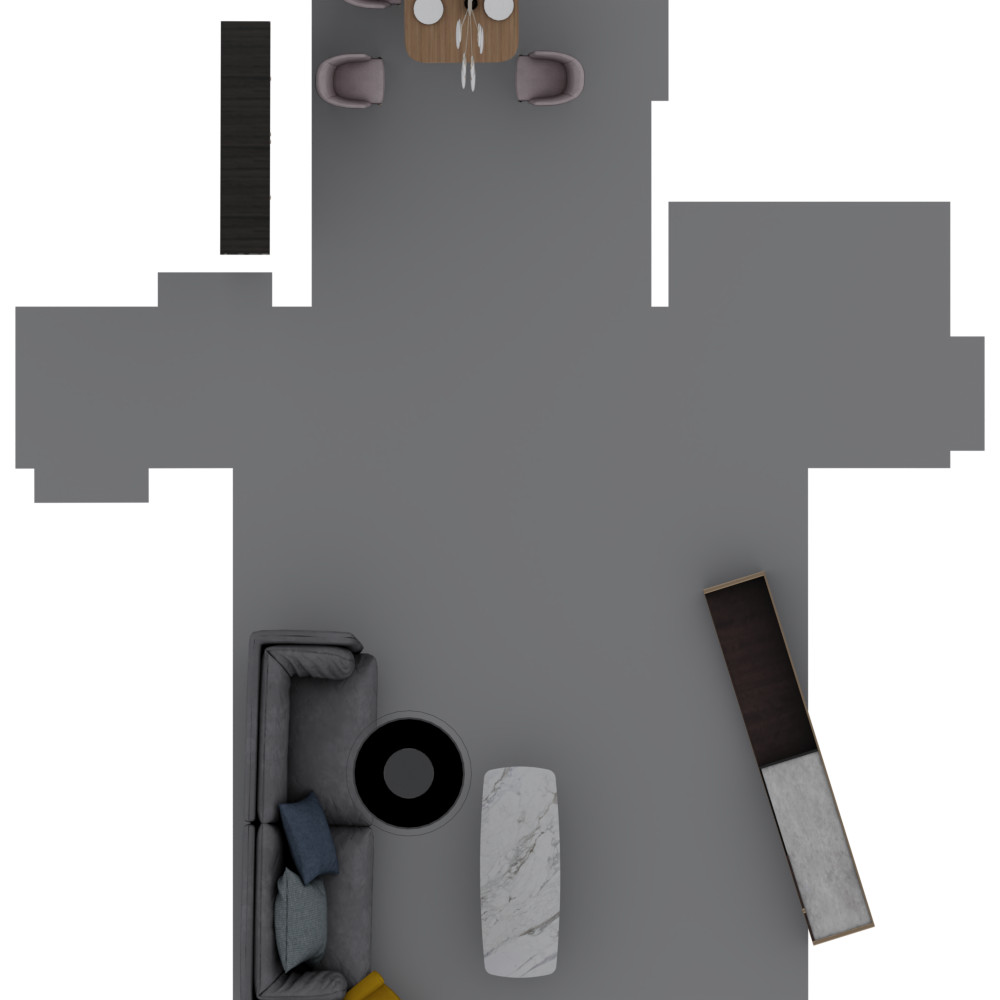}
	    \end{overpic}
    \end{subfigure}%
    \begin{subfigure}[b]{0.22\linewidth}
        \centering
        \begin{overpic}[width=0.8\textwidth, clip]{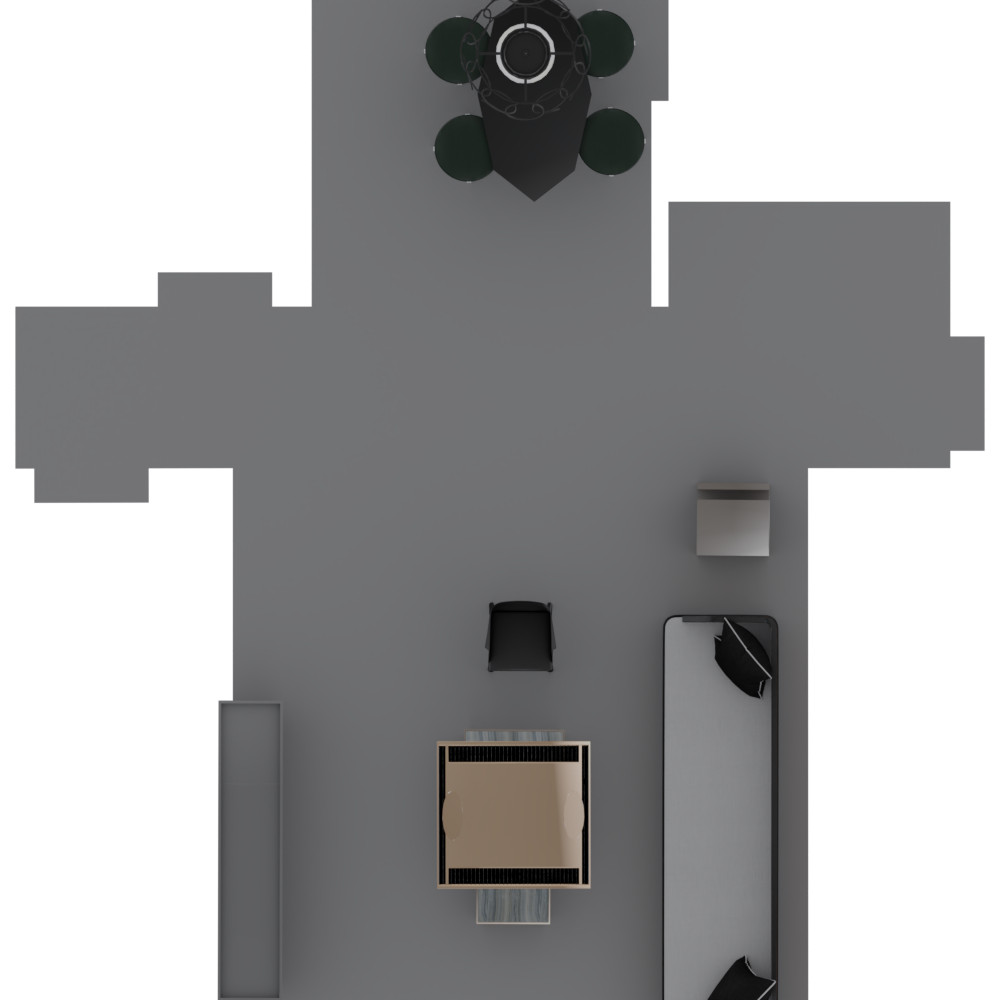}
        \end{overpic}
    \end{subfigure}%
    % \hfill%
    \vskip\baselineskip%
    \vspace{-0.75em}
    % \hfill
    %%%%%%%%%%%%%%%%%%%%%%%%%%%%%%%%%%%%%%%%%%%%%%%%%%%%%%%%%%%%%%%%%%%%%
    \begin{subfigure}[b]{0.22\linewidth}
        \centering
	    \includegraphics[width=\textwidth, clip]{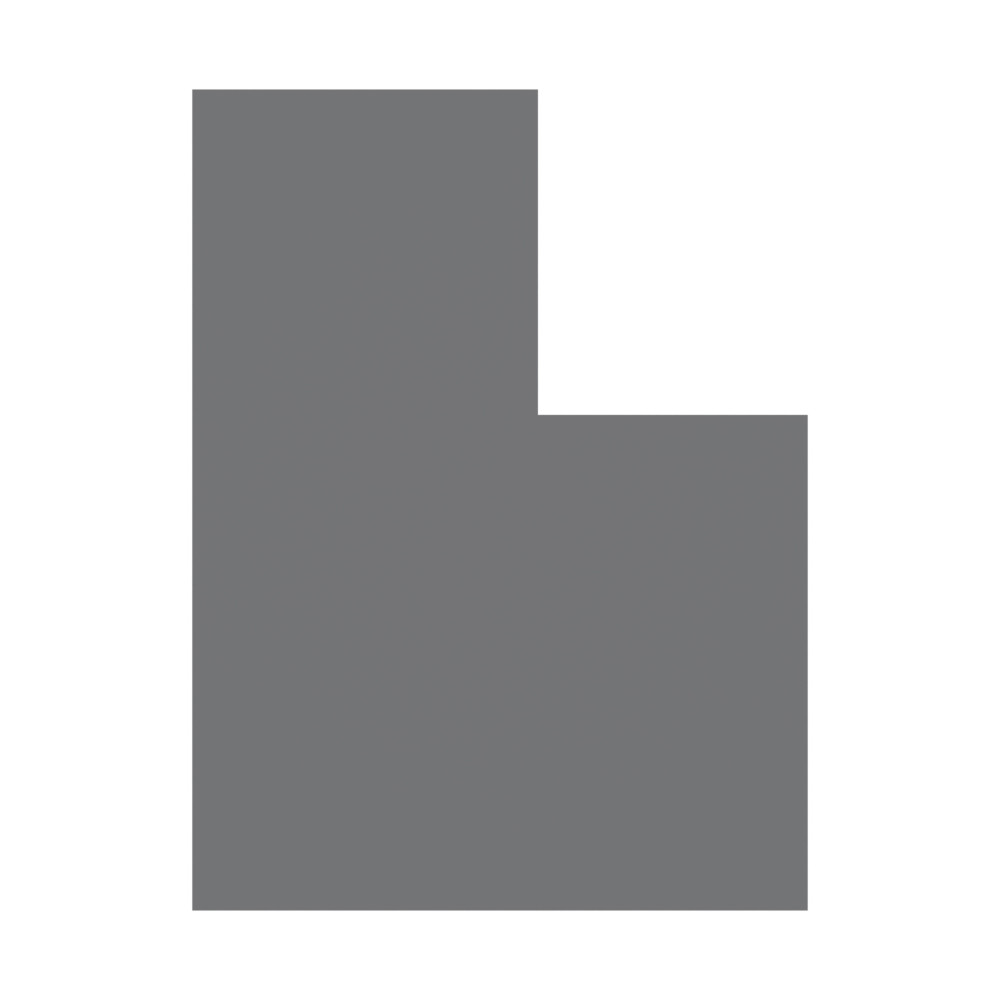}
    \end{subfigure}%
    \begin{subfigure}[b]{0.22\linewidth}
        \centering
        \begin{overpic}[width=\textwidth,  clip]{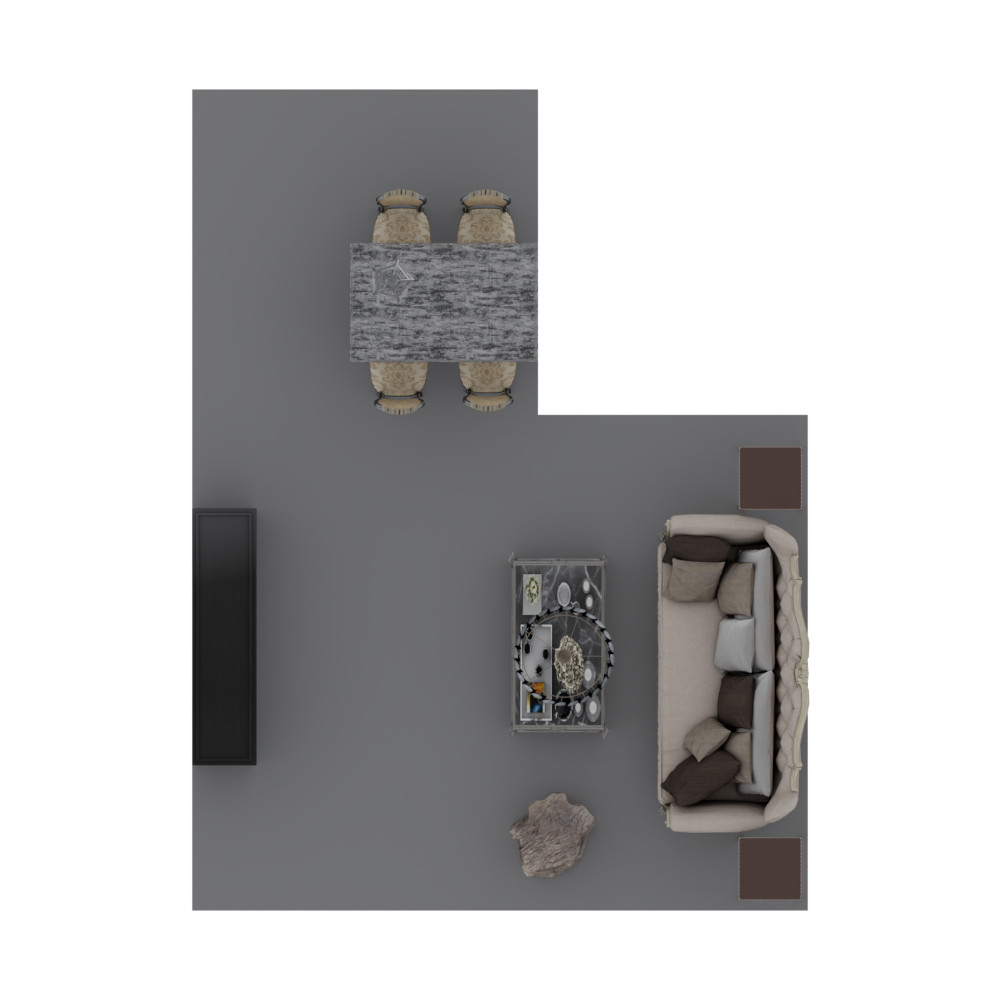}
        \end{overpic}
    \end{subfigure}%
    \begin{subfigure}[b]{0.22\linewidth}
        \centering
        \begin{overpic}[width=\textwidth, clip]{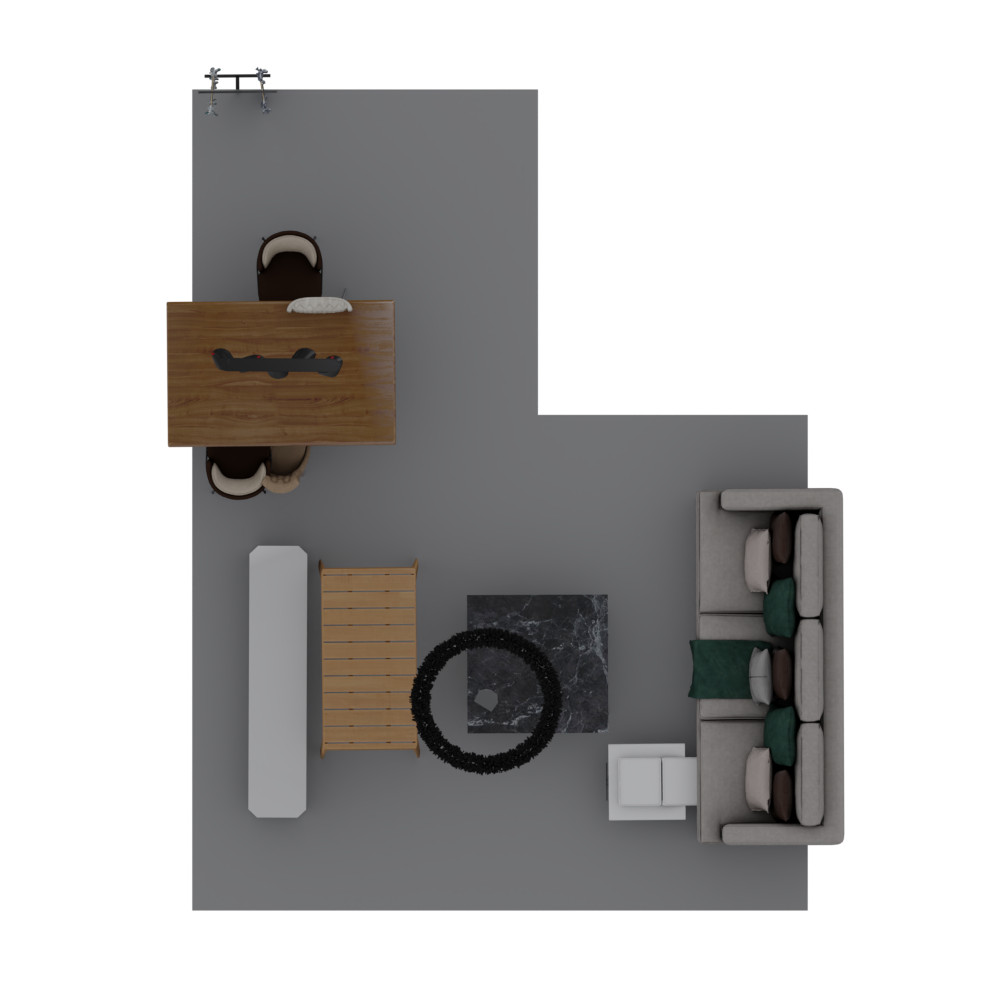}
	    \end{overpic}
    \end{subfigure}%
    \begin{subfigure}[b]{0.22\linewidth}
        \begin{overpic}[width=\textwidth,  clip]{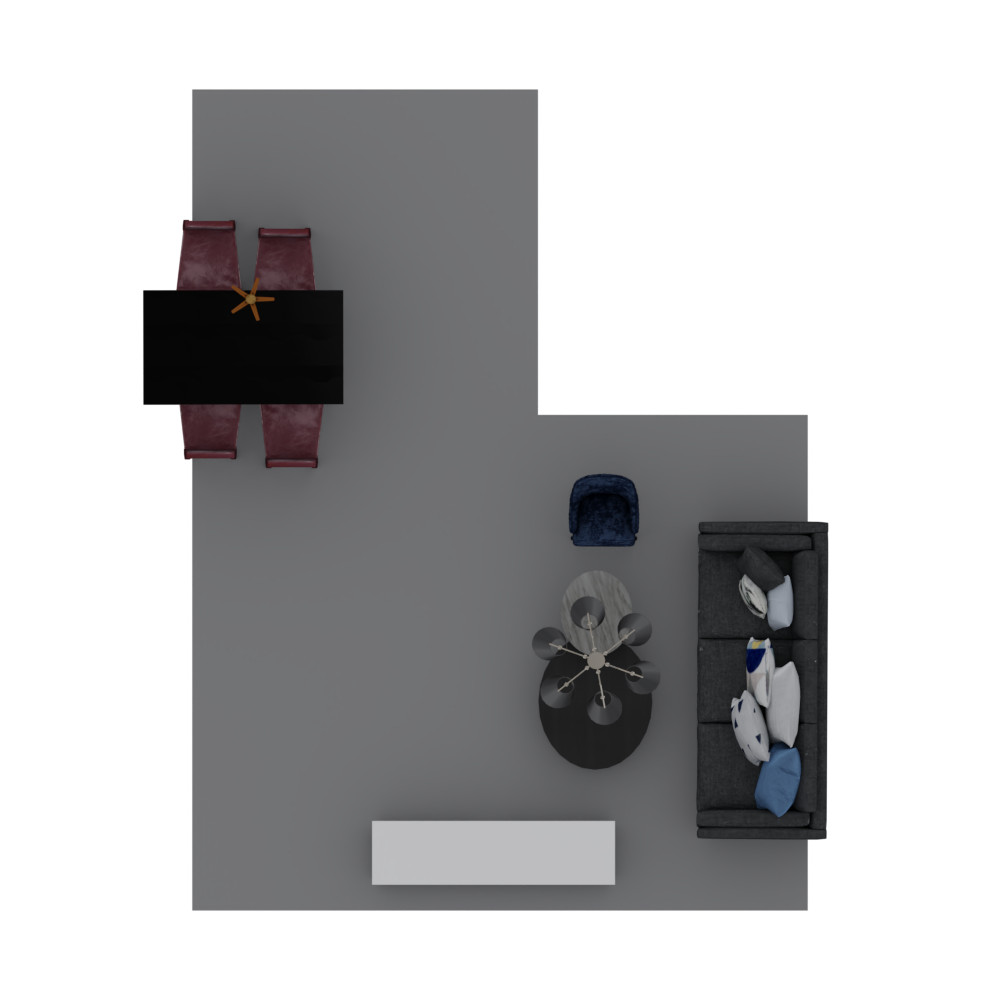}
        \end{overpic}
    \end{subfigure}%
    % \hfill%
    \vskip\baselineskip%
    \vspace{-0.75em}
    % \hfill
    %%%%%%%%%%%%%%%%%%%%%%%%%%%%%%%%%%%%%%%%%%%%%%%%%%%%%%%%%%%%%%%%%%%%%%%%%%%%%%%%%%%%%%%%%%%%%%%%%%%5
    \begin{subfigure}[b]{0.22\linewidth}
        \centering
	    \includegraphics[width=\textwidth, clip]{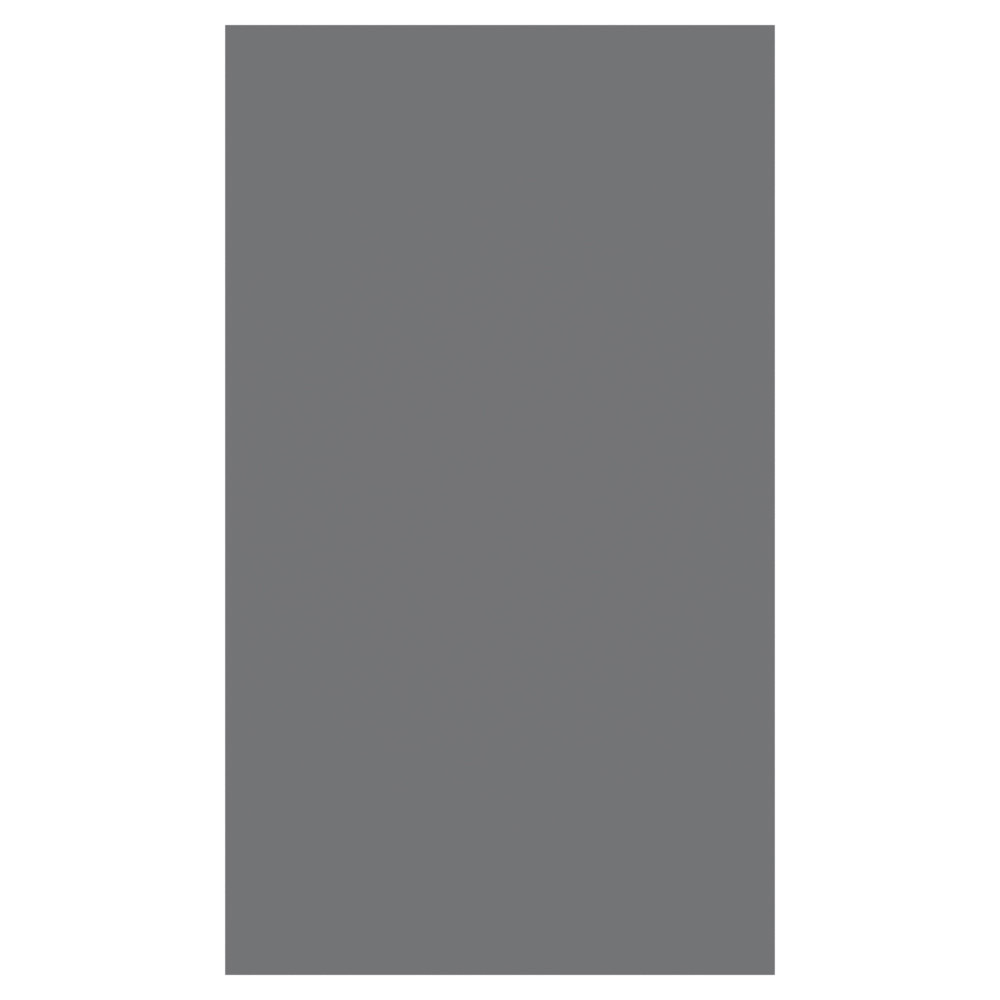}
    \end{subfigure}%
    \begin{subfigure}[b]{0.22\linewidth}
        \centering
        \begin{overpic}[width=\textwidth,  clip]{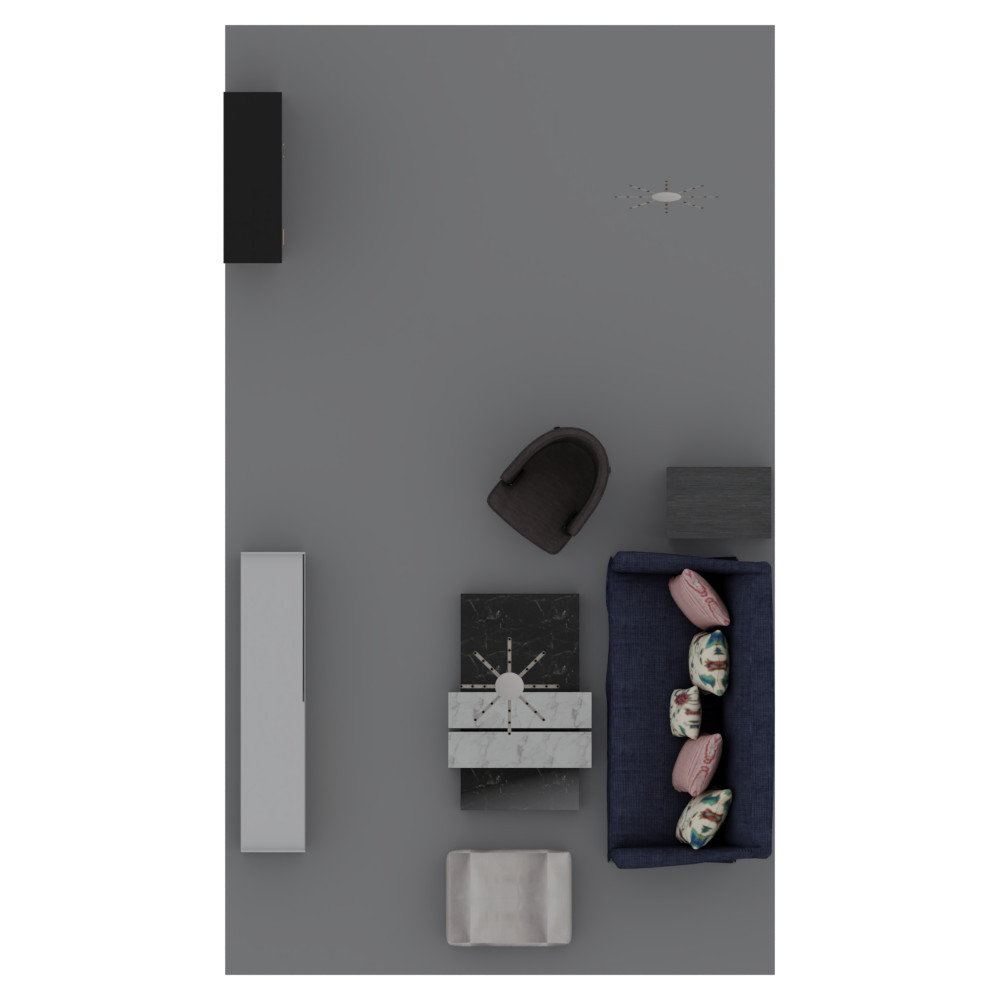}
        \end{overpic}
    \end{subfigure}%
    \begin{subfigure}[b]{0.22\linewidth}
        \centering
        \begin{overpic}[width=\textwidth, clip]{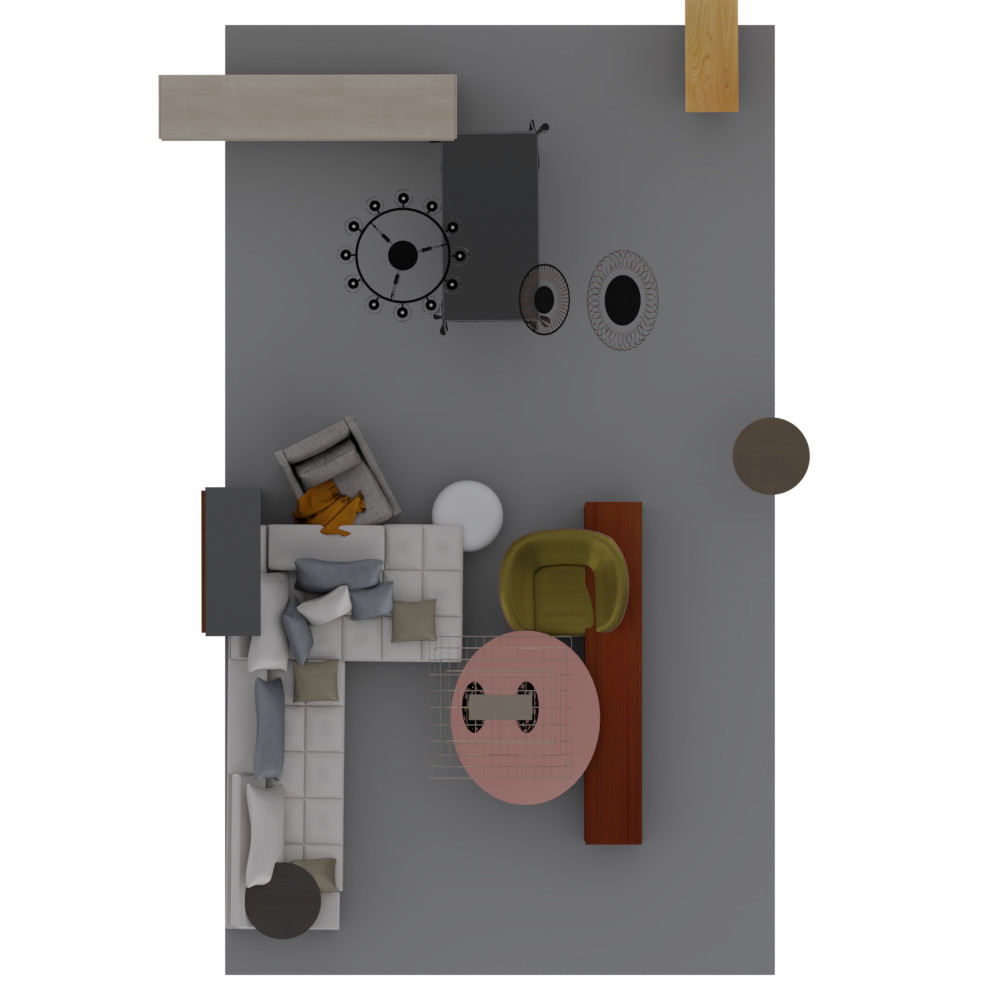}
	    \end{overpic}
    \end{subfigure}%
    \begin{subfigure}[b]{0.22\linewidth}
        \begin{overpic}[width=\textwidth,  clip]{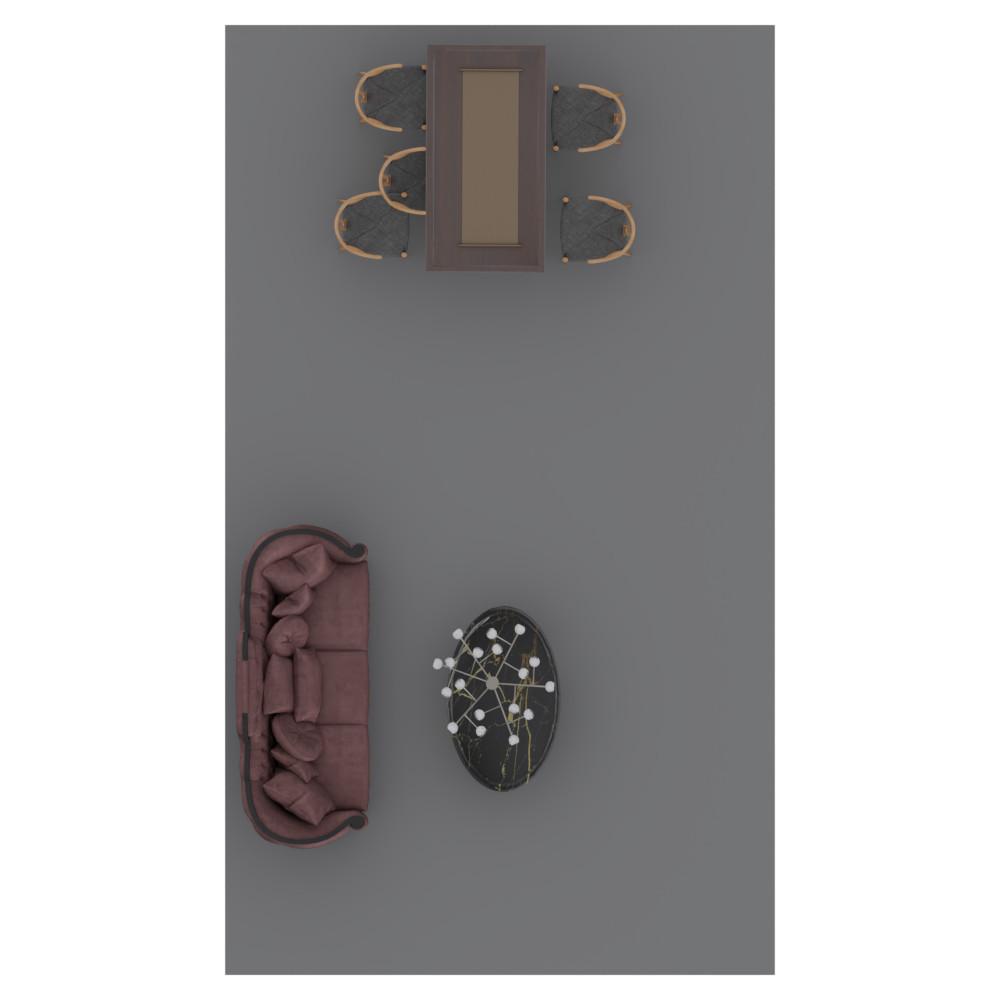}
        \end{overpic}
    \end{subfigure}%
    % \hfill%
    \vskip\baselineskip%
    \vspace{-0.75em}
    \begin{subfigure}[b]{0.22\linewidth}
        \centering
	    \includegraphics[width=\textwidth, clip]{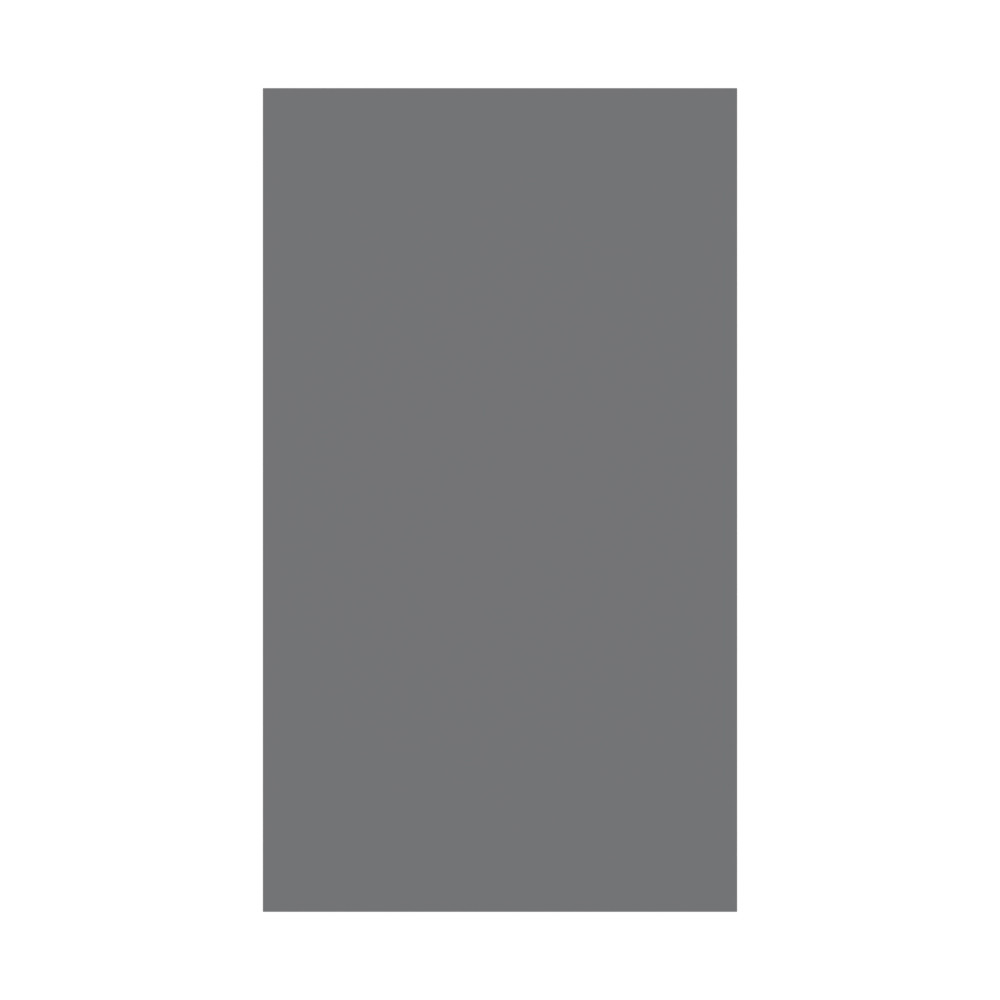}
    \end{subfigure}%
    \begin{subfigure}[b]{0.22\linewidth}
        \centering
        \begin{overpic}[width=\textwidth,  clip]{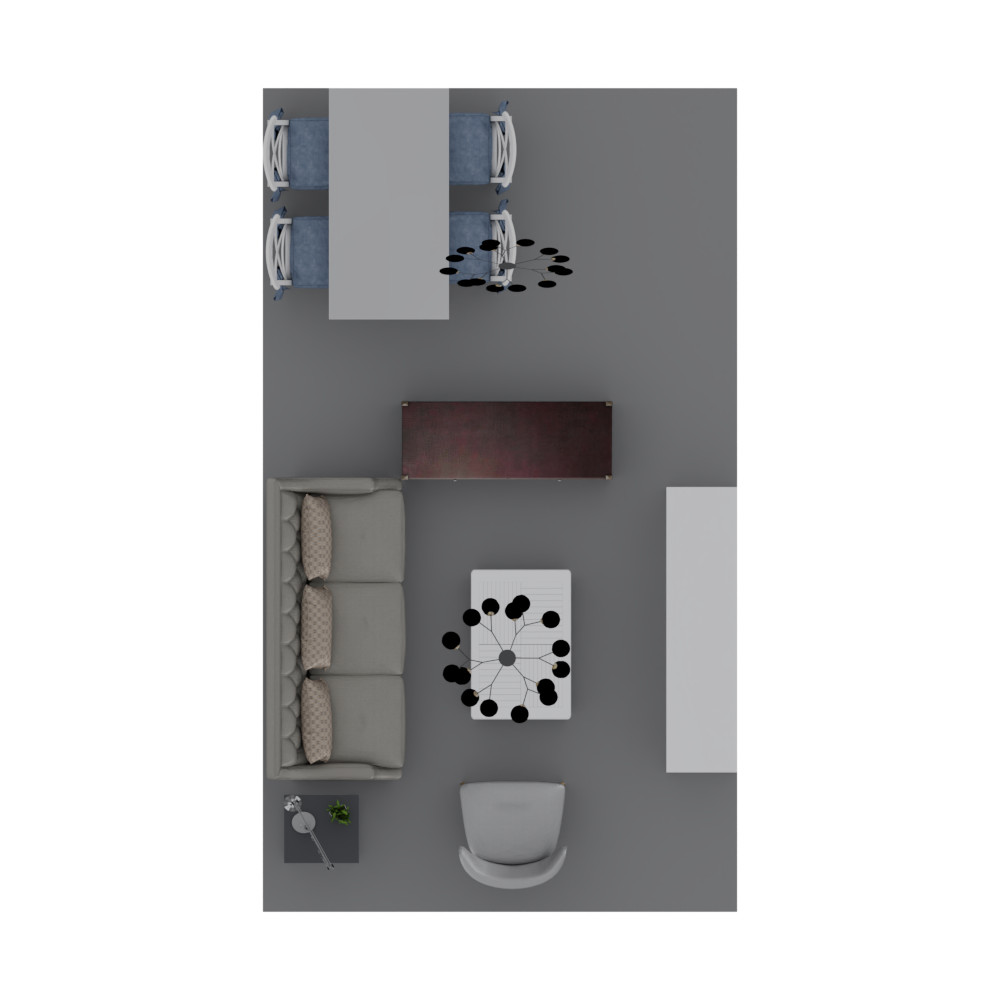}
        \end{overpic}
    \end{subfigure}%
    \begin{subfigure}[b]{0.22\linewidth}
        \centering
        \begin{overpic}[width=\textwidth,  clip]{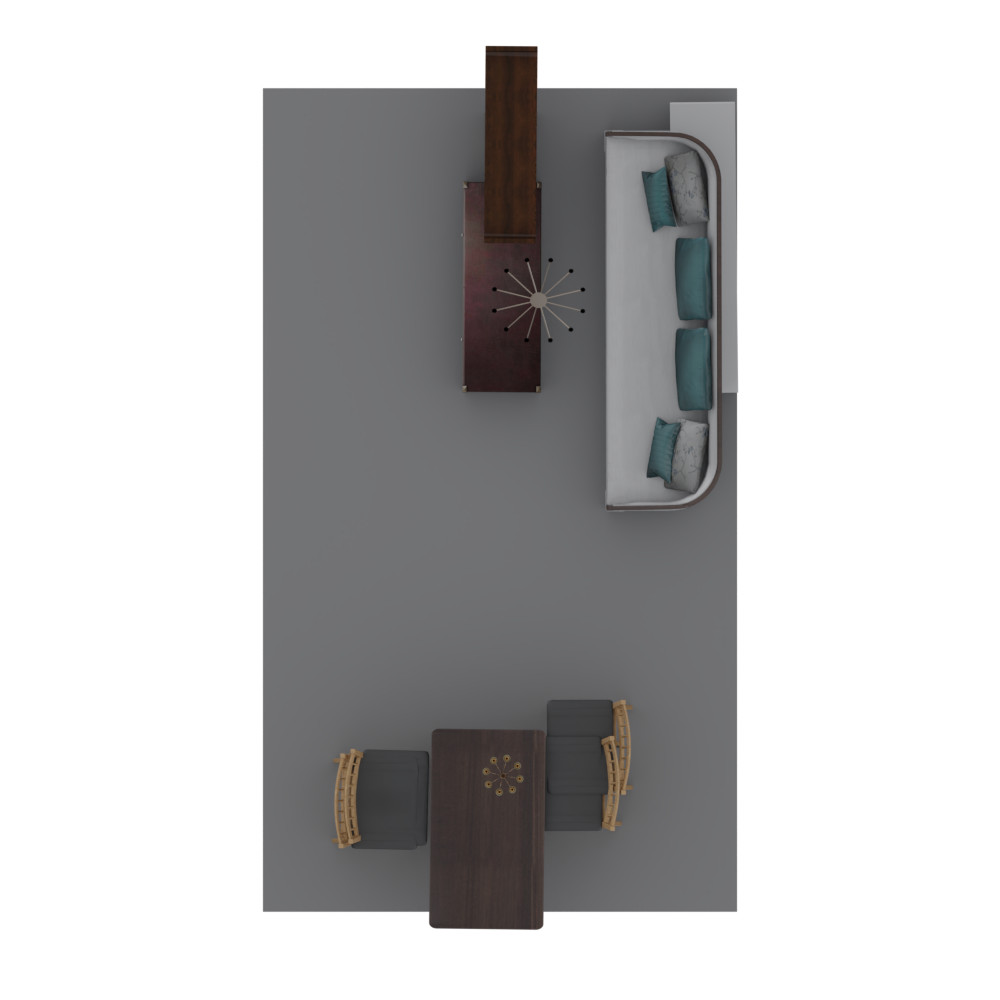}
	    \end{overpic}
    \end{subfigure}%
    \begin{subfigure}[b]{0.22\linewidth}
        \begin{overpic}[width=\textwidth,  clip]{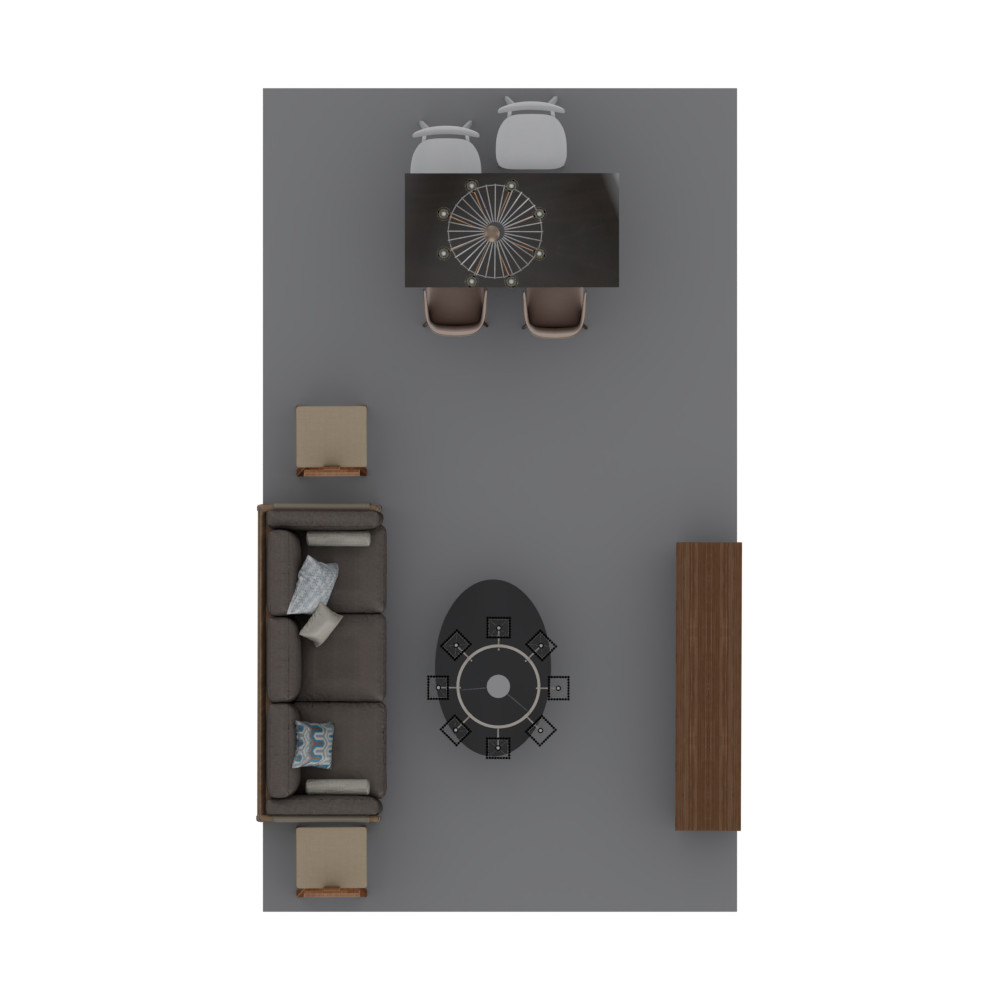}
        \end{overpic}
    \end{subfigure}%
    % \hfill%
    \vskip\baselineskip%
    \vspace{-0.75em}
    \begin{subfigure}[b]{0.22\linewidth}
        \centering
	    \includegraphics[width=0.8\textwidth, clip]{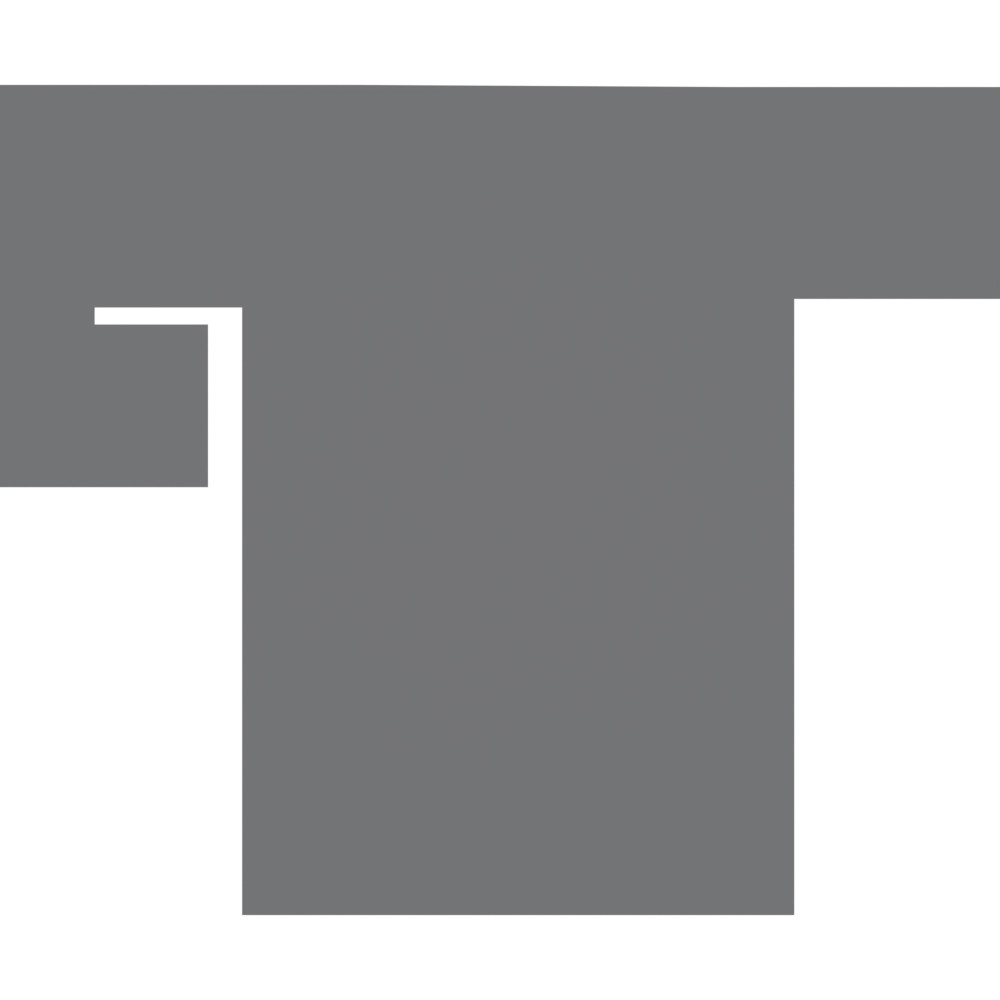}
    \end{subfigure}%
    \begin{subfigure}[b]{0.22\linewidth}
        \centering
        \begin{overpic}[width=0.8\textwidth,  clip]{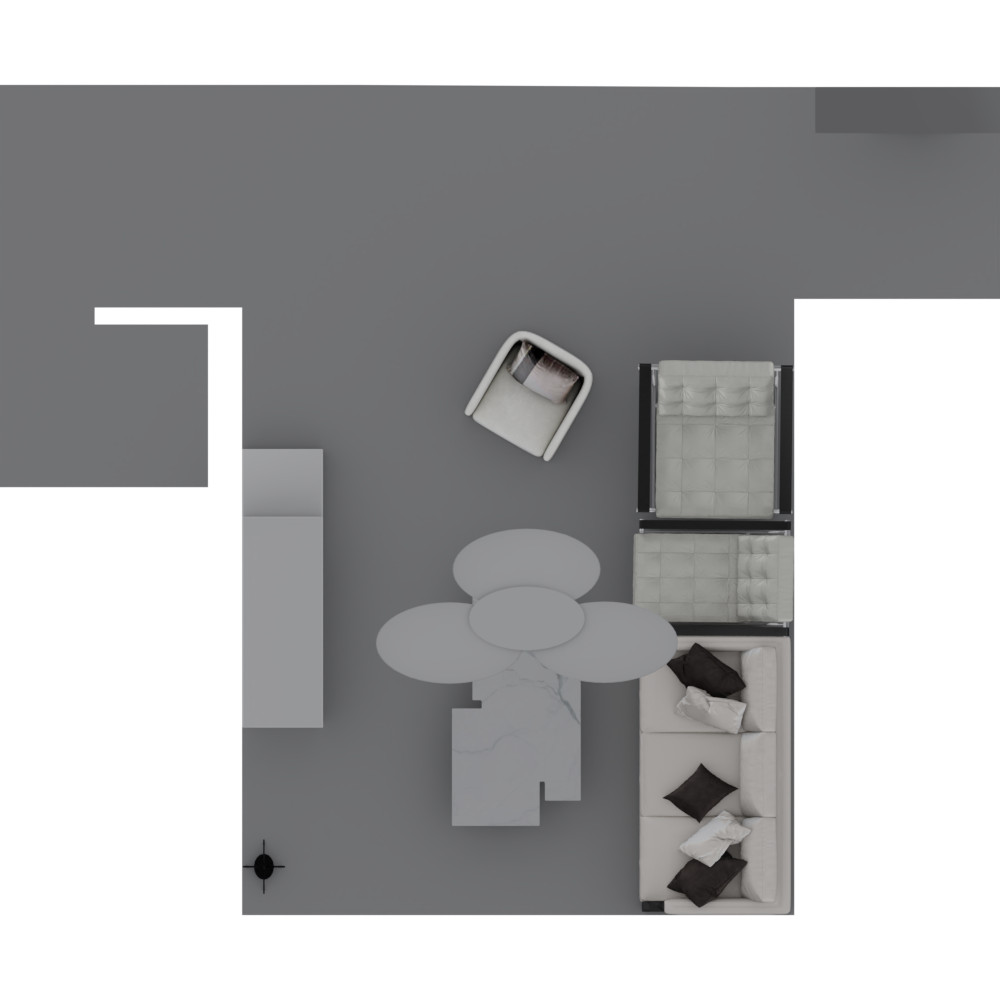}
        \end{overpic}
    \end{subfigure}%
    \begin{subfigure}[b]{0.22\linewidth}
        \centering
        \begin{overpic}[width=0.8\textwidth,  clip]{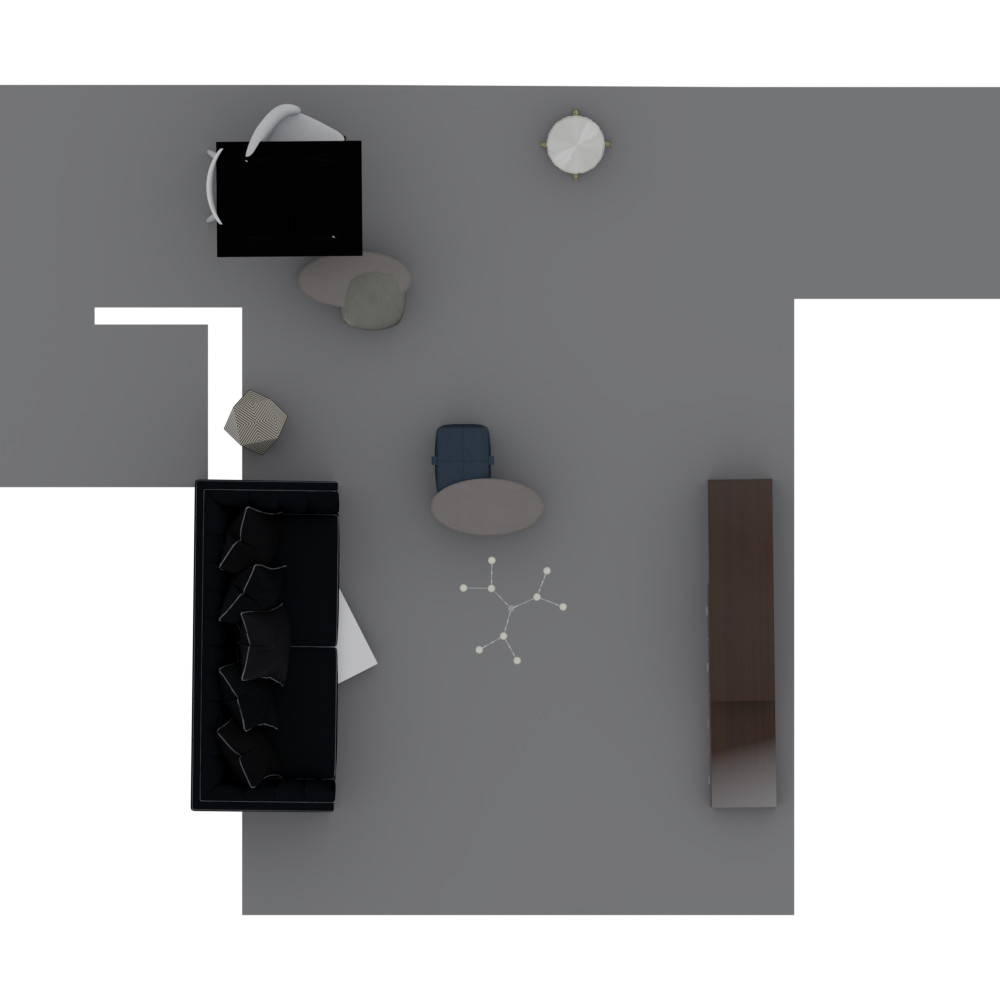}
	    \end{overpic}
    \end{subfigure}%
    \begin{subfigure}[b]{0.22\linewidth}
        \begin{overpic}[width=0.8\textwidth,  clip]{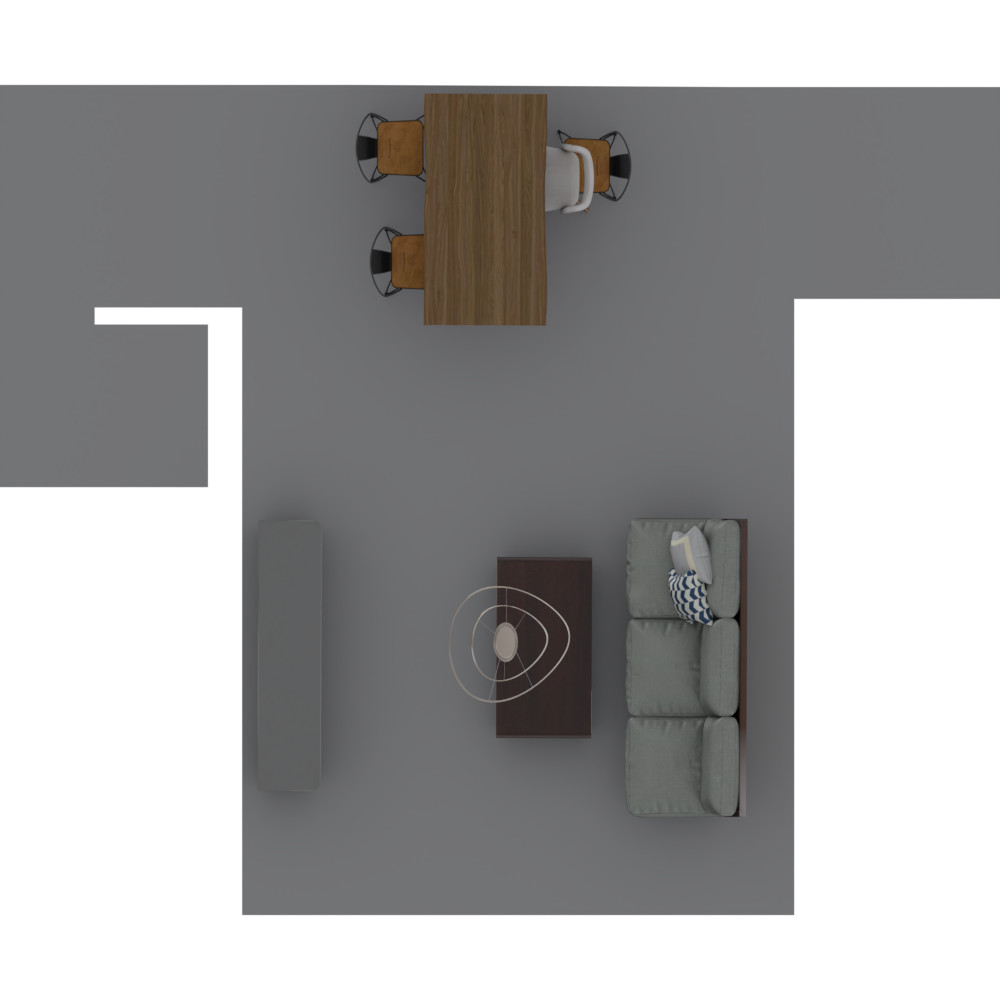}
        \end{overpic}
    \end{subfigure}%
    % \hfill%
    \vskip\baselineskip%
    \vspace{-0.75em}
    \begin{subfigure}[b]{0.22\linewidth}
        \centering
	    \includegraphics[width=\textwidth, clip]{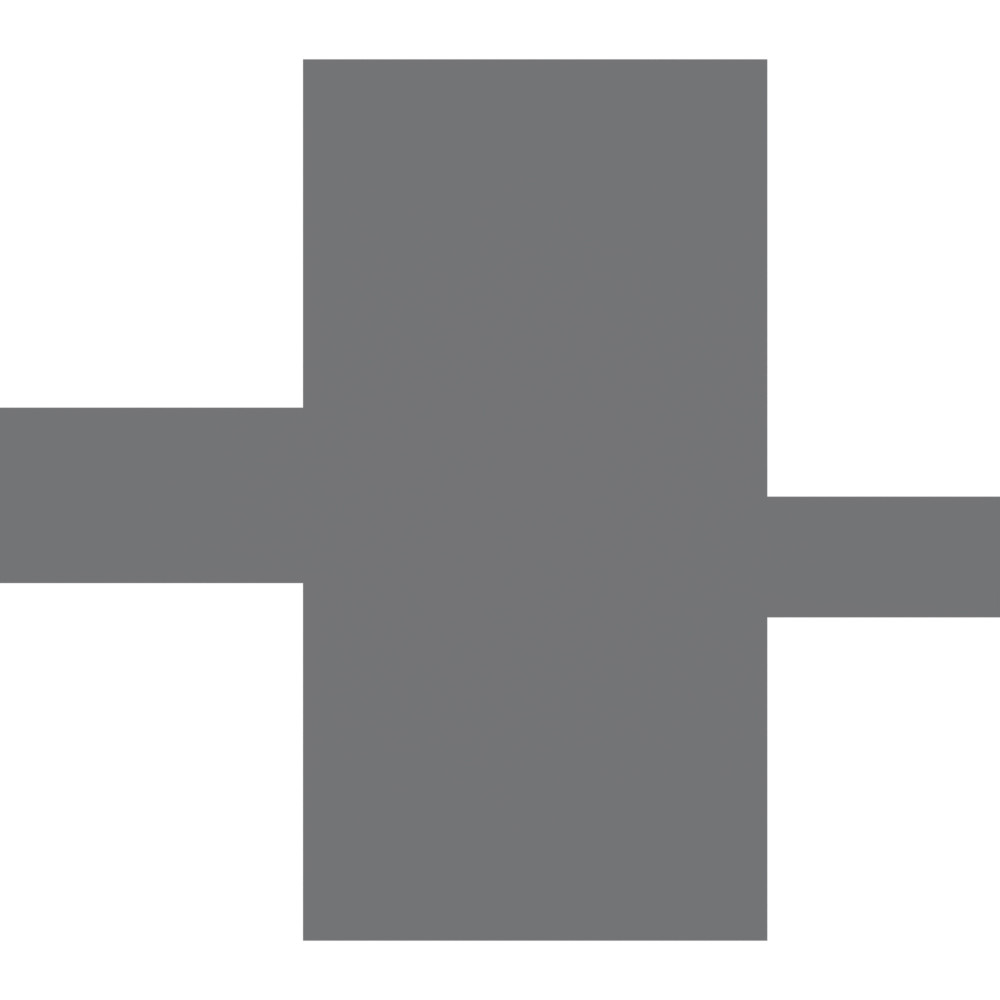}
    \end{subfigure}%
    \begin{subfigure}[b]{0.22\linewidth}
        \centering
        \begin{overpic}[width=\textwidth,  clip]{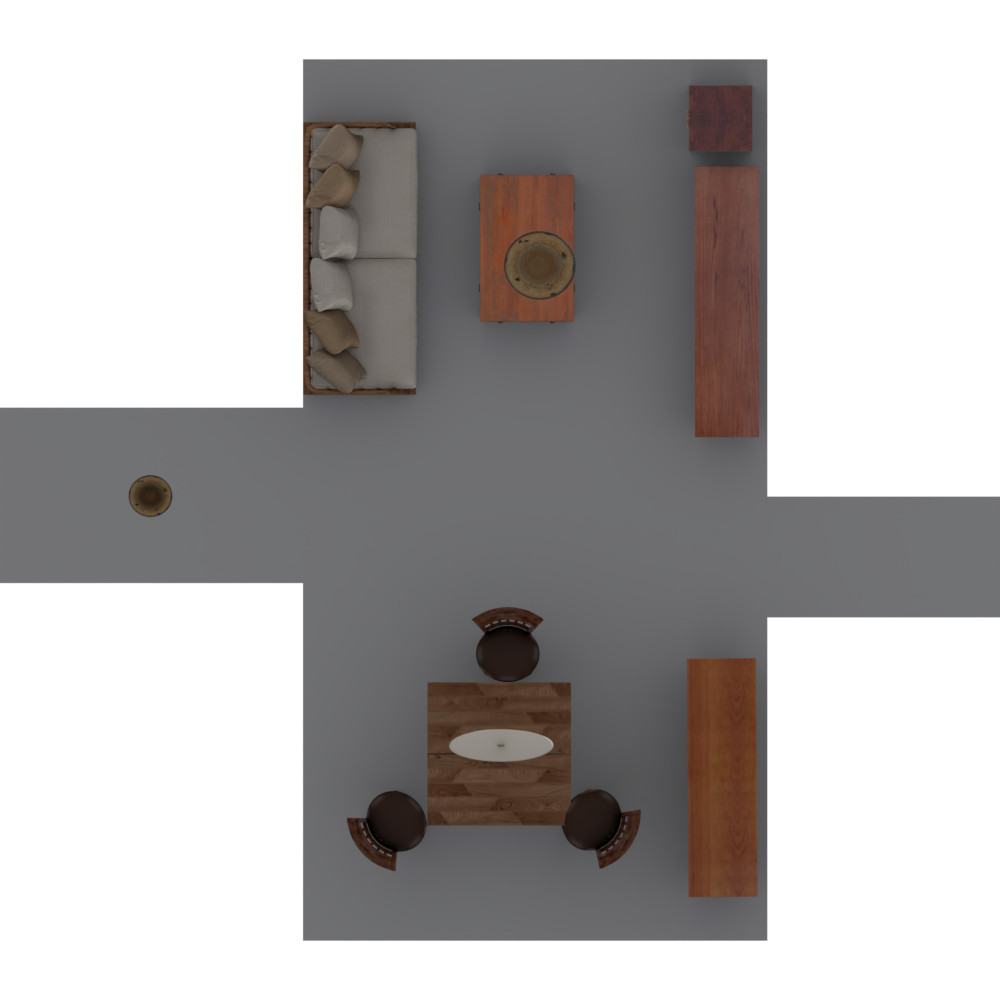}
        \end{overpic}
    \end{subfigure}%
    \begin{subfigure}[b]{0.22\linewidth}
        \centering
        \begin{overpic}[width=\textwidth,  clip]{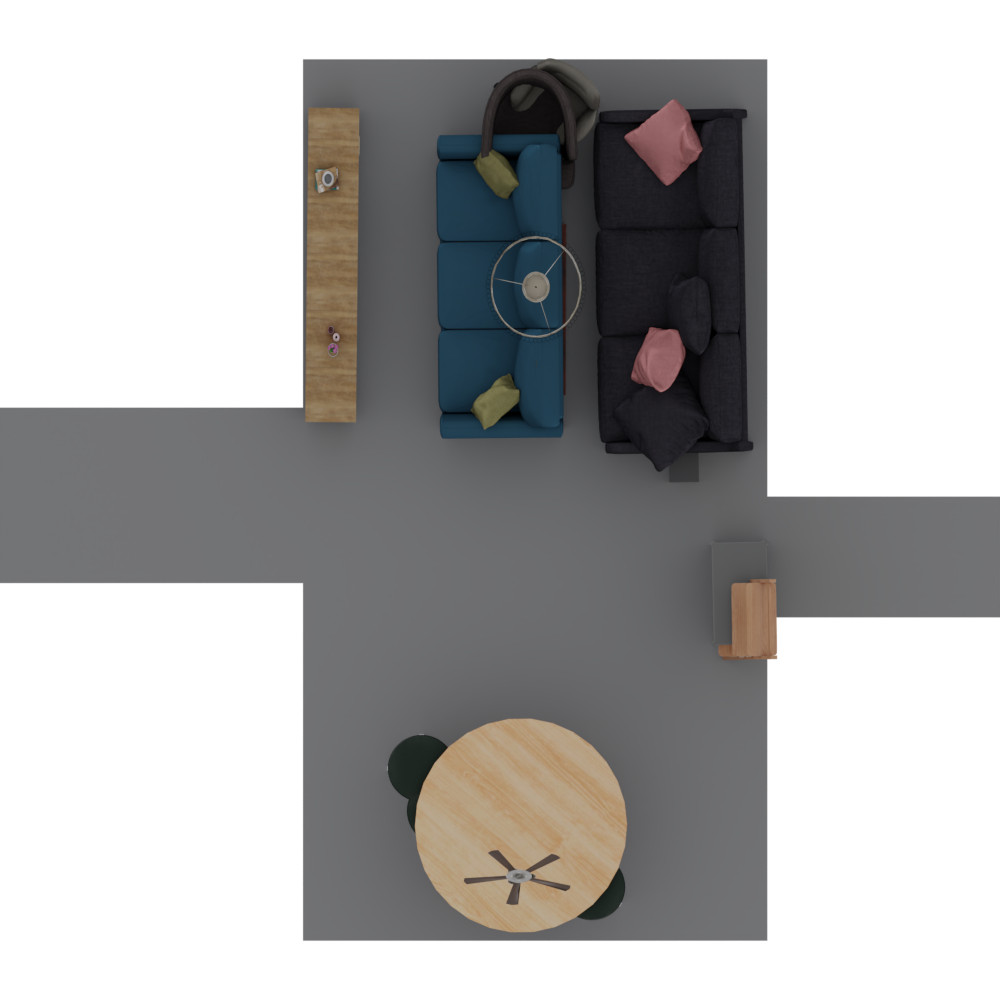}
	    \end{overpic}
    \end{subfigure}%
    \begin{subfigure}[b]{0.22\linewidth}
        \begin{overpic}[width=\textwidth, clip]{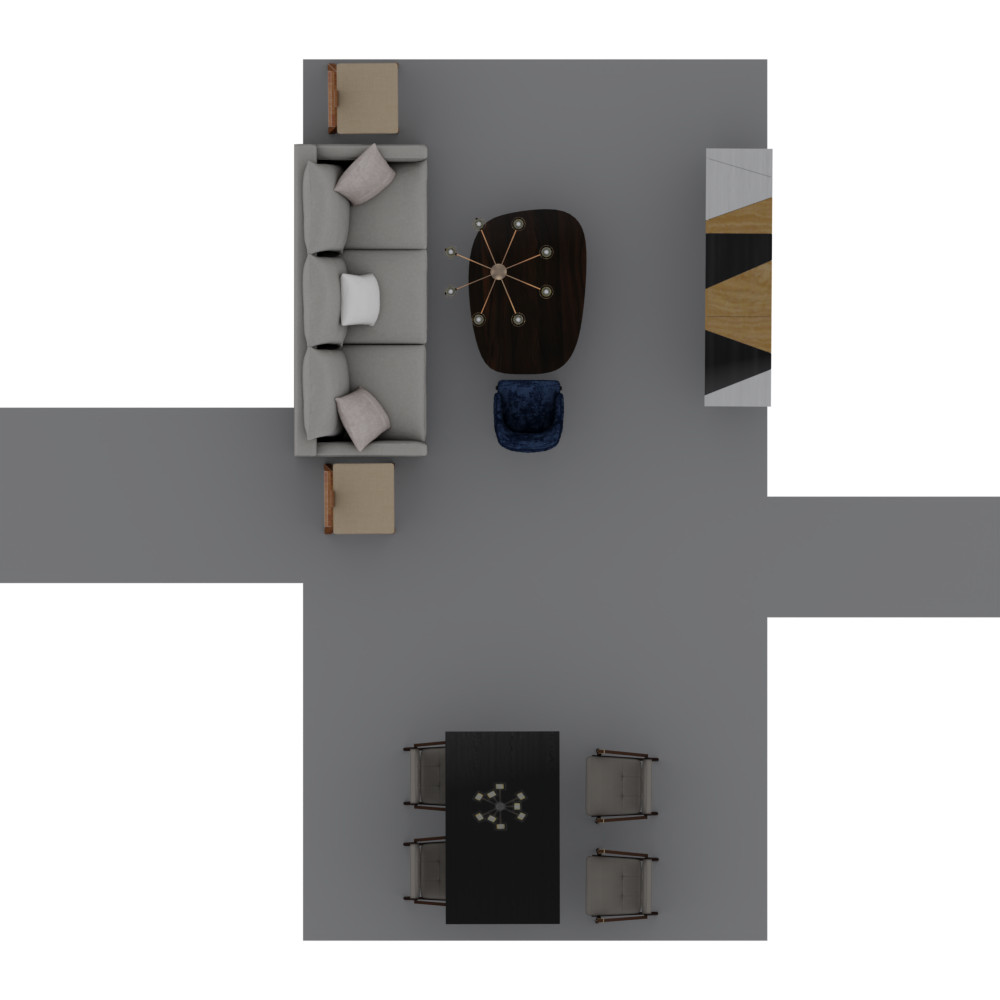}
        \end{overpic}
    \end{subfigure}%
    
    \caption{\textbf{Scene generation from scratch:} We compare generated scenes from GT, ATISS, and our model on \liv class.}
    \label{fig:addl_din}
\vspace{-0.75em}
\end{figure*}

\newpage
\FloatBarrier
\begin{table}[h]
  \centering
  \caption{Summary of key notation used in the paper.}
  \label{tab:notation}
  \begin{tabular}{@{}cl@{}}
  \toprule
  Symbol                    & Description                                                                                                                                                               \\ \midrule
  
  $\mathcal{I}$                       & \begin{tabular}{l} A binary image representation of the floorplan boundary. \end{tabular}                                                                                                                          \\ \midrule  
 
  $g_\phi$                       & \begin{tabular}{l} The Condition encoder. Implemented as Transformer Encoder with Bidirectional Attention. \end{tabular}                                                                                                                          \\ \midrule
  $f_\theta$                       & \begin{tabular}{l} The Generative Model. Implemented as Transformer Decoder with Causal Attention. \end{tabular}                                                                                                                          \\ \midrule
  
  $g_\psi^\mathcal{I}$                       & \begin{tabular}{l} The Boundary Encoder. An untrained ResNet-18 model. \end{tabular}                                                                                                                          \\ \midrule \midrule %%%% End of network
  $\mathcal{M}/\mask$                     & \begin{tabular}{l} A learnable token representing a missing value which the Generative Model tries to predict. \end{tabular}                                                                                                                    \\ \midrule
  $C$                     & \begin{tabular}{l} The sequence of tokens describing the condition.\\ It is the input to the Condition Encoder. \end{tabular}                                                                 \\ \midrule
  $c_i$                   & \begin{tabular}{l} The $i$-th element of $C$. \end{tabular}
                                                    \\ \midrule

  $C^g$                     & \begin{tabular}{l} The output of the last layer of the Condition Encoder. \\ Encodes conditions from $C$ and boundary $\mathcal{I}$. \end{tabular}                                                    \\ \midrule  %%%% end of sequences
  
  $S$                       & \begin{tabular}{l} The sequence representing the layout. \end{tabular}                                                    \\ \midrule  %%%% end of sequences
  
  $S^{GT}$                   & \begin{tabular}{l} The sequence representation of the Ground Truth layout. \end{tabular}
                                                    \\ \midrule        
  
  $s_i$                   & \begin{tabular}{l} The $i$-th element of $S$. \end{tabular}
                                                    % \\ \midrule 
\\ \bottomrule %%%%%%%%% end of model
  \end{tabular}
\end{table}

% Optionally include extra information (complete proofs, additional experiments and plots) in the appendix.
% This section will often be part of the supplemental material.

% \begin{figure*}[t!]

%     \fbox{\includegraphics[width=0.15\linewidth, trim=580  300 400 100 ]{figures/changed.jpg}}
% % \vspace{-0.5cm}
% \end{figure*}

\end{document}

%% file: symbols.tex
\newcommand{\bc}{\mathbf{c}}

\newcommand{\bq}{\mathbf{q}}
\newcommand{\br}{\mathbf{r}}
\newcommand{\bs}{\mathbf{s}}
\newcommand{\bt}{\mathbf{t}}

\newcommand{\cB}{\mathcal{B}}

\newcommand{\cI}{\mathcal{I}}

\newcommand{\cL}{\mathcal{L}}

%% file: sections/introduction.tex
\section{Introduction}
Automatic generation of realistic assets enables content creation at a scale that is not possible with traditional manual workflows. It is driven by the growing demand for virtual assets in both the creative industries, virtual worlds, and increasingly data-hungry deep model training. 3D scene and layout generation plays a central role in automatic asset generation, as much of the demand is for the types of real-world scenes we see and interact with every day, such as building interiors.

Deep generative models for assets like images, videos, 3D shapes, and 3D scenes have come a long way to meet this demand. 
%Generative models have come a long way to generate high quality images, videos, objects, and 3D scenes.
In the context of 3D scene and layout modeling, in particular auto-regressive models based on transformers enjoy great success. Inspired by language modeling, these architectures treat layouts as sequences of tokens and are particularly well suited for modeling spatial relationships between elements of a layout. For example, Para et al.~\cite{para} generate two-dimensional interior layouts with two transformers, one for furniture objects and one for spatial constraints between these objects, while SceneFormer~\cite{wang2021sceneformer} extends interior layout generation to 3D.

A main limitation of these approaches is that they do not support scene completion from arbitrary partial scenes, due to their need for a consistent sequence ordering. In bedroom layouts, for example, the bed always needs to be generated before the nightstands, which precludes completing scenes that already have nightstands, but are missing a bed. ATISS~\cite{Paschalidou2021NEURIPS}, which is the most current layout generation approach, tackles this problem by randomly permuting the token sequence during training, enabling scene completion from arbitrary subsets of objects.

%. In these methods, the sequence representation of a scene needs to have a consistent ordering (in bedroom layouts, for example, the bed always needs to be generated before the nightstands), therefore a scene that already has nightstands in it cannot 

%sequences need to 

%ATISS~\cite{ATISS} is the most current approach. 

%The main idea is to enable 

%The two main ideas are to shorten the token sequence by using one token per objects instead of 

%(Many of these architecture ideas are inspired by language modedling).
%One idea is to model objects and constraints with separate transformers (Wamiq ICCV 2021)

%MaskGit

%Scene Former

%ATIS is the most current model. The two main ideas are to encode each object as one token, opposed to enconding each attribute as separate token. Therefore the model uses fewer tokens to describe a scene. The second main idea is to enable sampling objects in arbitrary order and conditioning on objects given in arbitrary order (conditioning on arbitrary subset of objects).

While ATISS works well, we aim to improve on these results to enable more fine grained conditioning. We want to keep the advantage of conditioning on arbitrary subsets of objects, but we also want to extend to conditioning on arbitrary subsets of attributes. For example, a user might be interested to ask for a room with a table and two chairs, without specifying exactly where these objects should be located. Another example is to perform object queries for given geometry attributes. The user could specify the location of an object and query the most likely class, orientation, and size of an object at the given location. Our model thereby extends the baseline ATISS with new functionality while retaining all its existing properties and performance.

The main technical difficulty in achieving a more fine-grained conditioning is due to the autoregressive nature of the generative model. Tokens in the sequence that define a scene are generated iteratively, and each step only has information about the previously generated tokens. Thus, the condition can only be given at the start of the sequence, otherwise some generation steps will miss some of the conditioning information.
The main idea of our work is to allow for fine-grained conditioning using two mechanisms: (i)~Like ATISS, we train our generator to be approximately permutation-invariant and provide the condition as partial sequence that needs to be completed by the generative model.
Unlike previous work, the condition is not restricted to the start of the sequence, which means that some tokens do not have full information about the condition through the autoregressive generator alone. (ii)~To give our autoregressive model knowledge of the entire conditioning information in each step, we additionally use a transformer encoder that provides cross-attention over the complete conditioning information in each step. These two mechanisms allow us to accurately condition on arbitrary subsets of the token sequence, for example, only on tokens corresponding to specific object attributes.

%both through a transformer encoder and by manually fixing a subset of the object sequence.

%- add a transformer encoder that encodes the partial scene and allows the transformer to plan ahead for , while also manually 

In our experiments, we demonstrate four applications: (i) outlier detection, (ii) unconditional generation, (iii) traditional scene completion from a partial set of objects, and (iv) fine-grained conditioning on a subset of object attributes. We compare to three current state-of-the-art layout generation methods~\cite{fastsynth:ritchie,wang2021sceneformer,Paschalidou2021NEURIPS} and show performance that is on par or superior, while also enabling fine-grained conditioning, which, to the best of our knowledge, is currently not supported by any existing layout generation method.

%show that we can use this method to successfully complete scenes given a list of partial object attributes. ...

%(Insert Figure to compare ATISS and New Model)

% To summarize, our contributions are:
% \begin{enumerate}
%     \item Proposing a Masked Language Model for heterogeneous tokens instead of homogeneous tokens.
%     \item A strategy for augmenting the token sequence to let the transformer distinguish between different objects and different token types.
%     \item Using MLM scoring to identify outliers in a set which enables failure correction.
%     \item Breaking the Autoregressive nature of ATISS and enabling arbitrary conditioning order.
% \end{enumerate}

% \begin{figure*}[t]
%     \centering
%     \includegraphics[width=0.8\linewidth, height=1in]{figures/arch.png}
%     \caption{Schematic representation of our proposed method.}
%     \label{fig:architecture}
    
% \end{figure*}

% \newlength{\mywidth}
% \newsavebox{\mybox}
% \begin{figure*}
%     \centering
%     \savebox{\mybox}{\includegraphics[width=\linewidth]{figures/retrieval_clearcoat_transparent.png}}
%     % \settowidth{\mywidth}{\usebox{\mybox}}
%     \includegraphics[width=0.29\linewidth]{sections/retrieval_glossy_bsdf_walls_shadow_attempt2.png}
%     \includegraphics[width=0.29\linewidth]{sections/retrieval_glossy_bsdf_walls_shadow_attempt2.png}
%     \caption{Examples of our generated floorplans.}
%     \label{fig:my_label}

% \end{figure*}

% \begin{figure}
%     \centering
%     \fcolorbox{black}{gray!30}{\includegraphics{figures/blank.pdf}}
%     \caption{Caption}
%     \label{fig:my_label}
% \end{figure}

%% file: sections/related_work.tex
\section{Related Work}

\begin{table}[t]
\centering
\setlength\tabcolsep{2pt}
\resizebox{\columnwidth}{!}{
    \begin{tabular}{@{}lcccccc@{}} 
    \toprule
    % \multicolumn{1}{l}{} & \multicolumn{4}{c}{CAS ($\downarrow$)}    & \multicolumn{4}{c}{KL-Divergence $\times 10^3$ ($\downarrow$)}       \\ \cmidrule(l){2-5} \cmidrule(l){6-9}
        & \makecell{Generative \\ Model}      & Sampling  & Representation &  Throughput &  \makecell{Likelihood\\ Estimation } & \makecell{Arbitrary \\ Conditioning} \\ \midrule
      FastSynth &        CNN+VAE               &        AR   & Set+Sequence &      Low          &      No       &              \xmark                                 \\
      SceneFormer &      Transformer              &      AR  & Sequence     &      Low          &     Yes        &            \xmark                                  \\
      ATISS  &           Transformer             &       AR   & Set &          High      &        Yes     &              \xmark                                 \\
      Ours &             Transformer               &      NAR/AR     &  Set &       High        &     Yes        &          \cmark                 \\
    
    \bottomrule
    \end{tabular}
}
\vspace{0.1cm}
\caption{\textbf{Comparison of our proposed model to other state-of-the art models}: Our model is the only model that can perform non-autoregressive sampling, has a set representation, is lightweight, can estimate the likelihood of an existing scene and also allows for arbitrary conditioning on any object parameter. }
\vspace{-0.8cm}
\end{table}

We discuss recent work that we draw inspiration from. In particular, we build on previous work in Indoor Scene Synthesis, Masked Language Models, and Set Transformers.

\para{Indoor Scene Synthesis} Before the rise of deep-learning methods, indoor scene synthesis methods relied on layout guidelines developed by skilled interior designers, and an optimzation strategy such that the adherence to those guidelines is maximized~\cite{makeithome:yu, example_furniture:fisher, scalable:weiss}. Such optimization is usually based on sampling methods like simulated annealing, MCMC, or rjMCMC. Deep learning based methods, e.g.~\cite{planit:wang, fastsynth:ritchie, wang2021sceneformer, Paschalidou2021NEURIPS} are substantially faster and can better capture the variability of the design space. The state-of-the-art methods among them are autoregressive in nature. All of these  operate on a top-down view of a partially generated scene. PlanIT and FastSynth then autoregressively generate the rest of the scene. FastSynth uses separate CNNs+MLPs to create probability distributions over location, size and orientation and categories. PlanIT on the other hand generates graphs where nodes are objects and edges are constraints on those objects. Then a scene is instantiated by solving a CSP on that graph.

Recent methods, SceneFormer~\cite{wang2021sceneformer} and ATISS~\cite{Paschalidou2021NEURIPS} use transformer based architectures to sidestep the problem of rendering a partial scene which makes PlanIT and FastSynth slow. This is because using a transformer allows the model to accumulate information from previously generated objects using the attention mechanism. SceneFormer flattens the scene into a structured sequence of the object attributes, where the objects are ordered lexicographically in terms of their position. It then trains a separate model for each of the attributes. ATISS breaks the requirement of using a specific order by training on all possible permutations of the object order and removing the position encoding. In addition, it uses a single transformer model for all attributes and relies on different decoding heads which makes it substantially faster than other models while also having significantly fewer parameters.

\para{Masked Language Models} Masked Language Models (MLMs) like BERT~\cite{devlin-etal-2019-bert}, ROBERTa~\cite{liu2019roberta}, and BART~\cite{lewis-etal-2020-bart} have been very successful in pre-training for language models. These models are pretrained on large amounts of unlabeled data in an unsupervised fashion, and are then fine-tuned on a much smaller labeled dataset. These fine-tuned models show impressive performance on their corresponding downstream tasks. However, the generative capability of these models has not been much explored except by Wang et al. in \cite{wang2019bert}, which uses a Gibbs-sampling approach to sample from a pre-trained BERT model. Follow up work in Mansimov et al.~\cite{mansimov2020a}, proposes more general sampling approaches. However, the sample quality is still inferior to autoregressive models like GPT-2~\cite{radford2019language} and GPT-3~\cite{brown2020language}.
More recently, MLMs have received renewed interest especially in the context of image-generation~\cite{issenhuth2021edibert, chang2022maskgit}. MaskGit~\cite{chang2022maskgit} shows that with a carefully designed masking schedule, high quality image samples can be generated from MLMs with parallel sampling which makes them much faster than autoregressive models. Edi-BERT~\cite{issenhuth2021edibert} shows that the BERT masking objective can be succesfully used with a VQGAN~\cite{taming} representation of an image to perform high quality image editing. Our model most closely resembles BART when used as a generative model.

\para{Set Transfomers} Zaheer et al.~\cite{deepsets} introduced a framework called DeepSets providing a mathematical foundation for networks operating on set-structured data. A key insight is that operations in the network need to be permutation invariant. Methods based on such a formulation were extremely successful, especially in the context of point-could processing~\cite{pointnet, ravanbakhsh2016deep}. Transformer models without any form of positional encoding are permutation invariant by design. Yet, almost all the groundbreaking works in transformers use some from of positional encoding, as in objection detection~\cite{detr}, language generation~\cite{radford2019language, brown2020language}, and image-generation~\cite{chang2022maskgit}. One of the early attempts to use a truly permutation invariant set transformer was in Set Transformer~\cite{lee2019set}, who methodically designed  principled operations that are permutation invariant but could only achieve respectable performance in toy-problems. However, recent work based on \cite{lee2019set} shows impressive performance in 3d-Object Detection~\cite{he2022voxset}, 3d Pose Estimation~\cite{ugrinovic2022permutation}, and SFM~\cite{moransfm}. 

%% file: sections/method.tex
\section{Method}

\begin{figure*}[t!]
    \centering
    \includegraphics[width=\linewidth, height=1.67in]{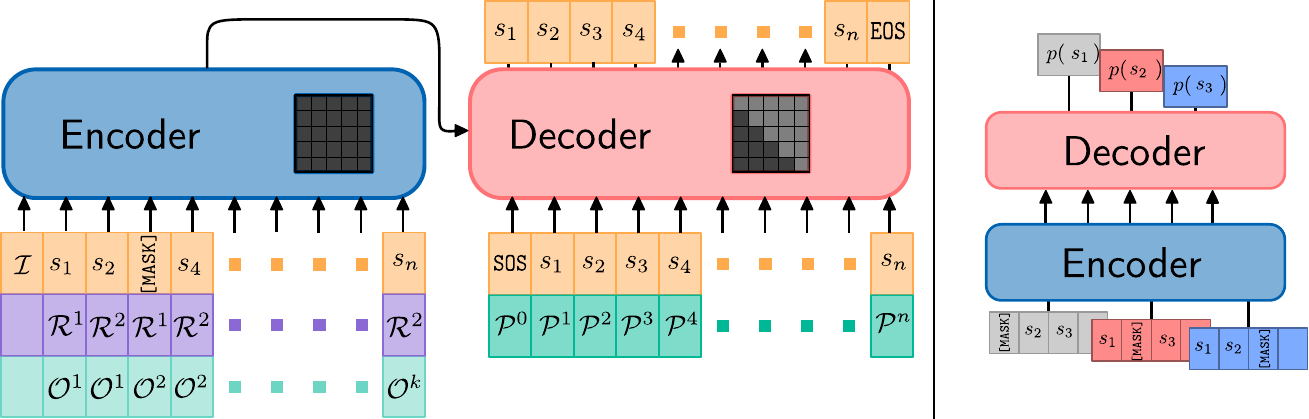}
 \caption{COFS Overview. (\textbf{Left}): The model is a BART-like encoder-decoder model, with a bidirectional encoder and an autoregressive decoder. Note the attention matrices shown in grey. At training time, a permutation $\pi$ permutes the objects in a scene. The encoder receives additional information in the form of \textit{Relative Position Tokens} $\mathcal{R}^i$, and the \textit{Object Index Tokens} $\mathcal{O}^i$.  A random proportion of tokens is replaced with a \mask token. The decoder is trained with \textit{Absolute Position Tokens} $\mathcal{P}^i$ and performs two tasks - 1. copy-paste: the attributes should be sampled at the proper location 2: mask-prediction: the decoder predicts the actual value of the token corresponding to a \mask token in the encoder sequence. (\textbf{Right}): During inference, to measure likelihood, we create a copy of the sequence with each token masked out. The decoder outputs a probability distribution over the possible values of the masked tokens.}
    \label{fig:architecture}
\end{figure*}

%In this example, we use 2 object attributes and $n$ objects.

Our goal is to design a generative model of object layouts that allows for fine-grained conditioning on individual object attributes. Fine-grained conditioning enables more flexible partial scene specification, for example specifying only the number and types of objects in a scene, but not their positions, or exploring suggestions for plausible objects at given positions in the layout.

% the following three paragraphs are similar to the intro, they can be moved to the intro to replace the current paragraphs in the intro, or be commented out if we need more space or think this is too much repetition

\paragraph{Generative model.} We use a transformer-based generative model, as these types of generative models have shown great performance in the current state of the art. Originally, proposed as a generative model for language, transformer-based generative models represents layouts as a sequence of tokens $S=(s_1, \dots, s_n)$ that are generated auto-regressively; one token is generated at a time, based on all previously generated tokens:
\begin{equation}
    p(s_i | S_{<i}) = f_\theta(S_{<i}),
\end{equation}
where $p(s_i | S_{<i})$ is the probability distribution over the value of token $s_i$, computed by the generative model $f_\theta$ given the previously generated tokens $S_{<i} = (s_1, \dots, s_{i-1})$.
We sample from $p(s_i | S_{<i})$ to obtain the token $s_i$.
Each token represents one attribute of an object, and groups of adjacent tokens correspond to objects. More details on the layout representation are described in Section~\ref{sec:layout_rep}.

\paragraph{Conditioning.} To condition a transformer-based generative model on a partial sequence $C$, we can replace tokens of $S$ with the corresponding tokens of $C$, giving us the sequence $S^C$. This is done after each generation step, so that the probability for the token in each step is conditioned on $S^C_{<i}$ instead of $S_{<i}$:
\begin{eqnarray}
    p(s_i | S^C_{<i}) = f_\theta(S^C_{<i}).
\end{eqnarray}
%where $\mathbf{1}$ is the indicator function, and $c_i$ is element $i$ of C.
Each generated token $s_i$ in $S^C$ (i.e. tokens that are not replaced by tokens in $C$) needs to have knowledge of the full condition during its generation step, otherwise the generated value may be incompatible with some part of the condition. Therefore, since each generated token $s_i$ only has information about the partial sequence $S^C_{<i}$, the condition can only be given as start of the sequence:
\begin{equation}
S^C =
\begin{cases}
c_i \text{ if } i \le |C| \\
s_i \text{ otherwise}.
\end{cases}
\end{equation}
%, , so that all generated tokens $S_{>|C|}$ have full knowledge of the condition.

\paragraph{Conditioning without permutation invariance.} Typically both the objects and the attributes of the objects in the sequence are consistently ordered according to some strategy, for example based on a raster order of the object positions~\cite{para}, or on the object size~\cite{planit:wang}. Therefore, a generative model $f^{\text{ordered}}_\theta$ that is only trained to generate sequences in that order cannot handle different orderings, so that in general:
\begin{equation}
    f^{\text{ordered}}_\theta(S^C_{<i}) \neq f^{\text{ordered}}_\theta(\pi_o(S^C_{<i})),
\end{equation}
where $\pi_o$ is a random permutations of the objects in sequence $S_{<i}$. The consistent ordering improves the performance of the generative model, but also presents a challenge for conditioning:
%since each generated token $s_i$ only has information about the previously generated tokens $S_{<i}$, a condition $C \subset S$, such as a partial set of objects, can only be given as start of the sequence $C = S_{<j}$, so that all subsequently generated tokens $S_{\ge j}$ have full access to the condition:
% \begin{equation}
%     p(s_i) = f_\theta(s_i | S_{<i}),
%     f^{\text{ordered}}_\theta(s_i | S_{<i}) \neq f^{\text{ordered}}_\theta(s_i | \pi_o(S_{<i})),
% \end{equation}
%Thus,
the consistent sequence ordering limits the information that can appear in the condition. In a bedroom layout, for example, if beds are always generated before nightstands in the consistent ordering, the layout can never be conditioned on nightstands only, as this would preclude the following tokens from containing a bed.

\paragraph{Permutation invariance for more general conditioning.} Recent work~\cite{Paschalidou2021NEURIPS} tackles this issue by forgoing the consistent object ordering, and instead training the generator to be approximately invariant to permutations $\pi_o$ of objects in the sequence:
\begin{equation}
    f_\theta(S^C_{<i}) \approx f_\theta(\pi_o(S^C_{<i})),
\end{equation}
This makes generation more difficult, but allows conditioning on arbitrary subset of objects, as now arbitrary objects can appear at the start of the sequence. However, since only objects are permuted and not their attributes, it does not allow conditioning on subsets of object attributes. Permuting object attributes to appear at arbitrary positions in the sequence is not a good solution to obtain a more fine-grained conditioning, as this would make it very hard for the generator to determine which attribute corresponds to which object.

\paragraph{Fine-grained conditioning.} We propose to extend previous work to allow for fine-grained conditioning by using two different conditioning mechanisms, in addition to the approximate object permutation invariance: First, similar to previous work, we provide the condition as partial sequence $C$. To allow conditioning on only a subset of the object attributes, we introduce special mask tokens $\mathcal{M}$ in $C$ that denote tokens that are not constrained by $C$. The constrained sequence $S^C$, is then defined as:
%by replacing mask tokens in $C$ with generated tokens:
\begin{equation}
S^C =
\begin{cases}
c_i \text{ if } i \le |C| \text{ and } c_i \neq \mathcal{M} \\
s_i \text{ otherwise}.
\end{cases}
\end{equation}
% However, 
% allow replacing arbitrary tokens in $C$, for example individual object attributes, with special mask tokens $\mu$. During sequence generation, these mask tokens are replaced by generated tokens:
% \begin{eqnarray}
%     p(s_i | S_{<i}, C) = \begin{cases}
%     \mathbf{1}_{\{c_i\}}(s_i) \text{ if } i \le |C| \text{ and } c_i \neq \mu\\
%     f_\theta(S_{<i} ) \text{ otherwise.}
%     \end{cases}
% \end{eqnarray}
Second, to provide information about the full condition to each generated token, including those that replace mask tokens, we modify $f_\theta$ to use a transformer encoder $g_\phi$ that encodes the condition $C$ into a set of feature vectors that each generated token has access to:
\begin{equation}
p(s_i | S^C_{<i}, C) = f_\theta(S^C_{<i}, C^g) \text{ where } C^g = \{g_\phi(c_1, C), \dots, g_\phi(c_{|C|}, C)\},
\label{eq:final_objective}
\end{equation}
where $C^g$ is the output of the encoder, a set of encoded condition tokens. We use a standard transformer encoder-decoder setup~\cite{vaswani} for $f_\theta$ and $g_\phi$, implementation details are provided in Section~\ref{sec:implementation}, and the complete architecture is described in detail in the appendix.

%The objective is to have a generative model of layouts. We desire the following properties - permutation invariance, fast sampling, fast likelihood evaluation. A normal language model will have a position encoding to account for the left-to-right (or the reverse) nature of language. If we remove this - we get a permutation invariant network as attention blocks themselves are permutation invariant. But this treats all tokens the same, which is not what we expect the network to model. There is a difference between the different parameters of a a layout and the network should have that built-in by design. Furthermore, the network should ensure that different token groups have different affinities - change in params of one object should affect the other params of that object more unless there is a global effect (which is modeled by the full attention matrix). As an example - making an object smaller should not change the positions of other objects, while making it larger might need movement of others to avoid intersections. To address each of these problems we introduce additional tokens to inject this bias into our representation and create an \emph{object-centric representation} instead of a \emph{token-centric representation} that is common in standard Language Modelling (LM). 

\subsection{Layout Representation}
\label{sec:layout_rep}
%\subsection{Problem Formulation}
\paragraph{Parameters.}

%WARNING
%\begin{wrapfigure}{l}{0.5\textwidth}
%   \centering
%     \includegraphics[width=0.5\textwidth]{figures/val_losses_v2.pdf}
%     \caption{TODO: INSERT CAPTION}
% \end{wrapfigure}

We focus on 3D layouts in our experiments. A 3D layout $\cL = (\cI, \cB)$ is composed of two elements - a top-down representation of the layout boundary $\cI$, such as the walls of a room, and a set of $k$ three-dimensional oriented bounding-boxes $\cB = \set{B}{i}{1}{k}$ of the objects in the layout. The boundary is given as a binary raster image and each bounding box is represented by four attributes: $B_i = (\tau_i, t_i, e_i, r_i)$, representing the object class, center position, size, and orientation, respectively. The orientation is a rotation about the up-axis, giving a total of 8 scalar values per bounding box. This setup is the same as ATISS.

\paragraph{Layout sequence.}
A layout sequence $S$ is defined by randomizing the order of the bounding boxes, concatenating their parameters, and adding special start and stop tokens $\texttt{SOS}$ and $\texttt{EOS}$ to mark the start and the end of a sequence: $S = [\texttt{SOS}; B_{\pi_1}; \dots; B_{\pi_k}; \texttt{EOS}]$, where $\pi$ is a permutation of the object indices $1, \dots, k$ and $[;]$ denotes concatenation. Note that the objects are permuted in $S$, while the parameters of each object have a consistent order.
The condition $C$ for a layout can be any sub-sequence $S_{<i}$ of $S$, where some tokens may be replaced by mask tokens $\mathcal{M}$, leaving them unconstrained. During training and sampling, these tokens are replaced by generated token.
%, and the model generates a probability distribution over the masked attribute..

\paragraph{Parameter probability distributions.}
The generative model outputs a probability distribution over one scalar component of the bounding box parameters in each step. Similar to previous work~\cite{Paschalidou2021NEURIPS, Salimans2017PixeCNN}, we represent probability distributions over continuous parameters, like the center position, size, and orientation, as mixture of $T$ logistic distributions. Probability distributions over the discrete object class $\tau$ are represented as vectors over logits $l_\tau$ over discrete choices that can be converted to probabilities with the $\text{softmax}$ function.
\begin{equation}
    p(b) = \frac{1}{\sum_i \pi_i} \sum_{i=1}^{T} \alpha_i \text{Logistic}(\mu_i, \sigma_i), \qquad\qquad p(\tau) = \text{softmax}(l_\tau),
    \label{eq:probab}
\end{equation}
where $b$ is a component of $t_i$, $e_i$, or $r_i$, and $\alpha$, $\mu$, $\sigma$, respectively are the mixture weight, mean and variance of the component logistic distributions. Each probability distribution over a continuous scalar component is parameterized by a $3T$-dimensional vector, and probability distributions over the object class are represented as $n_\tau$-dimensional vectors, where $n_\tau$ is the number of object classes.

\subsection{Implementation}
\label{sec:implementation}

%Now we describe how the transformer-based generative model $f_\theta$ and the transformer encoder $g_\phi$ are implemented.
%Now we describe how each of the parts in the objective Eq. \ref{eq:final_objective} are implemented.
%We use a BART~\cite{lewis-etal-2020-bart}-like architecture, where the encoder is permutation-invariant, and the generative model predicts the sequence one token at a time. BART is often used for machine translation, and in that setting usually receives the source language in the encoder, and outputs the target language using the generative model. In our setting, we receive the partial layout as condition in the encoder and complete the missing parts using the generative model.

%On a high level, we use a BART-like model for generation. We want the conditioning to be permutation-invariant, hence we remove absolute positional embeddings from the encoder. These are present in the decoder. BART is often used for machine translation, and in that setting usually receives the source language on the encoder, and target language on the decoder. We use BART for generation, and our encoder and decoder receive the same input, with some minor differences that we explain below. With our modifications the encoder represents a set (of conditions and masks), while the decoder samples a sequence. Thus, our task is a \textbf{Masked Set-to-Sequence prediction} task. 

\para{Condition encoder $g_\phi$} To encode the condition $C$ into a set of encoded condition tokens $C^g$, we use a Transformer encoder with full bidirectional attention. As positional encoding, we provide two additional sequences: \emph{object index tokens} $\mathcal{O}^i$ provide for each token the object index in the permuted sequence of objects; and \emph{relative position tokens} $\mathcal{R}^i$ provide for each token the element index inside the attribute tuple of an object. Since the attribute tuples are consistently ordered the index can be used to identify the attribute type of a token. These sequences are used as additional inputs to the encoder. The encoder architecture is based on BART~\cite{lewis-etal-2020-bart}, details are provided in the appendix.

\para{Boundary encoder $g^\mathcal{I}_\psi$} To allow conditioning on the layout boundary $\mathcal{I}$, we prepend a feature vector encoding of the boundary to the input of the condition encoder, as shown in Figure~\ref{fig:architecture}. Similar to ATISS, we use an untrained ResNet-18~\cite{resnet} to encode a top-down view of the layout boundary into an embedding vector.

% , which makes the encoder fully permutation-invariant. However, we found that giving the encoder some prior knowledge about the semantics of tokens is needed to yield good samples. We provide two additional sequences as positional embeddings to the encoder:
% %set representation $C^g$, we use a Transformer Encoder with full bidirectional attention.
% %Without any form of positional encoding, a standard transformer is fully permutation invariant. However, we find that some prior knowledge is needed to yield good samples. We provide two such sources of injecting information: 
% \emph{Object Index Tokens} ($\mathcal{O}^i$), consisting of We assign each object in the permuted set an index and 2. \emph{Relative Position Tokens} ($\mathcal{R}^i$): To each attribute of an object, we assign a canonical position. Note that all these additional tokens are permutation-invariant. The embeddings for one attribute are generated by summing up all the corresponding embeddings. %

\para{Generative model $f_\theta$} The generative model is implemented as a Transformer decoder with a causal attention mask.
%sequence decoder implements autoregressive decoding and is a Transformer Decoder with a causal attention mask.
Each block of the decoder performs cross-attention over the encoded condition tokens $C^g$. As positional encoding, we provide \emph{absolute position tokens} $\mathcal{P}$, which provide for each token the absolute position in the sequence $S$. This sequence is used as additional input to the generative model. The output of the generative model in each step is one of the parametric probability distributions described in Eq.~\ref{eq:probab}. Since the probability distributions for discrete and continuous values have a different numbers of parameters, we use a different final linear layer in the generative model for continuous and discrete parameters. Similar to the encoder, the architecture of the generative model is based on BART~\cite{lewis-etal-2020-bart}.
%activations of the final hidden layer of the Set Encoder.
%We reuse the \textit{unmasked} embeddings generated for encoder side, without embeddings for $\mathcal{R}^i$ and $\mathcal{O}^i$. Instead, we use full \textbf{Absolute Position Tokens} ($\mathcal{P}$) which is standard in Language Modelling. %The sequence is enclosed in \texttt{SOS} and \texttt{EOS} embeddings to denote the start and end of the sequence respectively.
% \para{Boundary encoder $g^\mathcal{I}_\psi$} To encode the representation of the floorplan boundary $\cI$, we follow ATISS and use an untrained ResNet-18~\cite{resnet} model to encode a top-down view of the layout boundary into a embedding vector.

\para{Training} During training, we create a ground truth sequence $S^\text{GT}$ with randomly permuted objects. We generate the condition $C$ as a copy of $S^\text{GT}$ and mask out a random percentage of the tokens by replacing them with the mask token $\mathcal{M}$. The boundary encoder $g^\mathcal{I}_\psi$, the condition encoder $g_\phi$ and the generative model $f_\theta$ are then trained jointly, with the task to generate the full sequence $S^\text{GT}$. For unmasked tokens in $C$, this is a copy task from $C$ to the output sequence $S$. For masked tokens, this is a scene completion task. We use the negative log-likelihood loss between the predicted probabilities $p(s_i)$ and ground truth values $s^\text{GT}_i$ for tokens corresponding to continuous parameters, and the cross-entropy loss for the object category $\tau$.
%using a ... \paul{TODO} loss.
The model is trained with teacher-forcing. 

\begin{figure*}[t!]
    \centering
    \includegraphics[width=\linewidth]{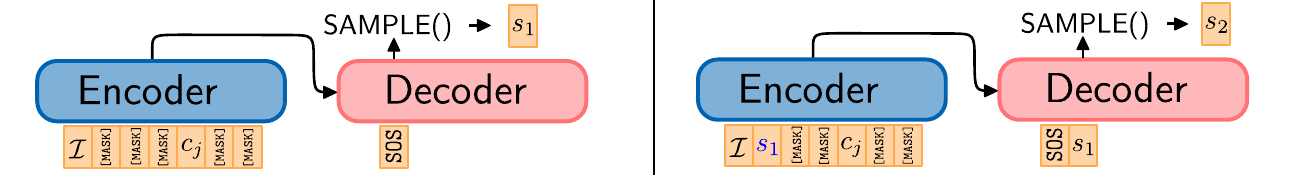}\caption{\textbf{Sampling Strategy}. (\textbf{Left}): We start out with a sequence of all \mask tokens on the encoder, and the \texttt{SOS} token on the decoder. Conditions, if any, are specified by replacing the \mask with the desired value of the attribute in the sequence. Sampling is performed autoregressively. (\textbf{Right}): After a token has been sampled, both the encoder and decoder sequences are updated, and atuoregressive generation continues.}
    \label{fig:sampling}
\vspace{-0.5cm}
\end{figure*}

\para{Sampling}
% \subsection{Sampling}
%Sampling from Masked Language Models is an open problem~(\cite{wang-cho-2019-bert}). But our structured sequences allow for easier sampling.
%Specifically,
% We formulate sampling as a Neural Machine Translation (NMT) problem~(\cite{klein-etal-2017-opennmt}).
We generate a sequence auto-regressively, one token at a time, by sampling the probability distribution predicted by the generative model (as defined in Eq.~\ref{eq:final_objective}) in each step. We use the same model for both conditional and unconditional generation. For unconditional generation, we start with a condition $C$ where all tokens are mask token $\mathcal{M}$. To provide more complete information about the partially generated scene to the encoded condition tokens $C^g$, we update the condition $C$ after each generation step by replacing mask tokens with the generated tokens.
% as shown in Algorithm~\ref{alg:decoding}.
Empirically, we observed that this improves generation performance. An illustration of this approach is shown in Fig. \ref{fig:sampling}.
%To perform unconditional sampling, we start with encoding the floorplan boundary $\cI$, and set all condition tokens $C$ to $\mathcal{M}$. Then we sample autoregressively on the decoder. The only difference is that in NMT, the encoder pass is made only once and the decoder sequence is updated, but in our case every new sample replaces a \mask token in both the encoder and the decoder. In order to perform, conditional sampling, we simply replace \mask tokens in $C$ with the attributes we want.
The function \textsf{SAMPLE} samples from the probability distributions described in Eq.~\ref{eq:probab}.
% \vspace{-0.5cm}
%
% \begin{minipage}[h]{0.48\linewidth}
%     \small
%     \begin{algorithm}[H]
%       \caption{Standard Cond. Sampling}\label{alg:decoding}
%       \begin{algorithmic}[1]
%         \Require{$C = (c_i)_{i=1}^k$, $S = (\texttt{SOS}), s = \phi$}
%         \State {${C^g = g_\phi([\cI, C])}$} \Comment{Only performed once}
%         \While{$s \neq \texttt{EOS}$}% \Comment{Loop over semantic group}
%             \State {$s = \textsf{SAMPLE}(f_\theta(S_{<i}, C^g))$}
%             \State {$S\textsf{.append($s$)}$} \Comment{$C$ not updated}
%         \EndWhile
%         \State \textbf{return} $S$
%       \end{algorithmic}
%     \end{algorithm}
% \end{minipage}
% \begin{minipage}[h]{0.48\linewidth}
%     \small
%     \begin{algorithm}[H]
%       \caption{Our Sampling}\label{alg:decoding}
%       \begin{algorithmic}[1]
%         \Require{$\cI$, $C = (\mask)_{i=1}^k$, $S = (\texttt{SOS})$}
%         \For{$i \gets 1$ to $k$}% \Comment{Loop over semantic group}
%             \State {${\color{blue} C^g = g_\phi([\cI, C])}$}
%             \State {$s = \textsf{SAMPLE}(f_\theta(S_{<i}, C^g))$}
%             \State {${\color{blue} C[i] = s}, \; S\textsf{.append($s$)}$}
%         \EndFor
%         \State \textbf{return} $S$
%       \end{algorithmic}
%     \end{algorithm}
% \end{minipage}
%
% differently for different tokens (attributes) - softmax for the class label and a mixture of discretized logistics~(\cite{Salimans2017PixeCNN}) for the other attributes (see Eq. \ref{eq:probab}). 

Once a layout has been generated, we can populate the bounding boxes with objects from the dataset. For each bounding box, we pick the object of the given category $\tau$ that best matches the size of the bounding box. In the supplementary, we will present an ablation of the tokens $\mathcal{O}^i$, $\mathcal{R}^i$, and $\mathcal{P}$ that we add to the conditional encoder and generative model.
%in practice we pick the object that best matches the 

%After we have sampled a layout conditioned on a given floorplan boundary, we populate the layout by assigning the labeled bounding box with an object from the dataset that is closest to the generated bounding box in size. 

%% file: sections/results.tex
\section{Results}

\begin{table}[b]
\centering
% \small
\caption{\textbf{Comparison on Unconditional Generation}: We provide floorplan boundaries from the Ground Truth as an input to the methods and compare the quality of generate scenes. We measure the CAS at a resolution of $256\times256$.We retrain the ATISS model and report the new numbers. The retrained model is called ATISS$^{*}$. The KL-Divergence follows from \cite{fastsynth:ritchie} and is the categorical KL-divergence between the distribution of classes in the GT and the layouts synthesised by our method.}
\setlength\tabcolsep{2pt}
\begin{tabular}{@{}lcccccccc@{}}
\toprule
\multicolumn{1}{l}{} & \multicolumn{4}{c}{CAS $\times 10^2$($\downarrow$)}    & \multicolumn{4}{c}{KL-Divergence $\times 10^3$ ($\downarrow$)}       \\ \cmidrule(l){2-5} \cmidrule(l){6-9}
                                    & \bed      & \liv & \din &  \lib                         & \bed & \liv & \din & \lib \\ \midrule
\multicolumn{1}{l}{FastSynth}       & 88.3        & 94.5  & 93.5  &  \multicolumn{1}{c|}{81.5}      & 6.4   & 17.6   & 51.8   & 43.1    \\
\multicolumn{1}{l}{SceneFormer}     & 94.5        & 97.2  & 94.1  &  \multicolumn{1}{c|}{88.0}      & 5.2     & 31.3   & 36.8   & 23.2    \\
\multicolumn{1}{l}{ATISS$^*$}       & 61.1       & \textbf{76.4}  & \textbf{69.1}  &  \multicolumn{1}{c|}{\textbf{61.77}}      & 8.6     & 14.1   & 15.6   & 10.1      \\%
%
% Our rows from here on
% \multicolumn{1}{l}{Ours+Cosine}     & -            &             &       & \multicolumn{1}{l|}{}     &                      &             &       &\\
\multicolumn{1}{l}{Ours}            & \textbf{61.0}            &    78.9         &    76.1   &  \multicolumn{1}{c|}{66.2}   &   \textbf{5.0}                   &   \textbf{8.1}          &    \textbf{9.3}   & \textbf{6.7} \\

\bottomrule
\end{tabular}
\vspace{0.1cm}
\end{table}

\begin{figure*}[t!]

    \centering
    \vspace{-1.5em}
    % \hfill
    \begin{subfigure}[b]{0.22\linewidth}
        \centering
	    \small Boundary
    \end{subfigure}%
    \begin{subfigure}[b]{0.22\linewidth}
        \centering
        \small GT
    \end{subfigure}%
    \begin{subfigure}[b]{0.22\linewidth}
        \centering
        \small ATISS
    \end{subfigure}%
    \begin{subfigure}[b]{0.22\linewidth}
        \centering
        \small Ours
    \end{subfigure}%
    % \hfill%
    \vskip\baselineskip%
    \vspace{-0.75em}
    %%%%%%%%%%%%%%%%%%%%%%%%%%%%%%%%%%%%
    % \hfill
    \begin{subfigure}[b]{0.22\linewidth}
        \centering
	    \includegraphics[width=\textwidth, trim=0 500 0 600, clip]{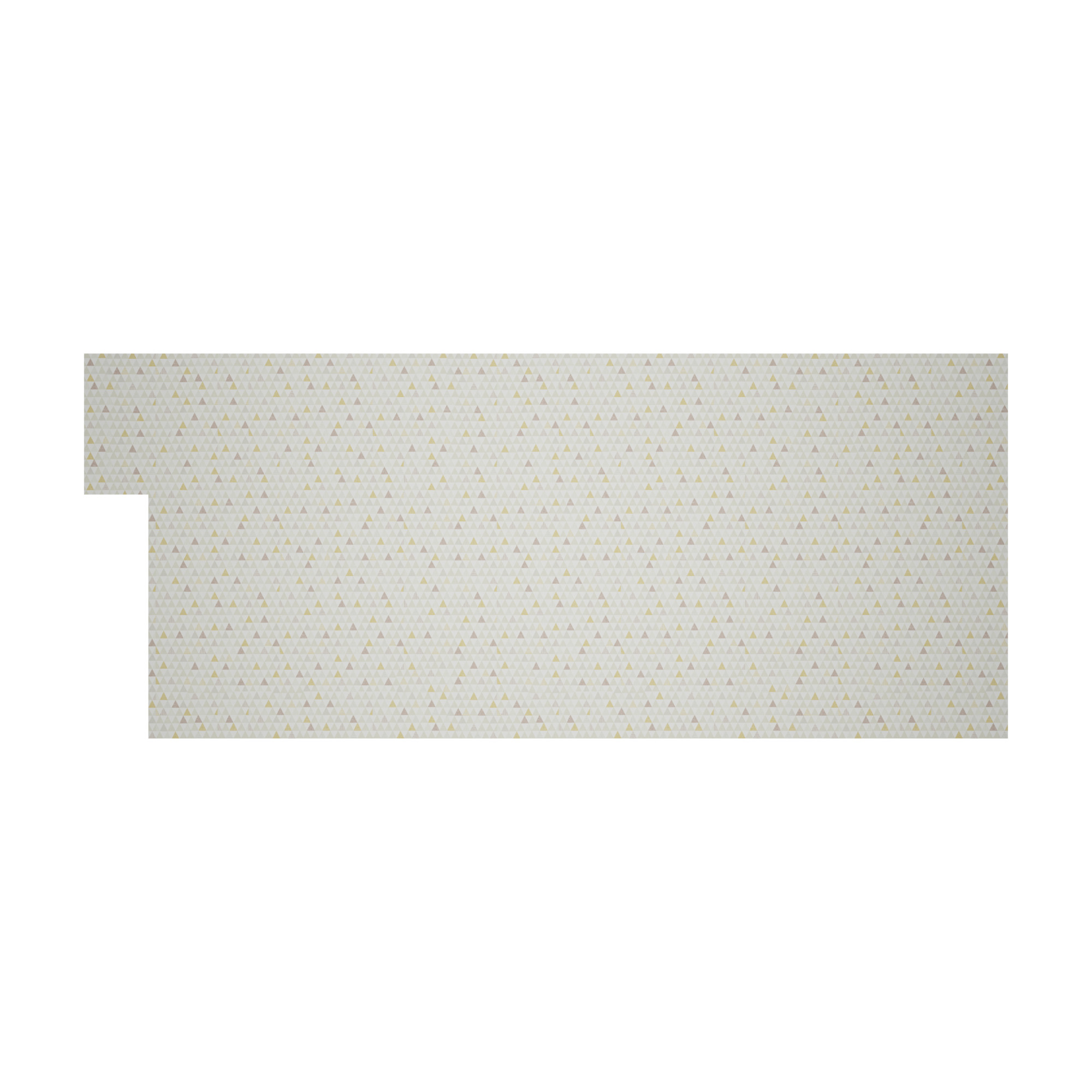}
    \end{subfigure}%
    \begin{subfigure}[b]{0.22\linewidth}
        \centering
        % \begin{mdframed}[backgroundcolor=yellow!25, linewidth=0]
            \begin{overpic}[width=\textwidth,  trim=0 500 0 600, clip]{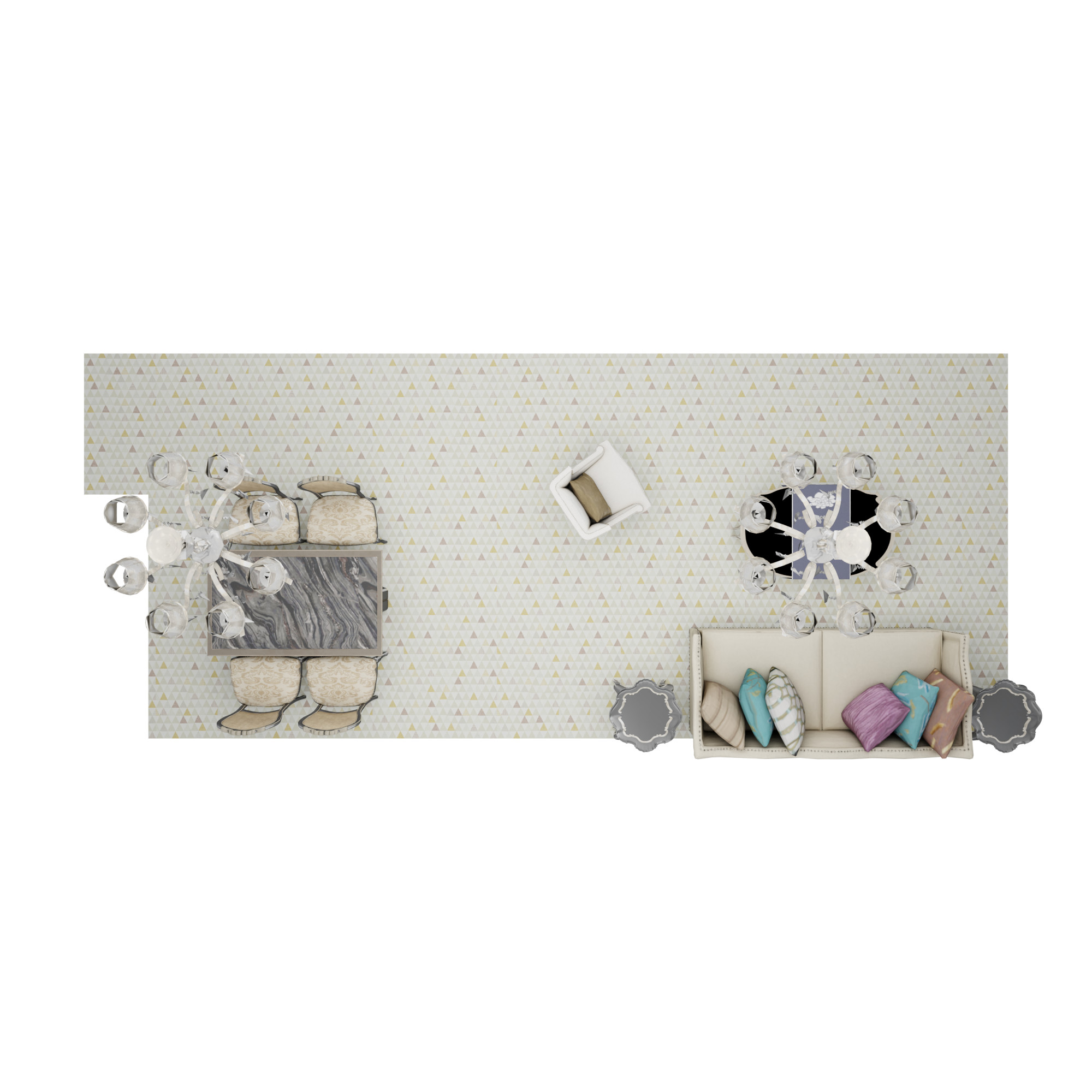}
        \end{overpic}
        % \end{mdframed}
    \end{subfigure}%
    \begin{subfigure}[b]{0.22\linewidth}
        \centering
        \begin{overpic}[width=\textwidth,  trim=0 500 0 600, clip]{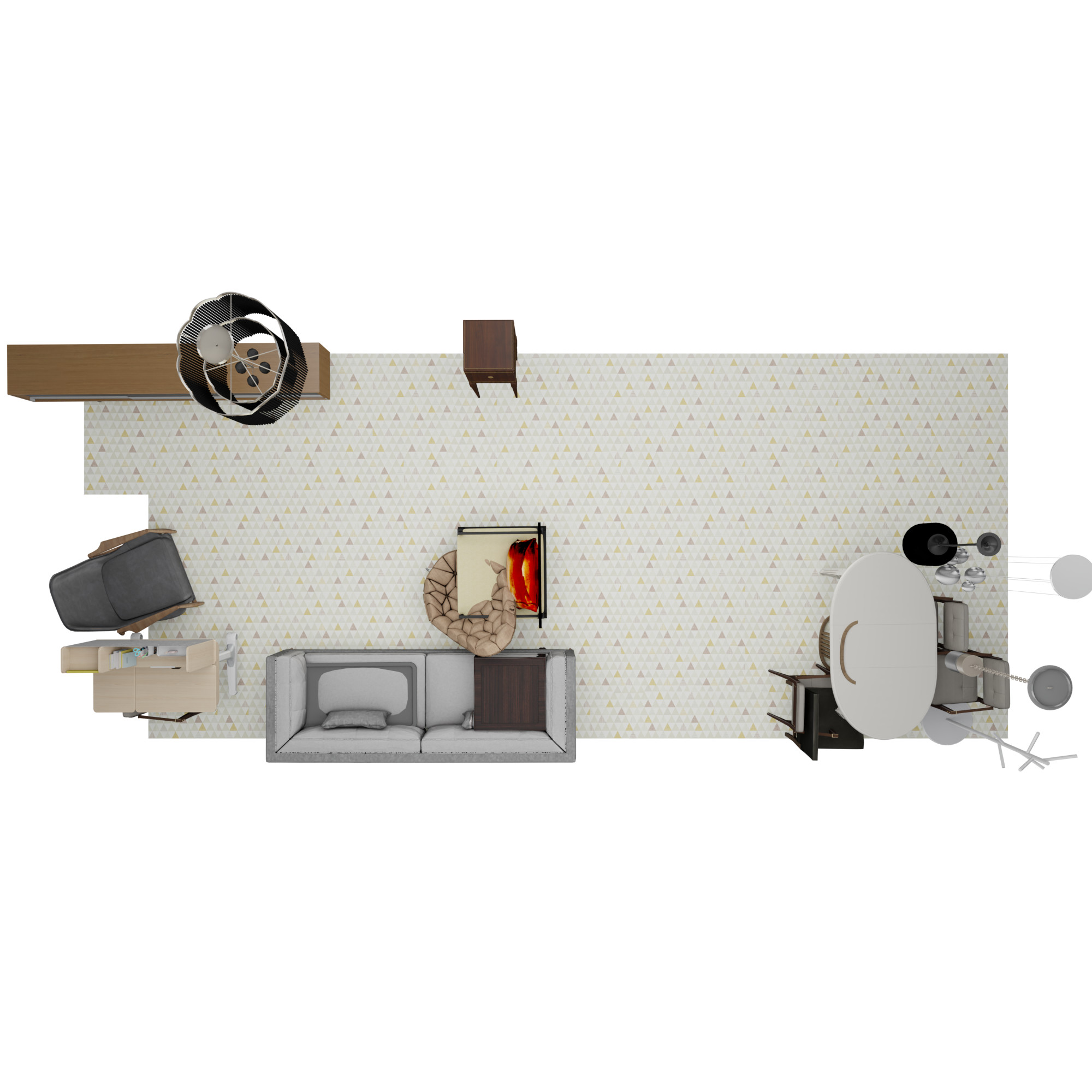}
	    \end{overpic}
    \end{subfigure}%
    \begin{subfigure}[b]{0.22\linewidth}
        \centering
        \begin{overpic}[width=\textwidth, trim=0 500 0 600, clip]{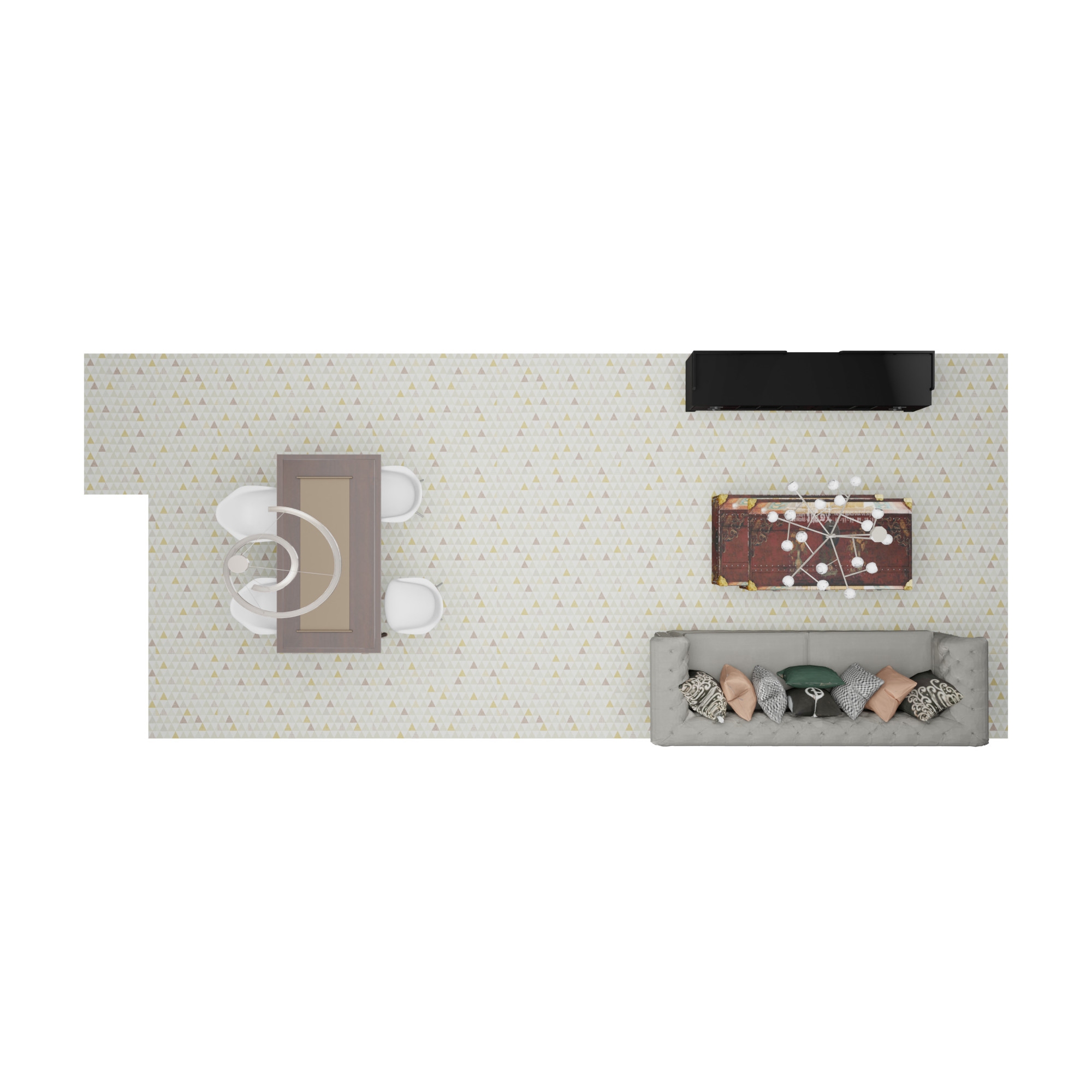}
        \end{overpic}
    \end{subfigure}%
    % \hfill%
    \vskip\baselineskip%
    \vspace{-0.75em}
    % \hfill
    %%%%%%%%%%%%%%%%%%%%%%%%%%%%%%%%%%%%%%%%%%%%%%%%%%%%%%%%%%%%%%%%%%%%%
    \begin{subfigure}[b]{0.22\linewidth}
        \centering
	    \includegraphics[width=\textwidth, trim=530 380 100 600, clip]{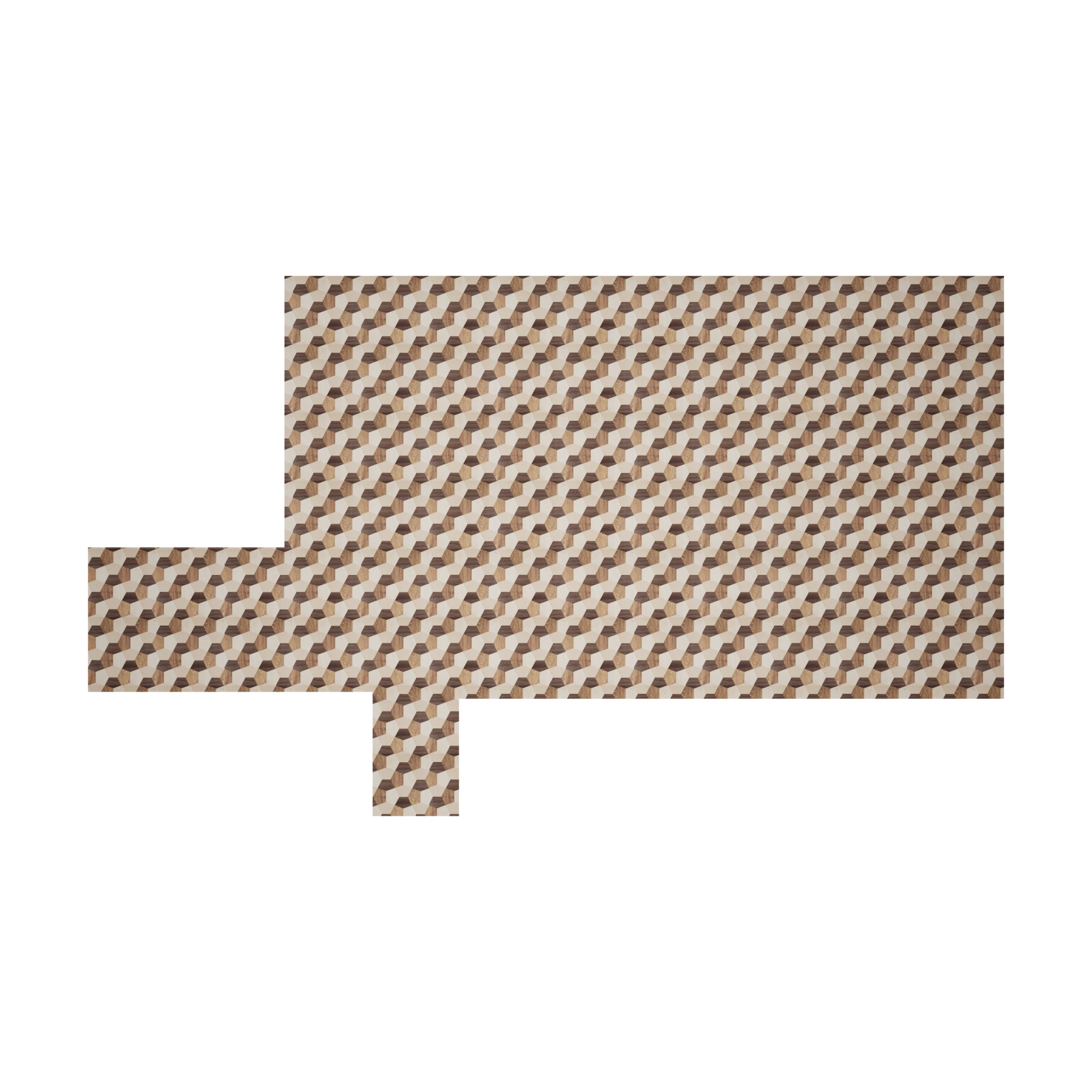}
    \end{subfigure}%
    \begin{subfigure}[b]{0.22\linewidth}
        \centering
        \begin{overpic}[width=\textwidth,  trim=530 380 100 600, clip]{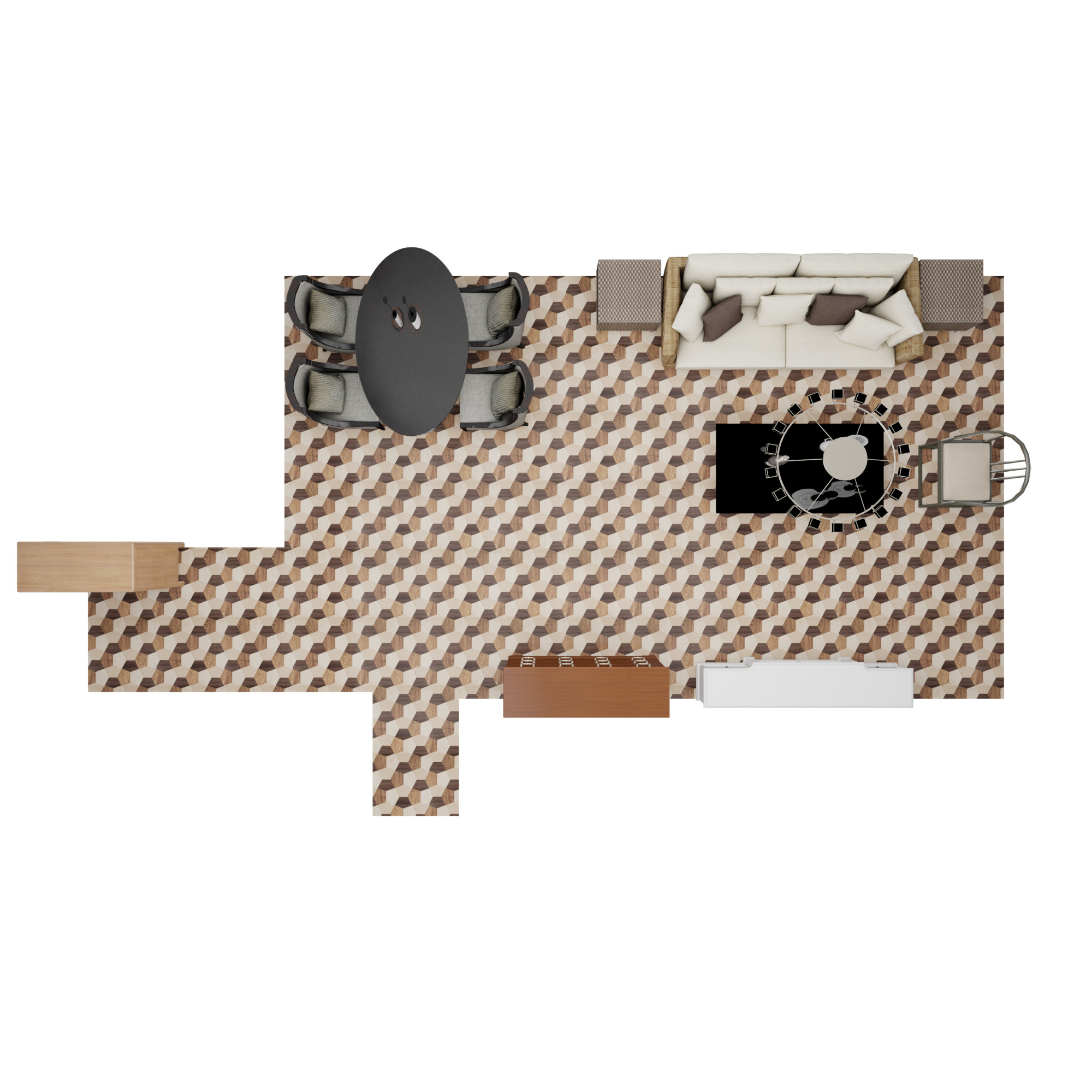}
        \end{overpic}
    \end{subfigure}%
    \begin{subfigure}[b]{0.22\linewidth}
        \centering
        \begin{overpic}[width=\textwidth,  trim=530 380 100 600,
        clip]{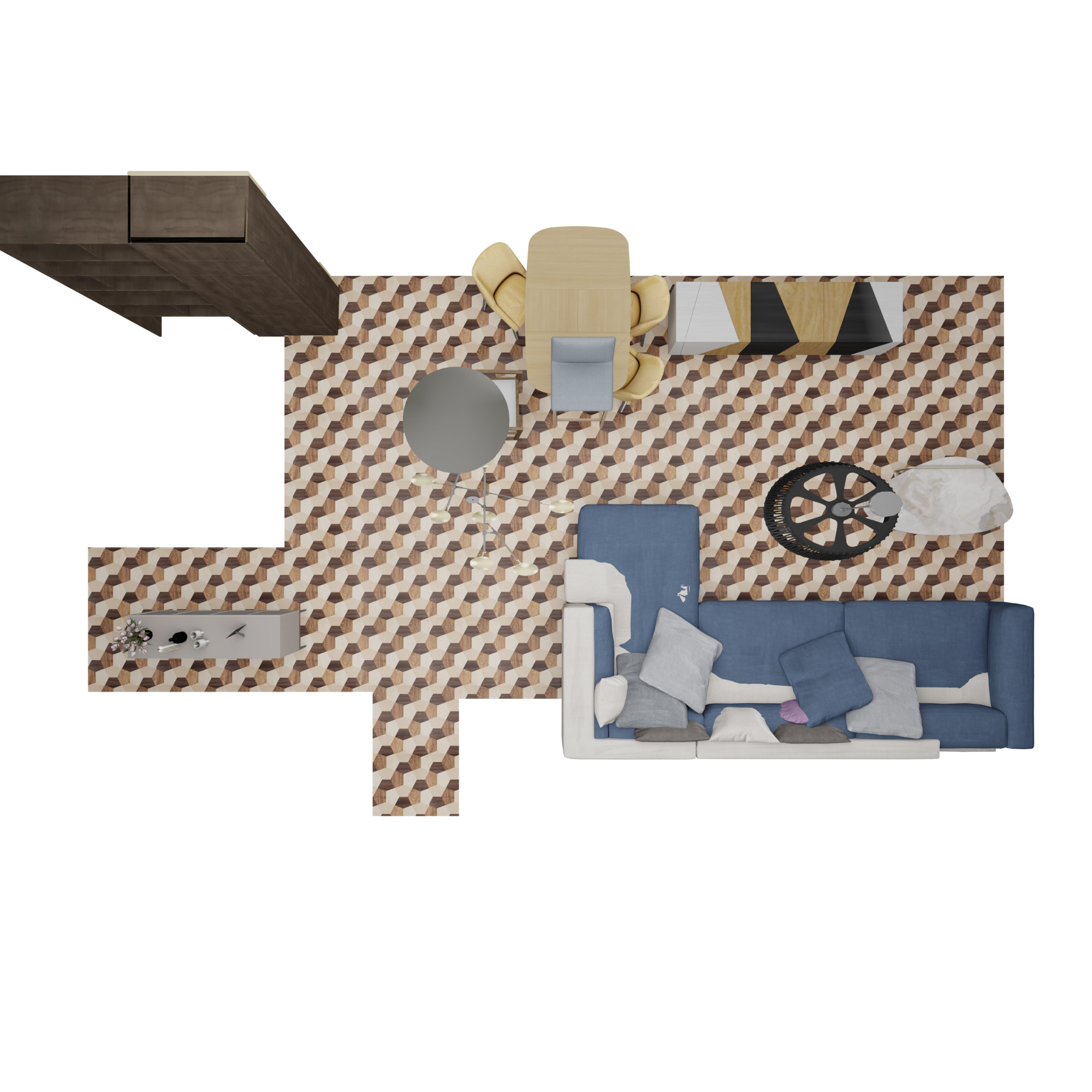}
	    \end{overpic}
    \end{subfigure}%
    \begin{subfigure}[b]{0.22\linewidth}
        \begin{overpic}[width=\textwidth,  trim=530 380 100 600, clip]{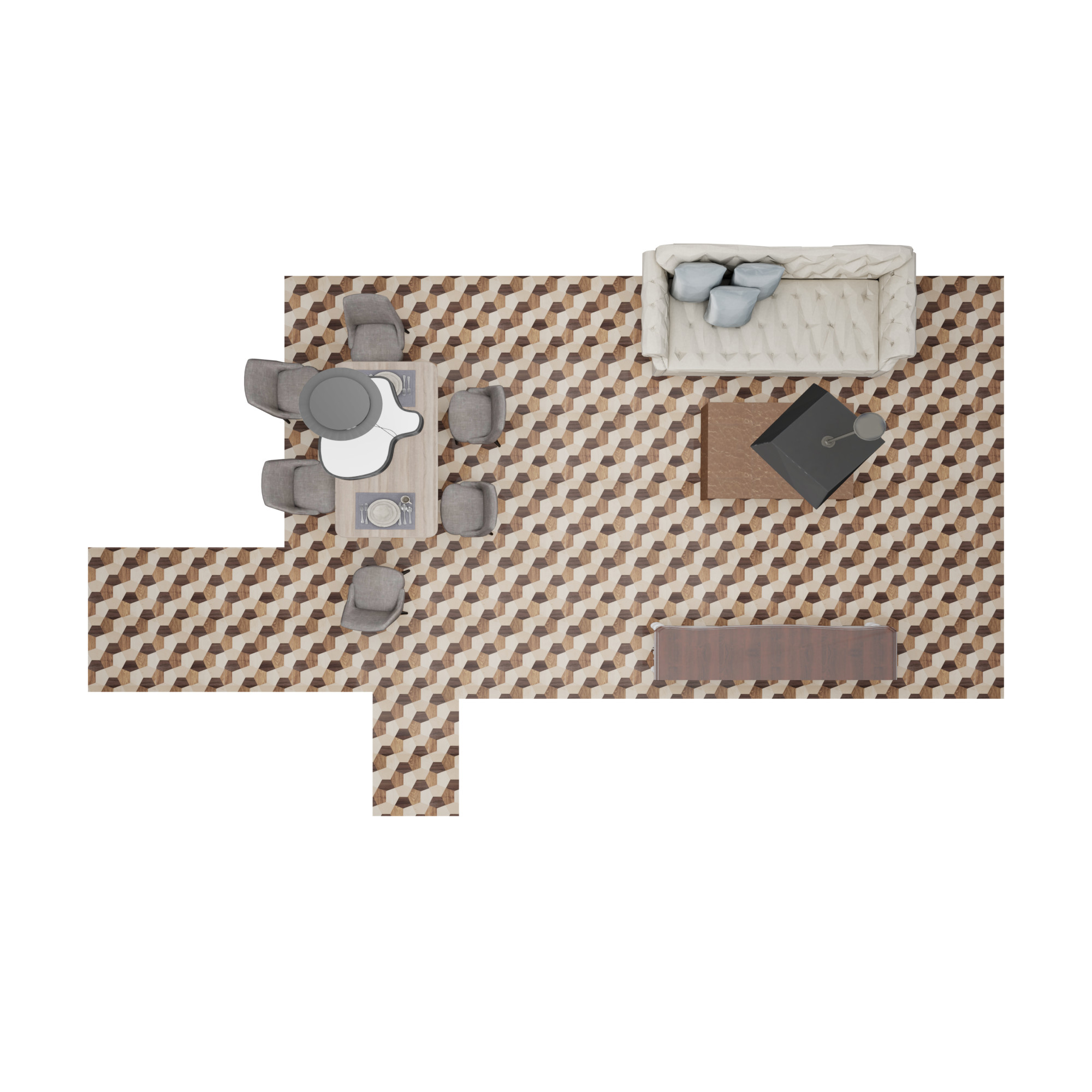}
        \end{overpic}
    \end{subfigure}%
    % \hfill%
    \vskip\baselineskip%
    \vspace{-0.75em}
    % \hfill
    %%%%%%%%%%%%%%%%%%%%%%%%%%%%%%%%%%%%%%%%%%%%%%%%%%%%%%%%%%%%%%%%%%%%%%%%%%%%%%%%%%%%%%%%%%%%%%%%%%%5
    \begin{subfigure}[b]{0.22\linewidth}
        \centering
	    \includegraphics[width=\textwidth,  trim=550 600 600 550, clip]{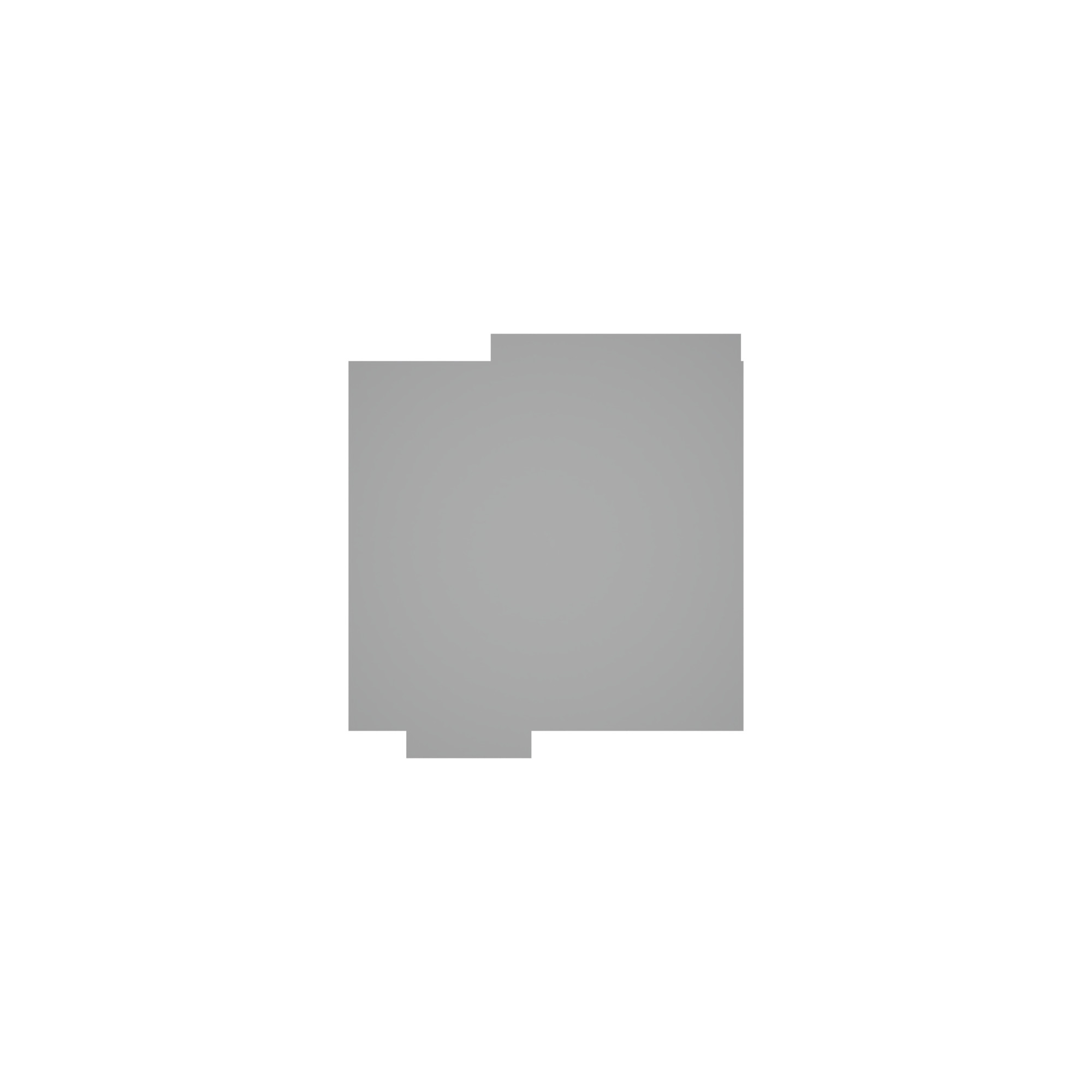}
    \end{subfigure}%
    \begin{subfigure}[b]{0.22\linewidth}
        \centering
        \begin{overpic}[width=\textwidth,  trim=550 600 600 550, clip]{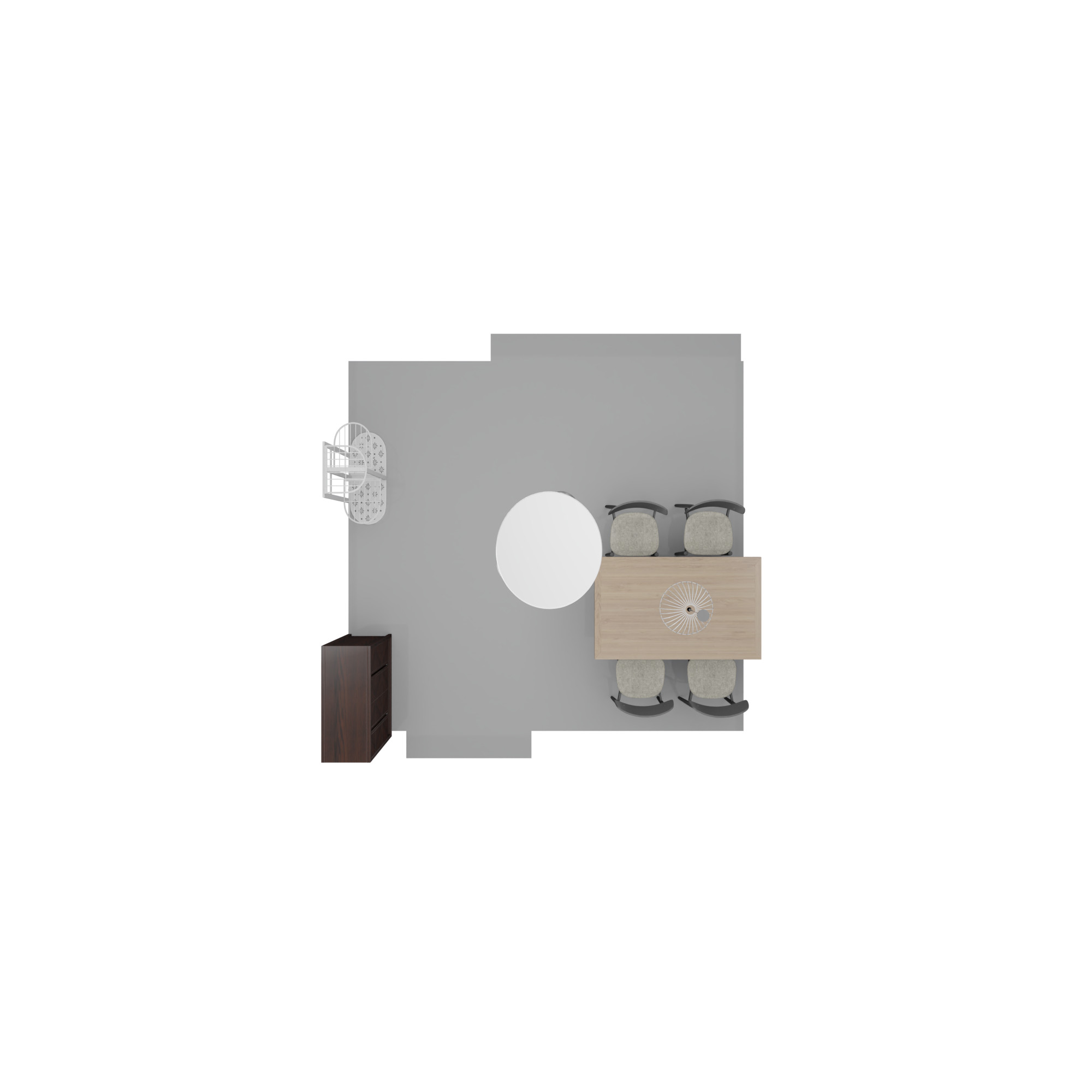}
        \end{overpic}
    \end{subfigure}%
    \begin{subfigure}[b]{0.22\linewidth}
        \centering
        \begin{overpic}[width=\textwidth,  trim=550 600 600 550,
        clip]{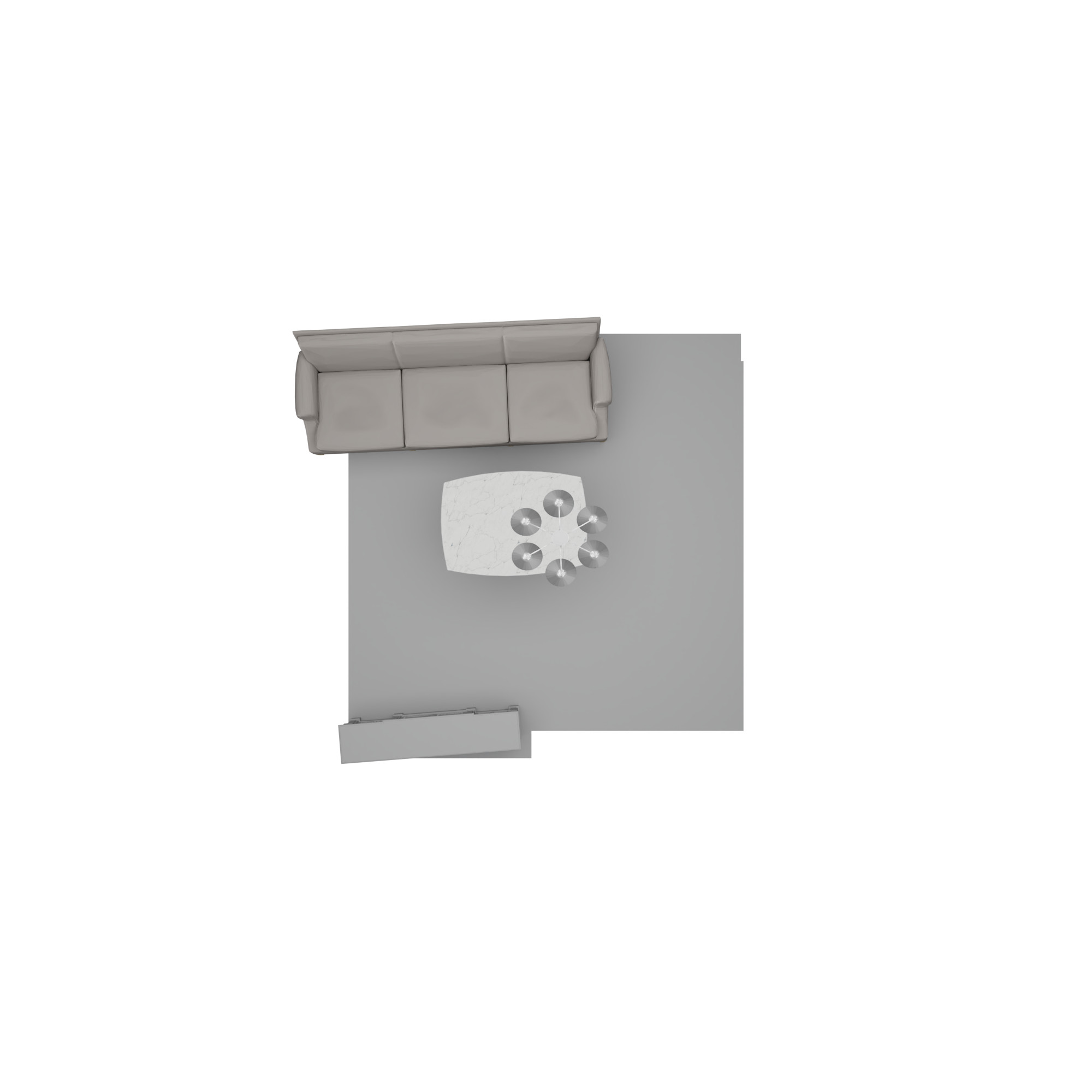}
	    \end{overpic}
    \end{subfigure}%
    \begin{subfigure}[b]{0.22\linewidth}
        \begin{overpic}[width=\textwidth,  trim=550 600 600 550, clip]{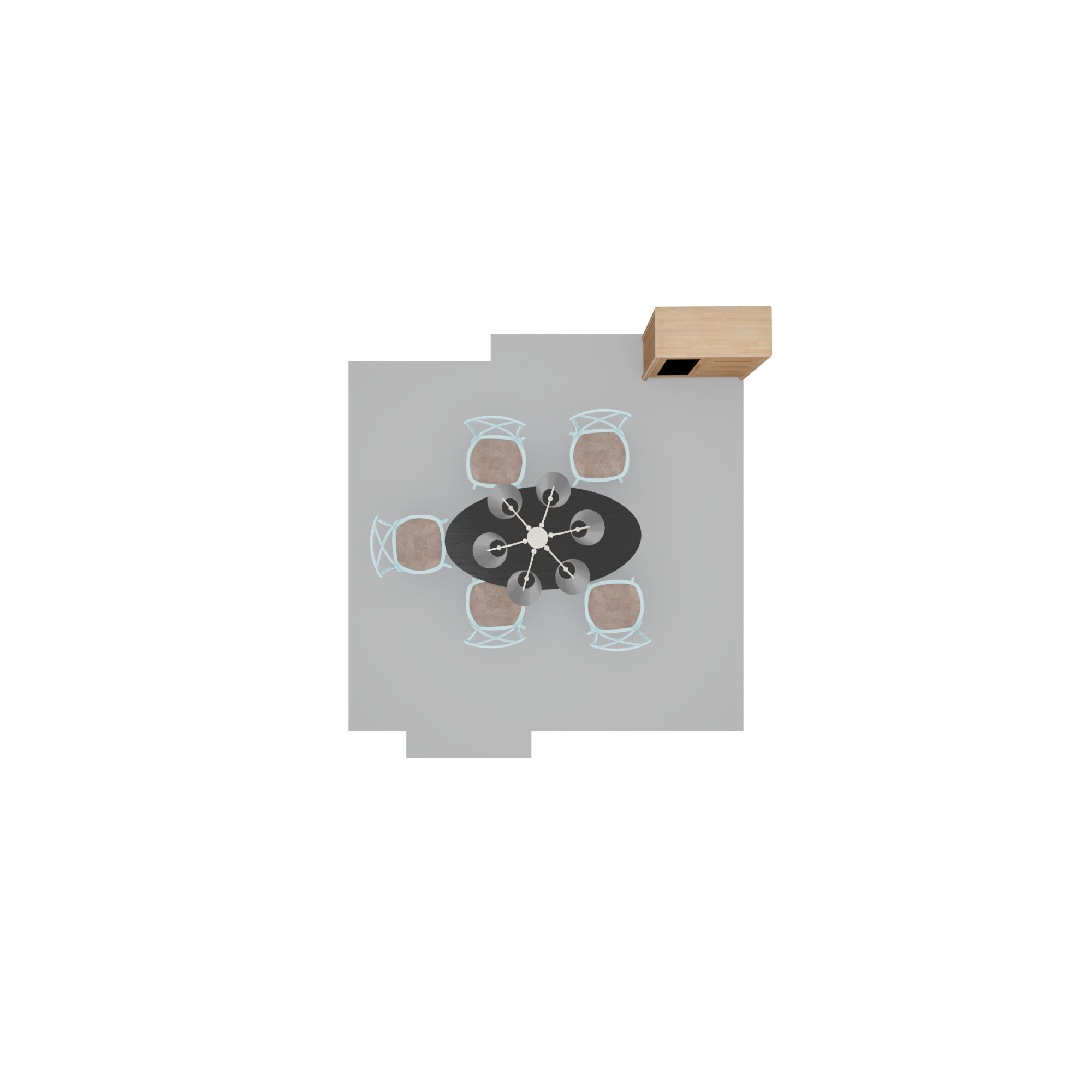}
        \end{overpic}
    \end{subfigure}%
    % \hfill%
    \vskip\baselineskip%
    \vspace{-0.75em}
    \begin{subfigure}[b]{0.22\linewidth}
        \centering
	    \includegraphics[width=\textwidth,  trim=530 380 300 600, clip]{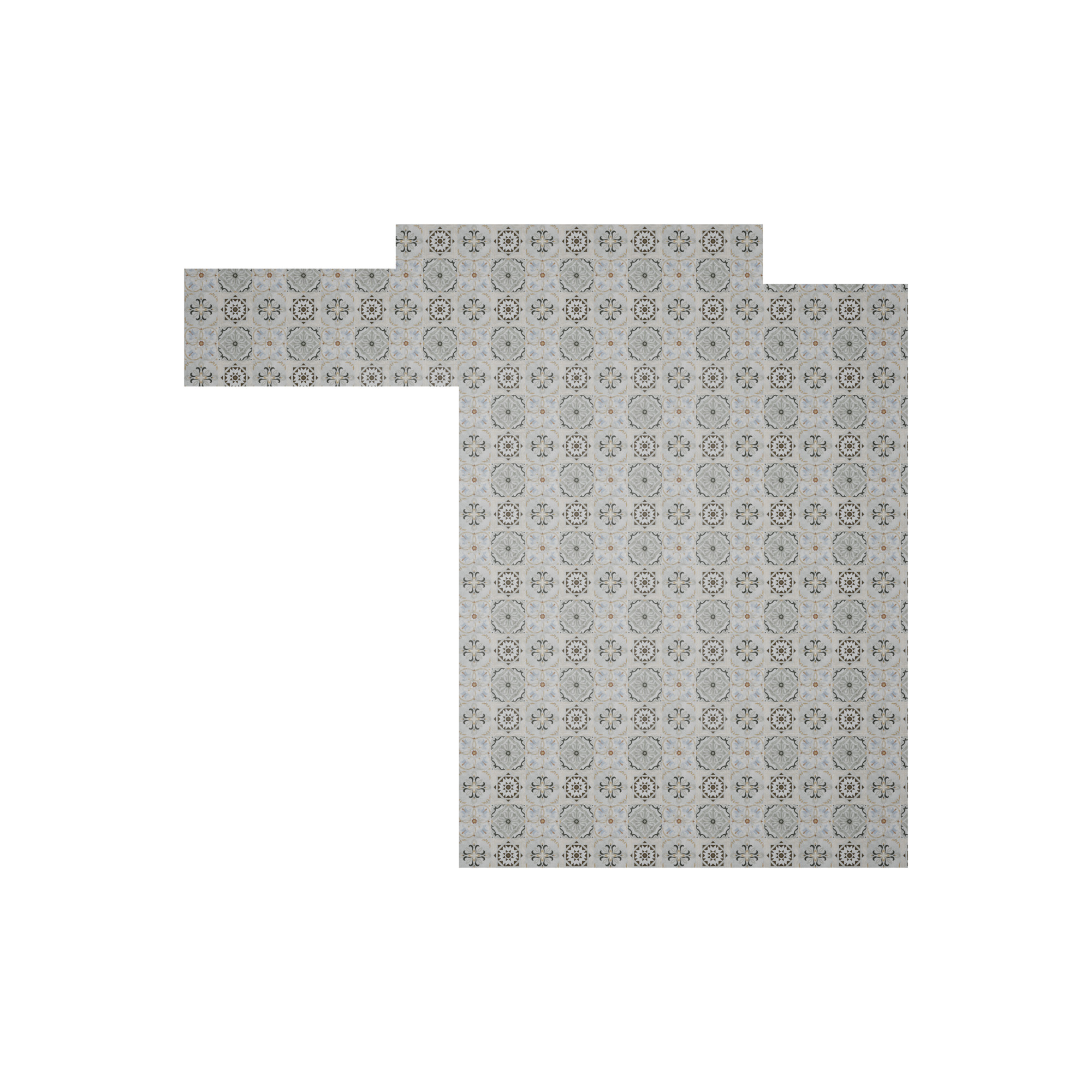}
    \end{subfigure}%
    \begin{subfigure}[b]{0.22\linewidth}
        \centering
        \begin{overpic}[width=\textwidth,  trim=530 380 300 600, clip]{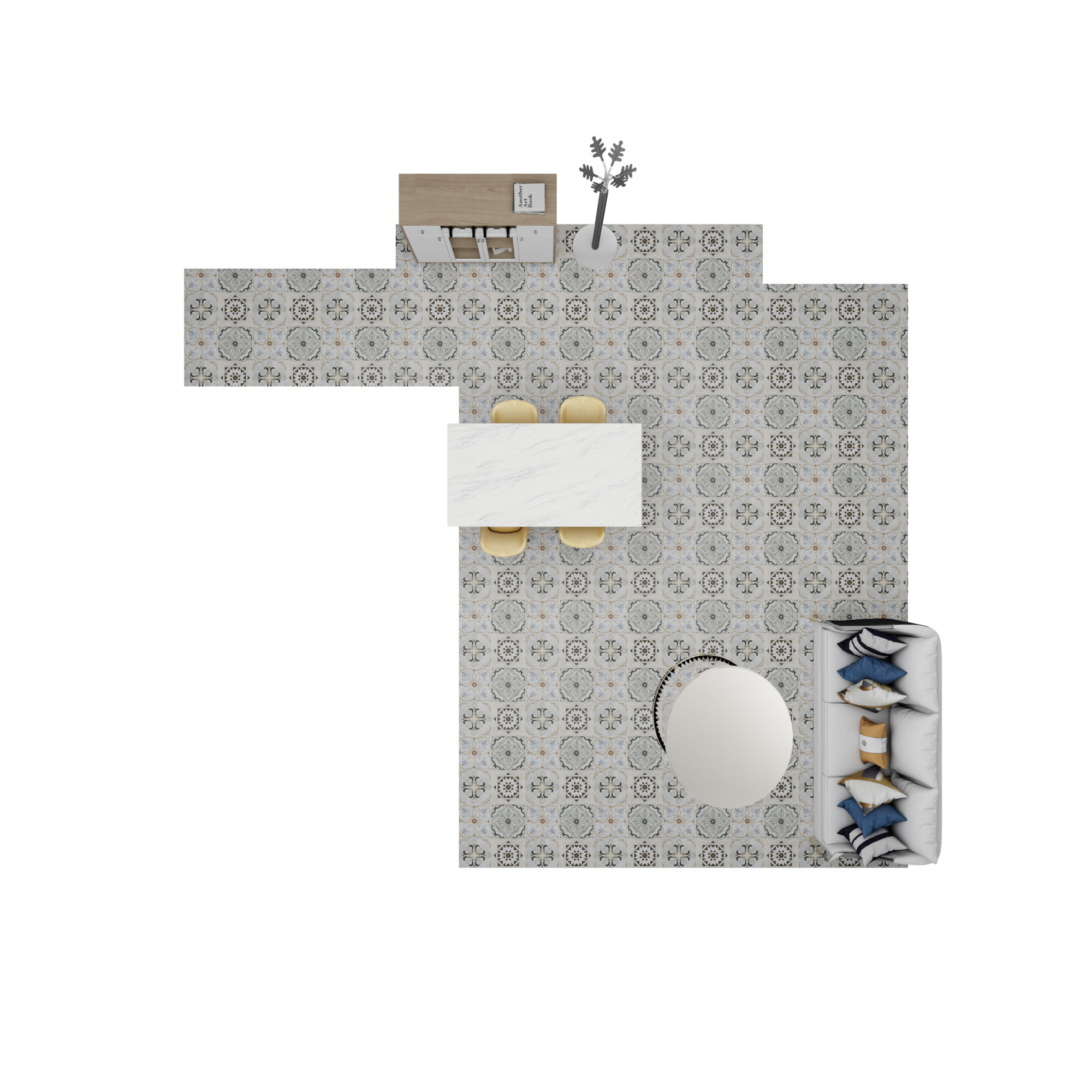}
        \end{overpic}
    \end{subfigure}%
    \begin{subfigure}[b]{0.22\linewidth}
        \centering
        \begin{overpic}[width=\textwidth,  trim=530 380 300 600,
        clip]{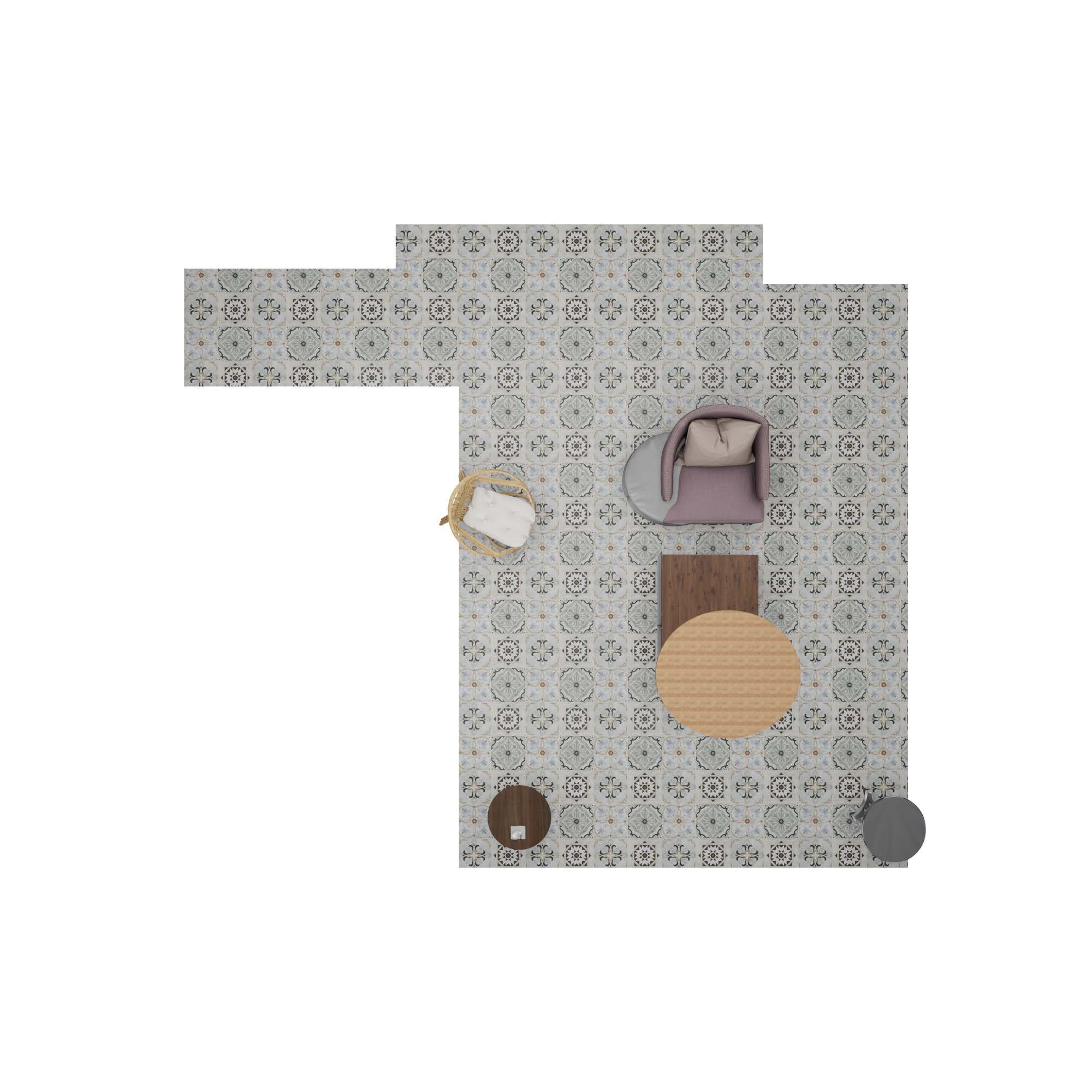}
	    \end{overpic}
    \end{subfigure}%
    \begin{subfigure}[b]{0.22\linewidth}
        \begin{overpic}[width=\textwidth,  trim=530 380 300 600, clip]{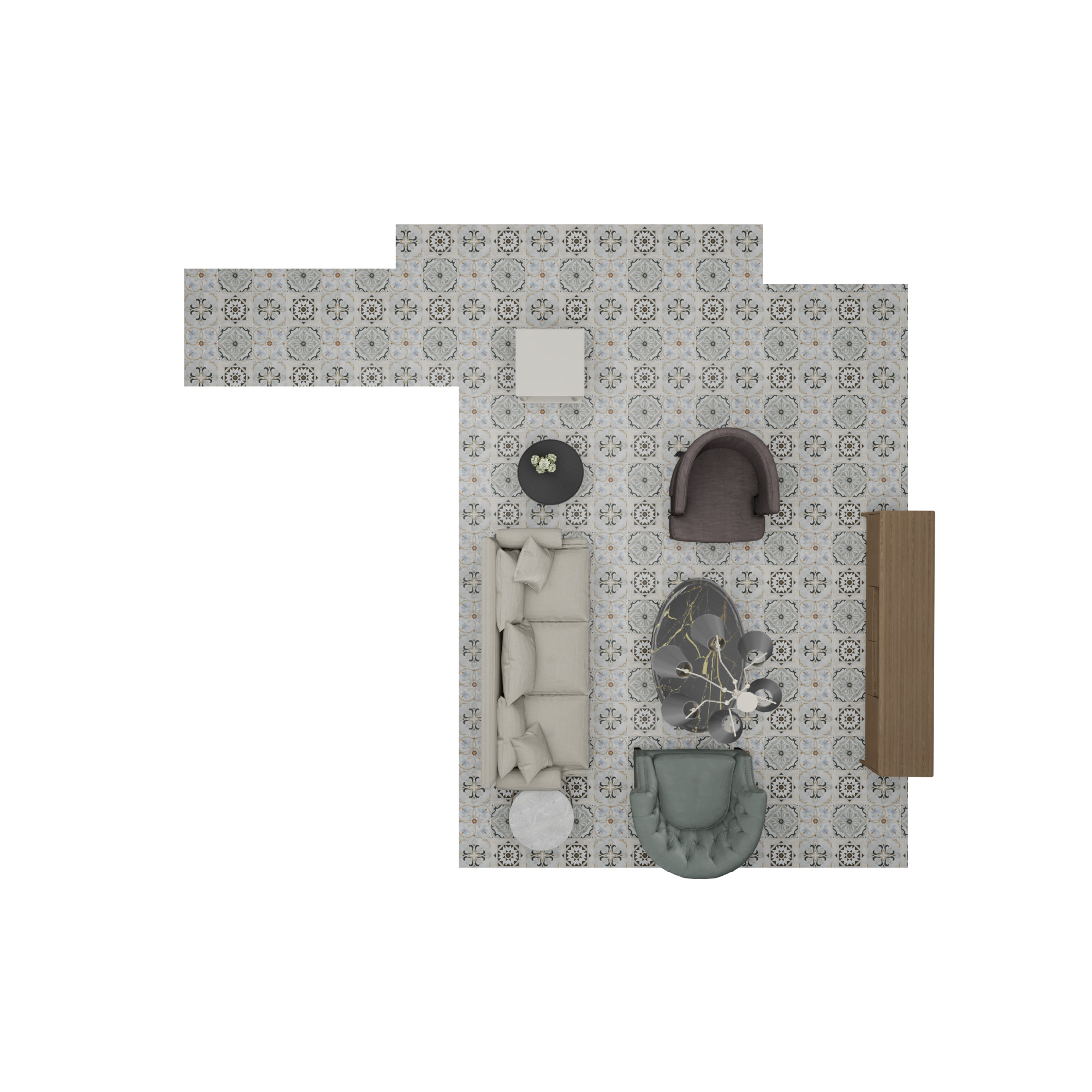}
        \end{overpic}
    \end{subfigure}%
    
    \caption{\textbf{Scene generation from scratch:} We compare generated scenes from GT, ATISS, and our model. Both ATISS and our model are conditioned on the floorplan boundary (first column). In contrast to ATISS, we can see that our model consistently creates plausible layouts within the floorplan boundary while avoiding unnatural object intersections. These are results on the challenging \liv (rows 1,3) and \din (row 2,4) categories. Additional results can be found in the supplementary. (Best viewed zoomed in, on a computer display)}
    \label{fig:qualitative_gen_from_scratch}
\vspace{-0.75em}
\end{figure*}

\subsection{Experimental Setup}

\para{Datasets} We train and evaluate our model on the 3D-FRONT dataset. It consists of of about $10k$ indoor scenes created by professional designers. We follow ATISS preprocessing which removes a few problematic scenes that have intersections between objects, or mislabeled objects, or scenes that have extremely large or small dimensions. For further details on the preprocessing, we refer the reader to ATISS~\cite{Paschalidou2021NEURIPS}. We train on the following classes of rooms - \bed, \lib, \din, \liv. We closely follow the preprocessing step of ATISS, yielding approximately $6k$, $0.6k$, $3k$ and $2.6k$ scenes for \bed, \lib, \din, \liv respectively. These classes have substantially different numbers of layouts and numbers of objects per layout. %Consequently, we train models of different sizes for each of the classes. %For detailed architectural details, please refer to the Appendix.

\para{Hyperparameters} We implement our models in PyTorch~\cite{pytorch} 1.7.0. We use standard transformer blocks for the encoder and decoder except that the ReLU activation is replaced with GeLU~\cite{gelu}. We use 4 encoder layers and 4 decoder layers, with a hidden dimension of 256 and 4 attention heads yielding a query vector of dimension 64. We use a batch size of 128 sequences and train on a single nVIDIA A100 GPU with the AdamW~\cite{loshchilov2018decoupled_adamw} optimizer which we found to be more stable than Adam~\cite{adam_kingma}. We use weight decay of 0.001 and clip the gradient norm to be a maximium of 30. We found that the networks begins to overfit very early, especially for classes other than \bed, because of the scarcity of data. Thus, for training networks on other classes, we pre-train on the \bed class, and then reuse those weights as initialization. We do not use any form of learning rate scheduling as our experiments did not suggest significant performance gains. We train for 1000 epochs and use early stopping.

\para{Baselines} Our main comparison is with respect to ATISS, as they are the only other method that considers the layout generation problem in a permutation invariant set-generation setting. We show that our model is competitive with ATISS and outperforms it on certain metrics. Moreover, we subsume all functionality of ATISS, including partial scene completion and outlier detection. Unlike ATISS, our model allows for fine-grained conditioning on arbitrary object parameters. 

%As an added advantage, our model sidesteps the autoregressive ordering for the generation of object parameters and allows for conditioning on any arbitrary object parameter.
ATISS does not provide pretrained models, hence we train their models using the official code \footnote{ \href{https://github.com/nv-tlabs/atiss}{https://github.com/nv-tlabs/atiss}, commit \texttt{0cce45b
}} and match their training settings as closely as possible. Further, we compare to previous state-of-the art methods, FastSynth~\cite{fastsynth:ritchie} and SceneFormer~\cite{wang2021sceneformer} quantitatively. % We additionally also compare against cite niessner cite ritchie and show that our model outperforms those baselines by a significant margin.

\para{Metrics} Our metrics mostly derive from \cite{fastsynth:ritchie, Paschalidou2021NEURIPS}. Following, \cite{fastsynth:ritchie}, we report the KL-divergence between the distribution of the classes of generated objects and the distribution of classes of the objects in the test set. We further report the Classification Accuracy Score (CAS)~\cite{Paschalidou2021NEURIPS}. We render the the populated layout from a top-down view using an orthographic camera at a resolution of $256 \times 256$. We report the FID computed between these rendered top down images of sampled layouts and the renders of the ground truth layouts.

\subsection{Applications}
In this section, we discuss the application of our model on multiple interactive tasks. We show that our formulation a) can perform the same tasks as existing models and b) additionally enables fine-grained conditioning on arbitrary sequence subsets, like individual object parameters. % To the best of our knowledge we propose the first layout generation method to do so.
% which no existing model in the literature can, to the best of our knowledge.

\begin{figure*}[t!]
    \centering
    \vspace{-1.5em}
    % \hfill
    \begin{subfigure}[b]{0.16\linewidth}
        \centering
	    \begin{overpic}[width=\linewidth,  trim=640 600 600 900, clip]{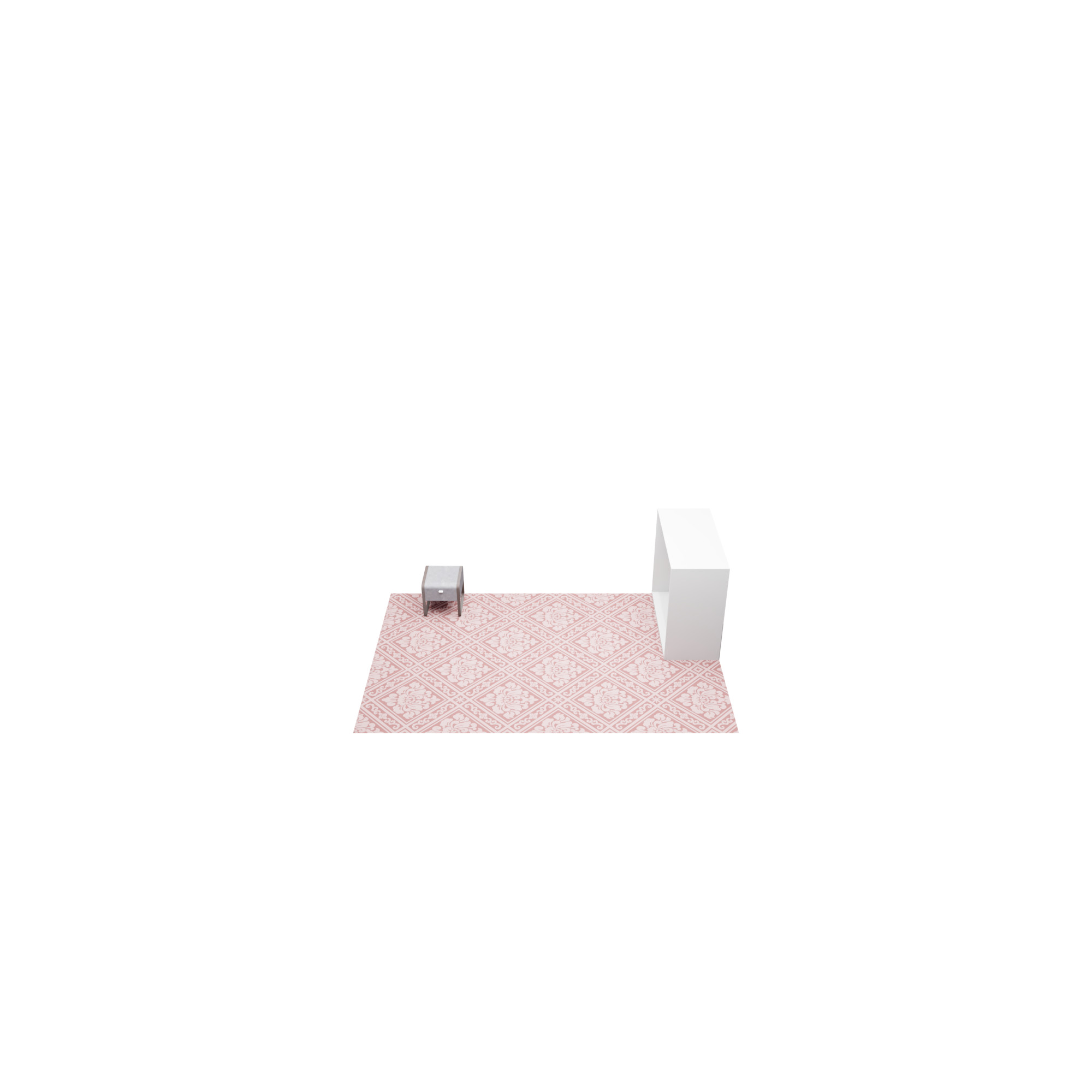}
	    \end{overpic}
    \end{subfigure}%
    \begin{subfigure}[b]{0.16\linewidth}
        \centering
        \begin{overpic}[width=\textwidth,  trim=600 600 600 760, clip]{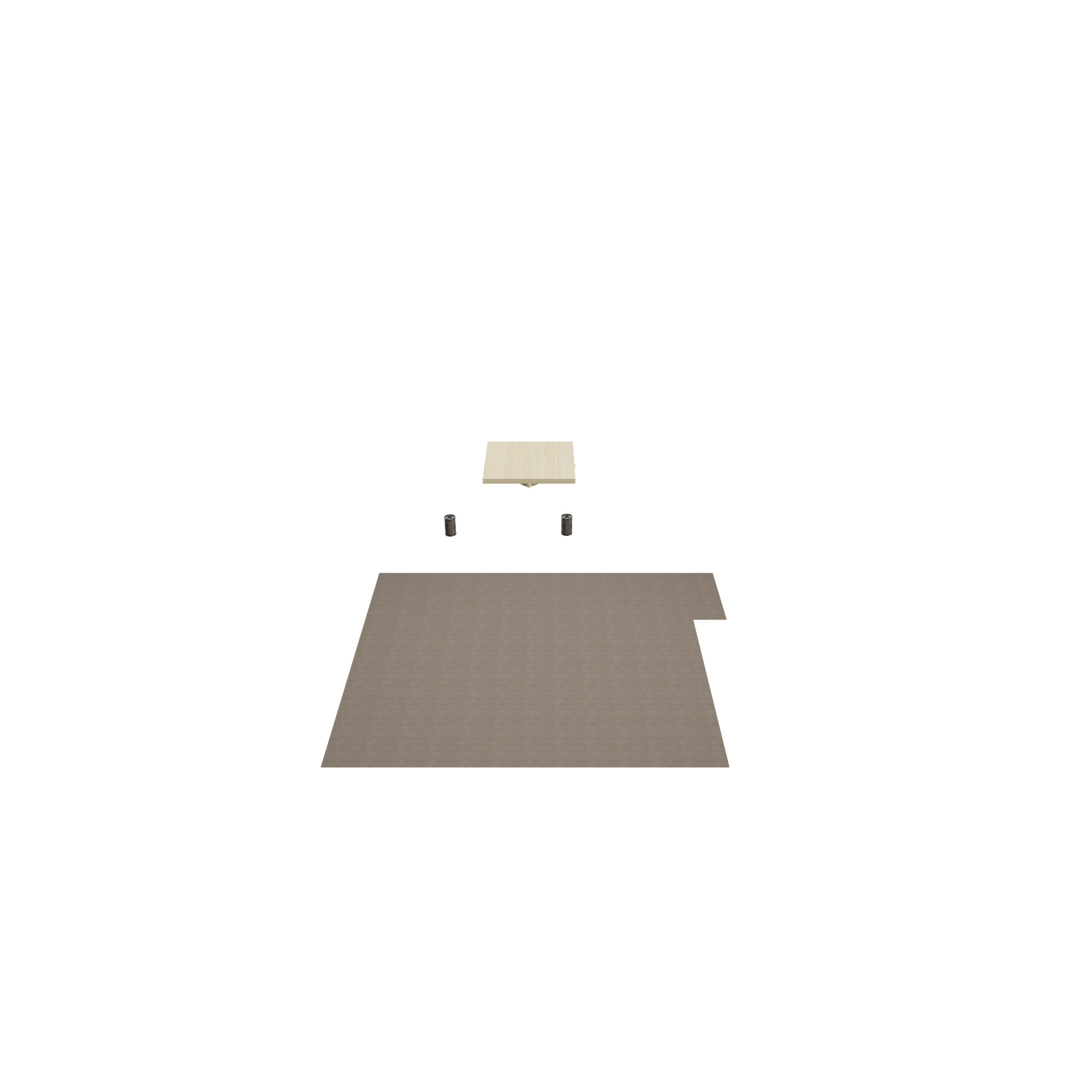}
	    \end{overpic}
    \end{subfigure}%
    \begin{subfigure}[b]{0.16\linewidth}
        \centering
        \vspace{-3mm}
        \begin{overpic}[width=\textwidth,  trim=600 600 600 700, clip]{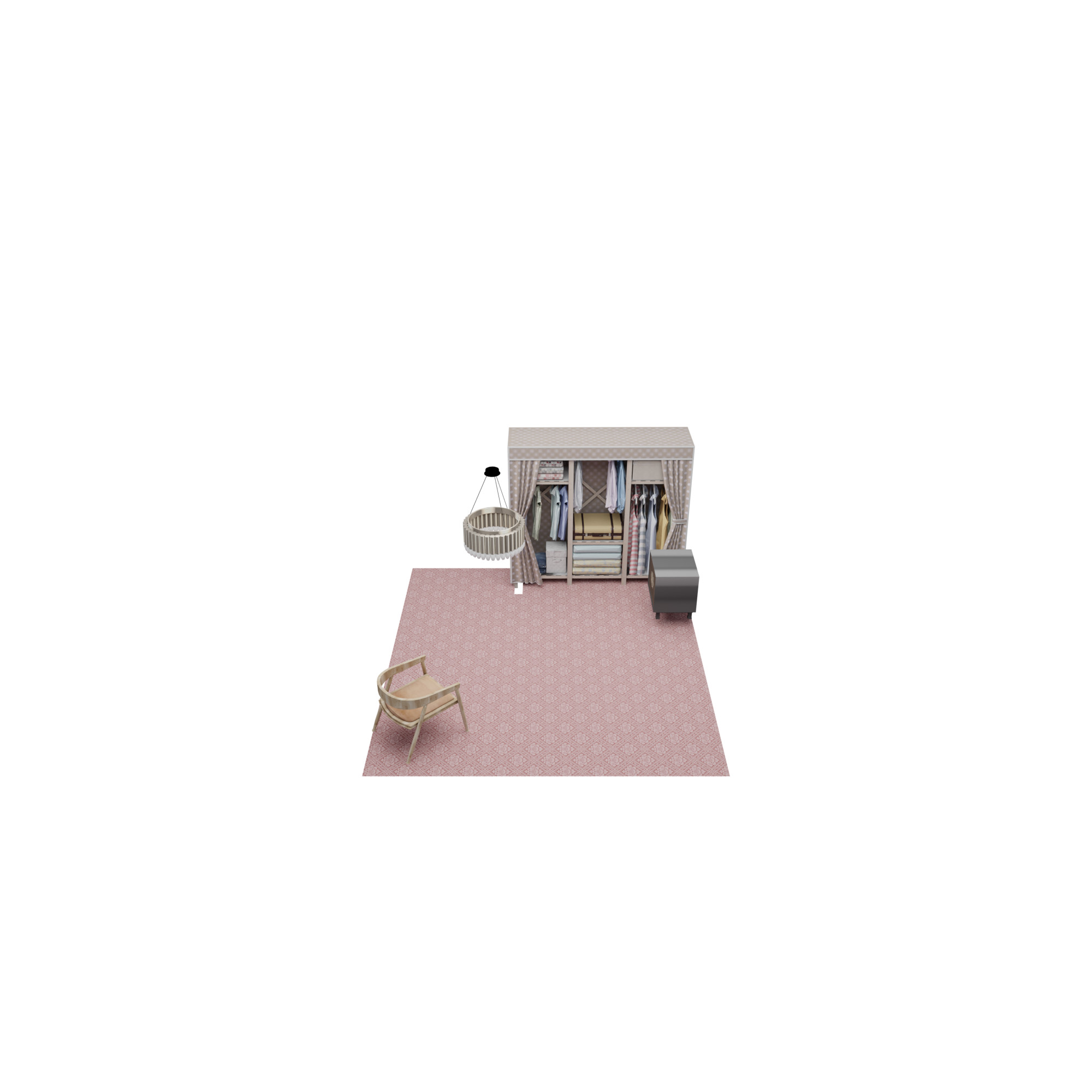}
	    \end{overpic}
    \end{subfigure}%
    \begin{subfigure}[b]{0.16\linewidth}
        \centering
        \begin{overpic}[width=\textwidth,  trim=550 700 500 500, clip]{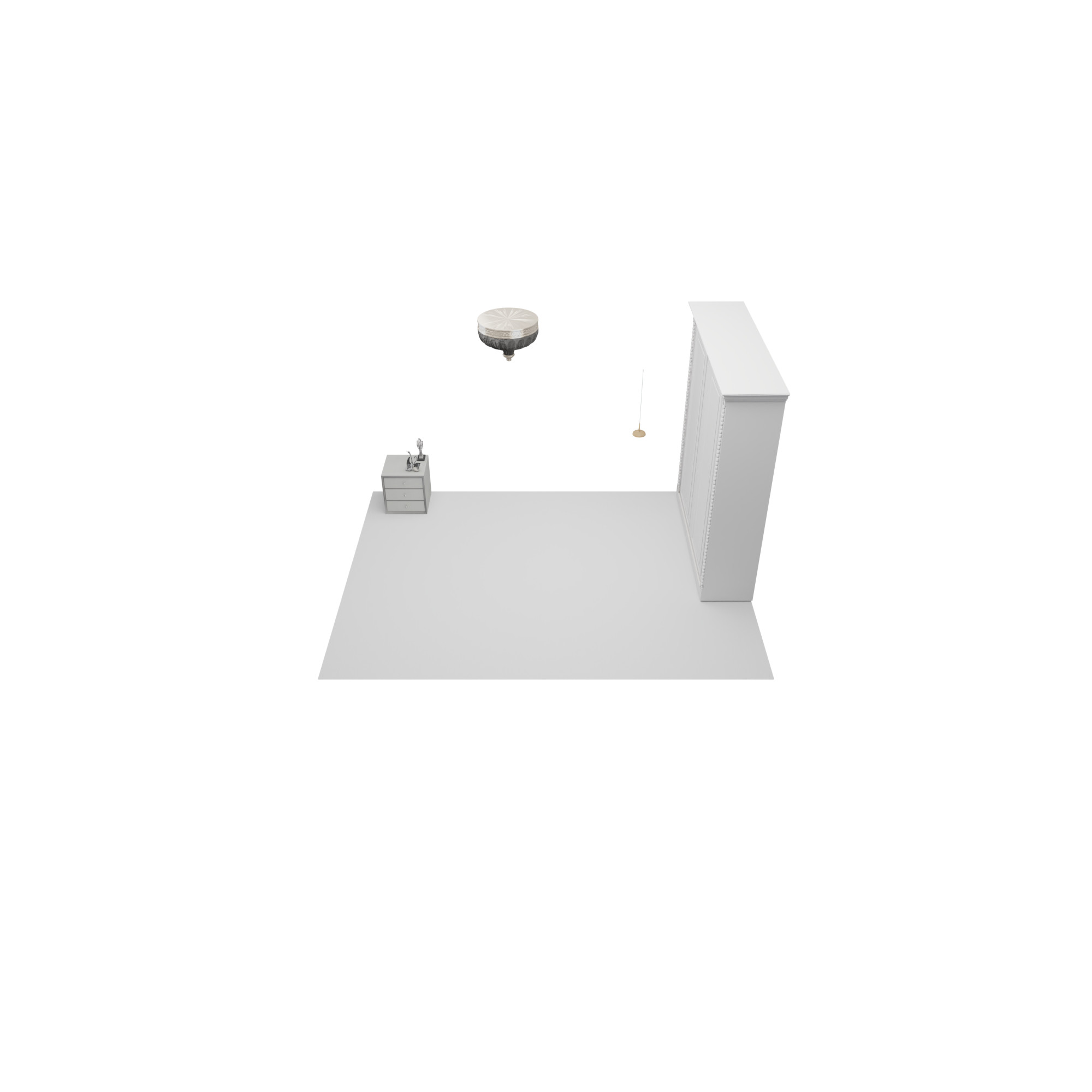}
	    \end{overpic}
    \end{subfigure}%
    \begin{subfigure}[b]{0.16\linewidth}
    \begin{overpic}[width=\textwidth,  trim=500 600 500 550 , clip]{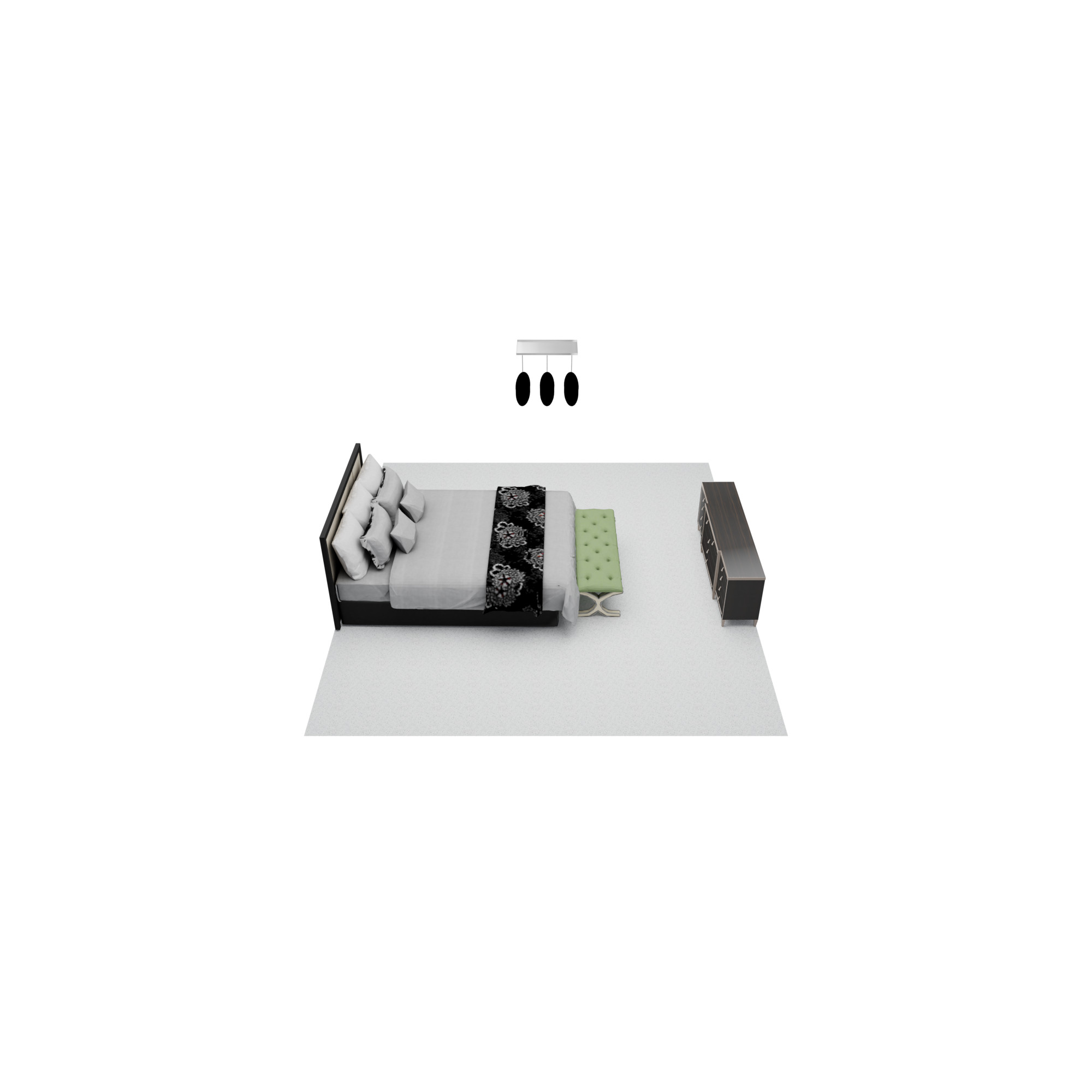}
	    \end{overpic}
    \end{subfigure}%
    \begin{subfigure}[b]{0.16\linewidth}
        \centering
            \begin{overpic}[width=\textwidth,  trim=600 700 600 550, clip]{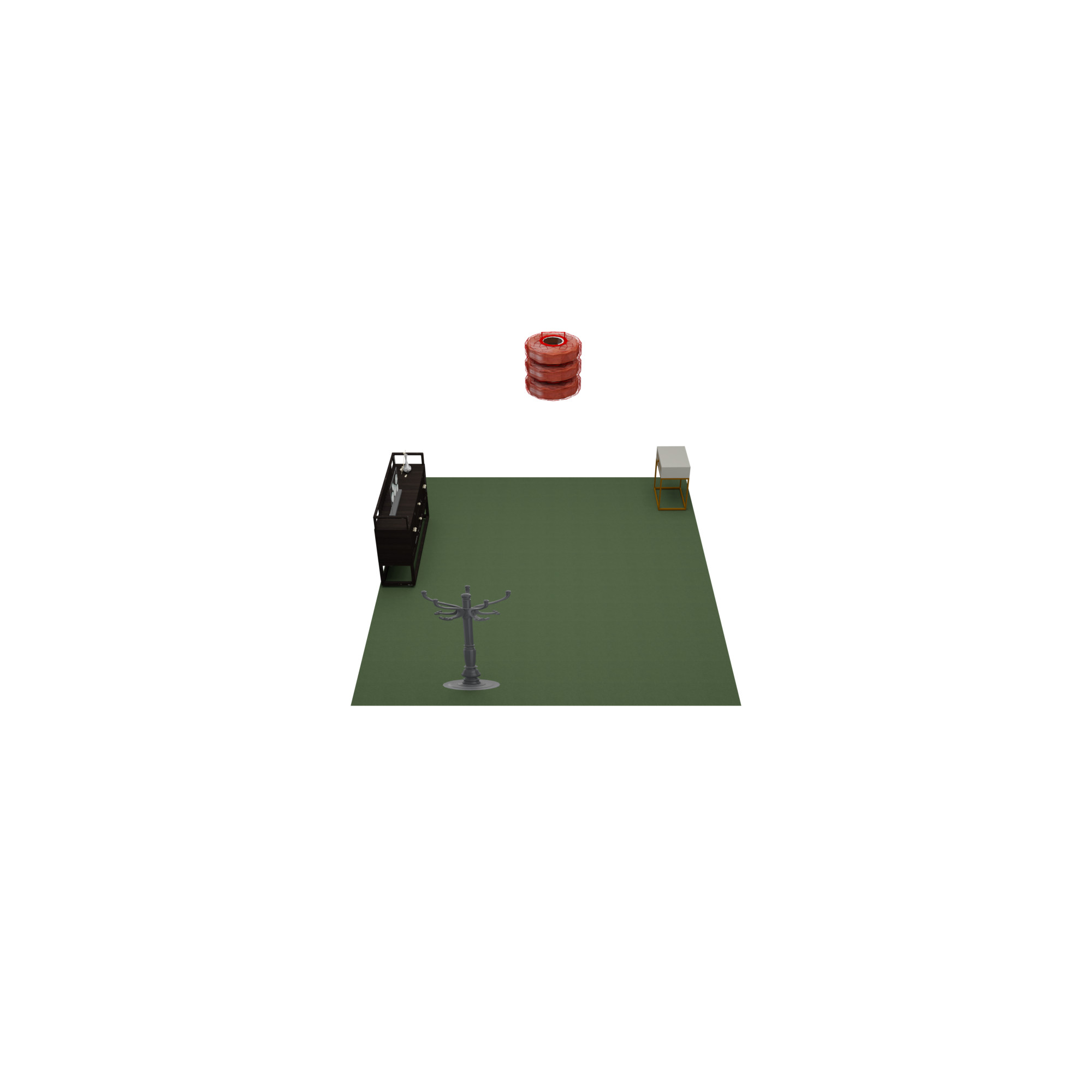}
        \end{overpic}
    \end{subfigure}%
    \vskip\baselineskip%
    \vspace{-1.75em}
    \begin{subfigure}[b]{0.16\linewidth}
        \centering
	    \begin{overpic}[width=\linewidth,  trim=640 600 600 900, clip]{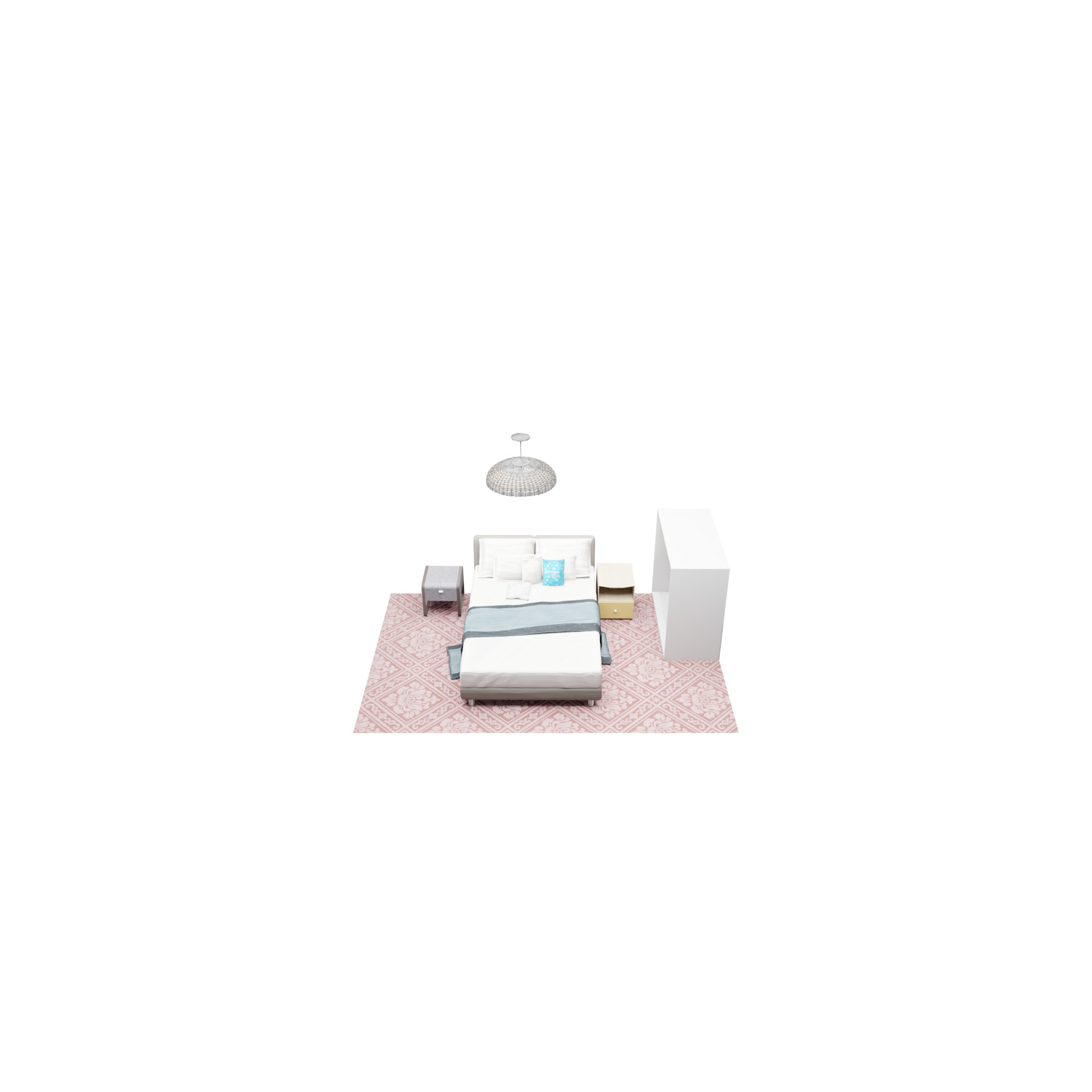}
	    \end{overpic}
    \end{subfigure}%
    \begin{subfigure}[b]{0.16\linewidth}
        \centering
        \begin{overpic}[width=\textwidth,  trim=600 600 600 760, clip]{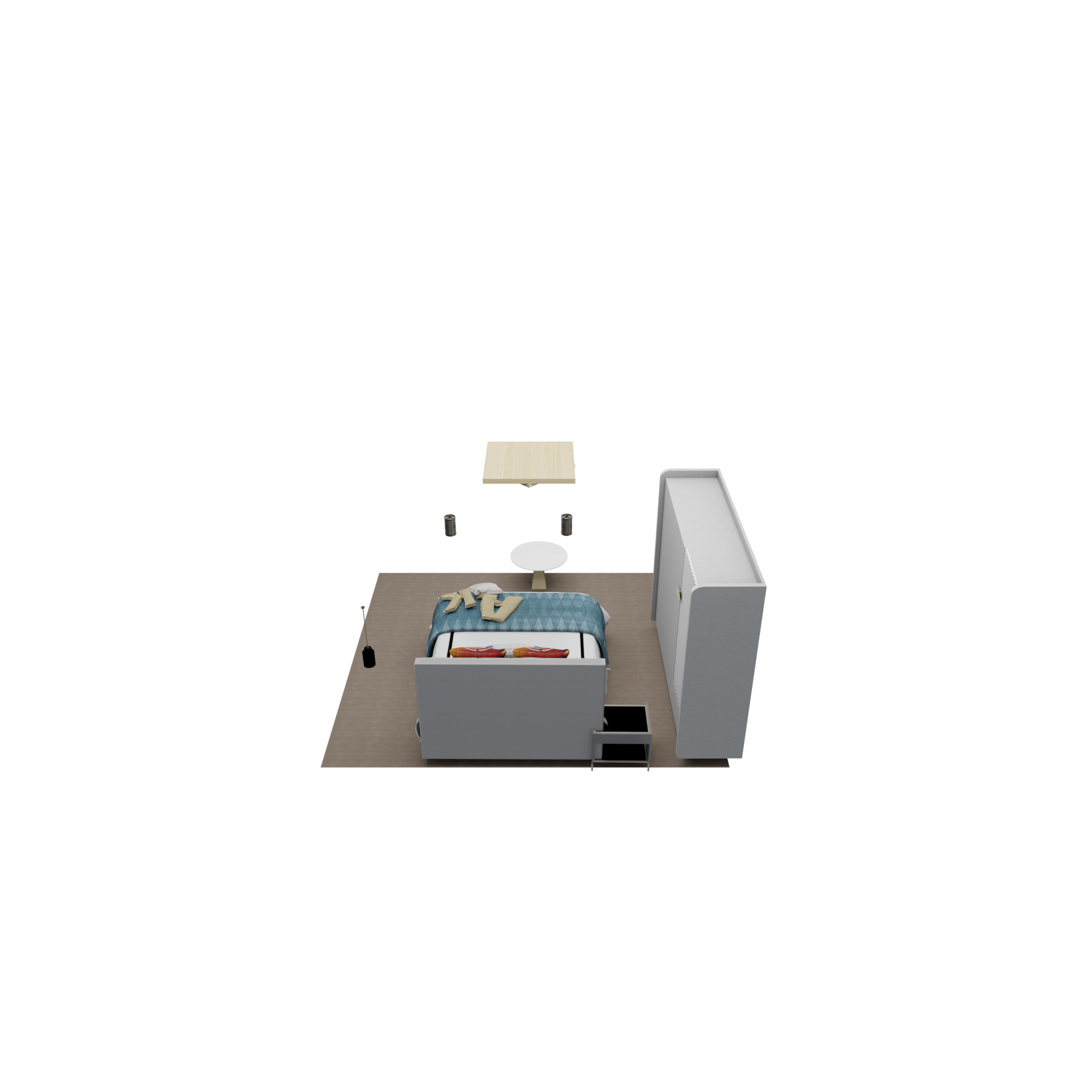}
	    \end{overpic}
    \end{subfigure}%
    \begin{subfigure}[b]{0.16\linewidth}
        \centering
        \begin{overpic}[width=\textwidth,  trim=600 600 600 700, clip]{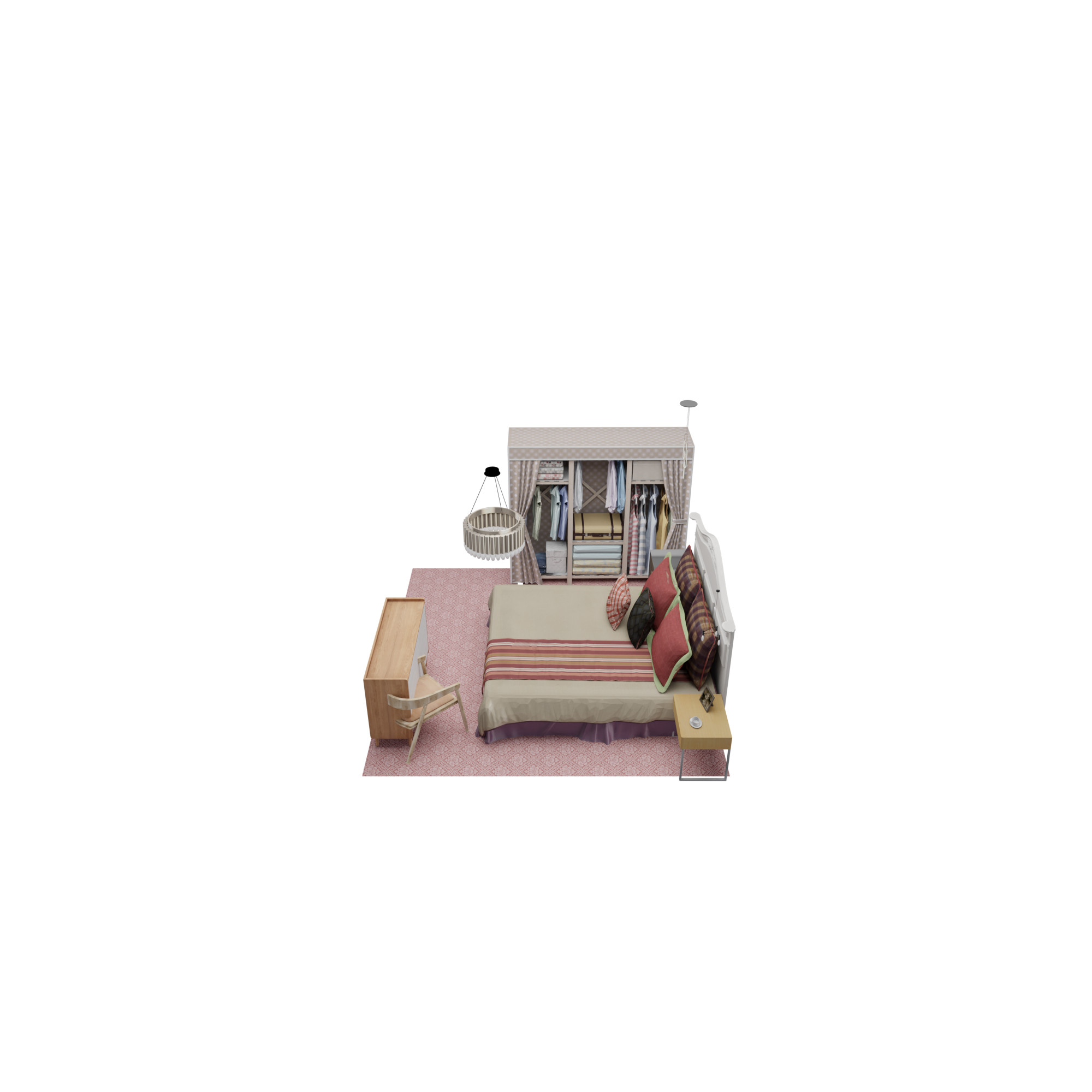}
	    \end{overpic}
    \end{subfigure}%
    \begin{subfigure}[b]{0.16\linewidth}
        \centering
        \begin{overpic}[width=\textwidth,  trim=550 700 500 500, clip]{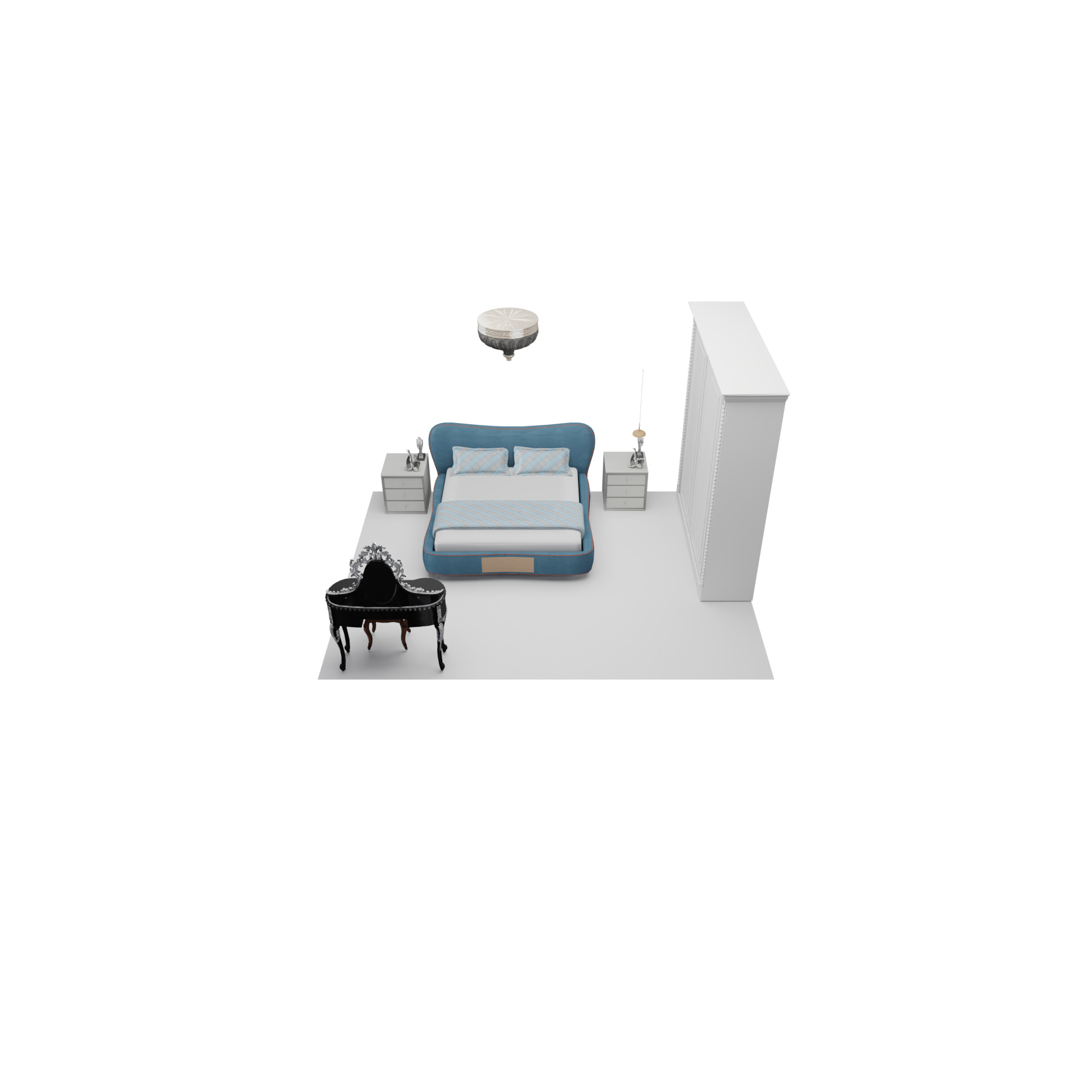}
	    \end{overpic}
    \end{subfigure}%
    \begin{subfigure}[b]{0.16\linewidth}
    \begin{overpic}[width=\textwidth,  trim=500 600 500 550, clip]{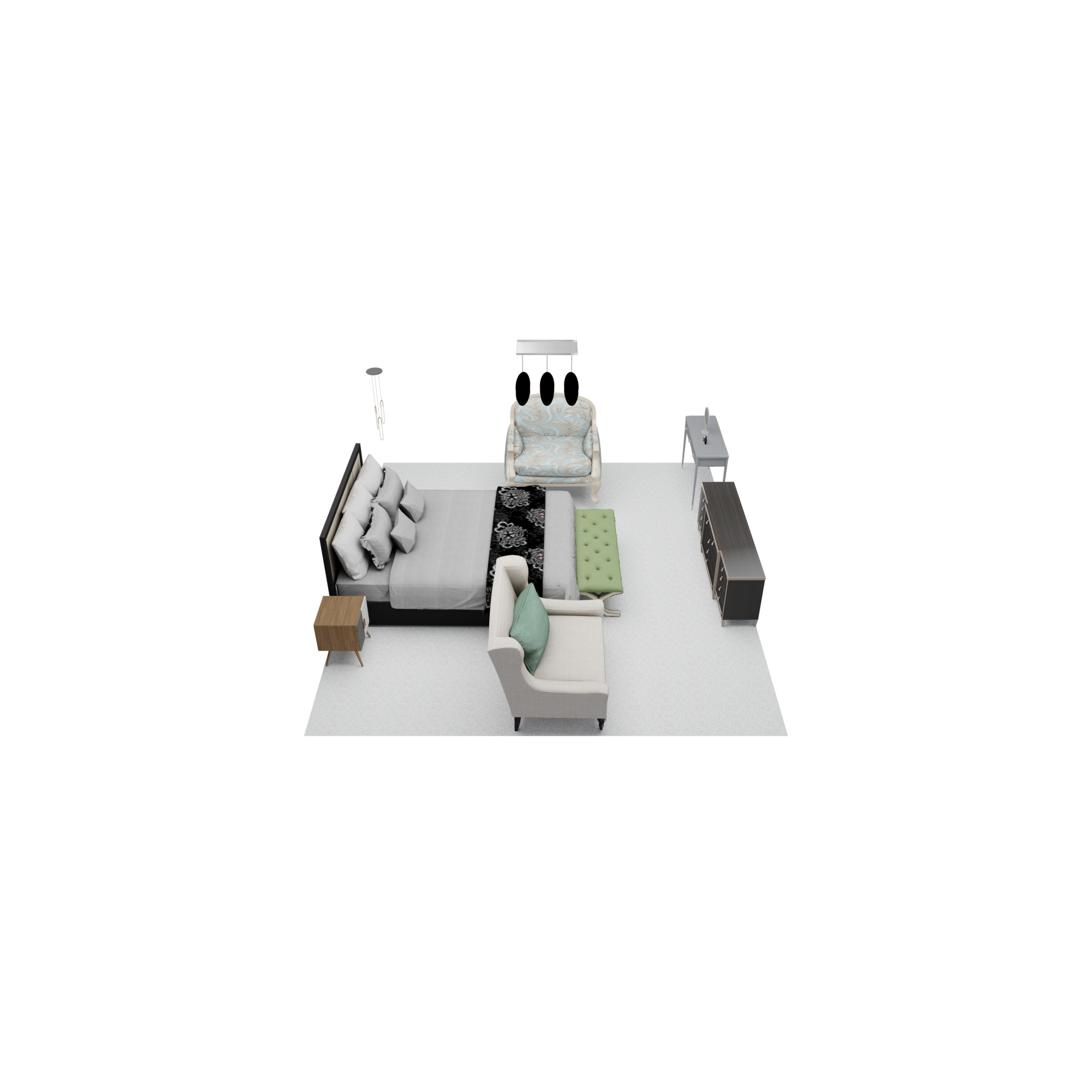}
	    \end{overpic}
    \end{subfigure}%
    \begin{subfigure}[b]{0.16\linewidth}
        \centering
            \begin{overpic}[width=\textwidth,  trim=600 700 600 550, clip]{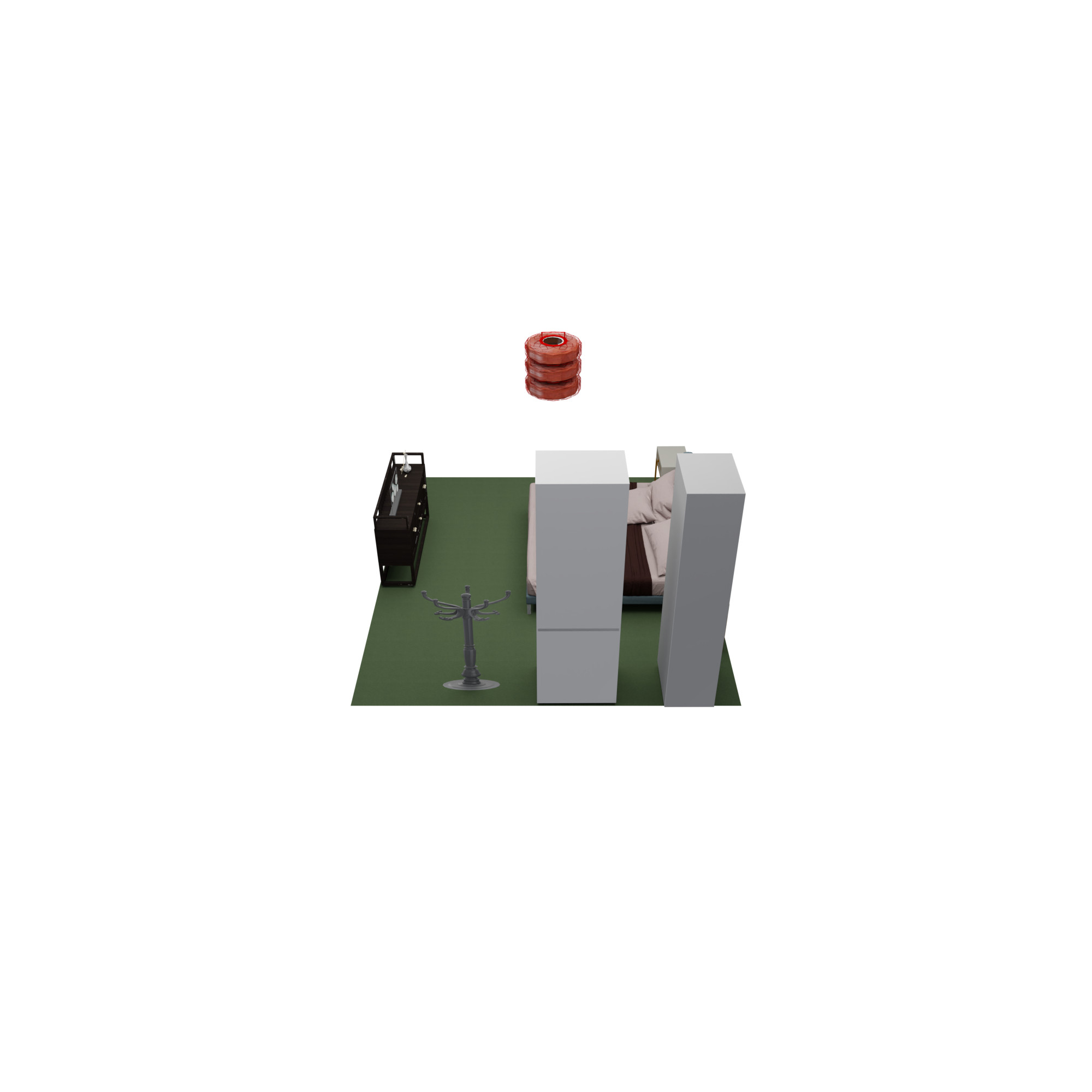}
        \end{overpic}
    \end{subfigure}%
    \caption{\textbf{Partial Scene Completion:} We show qualitative visualizations for the scene completion task. The generated layouts all respect the condition they were generated with (top row) and are all plausible room layouts.}
    \label{fig:partial_scene_completion}
\end{figure*}

%%%%%%%%%%%%%%%%%%%%%%%%%%%%%%%%%%%%%%%%%%%%%%%%%%%%%%%%%%%%%%%%%%%%%%%%%%%%%%%%%%%%%%%%%%%%%%%%%%%%%%%%
\para{Scene Completion} In order to perform scene completion, we prepend both $S$ and $C$ with the tokens of the objects that already exist in the scene. Examples are shown in Figure~\ref{fig:partial_scene_completion}. We can see that our method successfully generates plausible room layouts that respect the condition shown in the top row.

%%%%%%%%%%%%%%%%%%%%%%%%%%%%%%%%%%%%%%%%%%%%%%%%%%%%%%%%%%%%%%%%%%%%%%%%%%%%%%%%%%%%%%%%%%%%%%%%%%%%%%%%
\para{Outlier Detection} To estimate the likelihood of each token, we follow \cite{salazar-etal-2020-masked} and replace the token at $i$th position with \mask. This can be performed in parallel by creating a batch in which only one element is replaced with \mask. This is shown in Fig. \ref{fig:architecture}. The likelihood of one bounding box then is the product of likelihoods of all element of the bounding box. Objects with low likelihood can then be resampled. Based on user constraints, only one parameter may be changed as we allow for arbitrary conditioning as explained in the next section.

\begin{figure*}[b]
    \centering
    \vspace{-1.5em}
    % \hfill
    \begin{subfigure}[b]{0.16\linewidth}
        \centering
	    \begin{overpic}[width=\linewidth,  trim=500 500 500 500, clip]{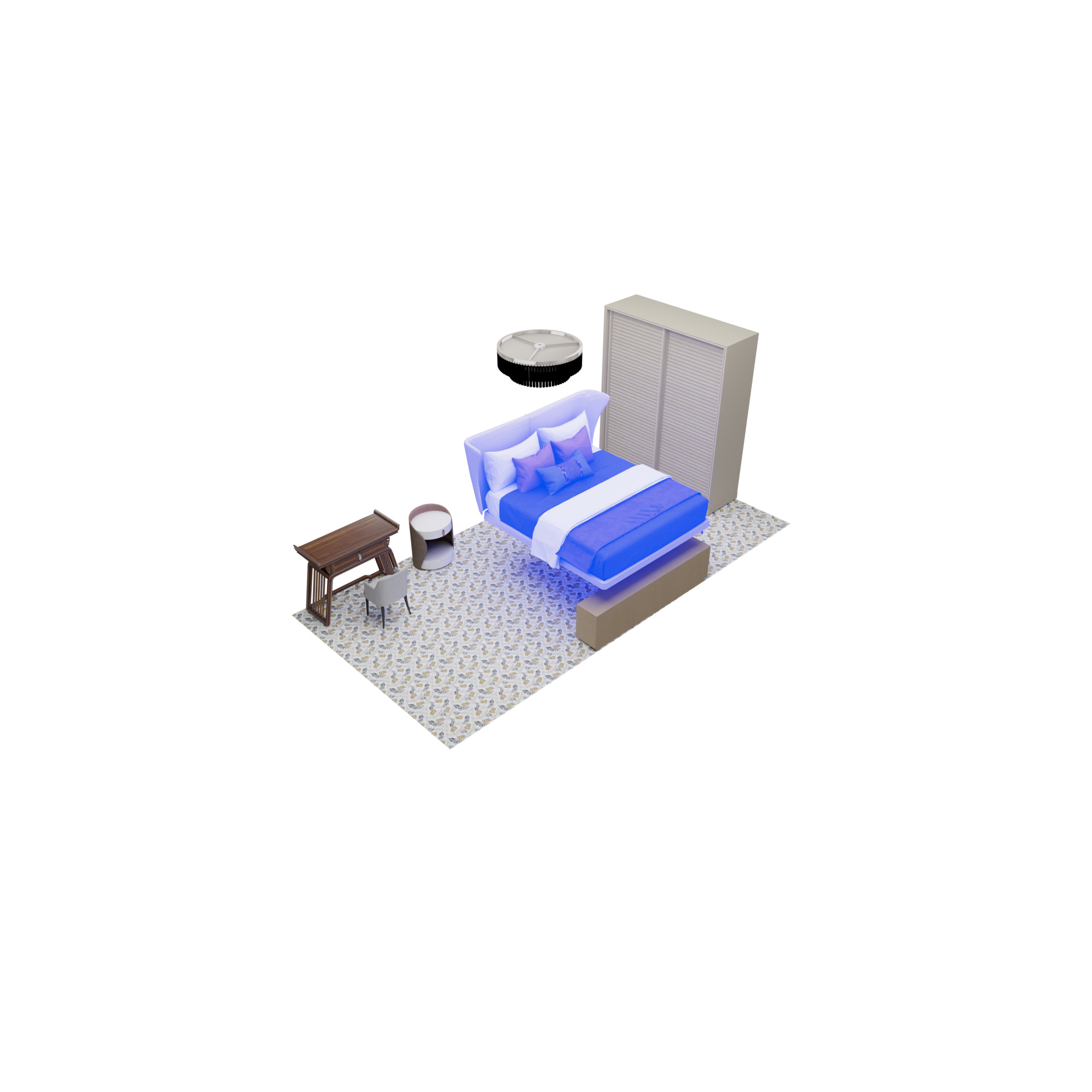}
	    \end{overpic}
    \end{subfigure}%
    \begin{subfigure}[b]{0.16\linewidth}
        \centering
        \begin{overpic}[width=\textwidth,  trim=500 500 500 500, clip]{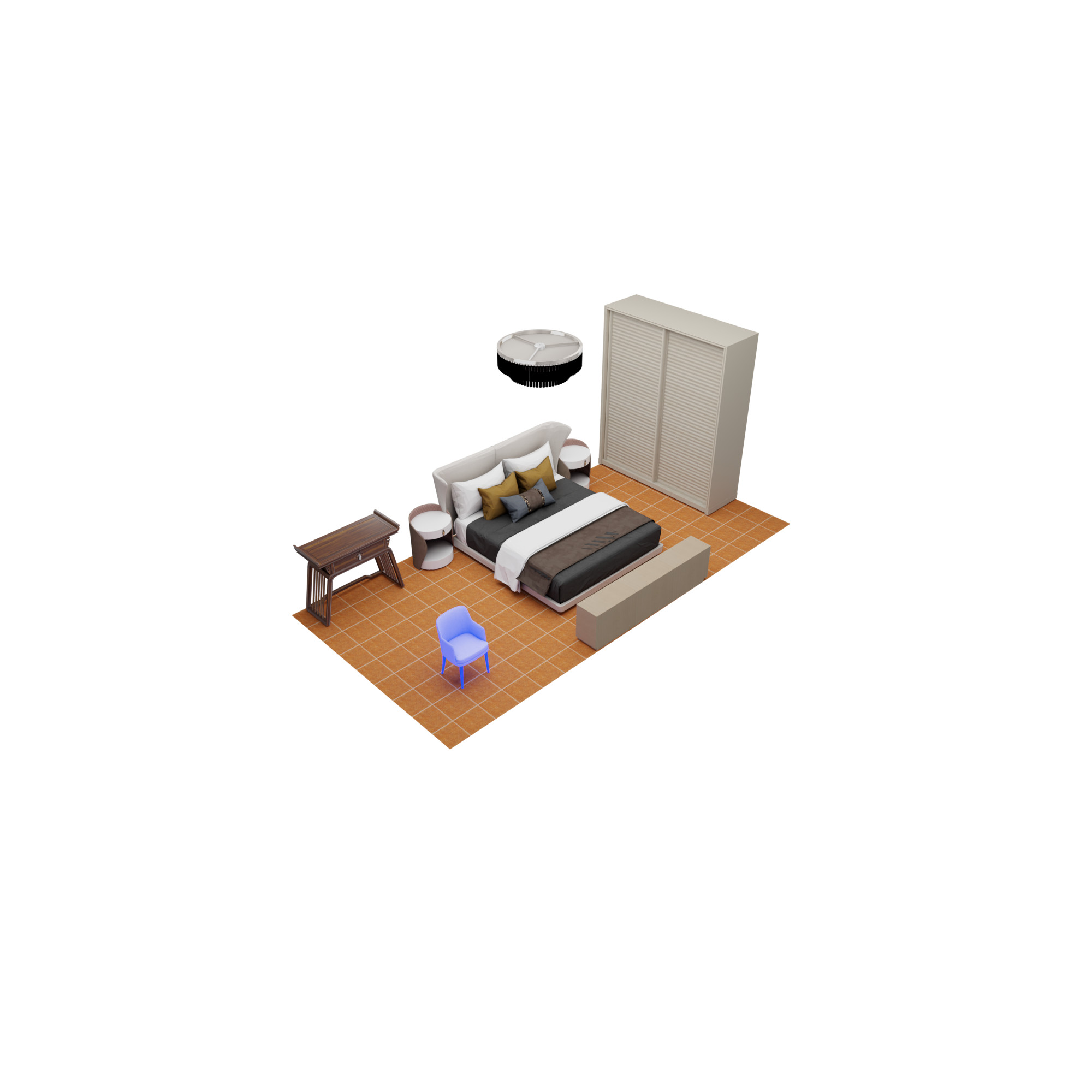}
	    \end{overpic}
    \end{subfigure}%
    \begin{subfigure}[b]{0.16\linewidth}
        \centering
        \vspace{-3mm}
        \begin{overpic}[width=\textwidth,  trim=100 300 400 700, clip]{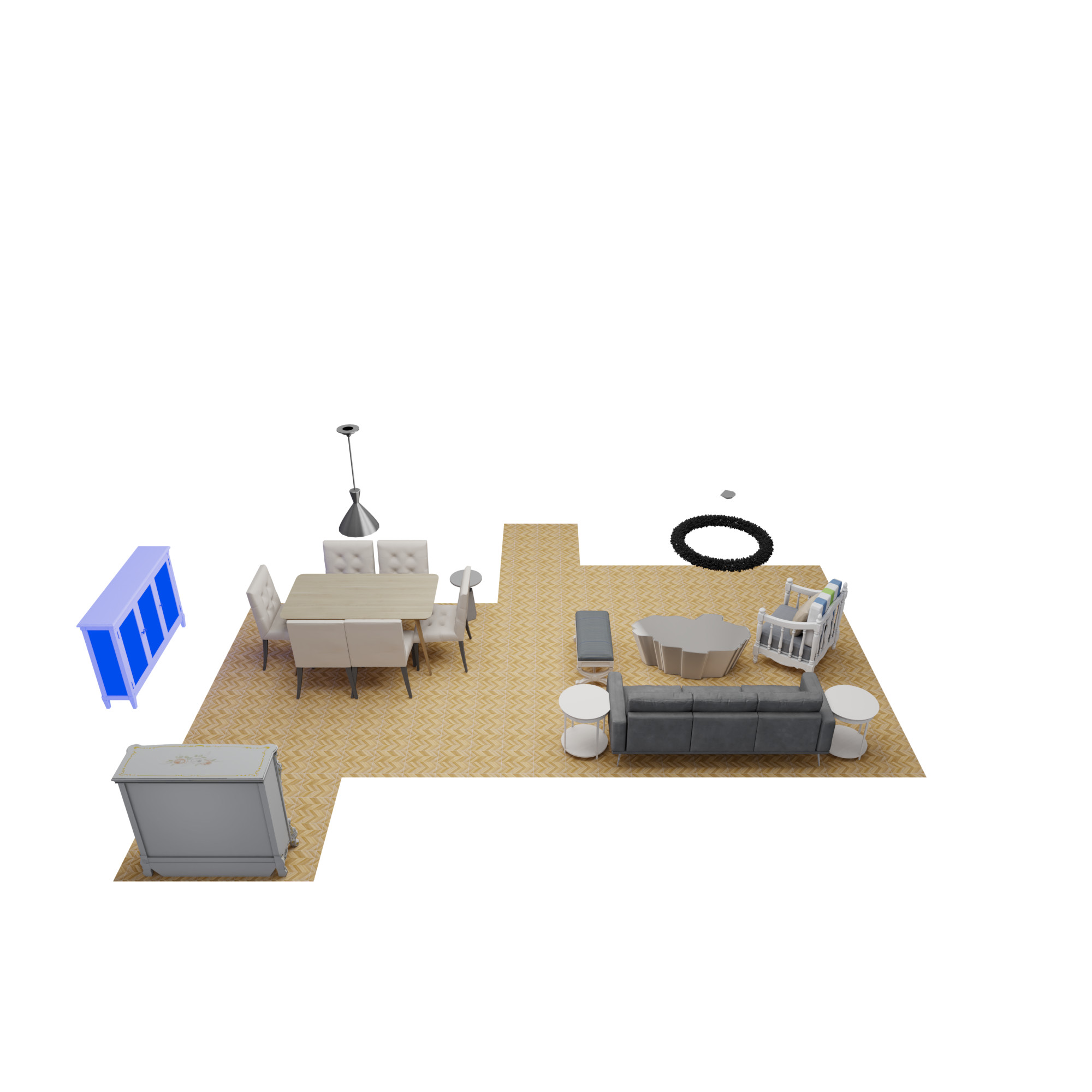}
	    \end{overpic}
    \end{subfigure}%
    \begin{subfigure}[b]{0.16\linewidth}
        \centering
        \begin{overpic}[width=\textwidth,  trim=400 300 400 650, clip]{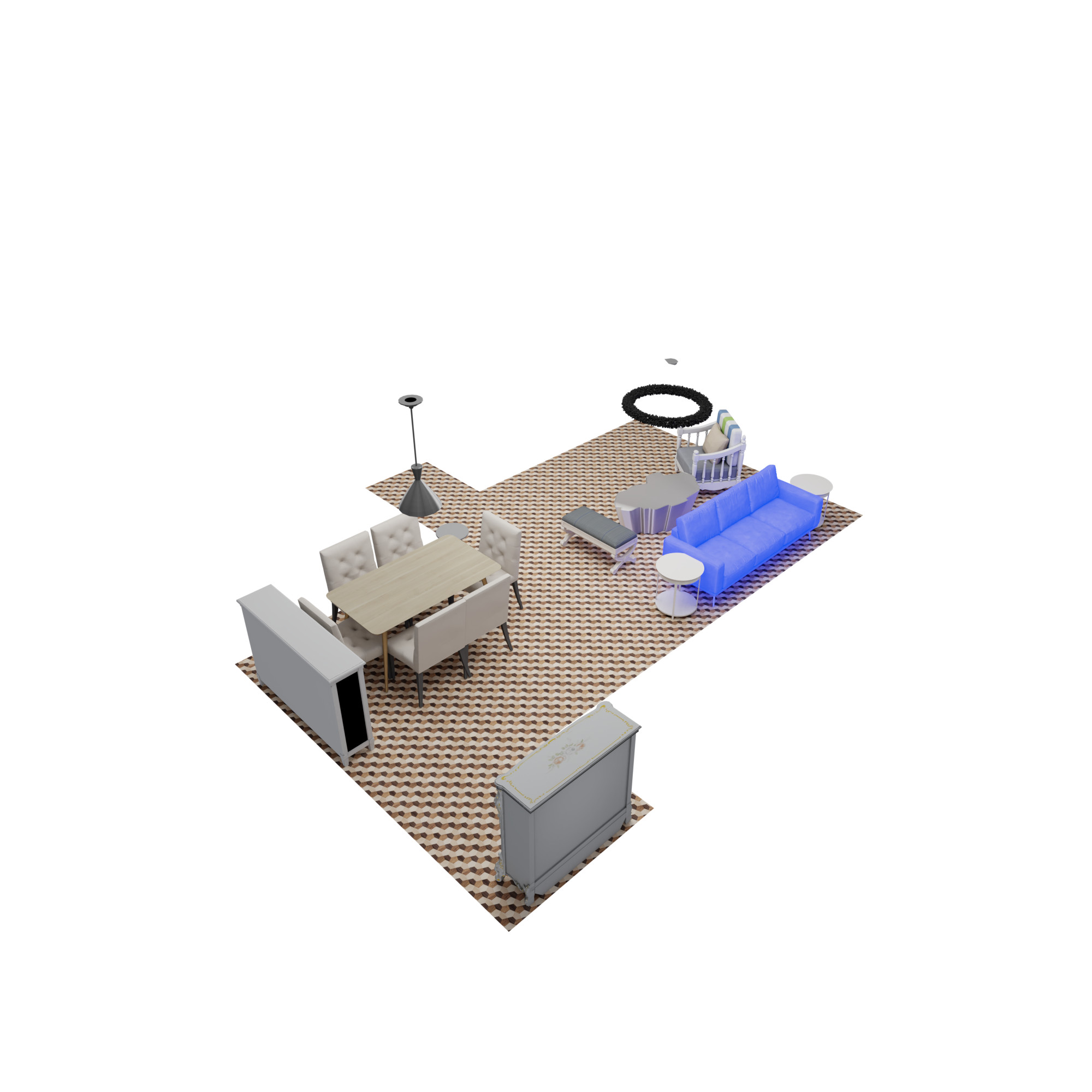}
	    \end{overpic}
    \end{subfigure}%
    \begin{subfigure}[b]{0.16\linewidth}
    \begin{overpic}[width=\textwidth,  trim=100 200 200 500, clip]{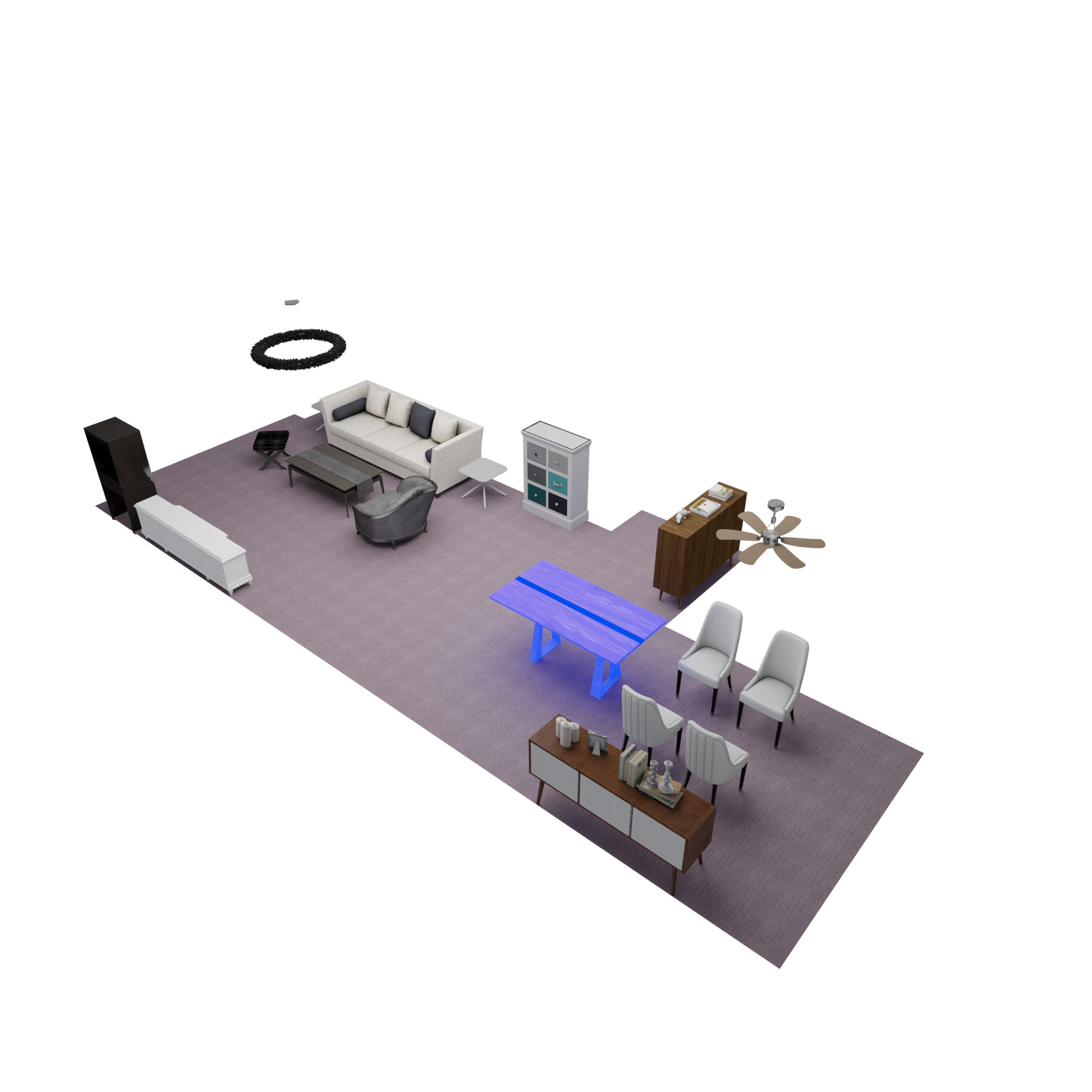}
	    \end{overpic}
    \end{subfigure}%
    \begin{subfigure}[b]{0.16\linewidth}
        \centering
            \begin{overpic}[width=\textwidth,  trim=100 400 400 550, clip]{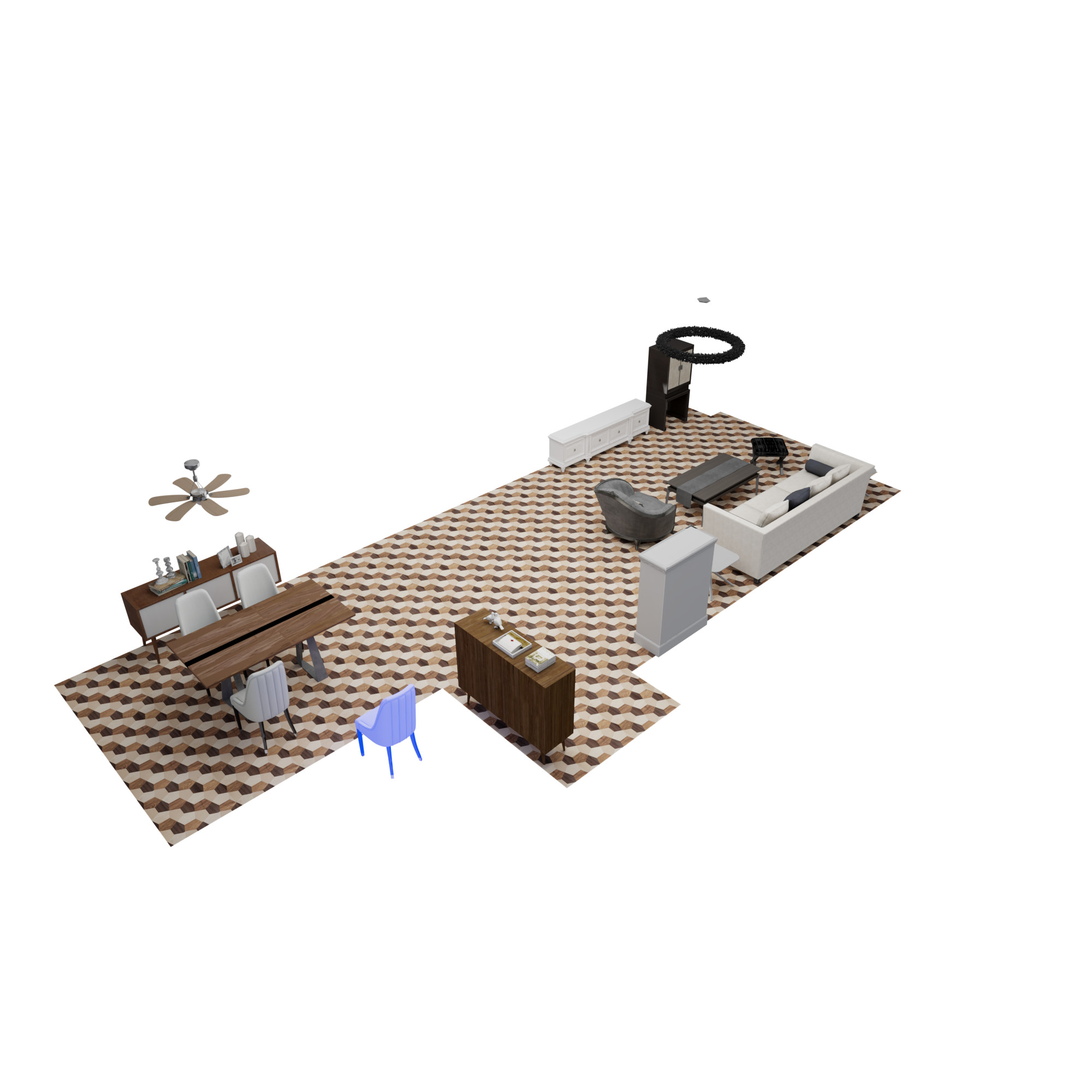}
        \end{overpic}
    \end{subfigure}%
    \vskip\baselineskip%
    \vspace{-1.75em}
    \begin{subfigure}[b]{0.16\linewidth}
        \centering
	    \begin{overpic}[width=\linewidth,  trim=500 500 500 500, clip]{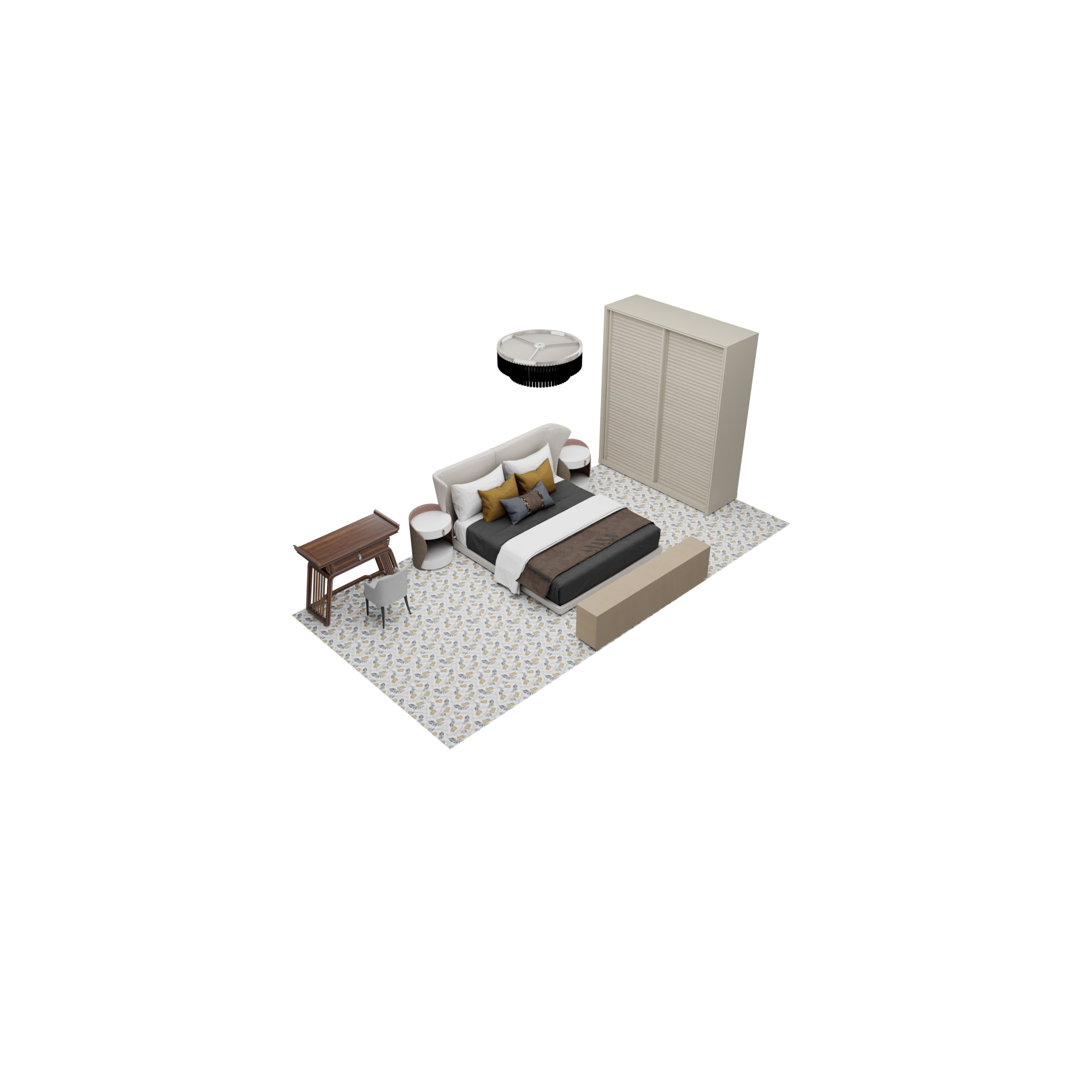}
	    \end{overpic}
    \end{subfigure}%
    \begin{subfigure}[b]{0.16\linewidth}
        \centering
        \begin{overpic}[width=\textwidth,  trim=500 500 500 500, clip]{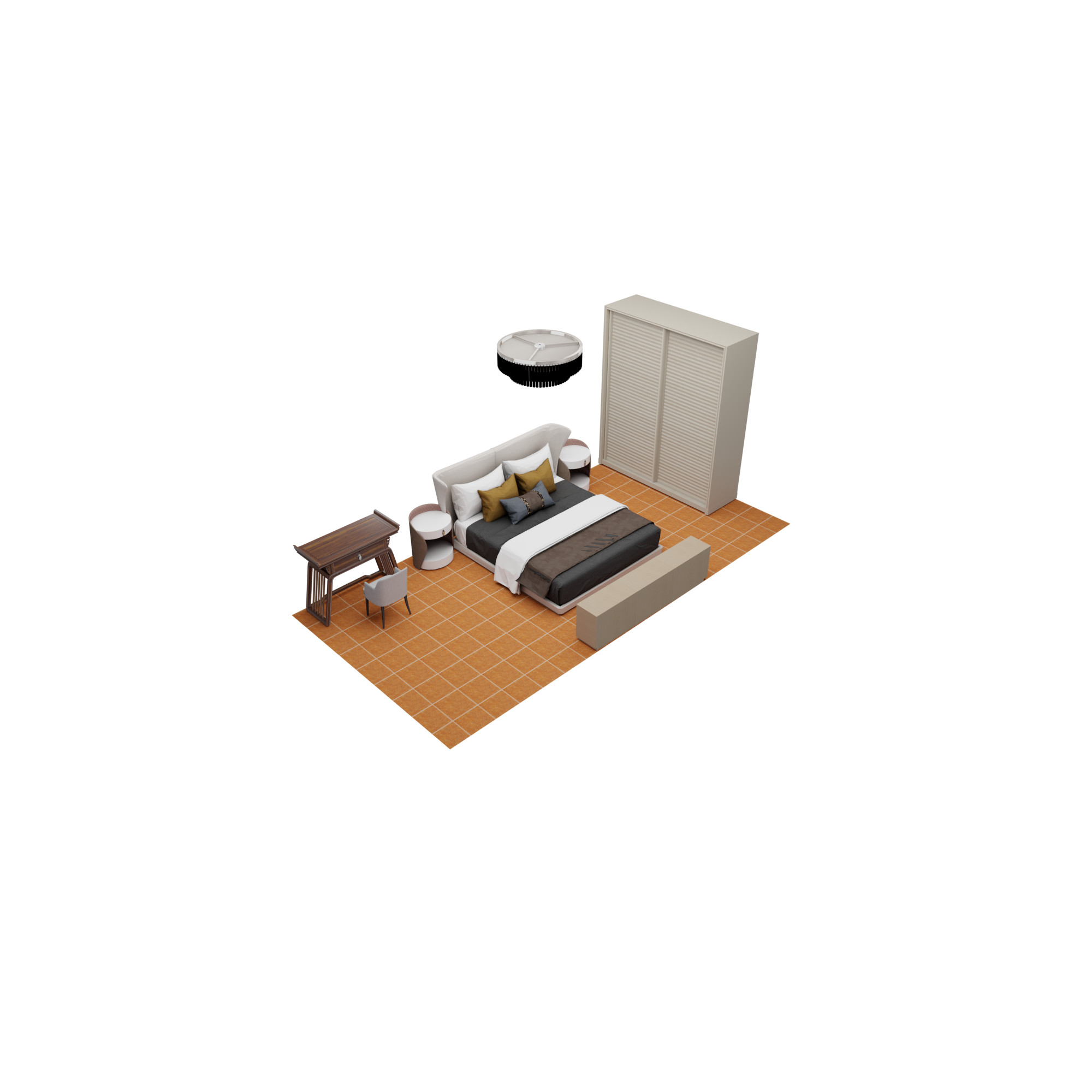}
	    \end{overpic}
    \end{subfigure}%
    \begin{subfigure}[b]{0.16\linewidth}
        \centering
        \begin{overpic}[width=\textwidth,  trim=100 300 400 700, clip]{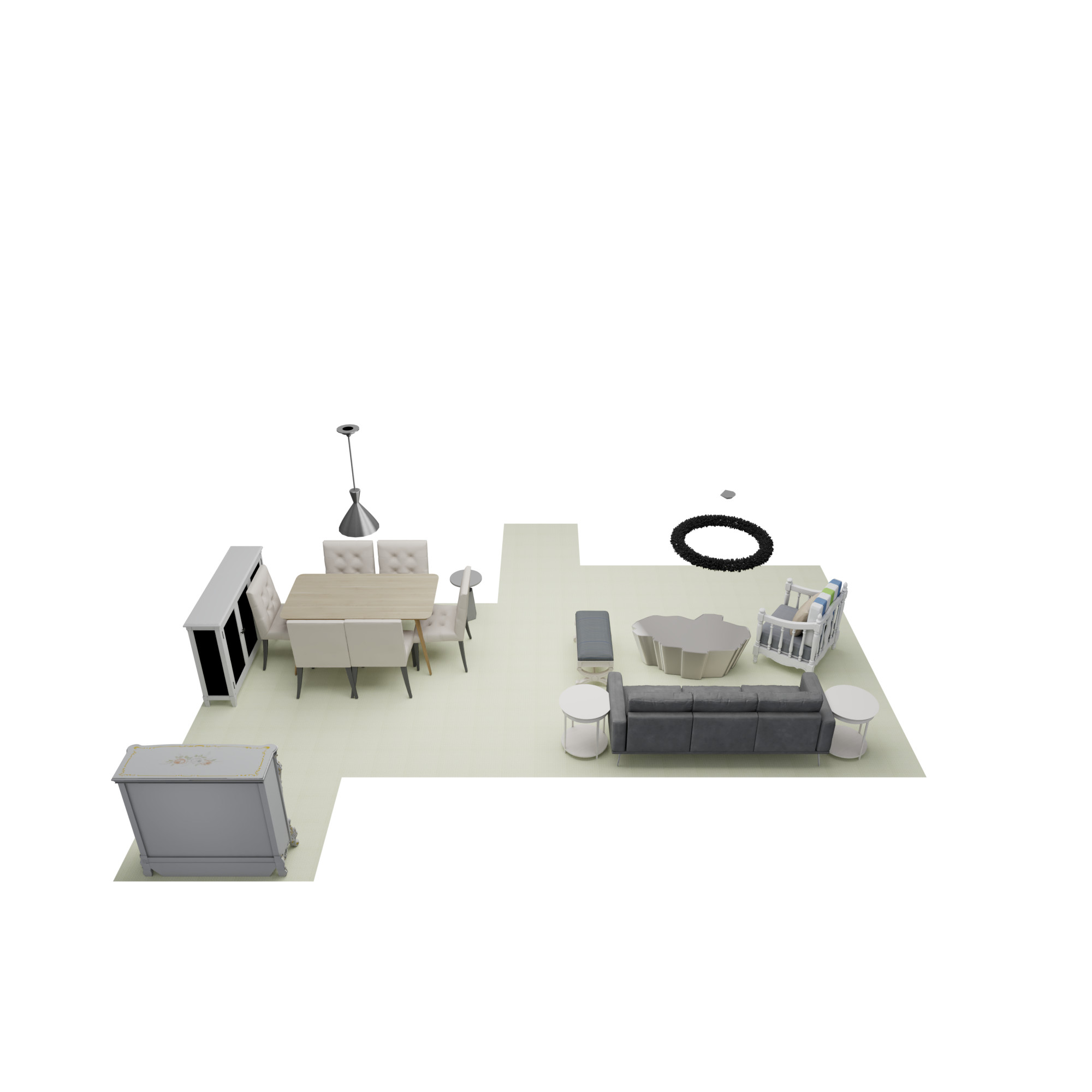}
	    \end{overpic}
    \end{subfigure}%
    \begin{subfigure}[b]{0.16\linewidth}
        \centering
        \begin{overpic}[width=\textwidth,  trim=400 300 400 650, clip]{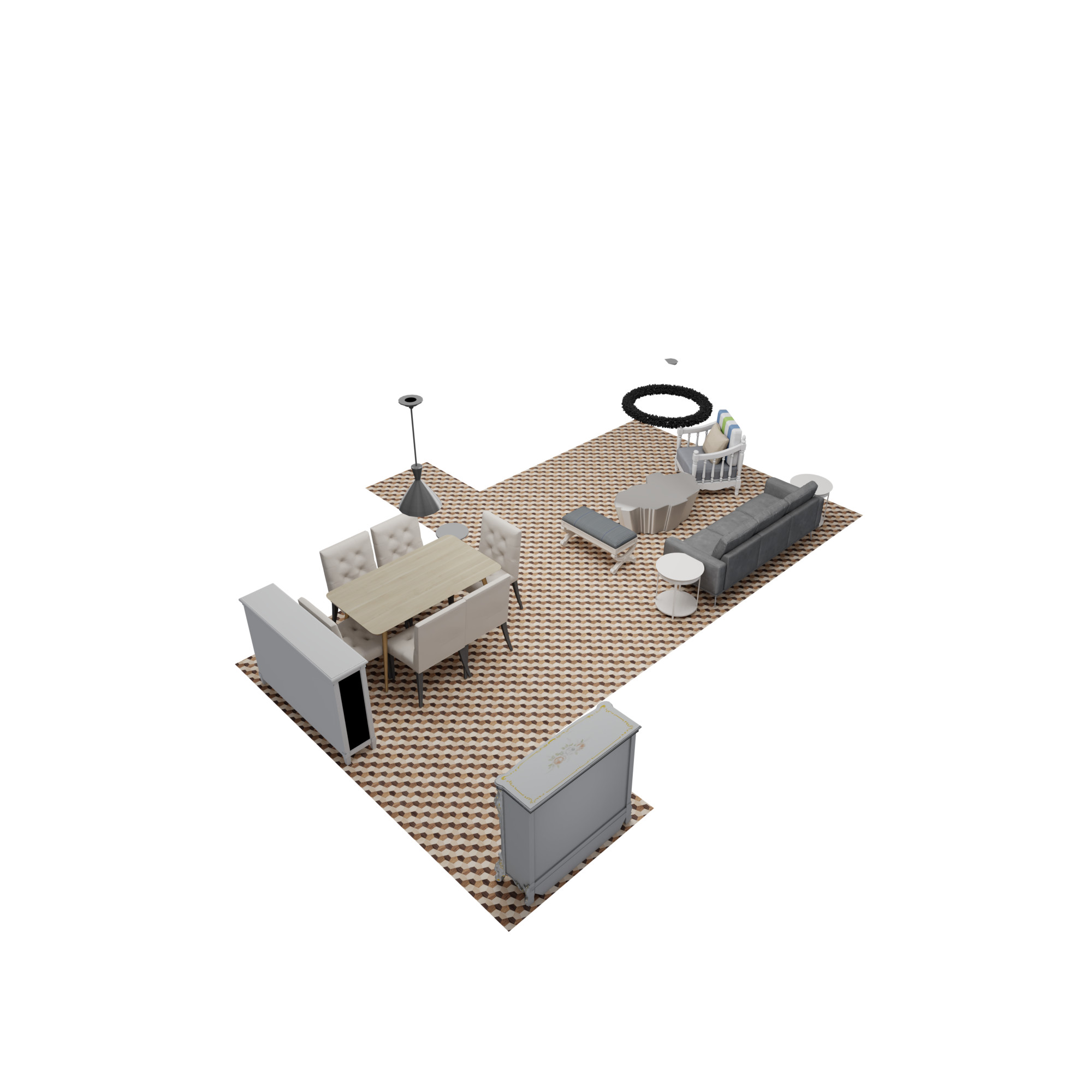}
	    \end{overpic}
    \end{subfigure}%
    \begin{subfigure}[b]{0.16\linewidth}
    \begin{overpic}[width=\textwidth,  trim=100 200 200 500, clip]{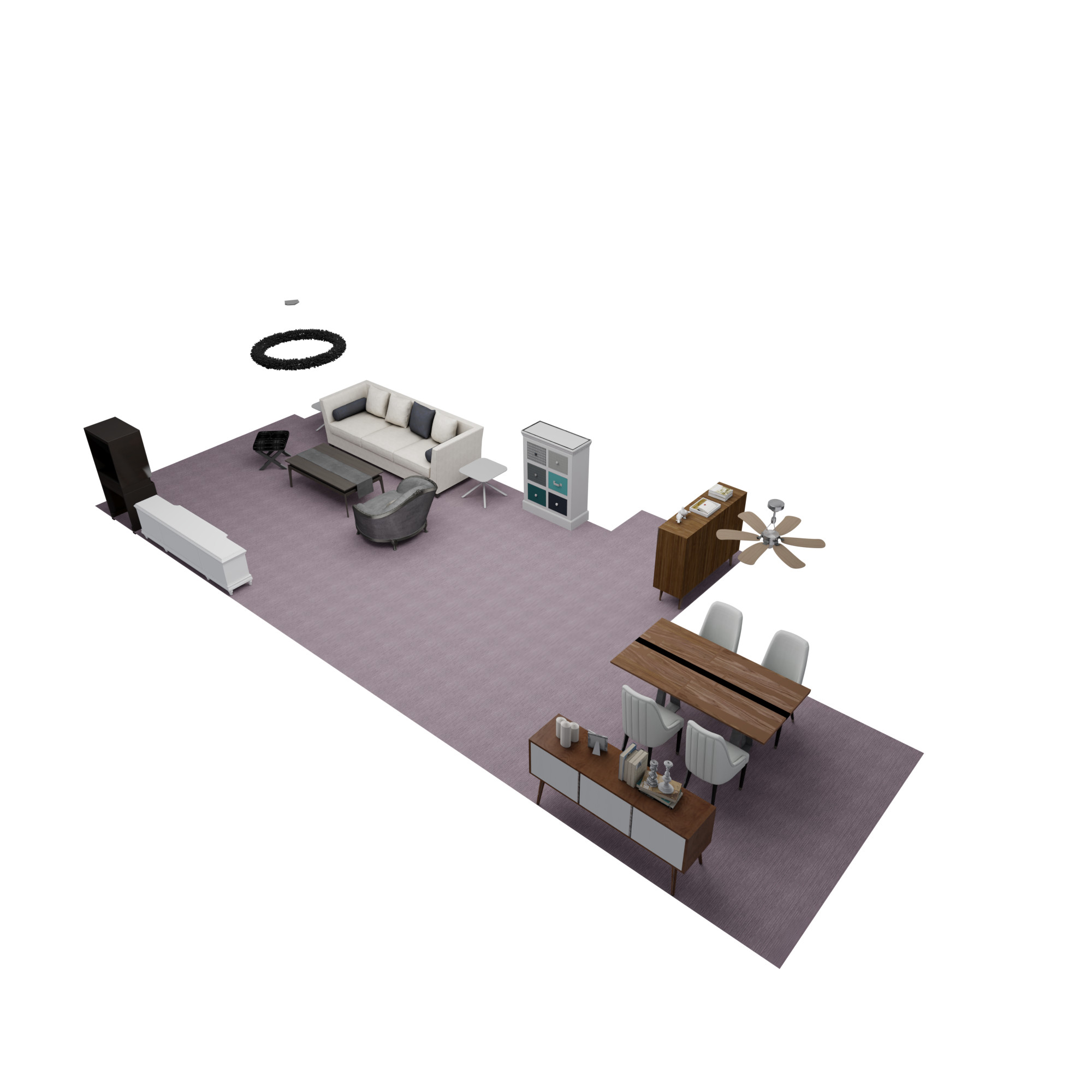}
	    \end{overpic}
    \end{subfigure}%
    \begin{subfigure}[b]{0.16\linewidth}
        \centering
            \begin{overpic}[width=\textwidth,  trim=100 400 400 550, clip]{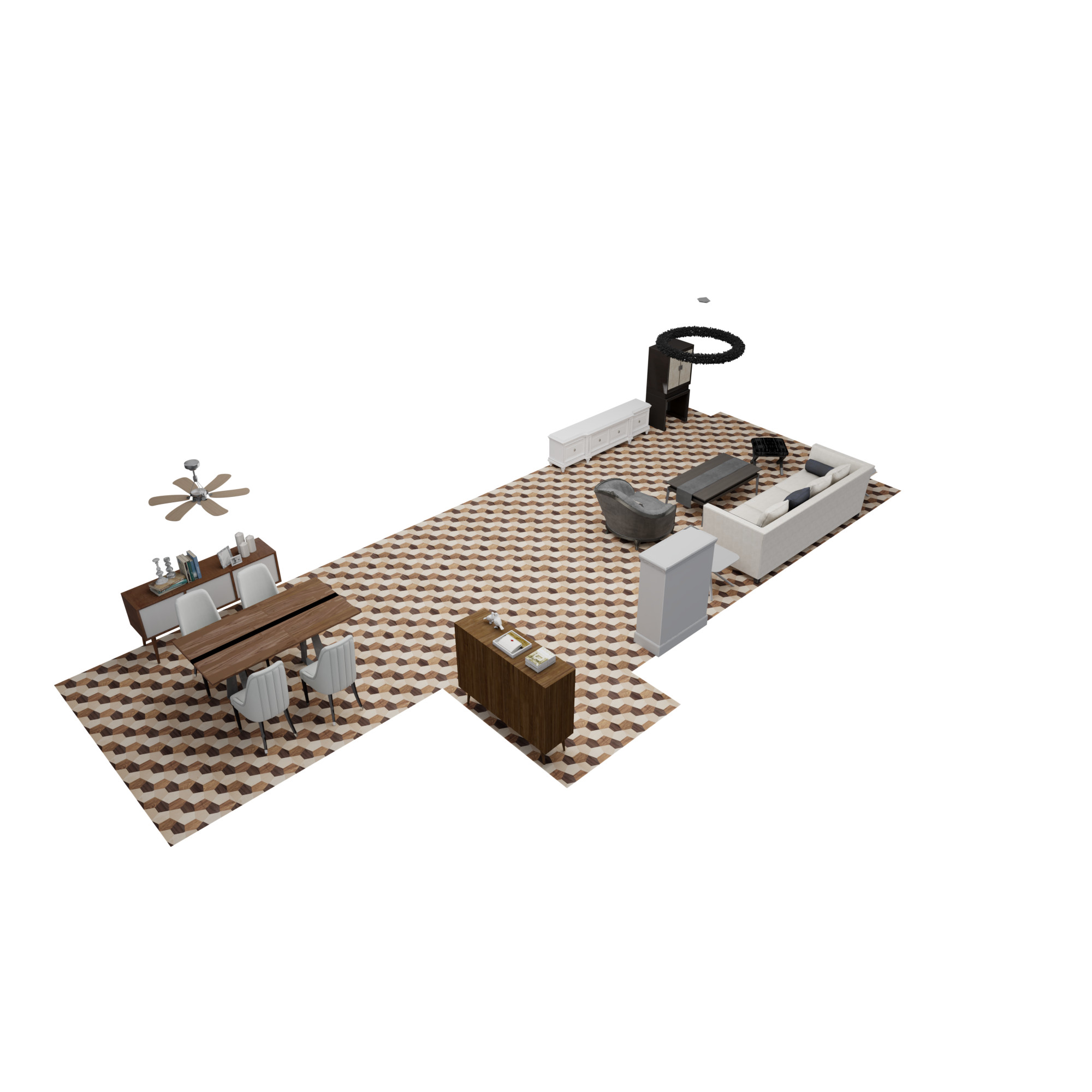}
        \end{overpic}
    \end{subfigure}%
    
    \caption{\textbf{Outlier detection:} Our model can utilize bidirectional attention to reason about unlikely arrangements of furniture. We can then sample new attributes that create a more likely layout. \textbf{Top row:} An object is perturbed to create an outlier (highlighted in blue). \textbf{Bottom row:} The object can be identified by its low likelihood, and new attributes sampled which place it more naturally. Please zoom in for best viewing.}
    \label{fig:outlier_detection}
\vspace{-0.75em}
\end{figure*}

\begin{figure*}[t!]
    \centering
    \begin{subfigure}[b]{0.22\linewidth}
        \centering
	    \includegraphics[width=\textwidth, trim=450 500 500 450, clip]{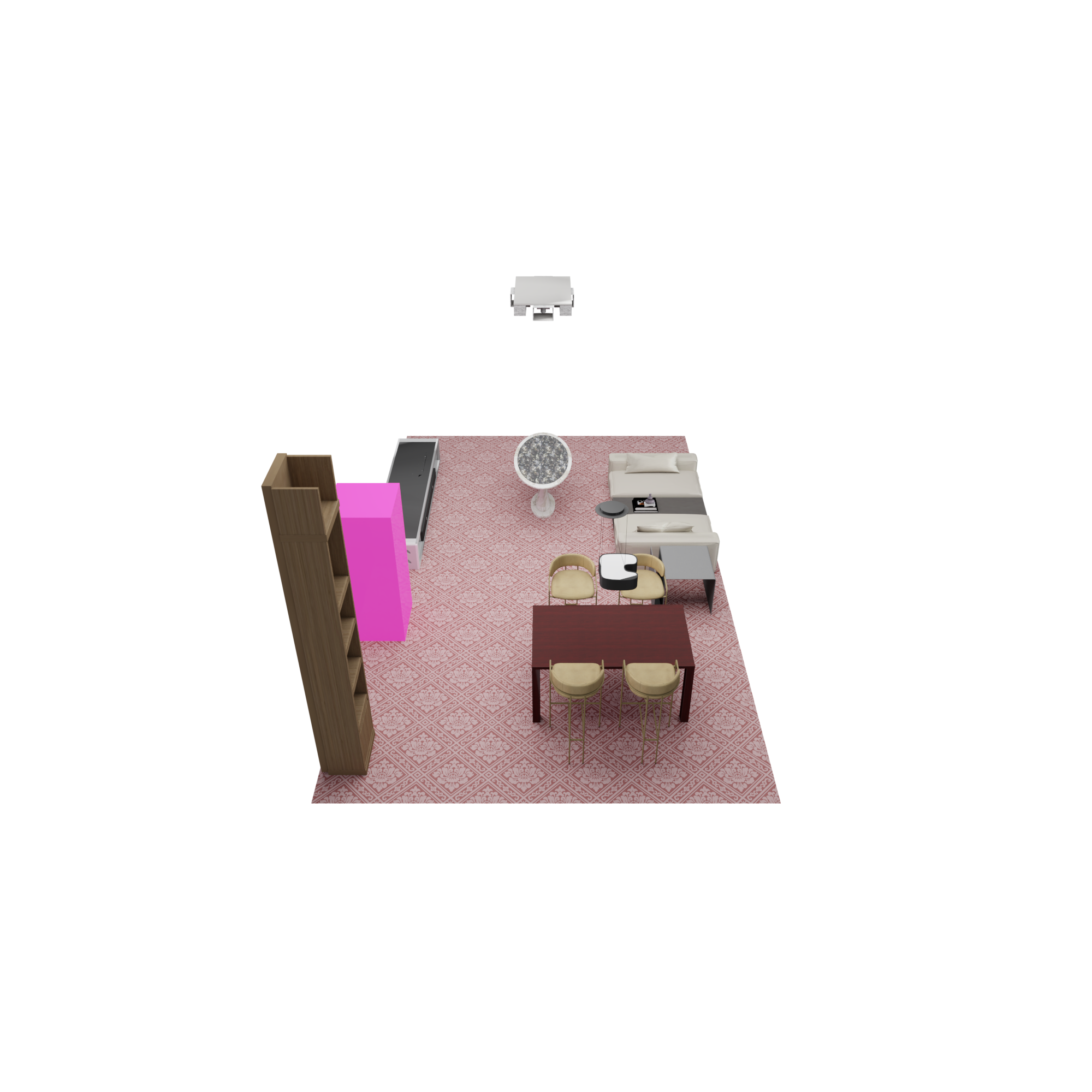}
    \end{subfigure}%
    \begin{subfigure}[b]{0.22\linewidth}
        \centering
            \begin{overpic}[width=\textwidth,  trim=450 500 500 450, clip]{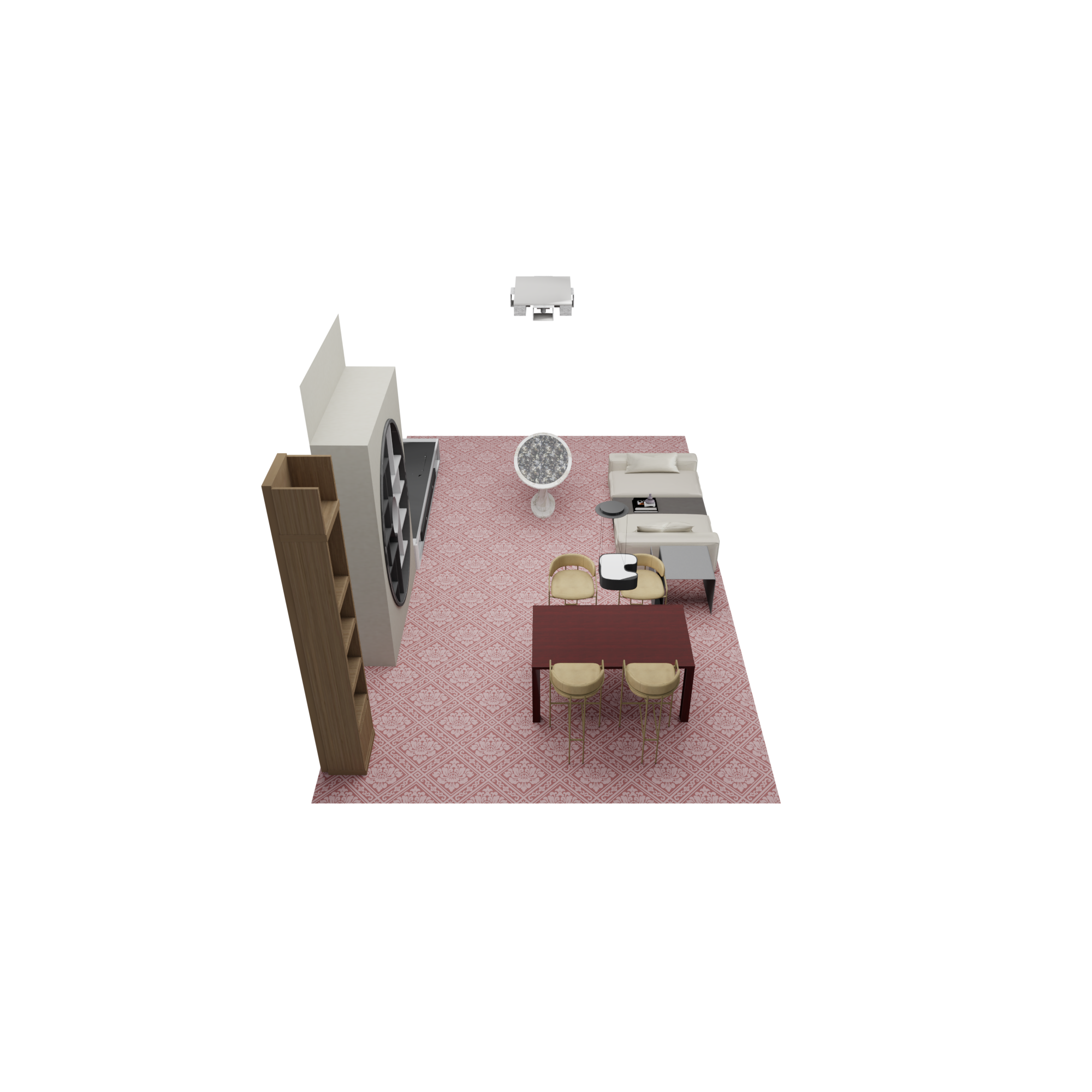}
        \end{overpic}
    \end{subfigure}%
    \hfill
    \begin{subfigure}[b]{0.22\linewidth}
        \centering
        \begin{overpic}[width=\textwidth,  trim=450 500 500 450, clip]{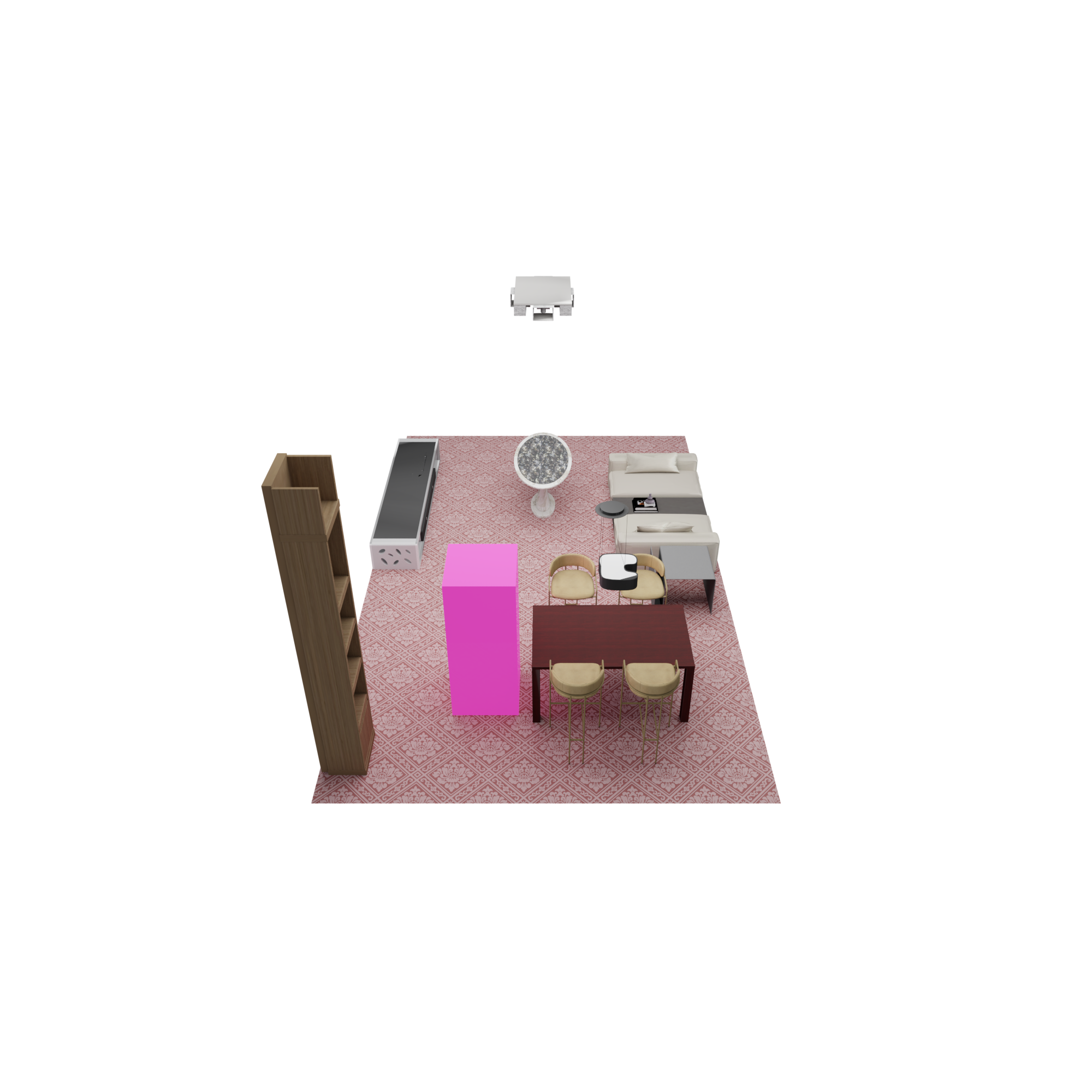}
	    \end{overpic}
    \end{subfigure}%
    \begin{subfigure}[b]{0.22\linewidth}
        \centering
        \begin{overpic}[width=\textwidth, trim=450 500 500 450, clip]{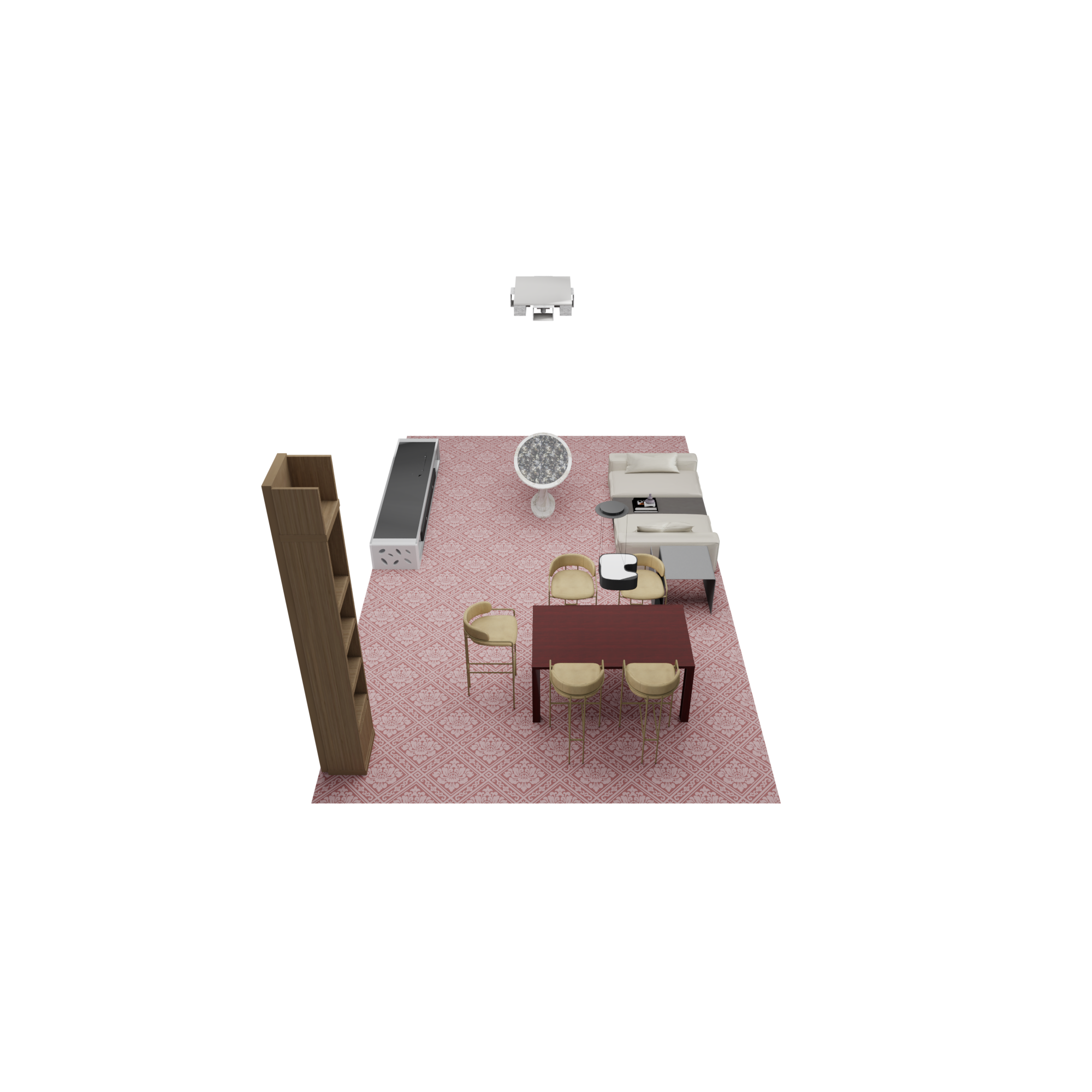}
        \end{overpic}
    \end{subfigure}%
    \vskip\baselineskip%
    \vspace{-0.75em}
    \caption{\textbf{Arbitrary conditioning:} We constrain the location of the next object to be sampled. The location is highlighted in pink. We sample location after the class. In a normal autoregressive model, we would not be able to constrain on the location. The bidirectional attention in the encoder allows the network to generate \textit{current} tokens based on \textit{future} tokens. In this example, the network automatically infers the proper class and size and even matches the style of the chairs in the example on the left.} 
    \label{fig:arbitrary_cond}
    
\end{figure*}

%%%%%%%%%%%%%%%%%%%%%%%%%%%%%%%%%%%%%%%%%%%%%%%%%%%%%%%%%%%%%%%%%%%%%%%%%%%%%%%%%%%%%%%%%%%%%%%%%%%%5
\para{Arbitrary Conditioning}  
To condition on an arbitrary sub-set of object attributes, we can construct a condition $C$ that is filled with \mask tokens, i.e. a fully unconstrained setting, and then only replace the tokens in $C$ corresponding to the attributes we want to constrain with constraint values. The non-mask tokens do not need to be contiguous in $C$. Any subset of $C$, for example only object class attributes or only object locations, can be used as unmasked condition. Given such a constraint $C$, the generator proceeds as described in Section~\ref{sec:implementation} and fills in all mask tokens with generated values. The encoder allows each generation step to have full knowledge of the condition $C$. See Fig.~\ref{fig:arbitrary_cond} for examples.
%
%We sample autoregressively - first class, then translation, size and angle. Yet, as described in Section 3, our model has bidirectional attention on all encoder tokens. Thus, at inference time, we can replace any token with a \mask token and allow the network to estimate the likelihood of the \mask token under the current parameters. This can be used for arbitrary conditioning in the following manner, say we want to add an object to a scene - we append a sequence of \mask tokens to the existing sequence. Then in order to specify the class, location, size or angle of the desired object we just replace the corresponding \mask token and allow the network to predict the distributions of the remaining parameters. We show a few examples in Fig. \cite{fig:arbitrary_cond}. This allows the network to \textit{look-ahead} and \textit{see} what parameters to generate in the future, enabling it to change the current probabilities accordingly.
To the best of our knowledge, ours is the first model that allows for this form of conditioning without imposing an order on the generation of parameters.

%% file: sections/conclusion.tex
\section{Conclusions}
We proposed a new framework to produce layouts with auto-regressive transformers with arbitrary conditioning information. While previous work was only able to condition on a set of complete objects, we extend this functionality and also allow for conditioning on individual attributes of objects. Our framework thereby enables several new modeling applications that cannot be achieved by any published framework.

\paragraph{Societal Impact.} We do not expect any negative social impact that is directly linked to our work. We do acknowledge possible concerns about the energy consumption of training auto-regressive transformers.

\paragraph{Limitations and Future Work.}
Our work inherits some limitations of auto-regressive models (or even generative models in general) in that we require long training times and that setting up the code is laborious, time-consuming and error-prone. In addition, it requires significant expert knowledge in training transformers. We will therefore release the code upon acceptance. Another limitation of the work is that all objects need to have the same number of attributes. An open challenge for future work is to extend our approach to layouts where each object is described by a token sequence of variable length.
Further, we would like to work on scaling auto-regressive models to much larger sequences, e.g., by using hierarchical auto-regressive models. In future, we would also like to explore parallel sampling strategies to sidestep the slow autoregressive generation process.

%\subsection{Limitations}
%While the 

\begin{table}[t]
\centering
\small
\caption{\textbf{Comparison on Synthesis Time}: We compare the time required to synthesize a single scene for different categories. Our model outperforms all other models on this metric, and is competitive on FID in comparison to ATISS. Our model achieves this while being parameter efficient. We only have about half the parameters of the next closest competitor. About 10m of these parameters come from the ResNet-18 encoder.}
\setlength\tabcolsep{2pt}
\begin{tabular}{@{}lcccccccc|c@{}}
\toprule
\multicolumn{1}{l}{} & \multicolumn{4}{c}{Synthesis Time ($\downarrow$)}                                 & \multicolumn{4}{c|}{FID ($\downarrow$)}                                              & \multirow{2}{*}{\begin{tabular}{c}
                                             Params  \\
                                             ($\downarrow$)
                                        \end{tabular}}   \\

\cmidrule(l){2-5} \cmidrule(l){6-9} 
                  & \bed             &  \liv         & \din           & \lib                           & \bed             & \liv          & \din              &  \lib                           & \\ \midrule
{FastSynth}       & 13193.77         &  30578.54     & 26596.08       & \multicolumn{1}{c|}{10813.87}  &  88.1             & 66.6           & 58.9             & \multicolumn{1}{c|}{ 86.6}         &  38.1  \\
{SceneFormer}     & 849.37           &  731.84       &  901.17        &  \multicolumn{1}{c|}{369.74}   &  90.6             & 68.1           & 60.1           & \multicolumn{1}{c|}{ 89.1}            & 129.29 \\
{ATISS}           & 102.38           &  201.59       &  201.84        & \multicolumn{1}{c|}{88.24}     &  \textbf{73.0 }  & 43.32         & 47.66           & \multicolumn{1}{c|}{\textbf{ 75.34}}       & 36.05      \\%
{Ours}            &  \textbf{33.69}  &\textbf{77.37} &\textbf{76.75}  & \multicolumn{1}{c|}{ \textbf{29.32}}  &  73.2            &\textbf{35.9} &\textbf{43.12}  & \multicolumn{1}{c|}{75.72}           & \textbf{19.44} \\
\bottomrule
\end{tabular}
\vspace{-0.1cm}
\end{table}